\documentclass[10pt,journal,compsoc]{IEEEtran}

\ifCLASSOPTIONcompsoc
  \usepackage{cite}
\else
  \usepackage{cite}
\fi
\ifCLASSINFOpdf
   \usepackage[pdftex]{graphicx}
   \graphicspath{{./media/}}
\else
   \usepackage[dvips]{graphicx}
\fi
\usepackage{amsmath}
\usepackage{float}

\usepackage{nidanfloat}

\usepackage{url}
\newcommand\MYhyperrefoptions{
	hypertexnames=false,
	bookmarks=true,bookmarksnumbered=true,
	pdfpagemode={UseOutlines},plainpages=false,pdfpagelabels=true,
	colorlinks=true,linkcolor={black},citecolor={black},urlcolor={black},
	pdftitle={Solving Inverse Problems With Deep Neural Networks - Robustness Included?},
	pdfsubject={Research Article},
	pdfauthor={Martin Genzel, Jan Macdonald, and Maximilian Maerz},
	pdfkeywords={inverse problems, image reconstruction, deep neural networks, adversarial robustness, medical imaging}}
\ifCLASSINFOpdf
\usepackage[\MYhyperrefoptions,pdftex]{hyperref}
\else
\usepackage[\MYhyperrefoptions,breaklinks=true,dvips]{hyperref}
\usepackage{breakurl}
\fi
\usepackage{xargs}
\usepackage{amsfonts}
\usepackage{mathrsfs}
\usepackage{dsfont}
\usepackage{amssymb}
\usepackage{units}
\usepackage{amsthm}
\usepackage{thmtools}
\usepackage[fixamsmath,disallowspaces]{mathtools}
\usepackage{etoolbox}
\usepackage{ifthen}
\usepackage[export]{adjustbox}
\usepackage{booktabs}
\usepackage{multirow}
\usepackage{siunitx}
\usepackage{tikz}
\usetikzlibrary{calc, fit}
\usepackage{rotating}
	{\clearpage\pagebreak[4]\global\pdfpageattr\expandafter{\the\pdfpageattr/Rotate 90}}%
	{\clearpage\pagebreak[4]\global\pdfpageattr\expandafter{\the\pdfpageattr/Rotate 0}}%
\usepackage{relsize}

\usepackage[textsize=tiny,english,colorinlistoftodos,
disable,
]{todonotes}
\usepackage{enumitem}
\newenvironment{listing}{\begin{itemize}[itemindent=0em,leftmargin=1.2em]}{\end{itemize}}

\usepackage{lipsum}
\usepackage{capt-of}

\newcommand{\opleft}[1]{\mathopen{}\left#1}
\newcommand{\opright}[1]{\right#1\mathclose{}}
\newcommandx{\braces}[4]{%
\ifstrequal{#3}{normal}{#1#4#2}{%
\ifstrequal{#3}{auto}{\left#1#4\right#2}{%
\ifstrequal{#3}{opauto}{\opleft#1#4\opright#2}{%
#3#1#4#3#2}}}%
}
\newcommand{\ifargdef}[3][{}]{\ifthenelse{\equal{#2}{}}{#1}{#3}}

\newcommand{\R}{\mathbb{R}} 
\newcommand{\C}{\mathbb{C}} 

\newcommand{\suchthat}[1][normal]{\ifstrequal{#1}{normal}{\mid}{#1|}} 

\newcommandx{\intvcl}[3][1=normal]{\braces{[}{]}{#1}{#2, #3}} 
\newcommandx{\intvop}[3][1=normal]{\braces{(}{)}{#1}{#2, #3}} 
\newcommandx{\intvclop}[3][1=normal]{\braces{[}{)}{#1}{#2, #3}} 
\newcommandx{\intvopcl}[3][1=normal]{\braces{(}{]}{#1}{#2, #3}} 
\DeclareMathOperator*{\argmin}{argmin} 
\DeclareMathOperator*{\argmax}{argmax} 

\newcommandx{\abs}[2][1=normal]{\braces{\lvert}{\rvert}{#1}{#2}} 
\newcommandx{\ceil}[2][1=normal]{\braces{\lceil}{\rceil}{#1}{#2}} 
\newcommandx{\floor}[2][1=normal]{\braces{\lfloor}{\rfloor}{#1}{#2}} 
\newcommandx{\round}[2][1=normal]{\braces{[}{]}{#1}{#2}} 
\newcommandx{\der}[1]{D^{#1}} 
\newcommandx{\gradient}{\nabla} 
\newcommandx{\partder}[4][1={},4={}]{\frac{\partial^{#4} #2}{\partial #3^{#4}}\ifargdef{#1}{\Big|_{#1}}} 
\newcommandx{\integ}[4][1={},2={}]{\int_{#1}^{#2} #3 \, #4} 
\newcommandx{\asympffaster}[2][1=normal]{o\braces{(}{)}{#1}{#2}} 
\newcommandx{\asympfaster}[2][1=normal]{O\braces{(}{)}{#1}{#2}} 
\newcommandx{\asympeq}[2][1=normal]{\Theta\braces{(}{)}{#1}{#2}} 
\newcommandx{\asympsslower}[2][1=normal]{\omega\braces{(}{)}{#1}{#2}} 
\newcommandx{\asympslower}[2][1=normal]{\Omega\braces{(}{)}{#1}{#2}} 

\newcommandx{\norm}[2][1=normal]{\braces{\|}{\|}{#1}{#2}} 
\renewcommandx{\sp}[3][1=normal]{\braces{\langle}{\rangle}{#1}{#2, #3}} 
\newcommand{\adj}[1]{{#1}^\ast} 
\newcommandx{\End}[2][2={}]{\mathcal{L}\opleft( #1 \ifargdef{#2}{, #2} \opright)} 
\renewcommand{\vec}[1]{\boldsymbol{#1}} 
\newcommandx{\opnorm}[2][1=normal]{\norm[#1]{#2}_{\operatorname{op}}} 
\newcommandx{\ball}[2][1={},2={}]{B_{#1}^{#2}} 

\newcommandx{\measure}[2][1=normal]{\operatorname{vol}\braces{(}{)}{#1}{#2}} 
\newcommandx{\Leb}[3][1={},3=normal]{L^{#2}\ifargdef{#1}{\braces{(}{)}{#3}{#1}}{}} 
\newcommandx{\Lebnorm}[4][1=normal,3={2},4={}]{\norm[#1]{#2}_{#3}} 
\renewcommandx{\l}[3][1={},3=normal]{\ell_{#2}\ifargdef{#1}{\braces{(}{)}{#3}{#1}}} 
\newcommandx{\lnorm}[4][1=normal,3={2},4={}]{\norm[#1]{#2}_{#3}} 
\newcommandx{\Smooth}[4][1={},3={},4=normal]{C_{#3}^{#2}\ifargdef{#1}{\braces{(}{)}{#4}{#1}}} 
\newcommandx{\Schwartz}[2][1={},2=normal]{\mathscr{S}\ifargdef{#1}{\braces{(}{)}{#2}{#1}}} 
\newcommandx{\Schwartzpoly}[2][1=normal]{\braces{\langle}{\rangle}{#1}{\abs[#1]{#2}} } 
\newcommandx{\Tempdistr}[2][1={},2=normal]{\mathscr{S}'\ifargdef{#1}{\braces{(}{)}{#2}{#1}}} 
\newcommandx{\distrinp}[3][1=normal]{\braces{\langle}{\rangle}{#1}{#2, #3}} 
\newcommandx{\ft}[3][1=default,2=auto]{
\ifstrequal{#1}{default}{\widehat{#3}}{
\ifstrequal{#1}{long}{{\braces{(}{)}{#2}{#3}}^{\wedge}}{}}} 
\newcommand{\ftop}{\mathcal{F}} 
\newcommandx{\ift}[3][1=default,2=auto]{
\ifstrequal{#1}{default}{\check{#3}}{
\ifstrequal{#1}{long}{{\braces{(}{)}{#2}{#3}}^{\vee}}{}}} 

\newcommandx{\prob}[2][1={},2=normal]{\mathbb{P}\ifargdef{#1}{\braces{[}{]}{#2}{#1}}}
\newcommandx{\mean}[2][1={},2=normal]{\mathbb{E}\ifargdef{#1}{\braces{[}{]}{#2}{#1}}}
\newcommandx{\var}[2][1={},2=normal]{\mathbb{V}\ifargdef{#1}{\braces{[}{]}{#2}{#1}}}

\newcommandx{\Unif}[2][1=normal]{\mathcal{U}\braces{(}{)}{#1}{#2}} 
\newcommandx{\Normdistr}[3][1=normal]{\mathcal{N}\braces{(}{)}{#1}{#2, #3}} 
\newcommandx{\Poi}[2][1=normal]{\mathrm{Poi}\braces{(}{)}{#1}{#2}} 
\newcommandx{\normsubg}[2][1=normal]{\norm[#1]{#2}_{\psi_2}} 

\newcommand{\adv}{\text{adv}}

\newcommand{\y}{\vec{y}} 
\newcommand{\ygrtr}{\vec{y}_0} 
\newcommand{\yadv}{\vec{y}_{\adv}}

\newcommand{\Noise}{\vec{e}} 
\newcommand{\Noiseadv}{\vec{e}_{\adv}}

\newcommand{\noisebnd}{\eta}
\newcommand{\jitterbnd}{\widetilde{\eta}}
\newcommand{\A}{\mathcal{A}} 

\newcommand{\x}{\vec{x}} 
\newcommand{\xsolu}{\hat{\vec{x}}} 
\newcommand{\xgrtr}{\vec{x}_0} 

\newcommand{\TV}{\textrm{TV}} 
\newcommand{\grad}{\nabla}
\newcommand{\tvnorm}[1]{\lnorm{\grad #1}[1]}

\newcommand{\TiraBB}{\mathcal{T}}
\newcommand{\UNetBB}{\mathcal{U}}
\newcommand{\Lin}{\mathcal{L}}
\newcommand{\DC}{\mathcal{DC}}
\newcommand{\dcparam}{\lambda}

\newcommand{\NNparam}{\vec{\theta}}
\newcommand{\wdparam}{\mu}

\newcommand{\ConvNet}{\textrm{ConvNet}}
\newcommand{\CompClass}{\textrm{CC}}
\newcommand{\UNet}{\textrm{UNet}}
\newcommand{\Net}{\textrm{Net}}
\newcommand{\Rec}{\textrm{Rec}}
\newcommand{\UNetFL}{\textrm{UNetFL}}
\newcommand{\Tira}{\textrm{Tira}}
\newcommand{\TiraFL}{\textrm{TiraFL}}
\newcommand{\ItNet}{\textrm{ItNet}}

\newcommand{\inv}[1]{#1^\ddagger}

\newcommand{\graphicwithlabel}[5]{%
  \adjustbox{valign=c}{%
    \begin{tikzpicture}%
      \node[inner sep=0em] (IMG) at (0,0) {\includegraphics[width=#2]{#1}};
      \coordinate (VERTICAL) at ($ (IMG.south)!#4!(IMG.north) $);
      \coordinate (HORIZONTAL) at ($ (IMG.west)!#5!(IMG.east) $);
      \node (LABEL) at (VERTICAL-|HORIZONTAL) {#3};
    \end{tikzpicture}
  }
}

\newcommand{\graphicwithbox}[6]{%
  \adjustbox{valign=c}{%
    \begin{tikzpicture}%
      \node[inner sep=0em] (IMG) at (0,0) {\includegraphics[width=#2]{#1}};
      \coordinate (VERTICAL_LO) at ($ (IMG.south)!#3!(IMG.north) $);
      \coordinate (HORIZONTAL_LO) at ($ (IMG.west)!#4!(IMG.east) $);
      \coordinate (VERTICAL_HI) at ($ (IMG.south)!#5!(IMG.north) $);
      \coordinate (HORIZONTAL_HI) at ($ (IMG.west)!#6!(IMG.east) $);
      \draw[ultra thick, color={rgb,1:red,0.6314;green,.7882;blue,0.9569}] (VERTICAL_LO-|HORIZONTAL_LO) rectangle (VERTICAL_HI-|HORIZONTAL_HI);
    \end{tikzpicture}
  }
}

\newcommand{\graphicwitharrows}[8]{%
  \adjustbox{valign=c}{%
    \begin{tikzpicture}%
      \node[inner sep=0em] (IMG) at (0,0) {\includegraphics[width=#2]{#1}};
      \coordinate (VERTICAL1) at ($ (IMG.south)!#3!(IMG.north) $);
      \coordinate (HORIZONTAL1) at ($ (IMG.west)!#4!(IMG.east) $);
      \draw[<-, >=stealth, ultra thick, color={rgb,1:red,0.6314;green,.7882;blue,0.9569}] (VERTICAL1-|HORIZONTAL1) -- ++ (#5:.4);
      \coordinate (VERTICAL2) at ($ (IMG.south)!#6!(IMG.north) $);
      \coordinate (HORIZONTAL2) at ($ (IMG.west)!#7!(IMG.east) $);
      \draw[<-, >=stealth, ultra thick, color={rgb,1:red,0.6314;green,.7882;blue,0.9569}] (VERTICAL2-|HORIZONTAL2) -- ++ (#8:.4);
    \end{tikzpicture}
  }
}

\begin{document}
%
\title{Solving Inverse Problems With Deep \\ Neural Networks -- Robustness Included?}
%
%
%
%

\author{Martin Genzel, Jan Macdonald,
        and Maximilian M\"{a}rz
\IEEEcompsocitemizethanks{\IEEEcompsocthanksitem M.~Genzel is with the Mathematical Institute of Utrecht University, Utrecht, Netherlands. 
\IEEEcompsocthanksitem J.~Macdonald and M.~M\"{a}rz are with the Institute of Mathematics of Technical University of Berlin, Berlin, Germany.}
\thanks{All authors contributed equally. Correspondence may be addressed to \href{mailto:maerz@math.tu-berlin.de}{\texttt{maerz@math.tu-berlin.de}}.}}

\IEEEtitleabstractindextext{%
\begin{abstract}
	In the past five years, deep learning methods have become state-of-the-art in solving various inverse problems.
	Before such approaches can find application in safety-critical fields, a verification of their reliability appears mandatory.
	Recent works have pointed out instabilities of deep neural networks for several image reconstruction tasks.
	In analogy to adversarial attacks in classification, it was shown that slight distortions in the input domain may cause severe artifacts.
	The present article sheds new light on this concern, by conducting an extensive study of the robustness of deep-learning-based algorithms for solving underdetermined inverse problems.
	This covers compressed sensing with Gaussian measurements as well as image recovery from Fourier and Radon measurements, including a real-world scenario for magnetic resonance imaging (using the NYU-fastMRI dataset).
	Our main focus is on computing adversarial perturbations of the measurements that maximize the reconstruction error.
	A distinctive feature of our approach is the quantitative and qualitative comparison with total-variation minimization, which serves as a provably robust reference method.
	In contrast to previous findings, our results reveal that standard end-to-end network architectures are not only resilient against statistical noise, but also against adversarial perturbations.
	All considered networks are trained by common deep learning techniques, without sophisticated defense strategies.
\end{abstract}

}

\maketitle

\IEEEdisplaynontitleabstractindextext

%
\IEEEpeerreviewmaketitle

\ifCLASSOPTIONcompsoc
\IEEEraisesectionheading{\section{Introduction}\label{sec:introduction}}
\else
\section{Introduction}
\label{sec:introduction}
\fi

%
%
%
%

\IEEEPARstart{S}{ignal} reconstruction from indirect measurements plays a central role in a variety of applications, including medical imaging \cite{ldsp08}, communication theory \cite{hbrn08}, astronomy \cite{spm02}, and geophysics \cite{tar81}.
Such tasks are typically formulated as an inverse problem, which in its prototypical, finite-dimensional form reads as follows:
\begin{equation}
	\left\{\begin{gathered}
	\textit{Given a linear forward operator $\A \in \R^{m \times N}$} \\[-.15\baselineskip]
	\textit{and corrupted measurements $\y = \A \xgrtr + \Noise$} \\[-.15\baselineskip]
	\textit{with $\lnorm{\Noise} \leq \noisebnd$, reconstruct the signal $\xgrtr$.}
	\end{gathered}\right\} \label{eq:intro:problem}
\end{equation}
The ubiquitous presence of noise makes it indispensable that a reconstruction method has to be \emph{robust} against additive perturbations $\Noise$.
Furthermore, the measurement process is often costly and potentially harmful.
Therefore, the underdetermined regime where $m \ll N$ has gained much attention during the last two decades.
This restriction turns \eqref{eq:intro:problem} into an \emph{ill-posed inverse problem}, which does not possess a unique solution.

Under the additional assumption of sparsity, the methodology of \emph{compressed sensing} has proven that accurate and robust reconstruction from incomplete measurements is still possible \cite{fh13}.
This means that a solution map $\Rec\colon \R^m \to \R^N$ for \eqref{eq:intro:problem} satisfies an error bound of the form
\begin{equation} \label{eq:intro:robust}
	\lnorm{\xgrtr - \Rec(\y)} \leq C \cdot \noisebnd,
\end{equation}
where $C > 0$ is a small constant.
Although state-of-the-art in various real-world applications, the practicability of the associated algorithms is often limited by computational costs, manual parameter tuning, and a mismatch between sparsity models and data.

Building on the recent success of artificial intelligence in computer vision \cite{ksh12,lbh15,gbc16}, there has been a considerable effort to solve the inverse problem \eqref{eq:intro:problem} by means of \emph{deep learning}, e.g., see \cite{kl10,yslx16,dlht16,kmy17,jmfu17,ham+17,che+17b,bjpd17,zlcrr18,bub+19} and \cite{amos19} for a recent survey.
This advance is primarily based on fitting an artificial \emph{neural network (NN)} model to a large set of data points in a supervised training procedure.
It is fair to say that such data-driven approaches can significantly outperform classical methods in terms of reconstruction accuracy and speed.
On the other hand, one may argue that the underlying mechanisms of NNs remain largely unclear \cite{ela17}.
Hence, in the absence of theoretical guarantees of the form \eqref{eq:intro:robust}, an empirical verification of their accuracy and robustness against measurement noise is crucial.

While a number of works report a remarkable resilience against noise \cite{che+17,zlcrr18,haao20}, several alarming findings indicate that deep-learning-based reconstruction schemes are typically unstable \cite{hua+18,arpah20,gaah20,rbk20}.
In particular, the recent study of Antun et al.~\cite{arpah20} suggests that deep learning for inverse problems comes at the cost of instabilities, in the sense that \textit{``[...] certain tiny, almost undetectable perturbations, both in the image and sampling domain, may result in severe artifacts in the reconstruction [...]''}.
In machine learning research on classification, such a sensitivity of NNs is a well-established phenomenon.
Initiated by Szegedy et al.~\cite{sze+14}, a substantial body of literature is devoted to \emph{adversarial attacks} (and their defenses), i.e., the computation of a visually imperceptible change to the input that fools the NN.
Typically, an ``attacker'' exploits gradient-based information in order to cross the discontinuous decision boundary of a classifier.
This can be a serious issue for sensitive applications where wrong predictions impose a security risk---imagine a misclassified stop sign in autonomous driving \cite{kgb17,eyk18}.

Despite these findings, it appears peculiar that solving inverse problems by deep-learning-based schemes might become unstable.
Learning a reconstruction algorithm can be seen as a regression task, where measurements are mapped to a high-dimensional signal manifold (e.g., medical images).
In contrast, a NN classifier maps to a low-dimensional, discrete output domain, resulting in a ``vulnerable'' decision boundary.
Moreover, it is well known that robust and accurate algorithms exist for many inverse problems.
Since these are often used as templates for NN architectures, it seems surprising that the latter should suffer from severe instabilities.
Clearly, the robustness against noise is quintessential for an application of deep learning in practice,  especially in sensitive fields such as biomedical imaging.
Therefore, we believe that a profound study of this topic is indispensable.

\subsection{Contributions}

This article is dedicated to a comprehensive numerical study of the robustness of NN-based methods for solving underdetermined inverse problems.
The primary objective of our experiments is to analyze how much the reconstruction error grows with the noise level $\noisebnd$.
We investigate this relationship in terms of statistical and adversarial noise: the former means that measurement noise is drawn from an appropriate probability distribution, while the latter explores worst-case perturbations that maximize the reconstruction error for fixed $\noisebnd$.
Similar to adversarial attacks in classification, computing worst-case noise is based on a non-convex formulation that is addressed by automatic differentiation and a gradient descent scheme.
In the absence of an empirical certificate of robustness, a central and distinctive component of our analysis is the systematic comparison with a classical benchmark method with provable guarantees, namely total-variation (TV) minimization.
In this case, evaluating the gradient is non-trivial and carried out by unrolling the underlying optimization problem.

Our experiments consider several prototypical inverse problems as use cases. This includes classical compressed sensing with Gaussian measurements as well as the reconstruction of phantom images from Radon and Fourier measurements. Furthermore, a real-world scenario for magnetic resonance imaging (MRI) is investigated, based on the NYU-fastMRI dataset \cite{zbo+19,kno+20}.
We examine a representative selection of learned reconstruction architectures, reaching from simple post-processing NNs to iterative schemes.
In total, this work presents a robustness analysis of more than 25 NNs, each of them trained in-house with publicly available code.\footnote{Our Python implementation, based on the \emph{PyTorch} package \cite{pas+17}, can be found under \url{https://github.com/jmaces/robust-nets}.}

Our main findings may be summarized as follows:

	(i) In every considered scenario, we find deep-learning-based methods that are at least as robust as TV minimization with respect to adversarial noise.
	This does not require sophisticated architectures or defense strategies.

	(ii) All trained NNs are remarkably robust against statistical noise.
	Although TV minimization may yield exact recovery for noiseless measurements, it is still outperformed by learned methods in mid- to high-noise regimes.

	(iii) The reconstruction performance is affected by the underlying NN architecture.
	For instance, promoting data consistency in iterative schemes may improve both accuracy and robustness.

	(iv) One should not commit the ``inverse crime'' of training a NN with \emph{noiseless} data, which may cause an unstable behavior for higher noise levels.
	We demonstrate that simply adding white Gaussian noise to the training measurements is an effective remedy---a regularization technique that is commonly known as \emph{jittering} in machine learning research.

Apart from these observations, our work is, to the best of our knowledge, the first to empirically characterize the performance gap between adversarial and statistical noise in the context of \eqref{eq:intro:problem}.
In particular, this gap is not exclusive to deep-learning-based schemes but also appears for classical methods such as TV minimization.
Our central conclusion is:
\begin{quote}\itshape
	The existence of adversarial examples in classification tasks does not naturally carry over to NN-based solvers for inverse problems.
	Such reconstruction schemes may not only supersede classical methods as state-of-the-art, but they can also exhibit a similar degree of robustness.
\end{quote}
Since our study as it is has required massive computational resources (\textgreater2 years of GPU computation time), some aspects have to remain unexplored, see Section~\ref{sec:discussion} for a discussion.
In particular, given the sheer number of NN architectures, we explicitly do \emph{not} claim that every deep-learning-based method is stable (cf.~Section~\ref{sec:additional:crime}).
Nevertheless, our findings suggest that fairly standard workflows allow for surprisingly robust reconstruction schemes.
This offers an alternative and novel perspective on the reliability of deep learning strategies in inverse problems.
Therefore, we believe that the present work takes an important step towards their safe use in practice.

\subsection{Organization of This Article}

Section~\ref{sec:literature} is devoted to relevant previous works, followed by a conceptual overview of our approach in Section~\ref{sec:methods}. The latter introduces all considered reconstruction methods, the associated NN architectures as well as our attack strategy to analyze their adversarial robustness.
The main results are then presented in Section~\ref{sec:results}, complemented by several additional experiments in Section~\ref{sec:additional}.
We conclude with a general discussion of our findings in Section~\ref{sec:discussion}.

\section{Related Work}
\label{sec:literature}

Initiated by Szegedy et al.~\cite{sze+14}, the vulnerability of deep NNs to adversarial examples has been the subject of more than 2500 publications \cite{car20}.
We refer to \cite{yhzl19,ommf20} for recent surveys of the field and further references.
The vast majority of existing articles is concerned with classification and related tasks, such as image segmentation \cite{amt20}. 
On the other hand, only few works have explicitly addressed the adversarial robustness of learned solvers for inverse problems.

To the best of our knowledge, Huang et al.~\cite{hua+18} have made the first effort to transfer adversarial attacks to NN-based reconstruction methods.
They demonstrate that a distortion of the network's input may result in the loss of small image features.
However, their initial findings are restricted to the specific problem of limited angle computed tomography, where the robust recovery of certain parts of the image is provably impossible \cite{qui93}. Moreover, the proposed perturbation model is non-standard and does not correspond to noise in the measurements.

More recently, the topic was brought to attention by the inspiring article of Antun et al.~\cite{arpah20}. 
Their numerical experiments show instabilities of existing deep NNs with respect to adversarial noise, out-of-distribution features, and changes in the number of measurements.
An important difference to our work is that adversarial noise is only computed for learned schemes.
We believe that a comparative ``attack'' of a classical benchmark method is crucial for a fair assessment of robustness.
Furthermore, the results of \cite{arpah20} are reported qualitatively by visualizing reconstructed images, as it is common in adversarial machine learning.
We argue that the mathematical setup of the inverse problem \eqref{eq:intro:problem} calls for a quantitative error analysis that is in line with the bound of \eqref{eq:intro:robust}.
Finally, the training stage of the networks in \cite{arpah20} does not seem to account for noise, which we have identified as a potential source of instability, see Section~\ref{sec:additional:crime}.
Note that our study also analyzes the FBPConvNet architecture \cite{jmfu17}, a relative of AUTOMAP~\cite{zlcrr18}, and an iterative scheme similar to DeepMRI~\cite{schpr17}. 
Nevertheless, a one-to-one comparison to \cite{arpah20} is subtle due to task-specific architectures and data processing.
A follow-up work of \cite{arpah20} presents a theoretical characterization of instabilities in terms of the kernel of the forward operator \cite{gaah20}.
Our results provide empirical evidence that the considered deep-learning-based schemes could be kernel aware (cf.~Section~\ref{sec:additional:badmeas}).

As a countermeasure to the outcome of \cite{arpah20}, Raj et al.~\cite{rbk20} suggest a sophisticated defense strategy resulting in robust networks.
This work also addresses shortcomings of the attack strategy in \cite{arpah20}, see Section~\ref{sec:methods:adv} for details.
Finally, in line with our findings, Kobler et al.~\cite{kekp20} propose the data-driven \emph{total deep variation} regularizer and demonstrate its adversarial robustness for image denoising.

\section{Methods and Preliminaries}
\label{sec:methods}

In this section, we briefly introduce the considered reconstruction schemes for solving the inverse problem \eqref{eq:intro:problem}.
This includes a representative selection of NN-based methods and total-variation minimization as a classical benchmark.
Furthermore, our attack strategy to analyze their adversarial robustness is presented.

\subsection{Neural Network Architectures}
\label{sec:methods:nnarch}

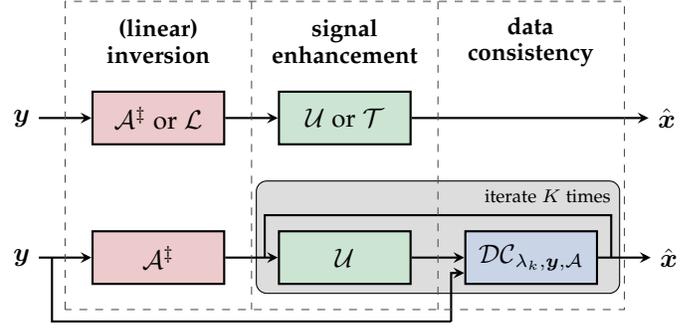
\begin{figure}
\centering
\begin{tikzpicture}
    \pgfdeclarelayer{background}
    \pgfsetlayers{background,main}

    \definecolor{deepred}{HTML}{C44E52}
    \definecolor{deepgreen}{HTML}{55A868}
    \definecolor{deepblue}{HTML}{4C72B0}
    \definecolor{deepgray}{HTML}{8C8C8C}

    \node[minimum height=.7cm] (y) at (0,0) {$\y$};
    \node[thick, draw, fill=deepred!30, minimum width=1.75cm, minimum height=.7cm, right=.7cm of y] (inv) {$\inv\A$ or $\Lin$};
    \node[thick, draw, fill=deepgreen!30, minimum width=1.75cm, minimum height=.7cm, right=.7cm of inv] (net) {$\UNetBB$ or $\TiraBB$};
    \node[minimum width=1.75cm, minimum height=.7cm, right=.7cm of net] (dc) {};
    \node[minimum height=.7cm, below=1.5cm of y] (y2) at (0,0) {$\y$};
    \node[thick, draw, fill=deepred!30, minimum width=1.75cm, minimum height=.7cm, right=.7cm of y2] (inv2) {$\inv\A$};
    \node[thick, draw, fill=deepgreen!30, minimum width=1.75cm, minimum height=.7cm, right=.7cm of inv2] (net2) {$\UNetBB$};
    \node[thick, draw, fill=deepblue!30, minimum width=1.75cm, minimum height=.7cm, right=.7cm of net2] (dc2) {$\DC_{\dcparam_k,\y,\A}$};

    \coordinate[above right=.2cm and .175cm of dc2] (abovedc);
    \coordinate[above left=.2cm and .175cm of net2] (abovenet);

    \draw[thick, ->, >=stealth] (y) -- (inv);
    \draw[thick, ->, >=stealth] (inv) -- (net);
    \draw[thick, ->, >=stealth] (net.east) -- ++ (3.15, 0) node[right] {$\xsolu$};
    \draw[thick, ->, >=stealth] (y2) -- coordinate[midway](y2inv) coordinate[pos=0.25](y2inv2) (inv2);
    \draw[thick, ->, >=stealth] (inv2) -- coordinate[midway](inv2net) (net2);
    \draw[thick, ->, >=stealth] (net2) -- coordinate[midway](net2dc) coordinate[pos=0.75](net2dc2) (dc2);
    \draw[thick, ->, >=stealth] (net2dc2)+(0,-.85) |- (dc2.193);
    \draw[thick] (net2dc2)+(0,-.85) -| (y2inv2);
    \draw[thick] (abovedc|-dc2) -- (abovedc) -- (abovenet) -- (abovenet|-net2);
    \draw[thick, ->, >=stealth] (dc2.east) -- coordinate[midway] (dc2x) ++ (.7, 0) node[right] {$\xsolu$};

    \node[inner sep=0cm, above=.25cm of abovedc, anchor=east] (it) {\scriptsize iterate $K$ times};
    \node[text height=.5cm, text depth=.3cm, minimum height=1cm, above=.15cm of inv] (invtext) {\parbox{2cm}{\centering\bfseries\small (linear)\\inversion}};
    \node[text height=.5cm, text depth=.3cm, minimum height=1cm, above=.15cm of net] (nettext) {\parbox{2cm}{\centering\bfseries\small signal\\enhancement}};
    \node[text height=.5cm, text depth=.3cm, minimum height=1cm, above=.15cm of dc] (dctext) {\parbox{2cm}{\centering\bfseries\small data\\consistency}};

    \draw[thin, dashed, draw=black!70] (y2inv|-invtext.north) ++ (0, -4.1) rectangle (dc2x|-invtext.north);
    \draw[thin, dashed, draw=black!70] (inv2net|-invtext.north) -- ++ (0, -4.1);
    \draw[thin, dashed, draw=black!70] (net2dc|-invtext.north) -- ++ (0, -4.1);

    \begin{pgfonlayer}{background}
    \node[thin, draw, rounded corners, fill=deepgray!30, fit=(net2) (dc2) (abovedc) (abovenet) (it)] {};
    \end{pgfonlayer}

\end{tikzpicture}
\caption{Schematic network reconstruction pipelines of $\UNet$, $\TiraFL$ (top), and $\ItNet$ (bottom).}
\label{fig:methods:pipeline}
\end{figure}

In the past five years, numerous deep-learning-based approaches for solving inverse problems have been developed;  see \cite{amos19,ojbmdw20} for overviews.
The present work focuses on a selection of widely used \emph{end-to-end network schemes} that define an explicit reconstruction map from $\R^m$ to $\R^N$, see also Fig.~\ref{fig:methods:pipeline}.

The first considered method is a \emph{post-processing network}:
\[
	\UNet \colon \R^m \to \R^N, \ \y \mapsto [\UNetBB\circ\inv\A](\y).
\]
It employs the U-Net architecture $\UNetBB \colon \R^N \to \R^N$ \cite{rfb15} as a residual network \cite{hzrs16} to enhance an initial, model-based reconstruction $\inv\A(\y)$. Here, $\inv\A\colon\R^m\to\R^N$ is an approximate inversion of the forward operator $\A$, e.g., the filtered back-projection for Radon measurements. Despite its simplicity, it has been demonstrated in \cite{jmfu17} that $\UNet$ is an effective solution method for \eqref{eq:intro:problem}; see also \cite{kmy17,che+17b,che+17,wlvi17,yan+18} for related approaches.

Our second reconstruction scheme is a \emph{fully-learned network:}
\[
	\TiraFL \colon \R^m \to \R^N, \ \y \mapsto [\TiraBB\circ\Lin](\y),
\]
which is closely related to $\UNet$, but differs in two aspects: It is based on the Tiramisu architecture $\TiraBB \colon \R^N \to \R^N$ \cite{jdvrb17} as a residual network, which can be seen as a refinement of the U-Net. While $\TiraBB$ shares the same multi-level structure, it is built from fully-convolutional dense-blocks \cite{hlvw17} instead of standard convolutional blocks.
More importantly, the fixed inversion $\inv\A$ is replaced by a learnable linear layer $\Lin \in \R^{N\times m}$, so that $\TiraFL$ does not contain fixed model-based components anymore.
The approach of $\TiraFL$ is similar to \cite{zlcrr18,sch+19}, which makes use of a fully-learned reconstruction map for MRI.
For the sake of completeness, we have also conducted experiments for $\Tira$, a Tiramisu-based post-processing network, as well as for $\UNetFL$, a U-Net-based fully-learned network, see Section~\ref{sec:supp:csA}--\ref{sec:supp:csC} in the supplementary material for results.

Finally, we also analyze an \emph{iterative network:}
\[
	\ItNet \colon \R^m \to \R^N, \ \y \mapsto \Big[\left(\bigcirc_{k=1}^K [\DC_{\dcparam_k,\y,\A} \circ \UNetBB ] \right)\circ \inv\A \Big] (\y)
\]
where
\[
	\DC_{\dcparam_k,\y,\A} \colon \R^N \to \R^N, \ \x \mapsto \x - \dcparam_k \cdot \adj\A\left(\A\x-\y\right).
\]
The scalar parameters $\dcparam_k$ are learnable and $\adj\A$ denotes the adjoint of $\A$.
Mathematically, $\DC_{\dcparam_k,\y,\A}$ performs a gradient step on the loss $\x \mapsto \frac{\dcparam_k}{2}\lnorm{\A\x - \y}^2$, promoting \emph{data consistent} solutions.
Therefore, the alternating cascade of $\ItNet$ can be seen as a proximal gradient descent scheme, where the proximal operator is replaced by a trainable enhancement network.
Here, the U-Net architecture is used again, due to its omnipresence in image-to-image processing tasks.
Unrolled methods in the spirit of $\ItNet$ are frequently used to solve inverse problems, e.g., see \cite{kl10,yslx16,ham+17,schpr17,amj18,ao18,hsqdsr19,chlf+20}.

\subsection{Neural Network Training}
\label{sec:methods:nntrain}

The learnable parameters of the networks are trained from sample data pairs $\{(\y^i=\A\xgrtr^i+\Noise^i, \xgrtr^i)\}_{i = 1}^M$ by minimizing an empirical loss function.
Depending on the use case, the signals $\xgrtr^i$ are either drawn from a fixed publicly available training dataset or according to a synthetic probability distribution.
If $\Net[\NNparam] \colon \R^m \to \R^N$ denotes a reconstruction network with all learnable parameters collected in $\NNparam$, then the training amounts to (approximately) solving
\begin{equation}\label{eq:methods:training}\textstyle
	\min_{\NNparam} \sum_{i=1}^M \ell\left(\Net[\NNparam](\y^i), \xgrtr^i\right) + \wdparam \cdot \lnorm{\NNparam}^2
\end{equation}
for some cost function $\ell \colon \R^N\times \R^N \to \R_{\geq 0}$, which is the squared distance unless stated otherwise.
Overfitting is addressed by $\l{2}$-regularization (weight decay) with a hyper-parameter $\wdparam \geq 0$.
In order to solve \eqref{eq:methods:training}, we utilize mini-batch stochastic gradient descent and the Adam optimizer \cite{kb14}.
We found that larger mini-batches were beneficial for the training performance during later epochs, which is achieved by gradient accumulation.

Due to the ubiquitous presence of noise in inverse problems, it is natural to account for it in the training data.
In many applications, measurement noise is modeled as an independent random variable, for instance, following a Gaussian distribution.
Therefore, the perturbation~$\Noise^i$ is treated as statistical noise during the training phase, i.e., a fresh realization is randomly drawn in each epoch. This technique is well known as \emph{jittering} in machine learning research, where it is primarily used to avoid overfitting \cite{sd91,hk92,bis95}; see also \cite{vllbm10}.
In Section~\ref{sec:additional:crime}, we relate jittering to the phenomenon of inverse crimes and demonstrate its importance for the robustness of learned reconstruction schemes.
Due to varying noise levels in the evaluation of our models, we design $\Noise^i$ as a centered Gaussian vector with random variance, such that its expected norm $\mean\lnorm{\Noise^i}$ is distributed uniformly in a range $[0,\jitterbnd]$ for some $\jitterbnd \geq 0$.

\subsection{Total-Variation Minimization}
\label{sec:methods:tv}

Dating back to the seminal work of Rudin et al.~\cite{rof92}, \emph{total-variation (TV) minimization} has become a standard tool for solving signal and image reconstruction tasks \cite{cl97,bb18}.
We apply it to the problem \eqref{eq:intro:problem} in the following form:
\begin{align}
	\TV[\noisebnd] \colon \R^{m} &\to \R^{N}, \label{eq:methods:tv} \\
	\y &\mapsto \argmin_{\x \in \R^N} \tvnorm{\x} \quad \text{s.t.} \quad \lnorm{\A \x - \y} \leq \noisebnd, \notag
\end{align}
where $\grad$ denotes a discrete gradient operator.
Crucial to the above optimization problem is the use of the $\l{1}$-norm, which is known to promote gradient-sparse solutions.
Indeed, under suitable assumptions on $\A$, compressed sensing theory suggests an error bound of the form \eqref{eq:intro:robust} for a gradient-sparse signal $\xgrtr$ and $\Rec = \TV[\noisebnd]$, e.g., see \cite{crt06a,nw13,poo15,gms20}.
In other words, TV minimization is provably robust with a near-optimal dependence on $\noisebnd$.
This particularly justifies its use as a reference method, allowing us to empirically characterize the robustness of learned reconstruction schemes.

In our numerical simulations, the problem of \eqref{eq:methods:tv} is solved by the \emph{alternating direction method of multipliers (ADMM)} \cite{gm75,gm76}.
For 1D signals, $\grad \in \R^{N \times N}$ is chosen as a forward finite difference operator with Neumann boundary conditions, extended by a constant row vector to capture the mean of the signal.
For image signals, $\grad \in \R^{2N \times N}$ corresponds to a forward finite difference operator with periodic boundary conditions. Finally, we emphasize that $\TV[\noisebnd]$ is explicitly adapted to the amount of perturbation of the measurements.

\subsection{Adversarial Noise}
\label{sec:methods:adv}

In the setup of \eqref{eq:intro:problem}, adversarial noise for a given reconstruction method $\Rec\colon \R^m \to \R^N$ can be computed by solving an optimization problem: for a fixed signal $\xgrtr \in \R^N$ and noise level $\noisebnd \geq 0$, find an additive perturbation $\Noiseadv \in \R^m$ of the noiseless measurements $\ygrtr = \A \xgrtr$ that maximizes the reconstruction error, i.e.,
\begin{equation}\label{eq:methods:findadv}
	\Noiseadv = \argmax_{\Noise\in\R^m} \lnorm{\Rec(\ygrtr + \Noise) - \xgrtr} \quad \text{s.t.} \quad \lnorm{\Noise} \leq \noisebnd.
\end{equation}
Such an attack strategy is a straightforward adaption of a common approach in adversarial machine learning \cite{yhzl19}.
In contrast to \cite{arpah20}, we consider a constrained optimization problem that avoids shortcomings of an unconstrained formulation; in particular, this allows for precise control over the noise level.
Moreover, \eqref{eq:methods:findadv} explores a natural perturbation model, operating directly in the measurement domain, cf.~the discussion in \cite{rbk20}.

In order to solve the problem \eqref{eq:methods:findadv}, we use the projected gradient descent algorithm in conjunction with the Adam optimizer, which was found to be most effective (cf.~\cite{cw17}).
The non-convexity of \eqref{eq:methods:findadv} is accounted for by choosing the worst perturbation out of multiple runs with random initialization.
Assuming a whitebox model (i.e., $\Rec$ is fully accessible), we use PyTorch's automatic differentiation \cite{pas+17} to compute gradients of the considered NN schemes.

A central aspect of our work is that the above perturbation strategy is also applied to $\TV[\noisebnd]$.
This is non-trivial, since the gradient of the implicit map $\y \mapsto \TV[\noisebnd](\y)$ has to be computed.
The large-scale nature of imaging problems prevents us from using the recent concept of differentiable convex optimization layers \cite{aabbdk19}.
Instead, we rely on unrolling the ADMM scheme for TV minimization, which again enables automatic differentiation.
However, a large number of iterations might be required to ensure convergence of ADMM. This leads to numerical difficulties when calculating the gradient of the unrolled algorithm, e.g., memory \& time constraints and error accumulation.
We address this issue by decreasing the number of ADMM iterations in combination with a pre-initialization of the primal and dual variables.

\section{Main Results}
\label{sec:results}

This section studies the robustness of NN-based solution methods for three different instances of the inverse problem \eqref{eq:intro:problem}.
The goal of our experiments is to assess the loss of reconstruction accuracy caused by noise.
To that end, we rely on two types of visualization:
\begin{listing}
\item
	\emph{Noise-to-error curves} are generated by plotting the relative noise level $\noisebnd / \lnorm{\A\xgrtr}$ against the relative reconstruction error $\lnorm{\xgrtr - \Rec(\A\xgrtr + \Noise)} / \lnorm{\xgrtr}$.
\item
	\emph{Individual reconstruction results} are shown for different relative noise levels and a randomly selected signal from the test set.
\end{listing}
In both cases, the perturbation vector $\Noise$ is either of \emph{statistical} or \emph{adversarial} type.
The former means that $\Noise$ is a random vector such that $\mean[\lnorm{\Noise}^2] = \noisebnd^2$, whereas the latter is found by \eqref{eq:methods:findadv}.
While noise-to-error curves are of quantitative nature, individual reconstructions facilitate a qualitative judgment of robustness.
Note that the sensitivity to noise is different in each considered scenario.
Therefore, we have selected the maximal level of adversarial noise such that the benchmark of TV minimization does not yield a (subjectively) acceptable performance anymore.
A specification of all empirically selected hyper-parameters can be found in the supplementary material (see Table~\ref{tab:networks:architectures}--\ref{tab:networks:attack}).


\subsection{Case Study A: Compressed Sensing With Gaussian Measurements}
\label{sec:results:A}

Our first study is devoted to sparse recovery of 1D signals from Gaussian measurements. This means that the entries of the forward operator $\A$ in \eqref{eq:intro:problem} are independent Gaussian random variables with zero mean and variance $1/m$.
Although a toy problem, this setup is a folkloric benchmark in the field of compressed sensing (CS) theory \cite{fh13}.

We consider two different scenarios based on (approximately) gradient-sparse signals; note that such a model is canonical for TV minimization and compatible with the local connectivity of our convolutional NN schemes.

	\vspace{3pt}
	\hypertarget{sec:results:pwc}{}
	\newcommand{\refAone}{\protect\hyperlink{sec:results:pwc}{A1}}
	\textsf{\textbf{Scenario~A1:}} We draw $\xgrtr$ from a synthetic distribution of \emph{piecewise constant signals} with zero boundaries and well-controlled random jumps, see Fig.~\ref{fig:tvsynth:example_adv} for an example.
	In this scenario, we choose $m = 100$, $N = 256$, and use $M = 200\text{k}$ training samples.

	\vspace{3pt}
	\hypertarget{sec:results:mnist}{}
	\newcommand{\refAtwo}{\protect\hyperlink{sec:results:mnist}{A2}}
	\textsf{\textbf{Scenario~A2:}} We sample $\xgrtr \in [0, 1]^{28 \times 28}$ from the widely used \emph{MNIST database} \cite{lbbh98} with $M = 60\text{k}$ training images of handwritten digits.
	In the context of \eqref{eq:intro:problem}, the images are treated as 1D signals of dimension $N = 28^2 = 784$.
	For visual purposes, all reconstructions are displayed as images, see Fig.~\ref{fig:mnist:example_adv}.
	The number of Gaussian measurements is $m = 300$.
	\vspace{3pt}

In both scenarios, we chose the model-based, linear inversion layer of the networks as a generalized Tikhonov matrix, i.e., $\inv\A = (\A^T \A + \alpha \cdot \grad^T \grad)^{-1} \A^T \in \R^{N \times m}$ with the empirically chosen regularization parameter $\alpha=0.02$. We were not able to train the NNs to a comparable reconstruction accuracy with other natural choices, such as $\inv\A = \A^T$. The above matrix is also used to initialize the inversion layer $\Lin \in \R^{N\times m}$ of the fully-learned schemes.

\begin{figure}
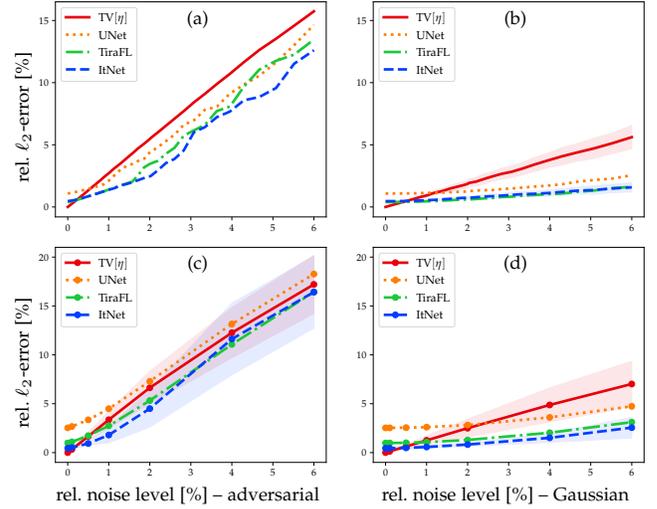

	\centering
	\scriptsize
	\setlength\tabcolsep{0pt}
	\begin{tabular}{cc@{\,\,}c}
		\rotatebox[origin=c]{90}{rel.~$\l{2}$-error [\%]} &
		\graphicwithlabel{tvsynth/results/attacks/fig_example_S6_adv_curve.pdf}{.22\textwidth}{(a)}{.9}{.55} &
		\graphicwithlabel{tvsynth/results/attacks/fig_example_S6_gauss_curve.pdf}{.22\textwidth}{(b)}{.9}{.55} \\
		\rotatebox[origin=c]{90}{rel.~$\l{2}$-error [\%]} &
		\graphicwithlabel{tvsynth/results/attacks/fig_table_adv.pdf}{.22\textwidth}{(c)}{.9}{.55} &
		\graphicwithlabel{tvsynth/results/attacks/fig_table_gauss.pdf}{.22\textwidth}{(d)}{.9}{.55} \\
		& rel.~noise level [\%] -- adversarial & rel.~noise level [\%] -- Gaussian
	\end{tabular}
	\caption{\textbf{Scenario~\refAone{} -- CS with 1D signals.} (a)~shows the adversarial noise-to-error curve for the randomly selected signal of Fig.~\ref{fig:tvsynth:example_adv}. (b)~shows the corresponding Gaussian noise-to-error curve, where the mean and standard deviation are computed over 200~draws of $\Noise$. (c)~and (d)~display the respective curves averaged over 50~signals from the test set. For the sake of clarity, we have omitted the standard deviations for $\UNet$ and $\TiraFL$, which behave similarly.}
	\label{fig:tvsynth:table}
\end{figure}

\begin{figure*}[b]
	\centering
	\scriptsize
	\begin{tabular}{l@{\,}c@{\,}c@{\,}c@{\,}c}
		& noiseless & 0.5\% rel.~noise -- adversarial & 2\% rel.~noise -- adversarial & 6\% rel.~noise -- adversarial \\
		\rotatebox[origin=c]{90}{$\TV[\noisebnd]$} &
		\includegraphics[valign=c,width=0.24\textwidth]{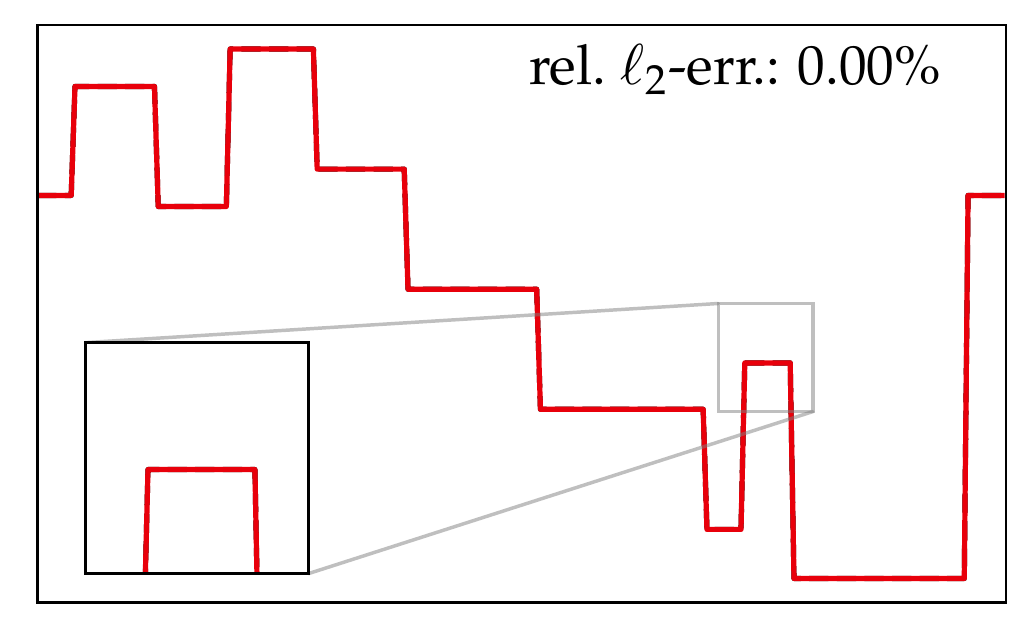} &
		\includegraphics[valign=c,width=0.24\textwidth]{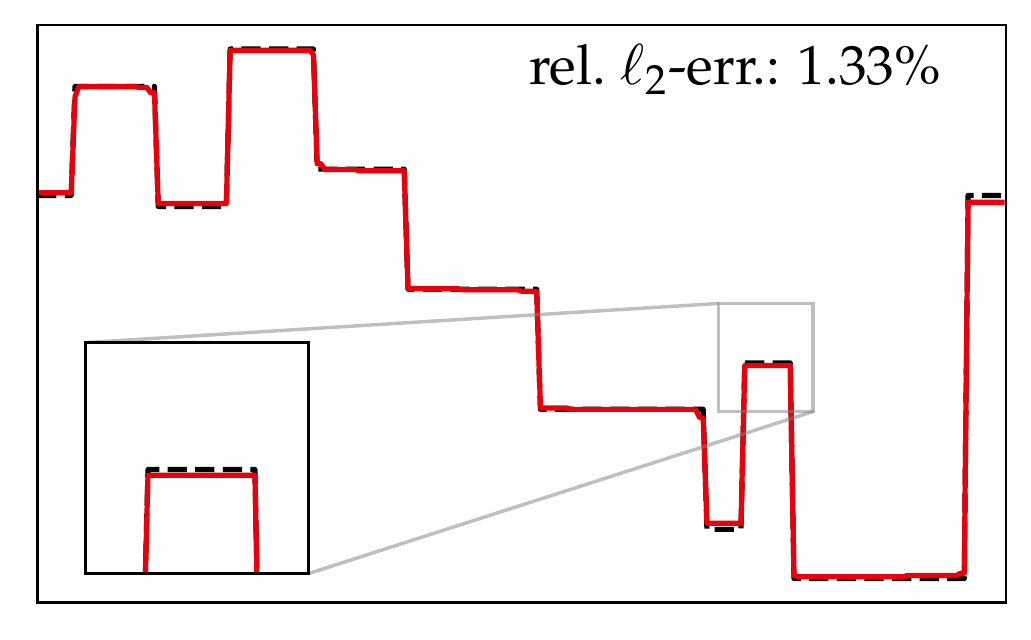} & \includegraphics[valign=c,width=0.24\textwidth]{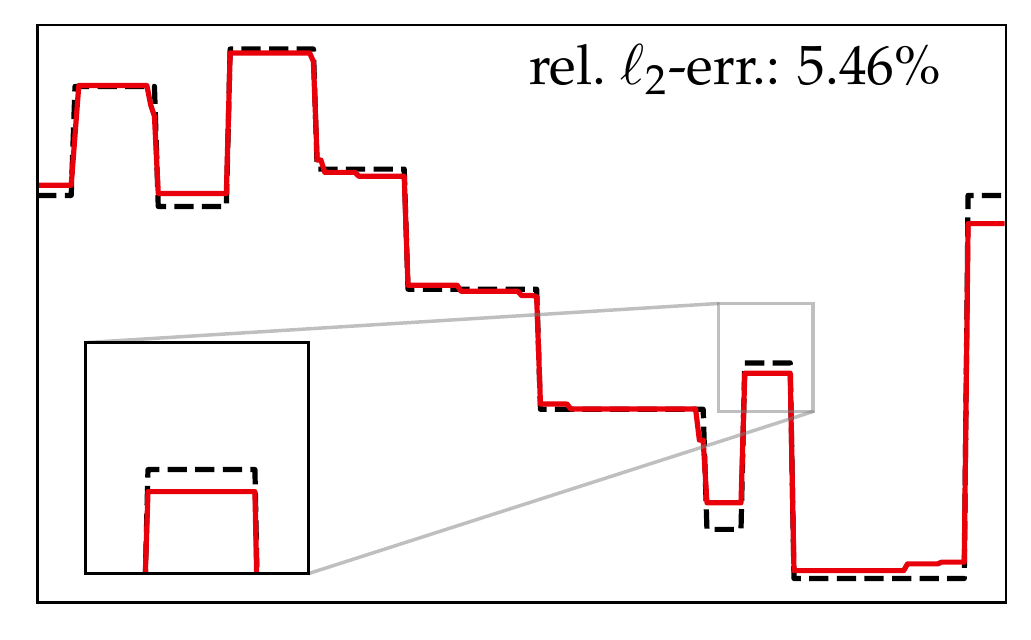} & \includegraphics[valign=c,width=0.24\textwidth]{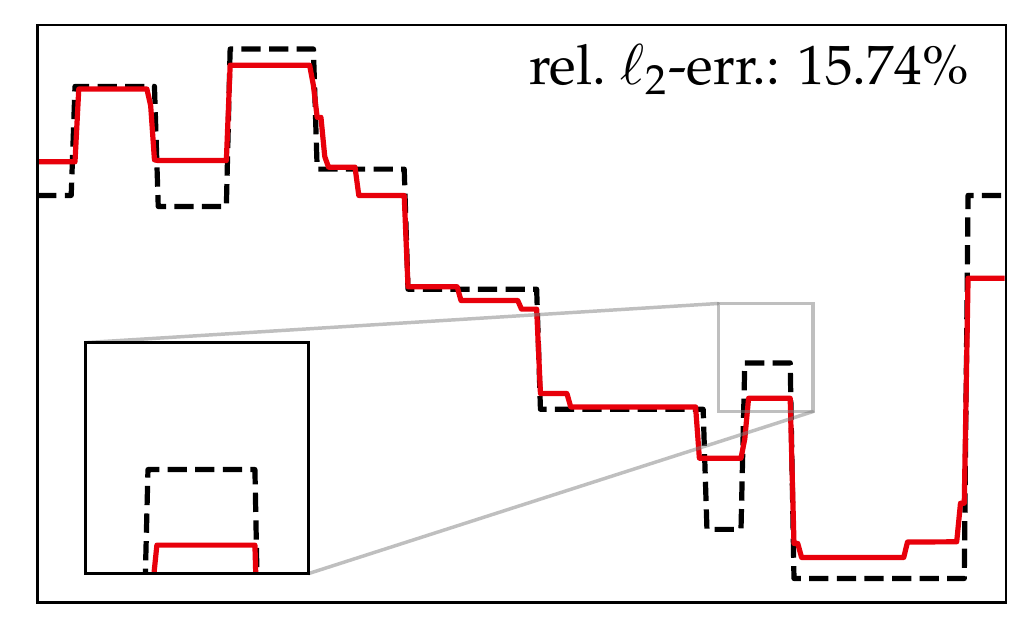} \\
		\rotatebox[origin=c]{90}{$\UNet$} &
		\includegraphics[valign=c,width=0.24\textwidth]{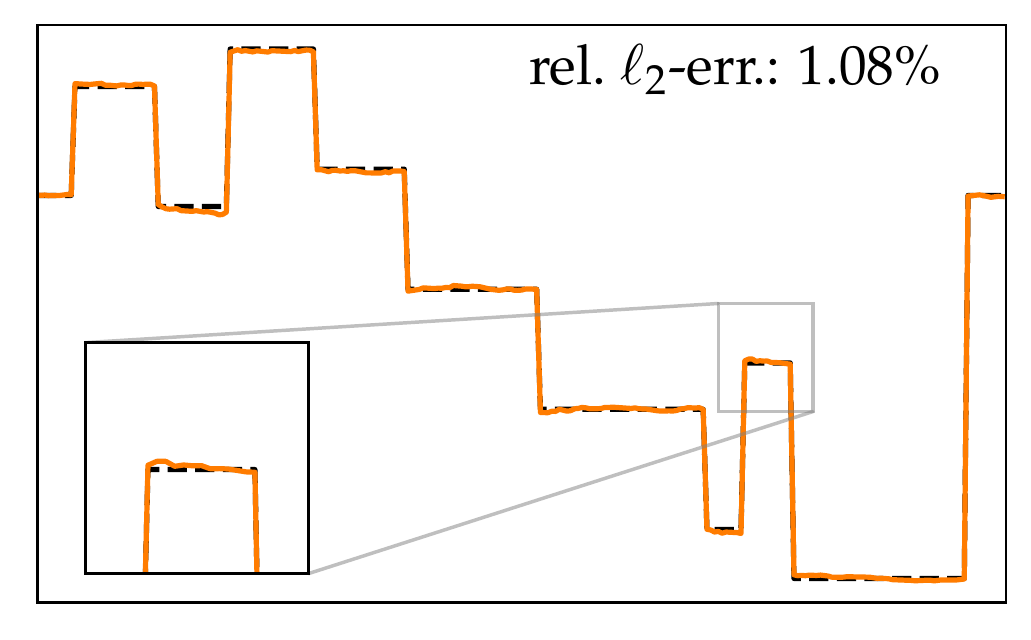} &
		\includegraphics[valign=c,width=0.24\textwidth]{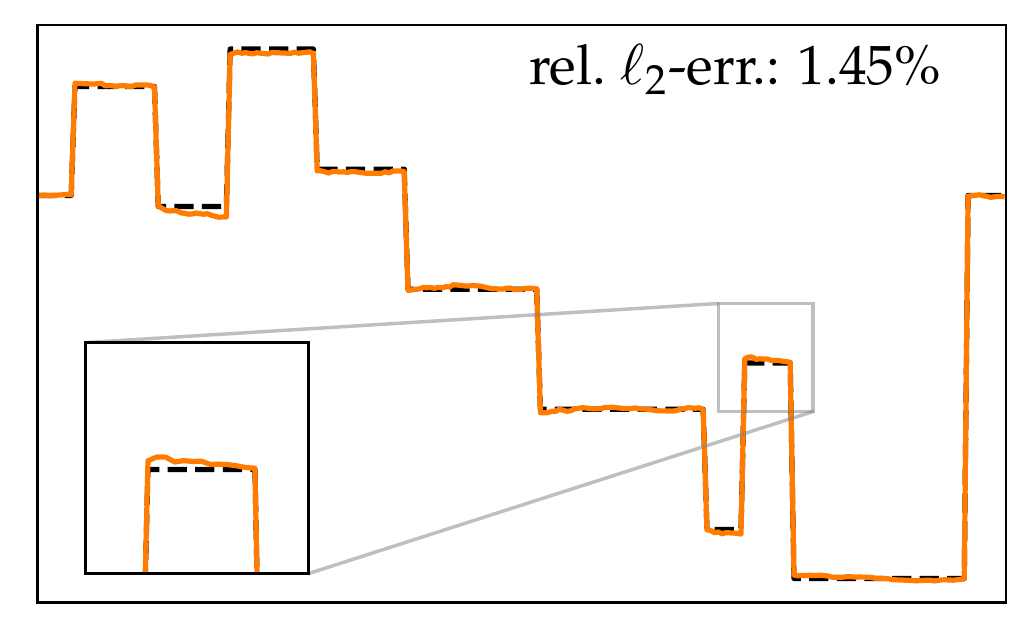} & \includegraphics[valign=c,width=0.24\textwidth]{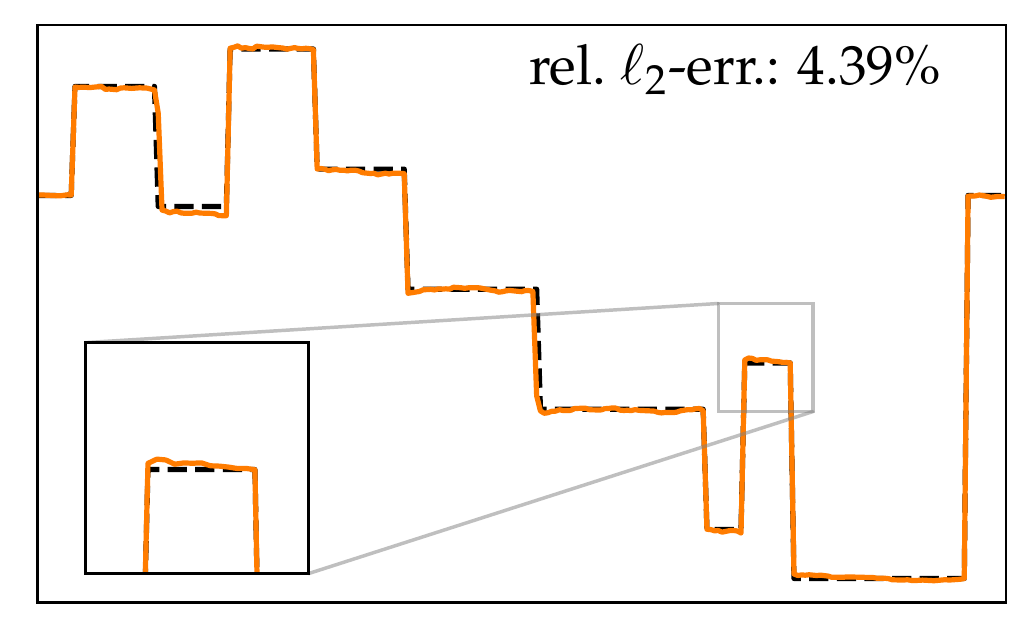} & \includegraphics[valign=c,width=0.24\textwidth]{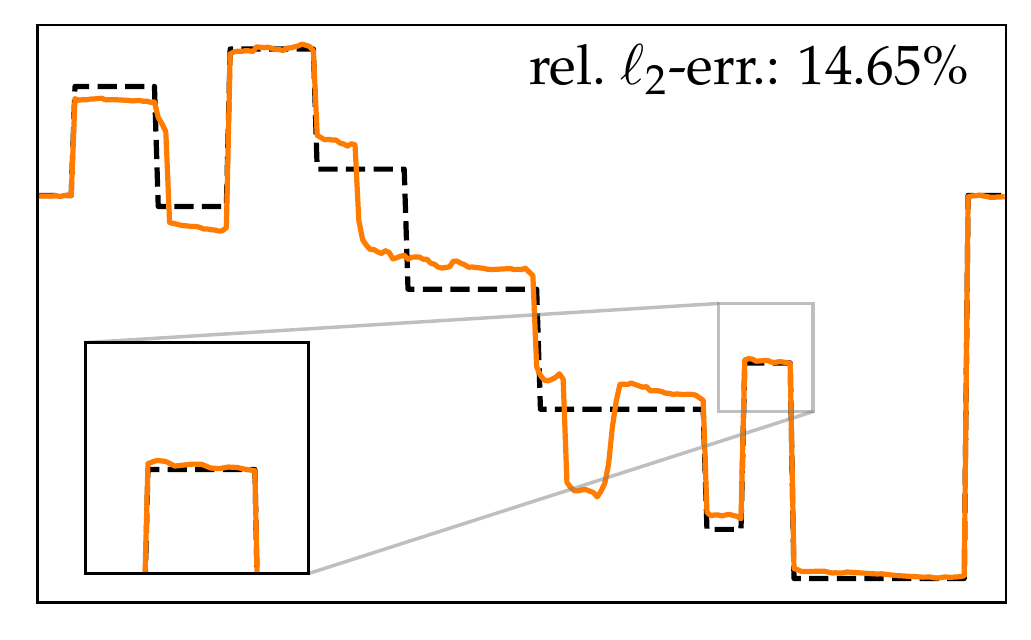} \\
		\rotatebox[origin=c]{90}{$\TiraFL$} &
		\includegraphics[valign=c,width=0.24\textwidth]{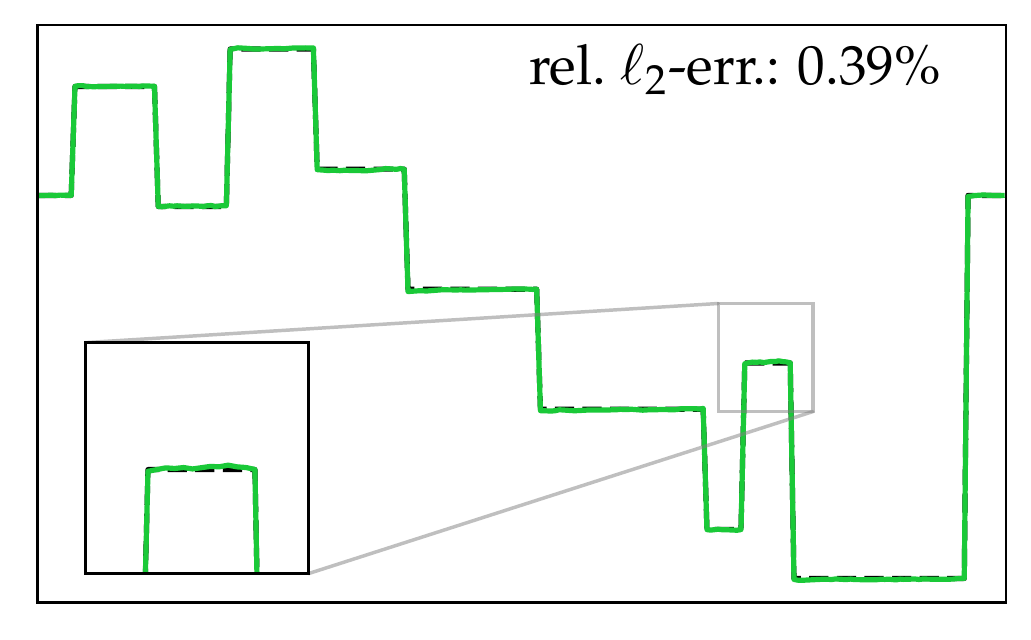} &
		\includegraphics[valign=c,width=0.24\textwidth]{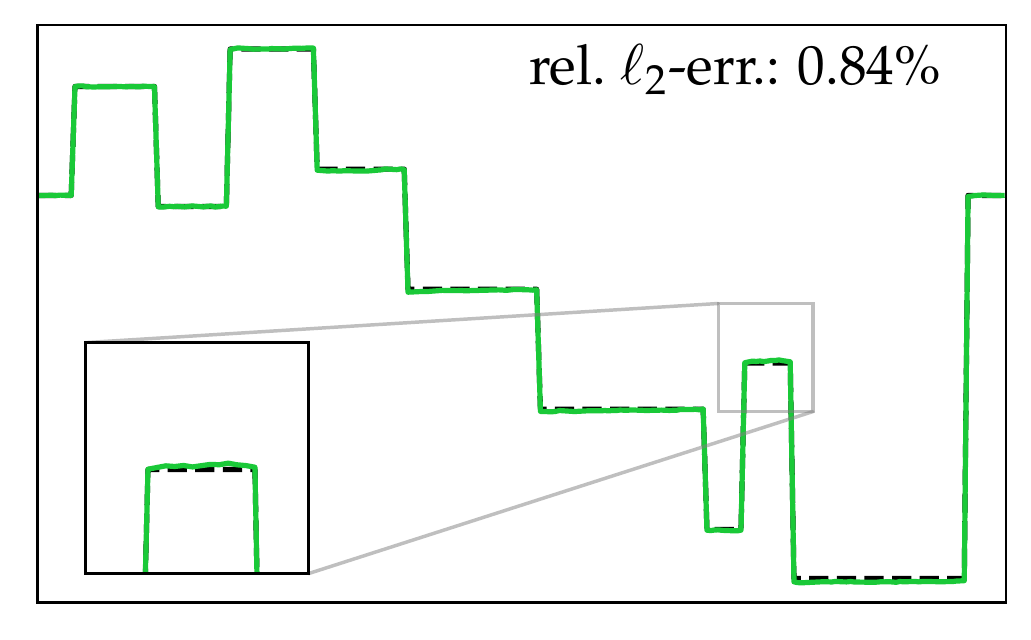} & \includegraphics[valign=c,width=0.24\textwidth]{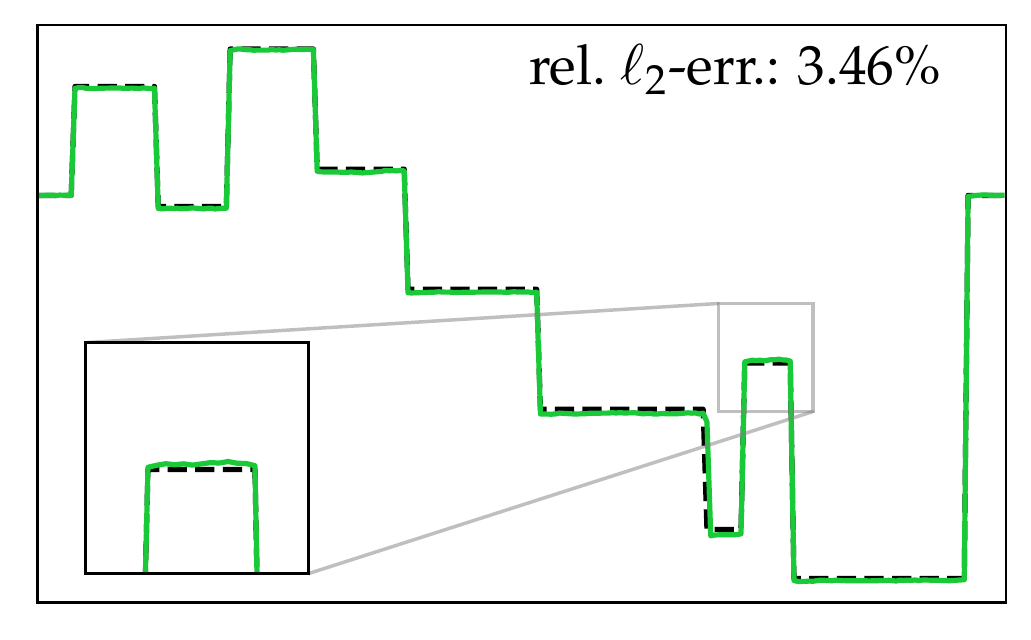} & \includegraphics[valign=c,width=0.24\textwidth]{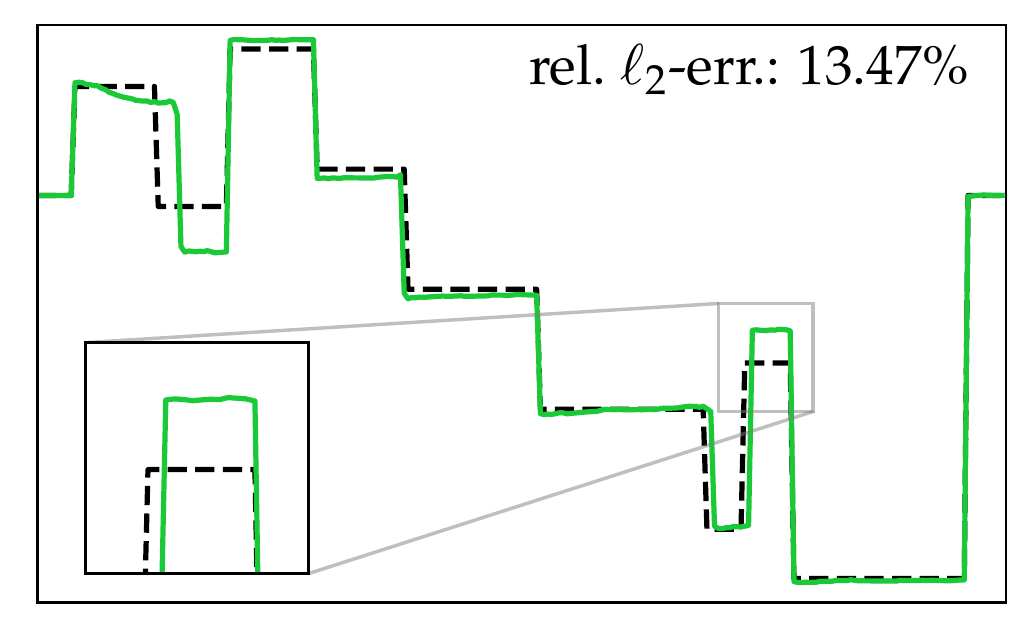} \\
		\rotatebox[origin=c]{90}{$\ItNet$} &
		\includegraphics[valign=c,width=0.24\textwidth]{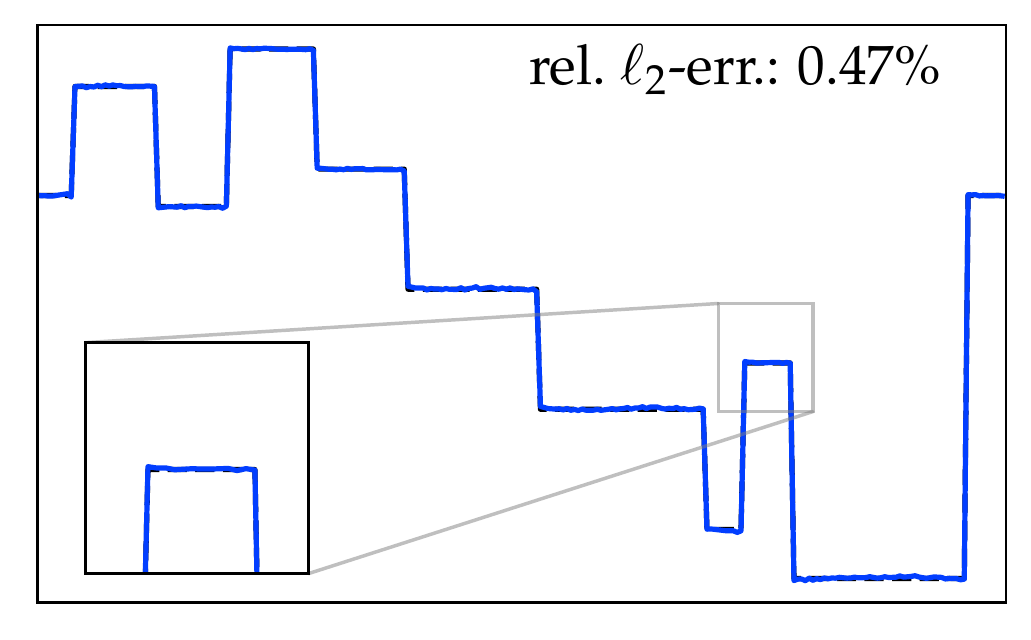} &
		\includegraphics[valign=c,width=0.24\textwidth]{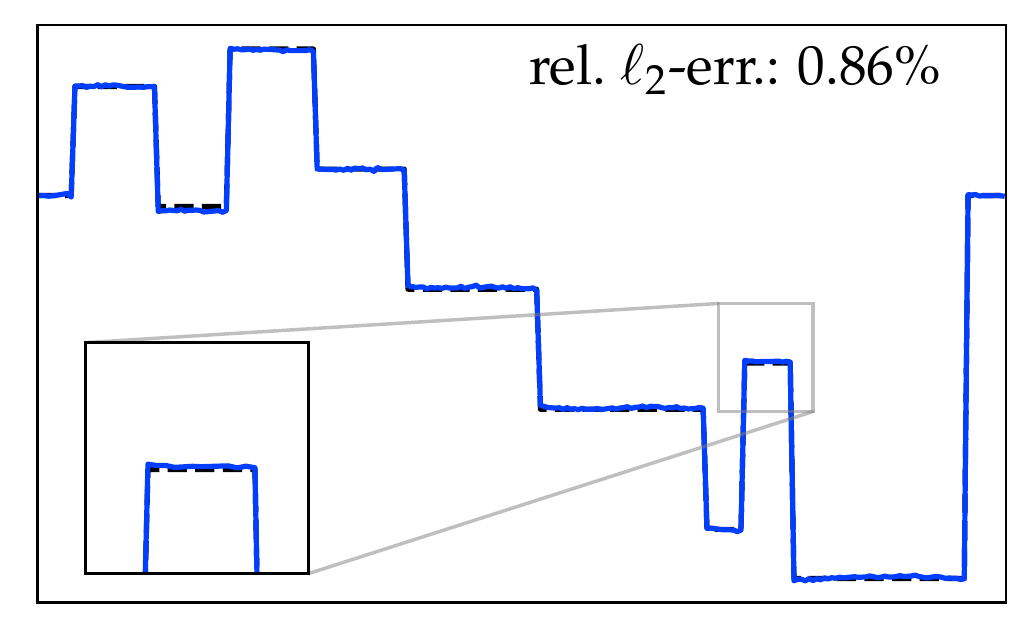} & \includegraphics[valign=c,width=0.24\textwidth]{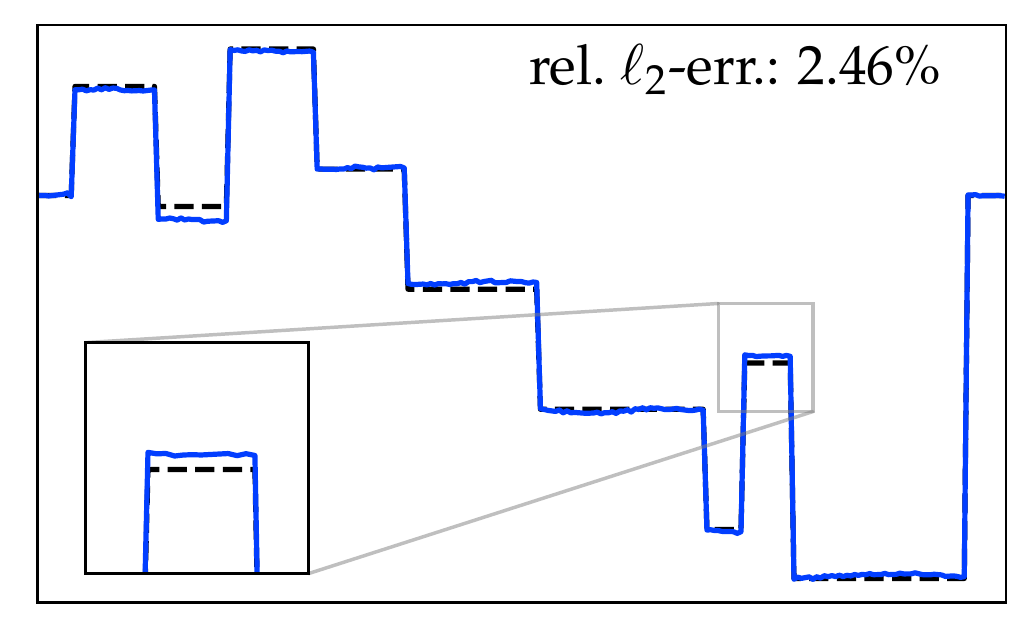} & \includegraphics[valign=c,width=0.24\textwidth]{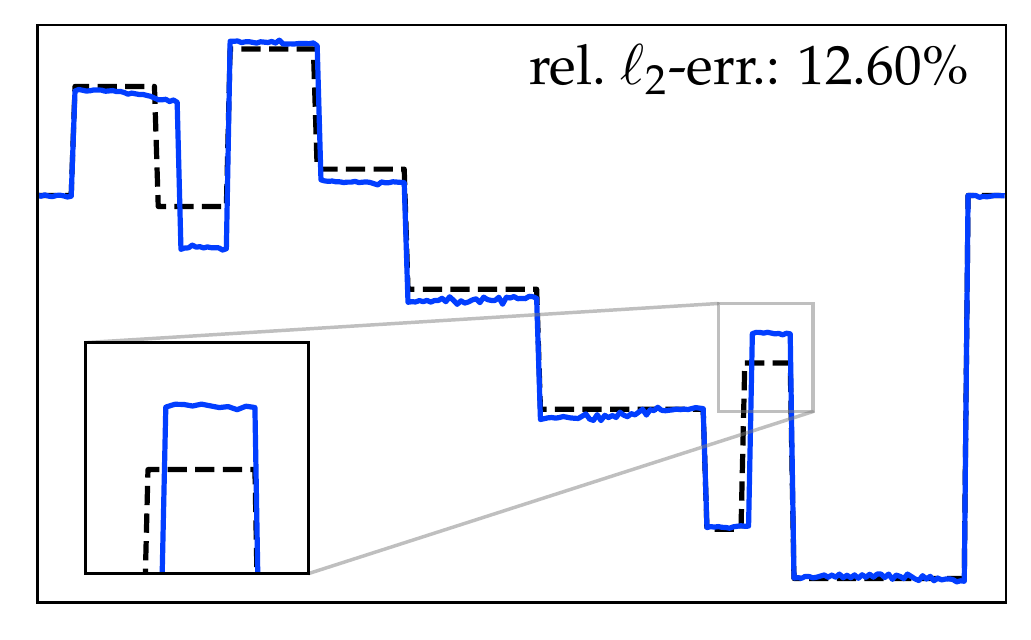}
	\end{tabular}
	\caption{\textbf{Scenario~\refAone{} -- CS with 1D signals.} Individual reconstructions of a randomly selected signal from the test set for different levels of adversarial noise. The ground truth signal is visualized by a dashed line.}
	\label{fig:tvsynth:example_adv}
\end{figure*}

\begin{figure}
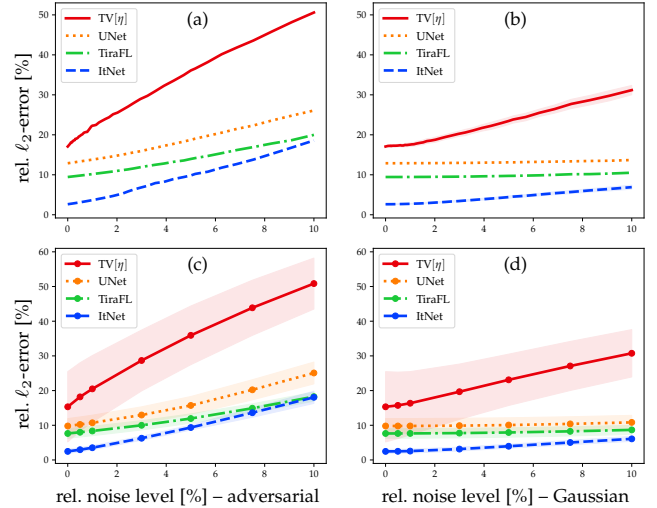

	\centering
	\scriptsize
	\setlength\tabcolsep{0pt}
	\begin{tabular}{cc@{\,\,}c}
		\rotatebox[origin=c]{90}{rel.~$\l{2}$-error [\%]} &
		\graphicwithlabel{mnist/results/attacks/fig_example_S0_adv_curve.pdf}{.22\textwidth}{(a)}{.9}{.55} &
		\graphicwithlabel{mnist/results/attacks/fig_example_S0_gauss_curve.pdf}{.22\textwidth}{(b)}{.9}{.55} \\
		\rotatebox[origin=c]{90}{rel.~$\l{2}$-error [\%]} &
		\graphicwithlabel{mnist/results/attacks/fig_table_adv.pdf}{.22\textwidth}{(c)}{.9}{.55} &
		\graphicwithlabel{mnist/results/attacks/fig_table_gauss.pdf}{.22\textwidth}{(d)}{.9}{.55} \\
		& rel.~noise level [\%] -- adversarial & rel.~noise level [\%] -- Gaussian
	\end{tabular}
	\caption{\textbf{Scenario~\refAtwo{} -- CS with MNIST.} (a)~shows the adversarial noise-to-error curve for the randomly selected digit \texttt{3} of Fig.~\ref{fig:mnist:example_adv}. (b)~shows the corresponding Gaussian noise-to-error curve, where the mean and standard deviation are computed over 200~draws of $\Noise$. (c)~and (d)~display the respective curves averaged over 50~signals from the test set.}
	\label{fig:mnist:table}
\end{figure}

Fig.~\ref{fig:tvsynth:table} shows the noise-to-error curves for \emph{Scenario~\refAone{} (CS with 1D signals)}; see also Table~\ref{tab:tvsynth:table_adv} and~\ref{tab:tvsynth:table_ref}.
The associated individual reconstructions for adversarial noise are displayed in Fig.~\ref{fig:tvsynth:example_adv}; see Fig.~\ref{fig:tvsynth:example_gauss} for the corresponding results with Gaussian noise.
Fig.~\ref{fig:tvsynth:table_2} supplements the simulation of Fig.~\ref{fig:tvsynth:table}(b) and~(d) by two additional types of random noise, drawn from the uniform and Bernoulli distribution. Both exhibit results that are virtually indistinguishable from the Gaussian case.
Fig.~\ref{fig:mnist:table} shows the noise-to-error curves for \emph{Scenario~\refAtwo{} (CS with MNIST)}; see also Table~\ref{tab:mnist:table_adv} and~\ref{tab:mnist:table_ref}.
The associated individual reconstructions for adversarial noise are displayed in Fig.~\ref{fig:mnist:example_adv}; see Fig.~\ref{fig:mnist:example_adv_supp} for two additional digits and Fig.~\ref{fig:mnist:example_gauss} for the corresponding results with Gaussian noise.

\textsf{\textbf{Conclusions:}} The above results confirm that the considered NN-based schemes are as least as robust to adversarial perturbations as the benchmark of TV minimization. Although $\TV[\noisebnd]$ is perfectly tuned to each noise level $\noisebnd$, it is clearly outperformed in the case of statistical noise.  The gap between statistical and adversarial perturbations is comparable for all methods.

TV minimization is a perfect match for Scenario~\refAone{}. In particular, exact recovery from noiseless measurements is guaranteed by CS theory \cite{almt14,gms20}.
Although this cannot be expected for NN-based solvers, they still come with an overall superior robustness against noise.
The situation is even more striking in Scenario~\refAtwo{}.
Here, TV minimization performs worse, since the signals are only approximately gradient-sparse.
In contrast, the NN-based reconstruction schemes adapt well to the simple MNIST database, leading to significantly better outcomes in every regard.
Hence, the increase in accuracy by learned methods does not necessarily imply a loss of robustness. 

The performance ranking of the considered deep NNs is as one might expect: First, data consistency as encouraged by the $\ItNet$-architecture is beneficial. Furthermore, Table~\ref{tab:tvsynth:table_adv}--\ref{tab:mnist:table_ref} reveal that the Tiramisu architecture is superior to a simple U-Net, and that a learnable inversion layer improves the recovery. The latter observation is not surprising, since Thikonov regularization is known to work poorly in conjunction with subsampled Gaussian measurements.

\begin{figure}
	\centering
	\begin{tabular}{c}
		\scriptsize
		\setlength\tabcolsep{1pt}
		\begin{tabular}{lccc}
			& 2\% rel.~noise -- adv. & 5\% rel.~noise -- adv. & 10\% rel.~noise -- adv. \\
			\rotatebox[origin=c]{90}{$\TV[\noisebnd]$} &
			\includegraphics[valign=c,width=0.14\textwidth]{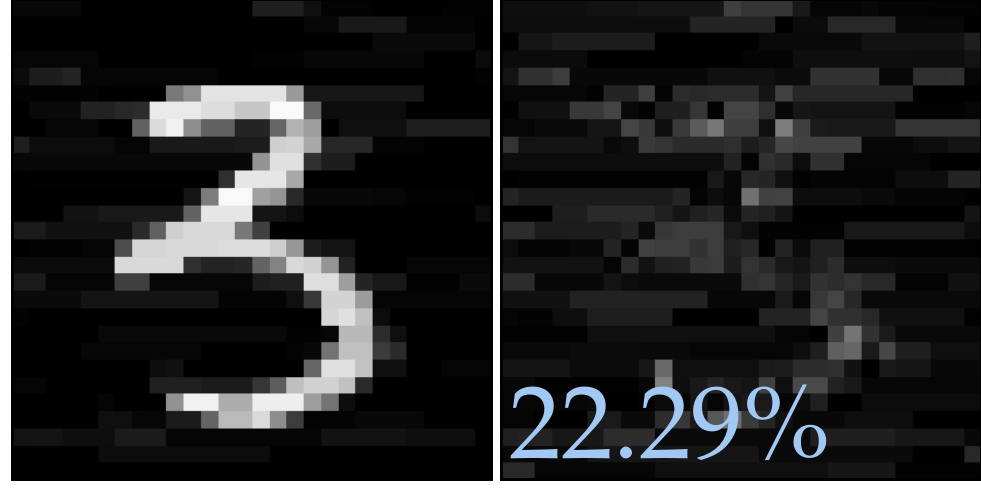} & \includegraphics[valign=c,width=0.14\textwidth]{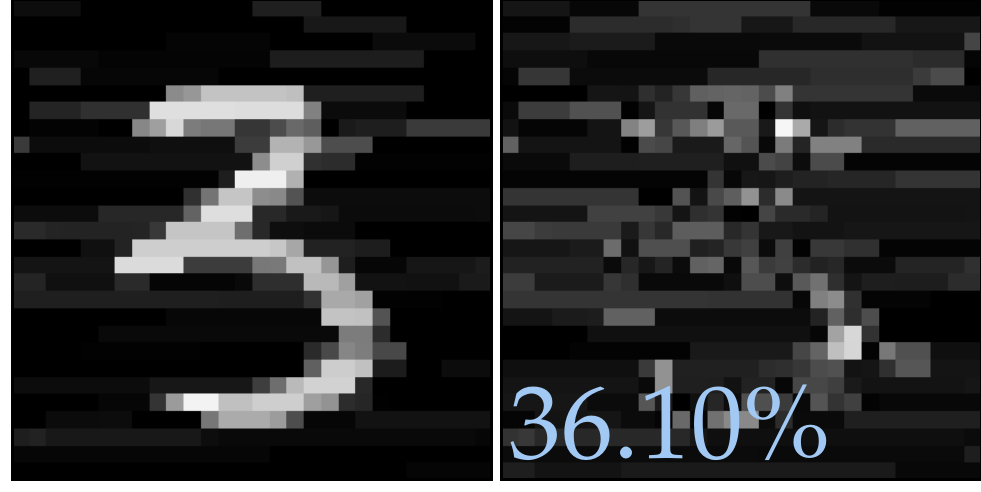} & \includegraphics[valign=c,width=0.14\textwidth]{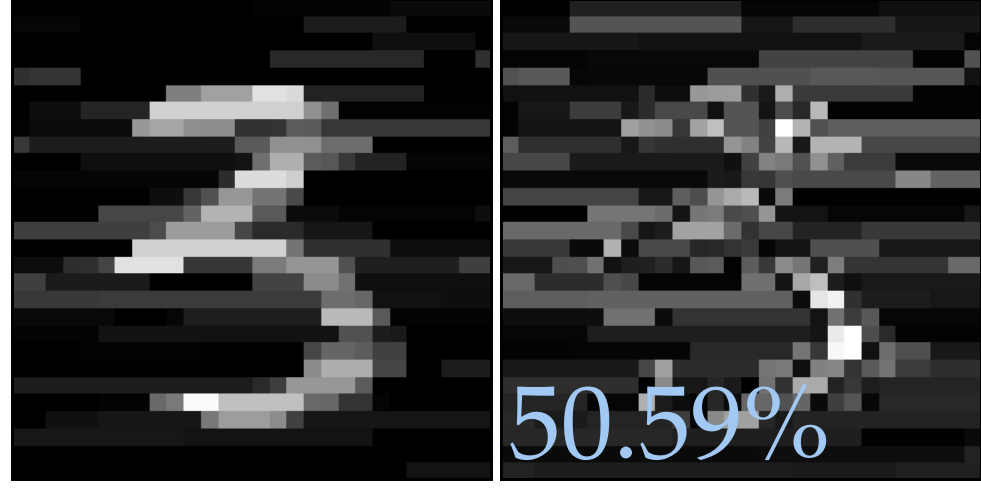} \\
			\rotatebox[origin=c]{90}{$\UNet$} &
			\includegraphics[valign=c,width=0.14\textwidth]{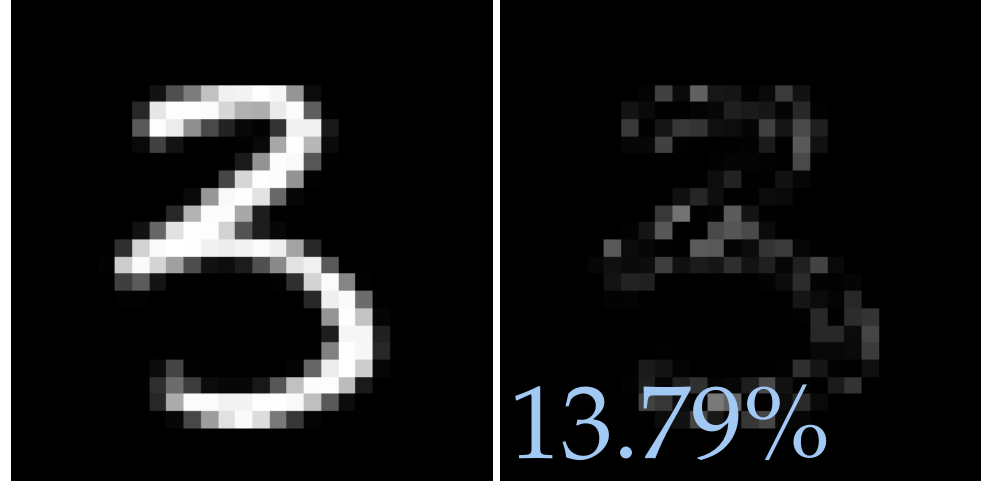} & \includegraphics[valign=c,width=0.14\textwidth]{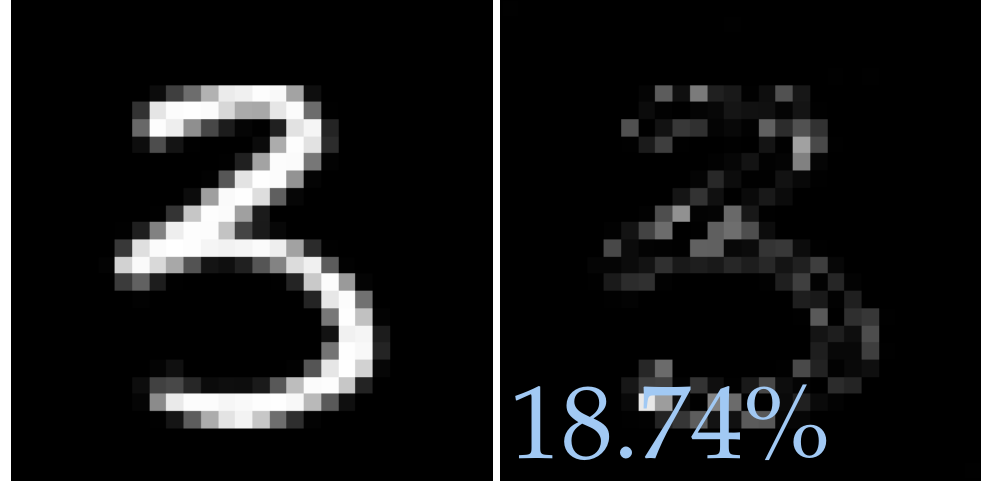} & \includegraphics[valign=c,width=0.14\textwidth]{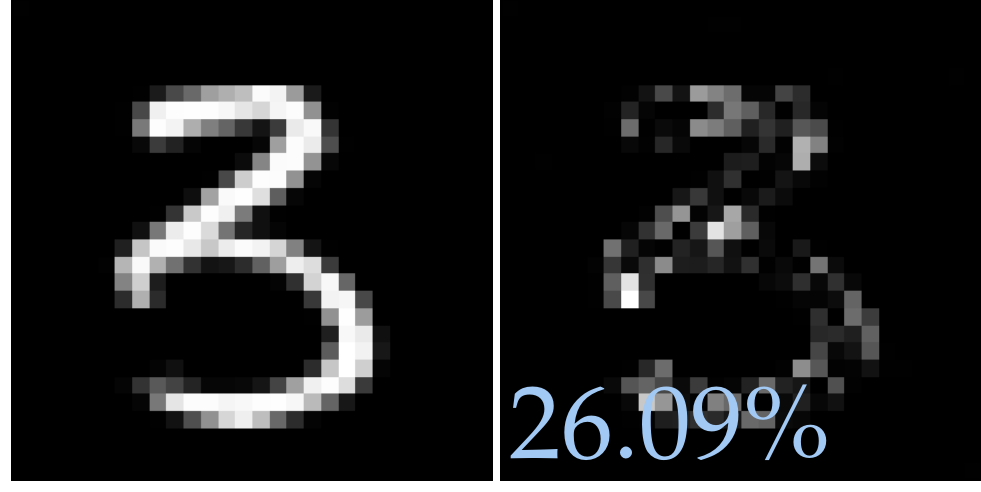} \\
			\rotatebox[origin=c]{90}{$\TiraFL$} &
			\includegraphics[valign=c,width=0.14\textwidth]{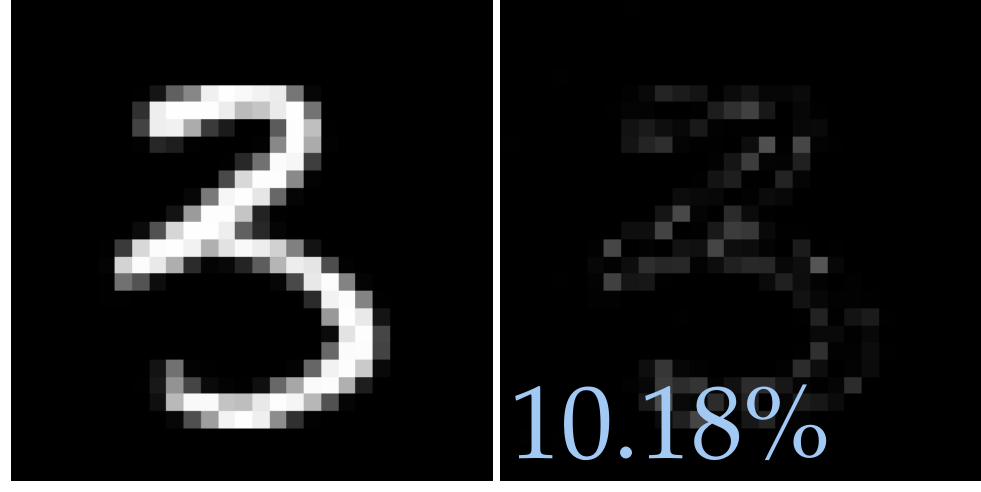} & \includegraphics[valign=c,width=0.14\textwidth]{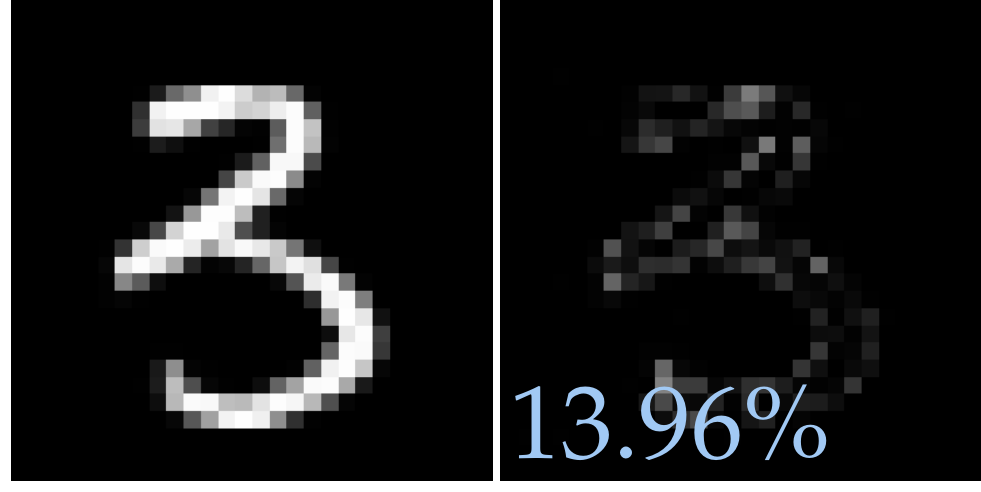} & \includegraphics[valign=c,width=0.14\textwidth]{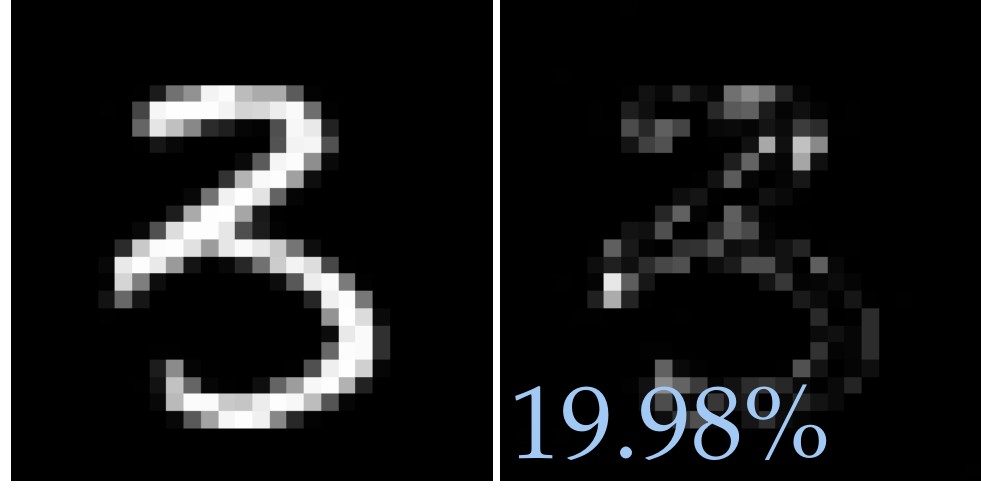} \\
			\rotatebox[origin=c]{90}{$\ItNet$} &
			\includegraphics[valign=c,width=0.14\textwidth]{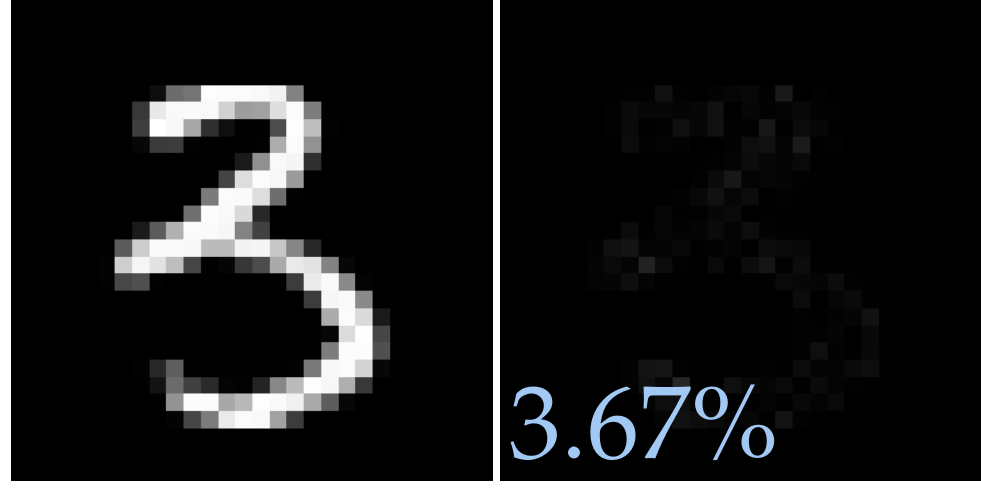} & \includegraphics[valign=c,width=0.14\textwidth]{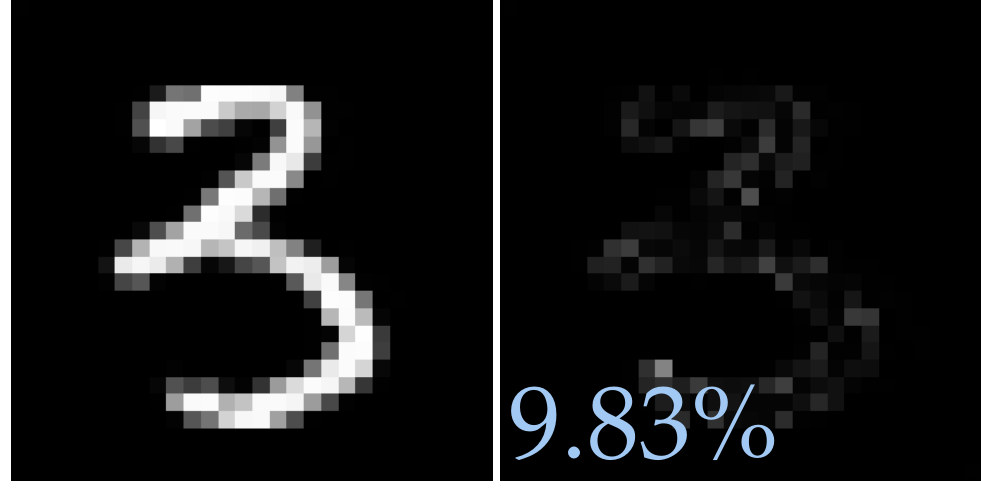} & \includegraphics[valign=c,width=0.14\textwidth]{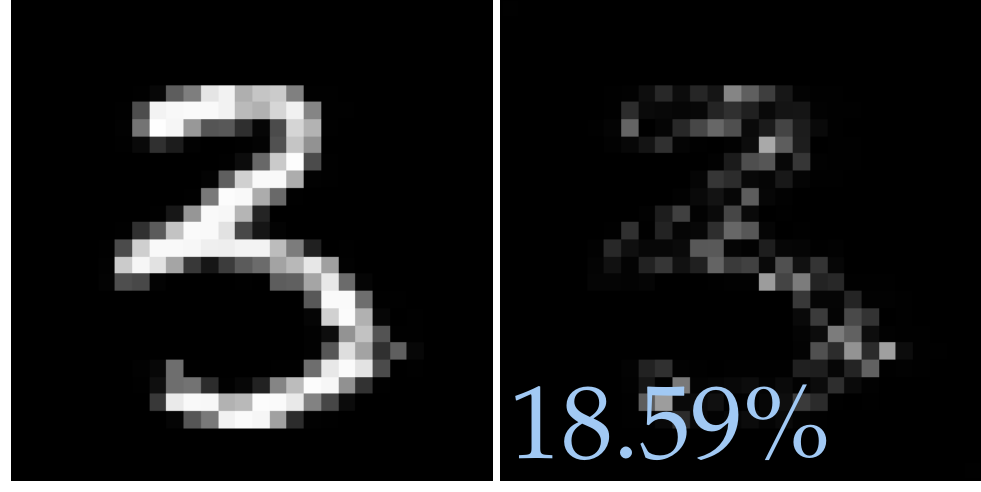} \\
		\end{tabular}
		\\
		\\
		\scriptsize
		\setlength\tabcolsep{1pt}
		\begin{tabular}{lccc}
			\rotatebox[origin=c]{90}{$\TV[\noisebnd]$} &
			\includegraphics[valign=c,width=0.14\textwidth]{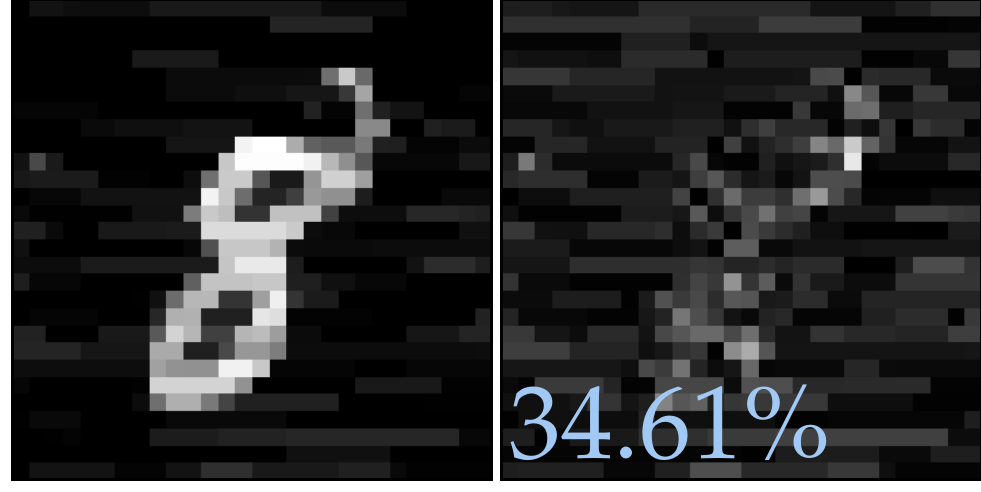} & \includegraphics[valign=c,width=0.14\textwidth]{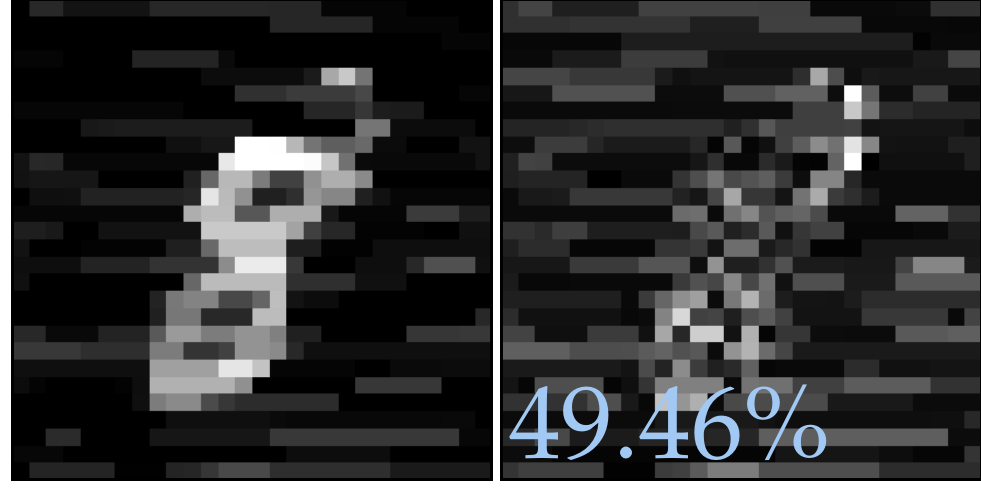} & \includegraphics[valign=c,width=0.14\textwidth]{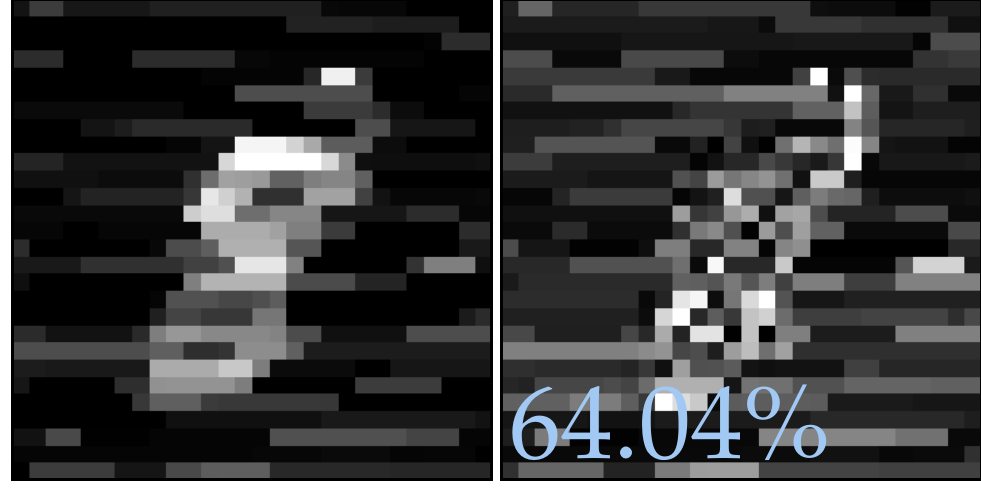} \\
			\rotatebox[origin=c]{90}{$\UNet$} &
			\includegraphics[valign=c,width=0.14\textwidth]{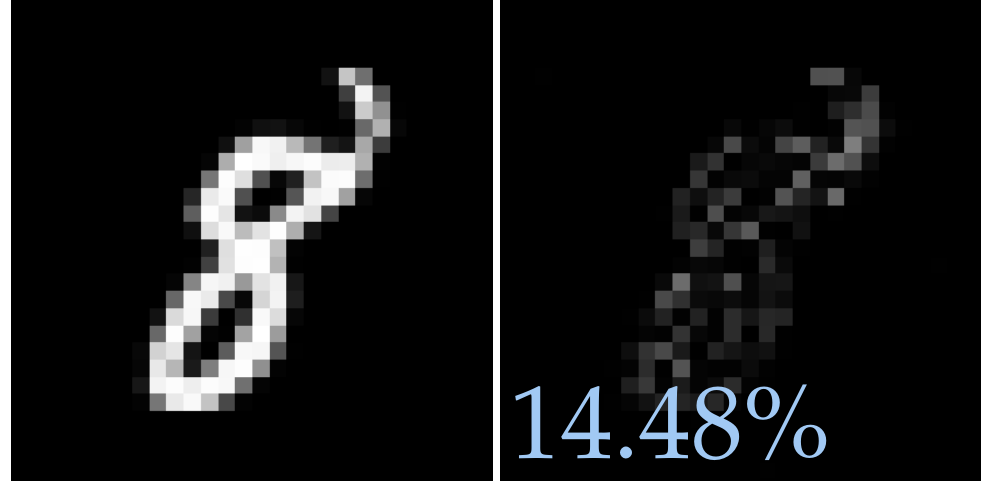} & \includegraphics[valign=c,width=0.14\textwidth]{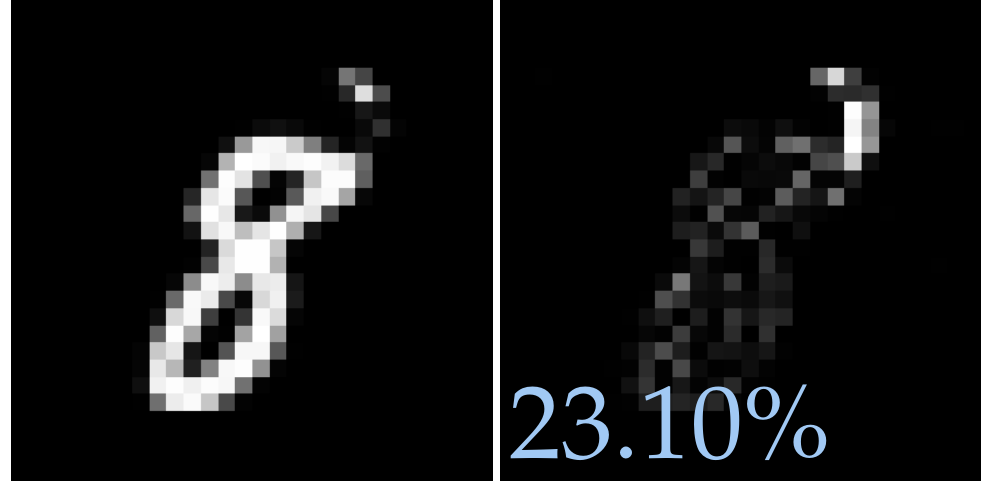} & \includegraphics[valign=c,width=0.14\textwidth]{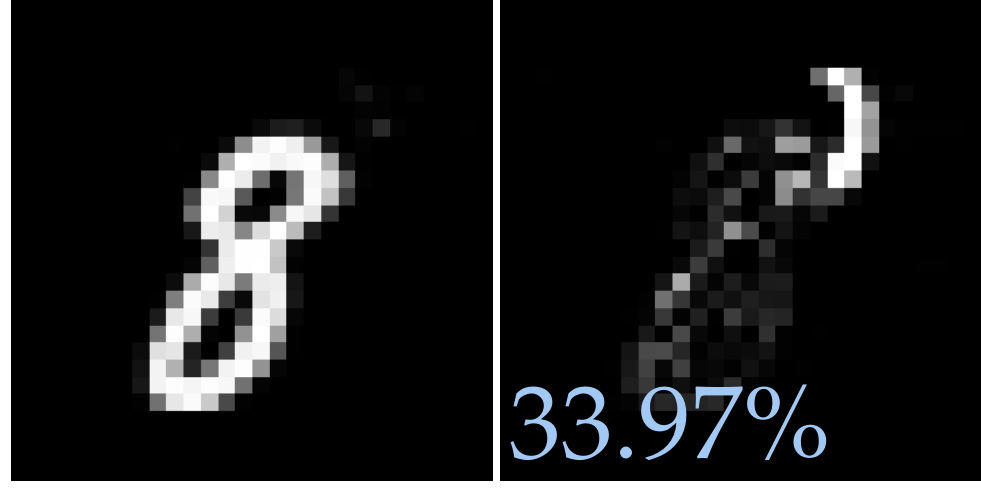} \\
			\rotatebox[origin=c]{90}{$\TiraFL$} &
			\includegraphics[valign=c,width=0.14\textwidth]{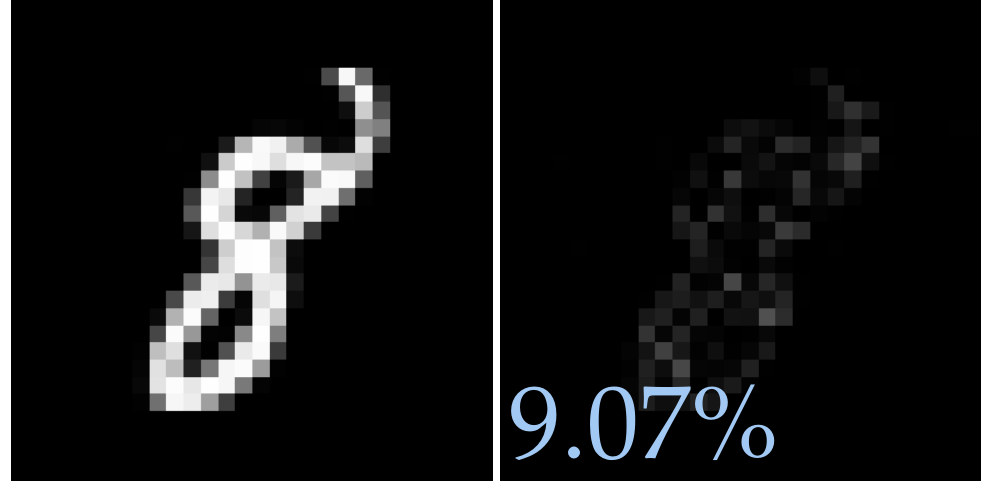} & \includegraphics[valign=c,width=0.14\textwidth]{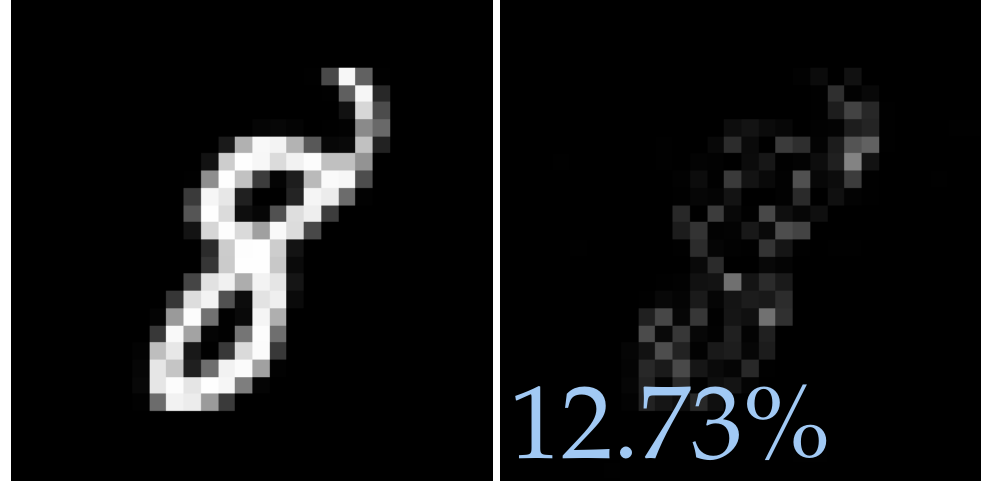} & \includegraphics[valign=c,width=0.14\textwidth]{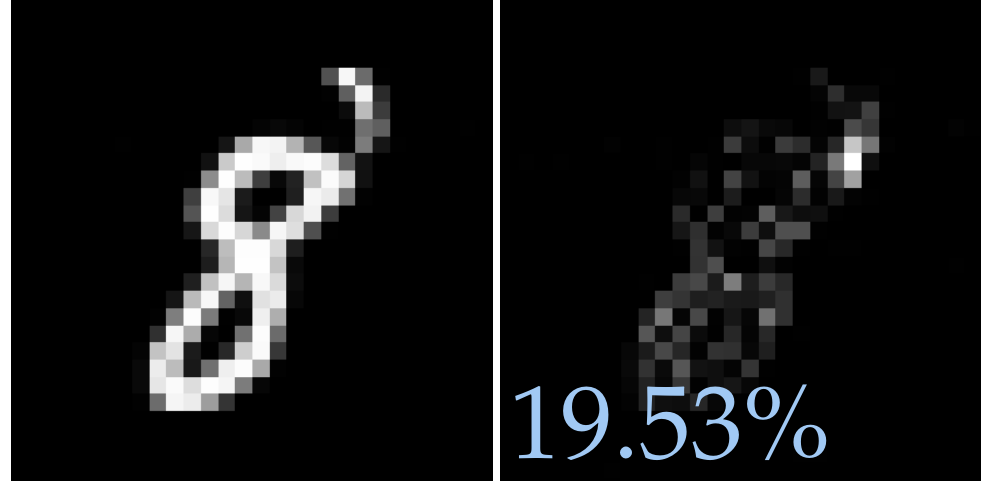} \\
			\rotatebox[origin=c]{90}{$\ItNet$} &
			\includegraphics[valign=c,width=0.14\textwidth]{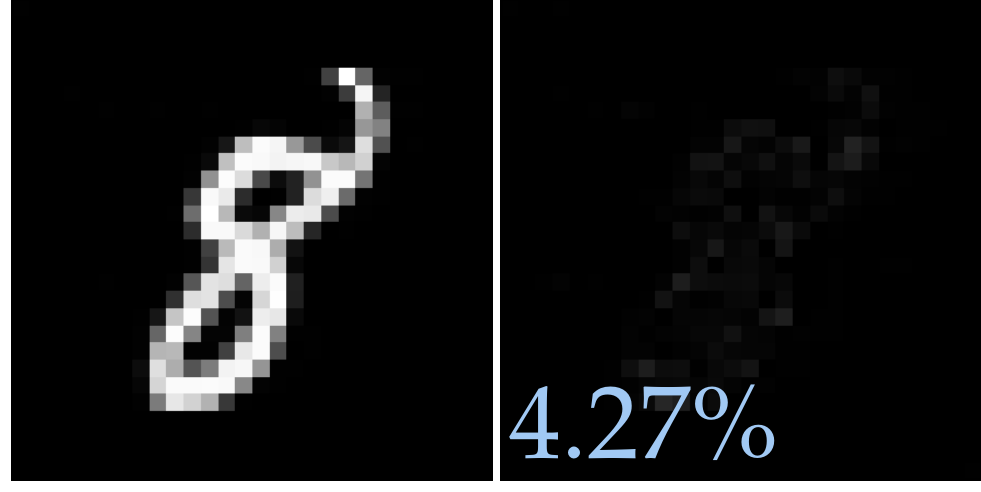} & \includegraphics[valign=c,width=0.14\textwidth]{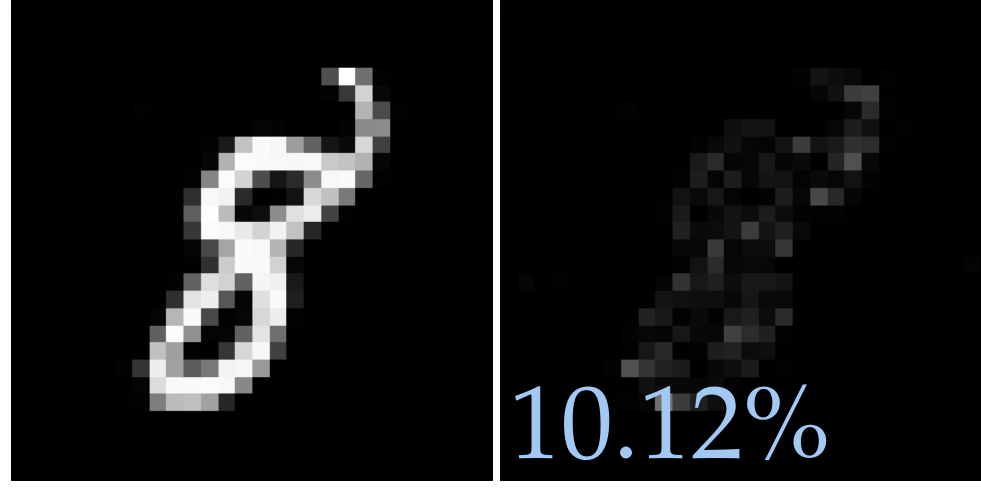} & \includegraphics[valign=c,width=0.14\textwidth]{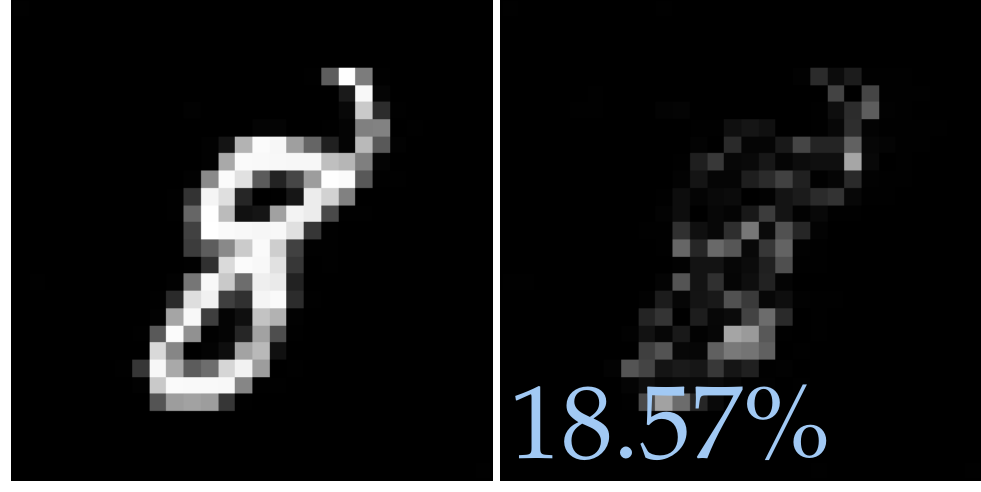} \\
		\end{tabular}
	\end{tabular}
	\caption{\textbf{Scenario~\refAtwo{} -- CS with MNIST.} Individual reconstructions of two randomly selected digits from the test set for different levels of adversarial noise. The reconstructed digits and their error plots (with relative $\l{2}$-error) are displayed in the windows $[0,1]$ and $[0, 0.6]$, respectively. The horizontal line artifacts in the $\TV[\noisebnd]$-solutions are due to the fact that the MNIST images are treated as vectorized 1D signals. Remarkably, although relying on 1D convolutional filters, the NN-based reconstructions do not suffer from these artifacts.}
	\label{fig:mnist:example_adv}
\end{figure}

\subsection{Case Study B: Image Recovery of Phantom Ellipses}
\label{sec:results:B}

Our second set of experiments concerns the recovery of phantom ellipses from Fourier or Radon measurements.
These tasks correspond to popular simulation studies for biomedical imaging, e.g., see \cite{jmfu17,ao17,ao18,bub+19}.
We sample $\xgrtr \in [0, 1]^{256 \times 256}$ from a distribution of superimposed random ellipses with mild linear intensity gradients and well-controlled geometric properties, see Fig.~\ref{fig:ellipses:example_adv} for an example.
The training is performed on $M = 25\text{k}$ images.
We consider the following two measurement scenarios for \eqref{eq:intro:problem}, associated with the problems of \emph{compressed sensing MRI} \cite{ldsp08} and \emph{low-dose computed tomography (CT)} \cite{sp08,jmfu17}, respectively:


	\vspace{3pt}
	\hypertarget{sec:results:fourier}{}
	\newcommand{\refBone}{\protect\hyperlink{sec:results:fourier}{B1}}
	\textsf{\textbf{Scenario~B1:}} The forward operator takes the form $\A = P \ftop \in \C^{m \times N}$, where $\ftop \in \C^{N \times N}$ is the \emph{2D discrete Fourier transform} and $P \in \{0,1\}^{m \times N}$ is a subsampling operator defined by a golden-angle radial mask with 40 lines ($m = 10941$ and $N = 256^2 = 65536$).
	Note that the entire data processing is complex-valued, while the actual reconstructions are computed as real-valued magnitude images, as common in MRI.
	We use the canonical inversion layer $\inv\A = \adj\A = \ftop^{-1} \adj{P} \in \C^{N \times m}$.

	\vspace{3pt}
	\hypertarget{sec:results:radon}{}
	\newcommand{\refBtwo}{\protect\hyperlink{sec:results:radon}{B2}}
	\textsf{\textbf{Scenario~B2:}} The forward operator $\A \in \R^{m \times N}$ is given by a sparse-angle \emph{Radon transform} with 60 views ($m = 21780$ and $N = 65536$).
	The non-linear inversion layer $\inv\A \colon \R^m \to\nobreak \R^N$ is chosen as the filtered back-projection algorithm (FBP) with a Hann filter.
	\vspace{3pt}
	
\begin{figure}
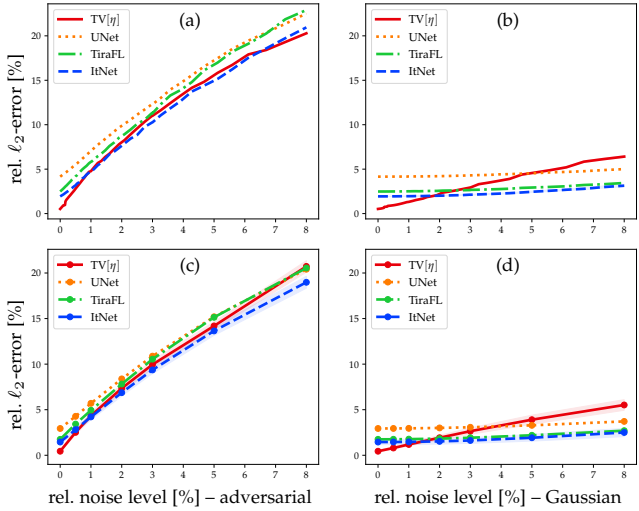

	\centering
	\scriptsize
	\setlength\tabcolsep{0pt}
	\begin{tabular}{cc@{\,\,}c}
		\rotatebox[origin=c]{90}{rel.~$\l{2}$-error [\%]} &
		\graphicwithlabel{ellipses/results/attacks/fig_example_S66_adv_curve.pdf}{.22\textwidth}{(a)}{.9}{.55} &
		\graphicwithlabel{ellipses/results/attacks/fig_example_S66_gauss_curve.pdf}{.22\textwidth}{(b)}{.9}{.55} \\
		\rotatebox[origin=c]{90}{rel.~$\l{2}$-error [\%]} &
		\graphicwithlabel{ellipses/results/attacks/fig_table_adv.pdf}{.22\textwidth}{(c)}{.9}{.55} &
		\graphicwithlabel{ellipses/results/attacks/fig_table_gauss.pdf}{.22\textwidth}{(d)}{.9}{.55} \\
		& rel.~noise level [\%] -- adversarial & rel.~noise level [\%] -- Gaussian
	\end{tabular}
	\caption{\textbf{Scenario~\refBone{} -- Fourier meas.~with ellipses.} (a)~shows the adversarial noise-to-error curve for the randomly selected image of Fig.~\ref{fig:ellipses:example_adv}. (b)~shows the corresponding Gaussian noise-to-error curve, where the mean and (almost imperceptible) standard deviation are computed over 50~draws of $\Noise$. (c)~and (d)~display the respective curves averaged over 50~images from the test set. For the sake of clarity, we have omitted the standard deviations for $\UNet$ and $\TiraFL$, which behave similarly.}
	\label{fig:ellipses:table}
\end{figure}

\begin{figure*}[p]
	\centering
	\scriptsize
	\begin{tabular}{l@{\,}c@{\,}c@{\,}c@{\,}c@{\enspace}c}
		& noiseless & 1\% rel.~noise -- adversarial & 3\% rel.~noise -- adversarial & 8\% rel.~noise -- adversarial \\
		\rotatebox[origin=c]{90}{$\TV[\noisebnd]$} &
		\includegraphics[valign=c,width=0.2\textwidth]{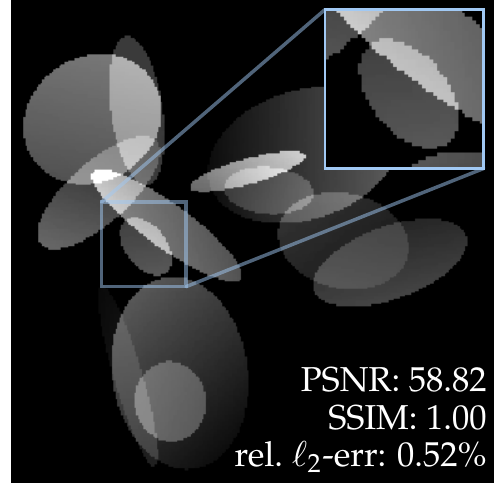} &
		\includegraphics[valign=c,width=0.2\textwidth]{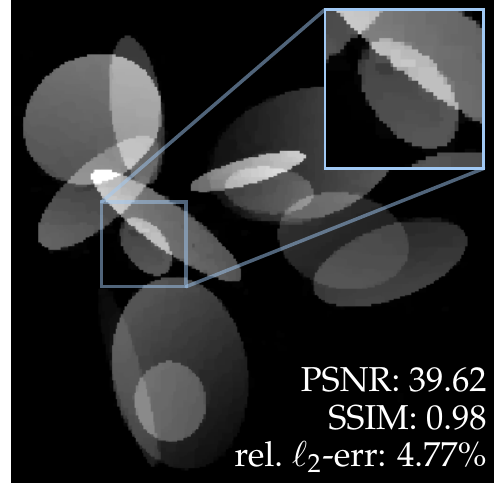} & \includegraphics[valign=c,width=0.2\textwidth]{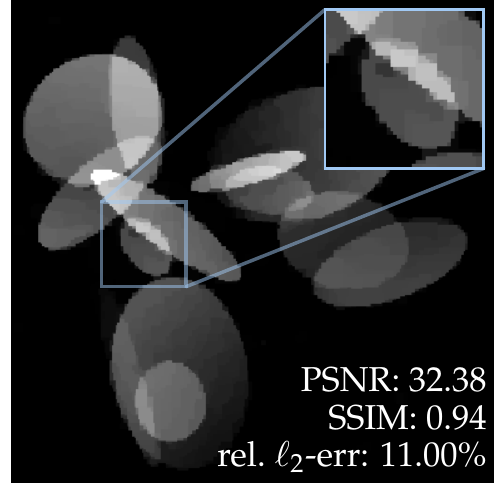} & \includegraphics[valign=c,width=0.2\textwidth]{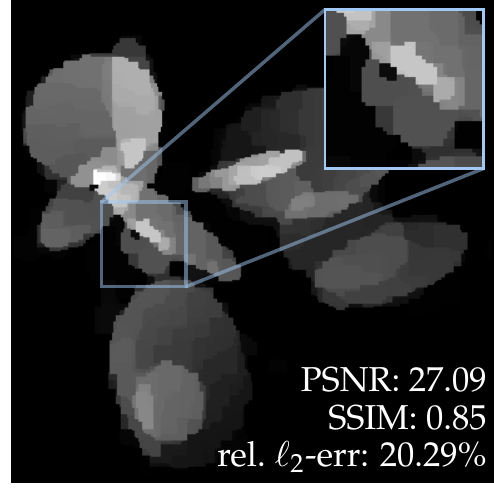} & \\
		\rotatebox[origin=c]{90}{$\ItNet$} &
		\includegraphics[valign=c,width=0.2\textwidth]{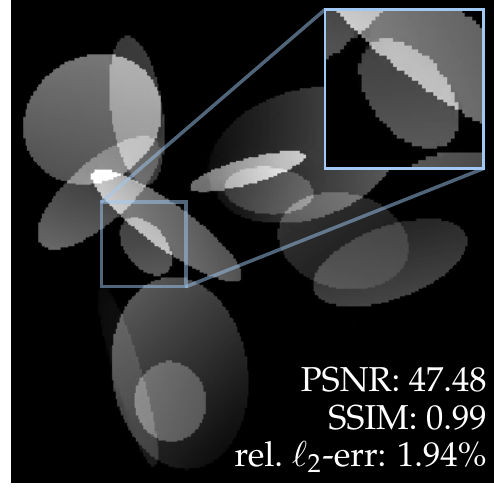} &
		\includegraphics[valign=c,width=0.2\textwidth]{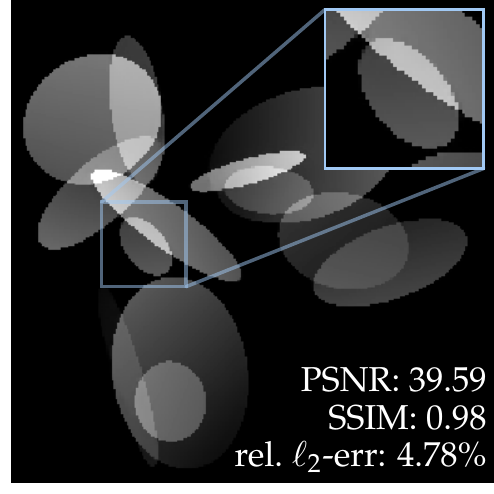} & \includegraphics[valign=c,width=0.2\textwidth]{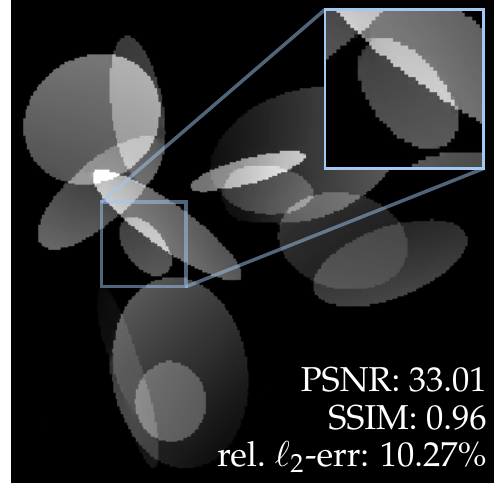} & \includegraphics[valign=c,width=0.2\textwidth]{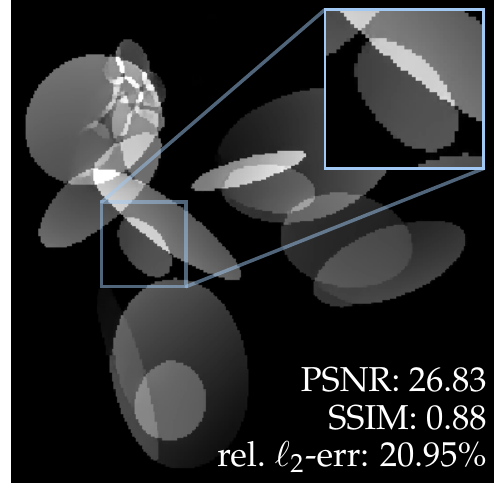} & \\
		\multicolumn{4}{l}{\parbox{.55\textwidth}{\captionof{figure}{\label{fig:ellipses:example_adv}\textbf{Scenario~\refBone{} -- Fourier meas.~with ellipses.} Individual reconstructions of a randomly selected image from the test set for different levels of adversarial noise. The reconstructed images are displayed in the window $[0,0.9]$, which is also used for the computation of the PSNR and SSIM. For error plots and the results of $\UNet$ and $\TiraFL$, we refer to Fig.~\ref{fig:ellipses:example_adv_supp}. The bottom right figure concerns the \emph{transferability} of adversarial noise: it shows the reconstruction $\TV[\noisebnd](\yadv)$, where $\yadv$ is the perturbation found for $\ItNet$ with 8\% noise; see Fig.~\ref{fig:ellipses:transfer} for additional experiments. The ground truth image $\xgrtr$ has been omitted, as it is visually indistinguishable from the noiseless reconstruction by $\TV[\noisebnd]$.}}} &
		\includegraphics[valign=c,width=0.2\textwidth]{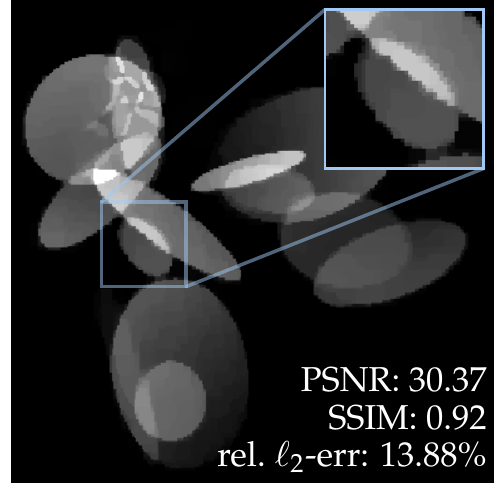} &
		\rotatebox[origin=c]{90}{\parbox{0.2\textwidth}{\centering Transferability \\ $\ItNet \longrightarrow \TV[\noisebnd]$}}
	\end{tabular}
\end{figure*}

\begin{figure*}[p]
	\centering
	\scriptsize
	\begin{tabular}{l@{\,}c@{\,}c@{\,}c@{\,}c@{\enspace}c}
		& noiseless & 0.5\% rel.~noise -- adversarial & 1\% rel.~noise -- adversarial & 2\% rel.~noise -- adversarial & \\
		\rotatebox[origin=c]{90}{$\TV[\noisebnd]$} &
		\includegraphics[valign=c,width=0.2\textwidth]{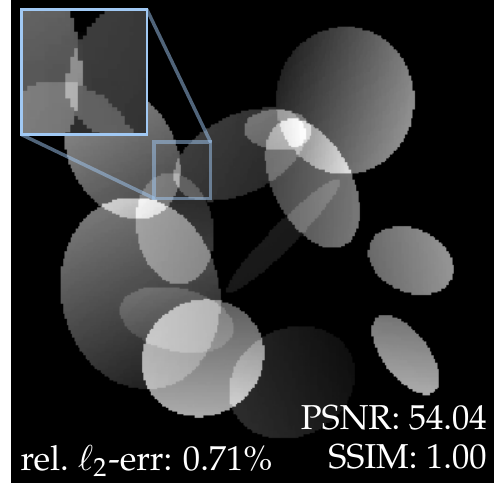} &
		\includegraphics[valign=c,width=0.2\textwidth]{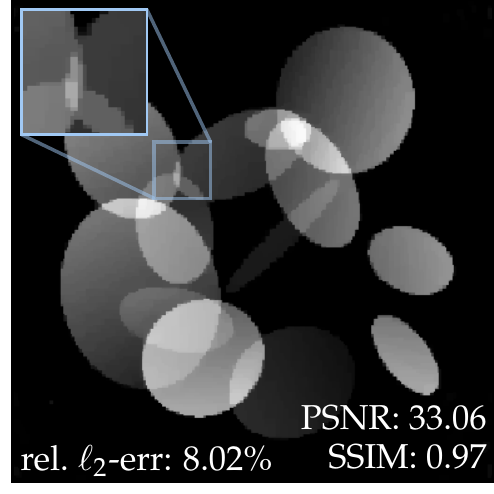} & \includegraphics[valign=c,width=0.2\textwidth]{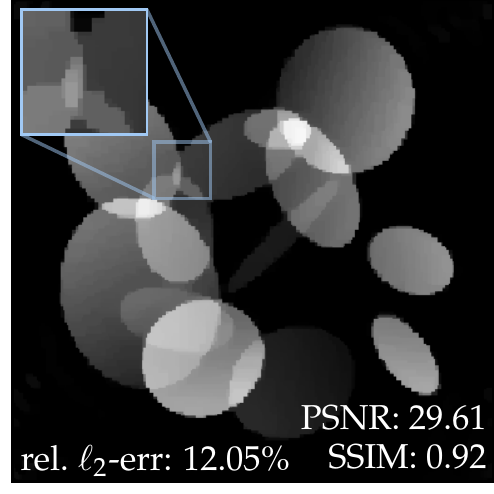} & \includegraphics[valign=c,width=0.2\textwidth]{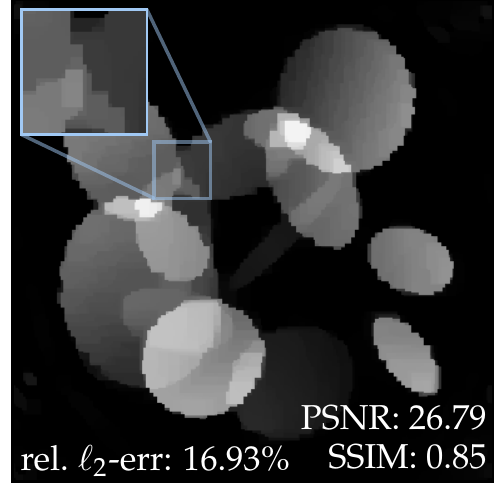} & \\
		\rotatebox[origin=c]{90}{$\UNet$} &
		\includegraphics[valign=c,width=0.2\textwidth]{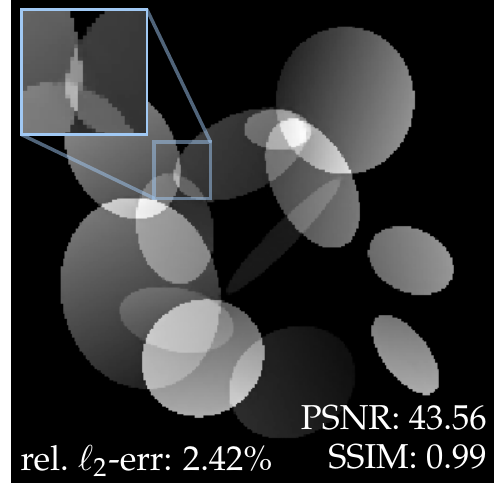} &
		\includegraphics[valign=c,width=0.2\textwidth]{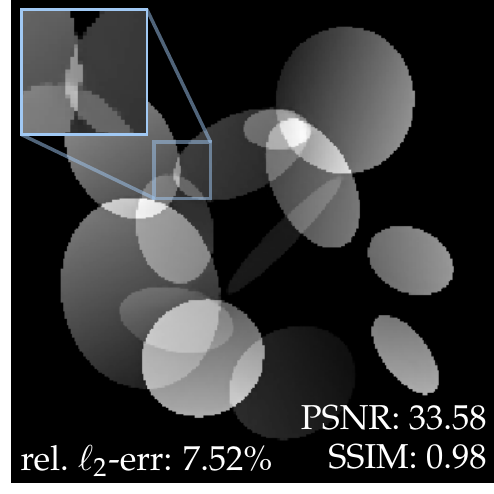} & \includegraphics[valign=c,width=0.2\textwidth]{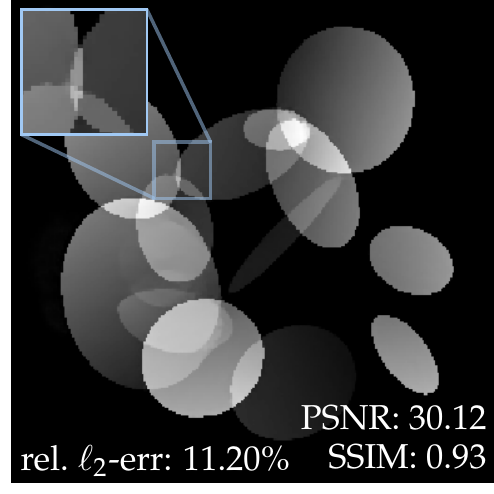} & \includegraphics[valign=c,width=0.2\textwidth]{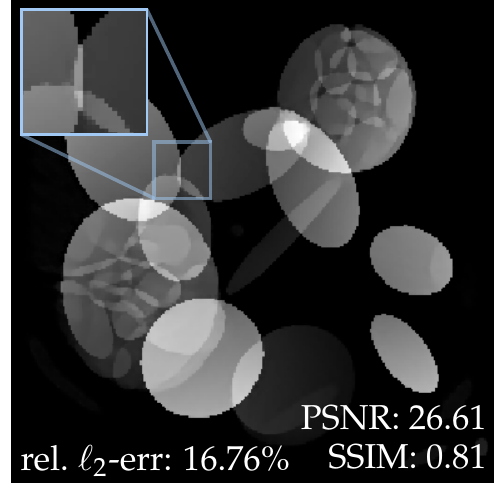} & \\
		\multicolumn{4}{l}{\parbox{.55\textwidth}{\captionof{figure}{\label{fig:ellipses:example_adv_radon}\textbf{Scenario~\refBtwo{} -- Radon meas.~with ellipses.} Individual reconstructions of a randomly selected image from the test set for different levels of adversarial noise. The reconstructed images are displayed in the window $[0,1]$, which is also used for the computation of the PSNR and SSIM. The bottom right figure shows the FBP inversion of the 2\%-adversarial perturbation found for $\UNet$. The ground truth image $\xgrtr$ has been omitted, as it is visually indistinguishable from the noiseless reconstruction by $\TV[\noisebnd]$.}}} &
		\includegraphics[valign=c,width=0.2\textwidth]{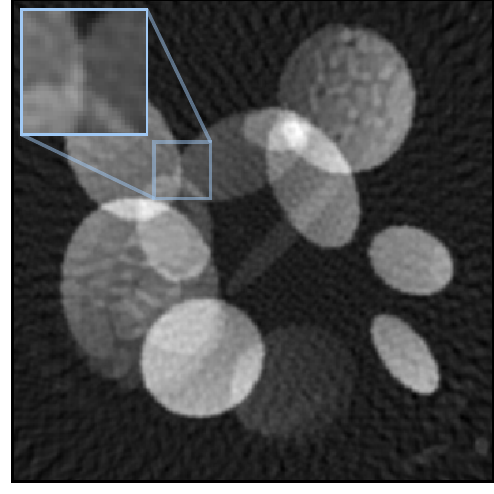} &
		\rotatebox[origin=c]{90}{$\inv{\A}\yadv$}
	\end{tabular}
\end{figure*}

In contrast to Case Study~A, the aforementioned problems are of significantly higher dimensionality, requiring several adaptions.
First, fully-learned schemes are difficult to realize, since the size of the inversion layer scales multiplicatively in the image dimensions.
In the Fourier case, the number of free parameters can be reduced by enforcing a Kronecker product structure on $\Lin \in \C^{N\times m}$; this exploits the fact that $\ftop$ is a tensor product of two 1D Fourier transforms, cf.~\cite{sch+19}.
Furthermore, due to the non-separability of $\tvnorm{(\cdot)}$, the formulation of $\TV[\noisebnd]$ in \eqref{eq:methods:tv} becomes computationally infeasible for finding adversarial noise.
Hence, we solve the unconstrained version of $\TV[\noisebnd]$ instead, i.e., the objective function is changed to $\x \mapsto \lambda \cdot \tvnorm{\x} + \lnorm{\A \x - \y}^2$. Note that this strategy is theoretically equivalent \cite[Appx.~B]{fh13}, but requires an appropriate choice of the regularization parameter $\lambda > 0$. A near-optimal selection with respect to the relative $\l{2}$-error is determined by grid searches over the test set and a densely sampled range of noise levels $\noisebnd$.

Fig.~\ref{fig:ellipses:table} shows the noise-to-error curves for \emph{Scenario~\refBone{} (Fourier meas.~with ellipses)}; see also Table~\ref{tab:ellipses:table_adv} and~\ref{tab:ellipses:table_ref}.
The associated individual reconstructions for $\TV[\noisebnd]$ and $\ItNet$ with adversarial noise are displayed in Fig.~\ref{fig:ellipses:example_adv}; see Fig.~\ref{fig:ellipses:example_adv_supp} for the remaining networks and Fig.~\ref{fig:ellipses:example_gauss} for the corresponding results with Gaussian noise.
In the tables and individual reconstructions, we have also reported the \emph{peak signal-to-noise ratio (PSNR)} and \emph{structural similarity index measure (SSIM)} \cite{wbss04}.
In the case of \emph{Scenario~\refBtwo{} (Radon meas.~with ellipses)}, we only present individual reconstructions based on $\TV[\noisebnd]$ and $\UNet$; see Fig.~\ref{fig:ellipses:example_adv_radon} for adversarial noise and Fig.~\ref{fig:ellipses:example_poisson_radon} for the common Poisson noise model.
This restriction is due to the more complicated nature of the Radon transform, and in particular, the need for automatic differentiation.
The used implementation \cite{ern20} requires significantly more computational effort, compared to the fast Fourier transform.

\textsf{\textbf{Conclusions:}} The main findings of Case Study~A remain valid: (i) the adversarial robustness of NN-based methods and TV minimization is similar with respect to the $\l{2}$-error; (ii) NNs are more resilient against statistical perturbations in mid- to high-noise regimes (see also the individual reconstructions in Fig.~\ref{fig:ellipses:example_gauss} and~\ref{fig:ellipses:example_poisson_radon}); (iii) there is a clear gap between adversarial and statistical noise that is comparable for model-based and learned schemes.

The individual reconstruction results in Fig.~\ref{fig:ellipses:example_adv} and~\ref{fig:ellipses:example_adv_radon} allow for further insights.
First, the effect of adversarial noise for $\TV[\noisebnd]$ manifests itself in the well-known staircasing phenomenon, a considerable loss of resolution as well as point-like artifacts (see the zoomed region in Fig.~\ref{fig:ellipses:example_adv}).
In contrast, NN-based methods always produce sharp images, with almost imperceptible visual errors up to 3\% relative noise in the case of Fourier measurements (1\% noise in the case of Radon measurements).
For the highest noise level, on the other hand, they exhibit unnatural ellipsoidal artifacts.

At first sight, this observation might indicate a vulnerability to adversarial noise.
However, a simple \emph{transferability test} refutes this conclusion (cf.~\cite{pmg16}): plugging the perturbed measurements for $\ItNet$ into $\TV[\noisebnd]$ leads to the same ellipsoidal artifacts; see Fig.~\ref{fig:ellipses:example_adv} and Fig.~\ref{fig:ellipses:transfer}. Furthermore, Fig.~\ref{fig:ellipses:example_adv_radon} reveals that the corresponding artifacts are already present in the FBP inversion and are not caused by the post-processing network.
This shows that the learned solvers do not suffer from undesired instabilities, but the observed artifacts are due to actual features in the corrupted measurements.
Interestingly, adversarial perturbations found for $\TV[\noisebnd]$ do not transfer to NN-based methods, see Fig.~\ref{fig:ellipses:transfer}.
Overall, the attack strategy of \eqref{eq:methods:findadv} has different qualitative effects on each reconstruction paradigm: while known flaws of TV minimization are amplified, the NNs are perturbed by adding ``real'' ellipsoidal features to the measurements.

On a final note, we confirm the ranking of architectures as pointed out in Case Study~A. Nevertheless, there is no clear superiority of the fully learned schemes as in case of Gaussian measurements, since the inverse Fourier transform appears to be a near-optimal choice of model-based inversion layer.

\subsection{Case Study C: MRI on Real-World Data (fastMRI)}
\label{sec:results:C}

The third case study of this article is devoted to a real-world MRI scenario.
To this end, we use the publicly available \emph{fastMRI} knee dataset, which consists of 1594 multi-coil diagnostic knee MRI scans.\footnote{Data used in the preparation of this article were obtained from the NYU fastMRI Initiative database \cite{zbo+19,kno+20} (\url{https://fastmri.med.nyu.edu}). As such, NYU fastMRI investigators provided data but did not participate in analysis or writing of this article. The primary goal of fastMRI is to test whether machine learning can aid in the reconstruction of medical images.}
Our experiments are based on the subset of 796 coronal proton-density weighted scans without fat-suppression, resulting in $M \approx 17\text{k}$ training images.
We draw magnitude images $\xgrtr\in\R^{320\times 320}$, obtained from fully-sampled multi-coil\footnote{Note that our measurement model actually corresponds to the simpler modality of subsampled single-coil MRI. While the fastMRI challenge also provides single-coil data, it is based on retrospective masking of \emph{emulated} Fourier measurements. The subsampling is done by omitting k-space lines in the phase-encoding direction, which we found less suitable for our robustness analysis; see Section~\ref{sec:additional:badmeas} for an experiment with the original setup. Since emulating single-coil measurements is unavoidable, we have decided to sample from the multi-coil magnitude reconstructions in favor of higher image quality. This was found to be particularly important to ensure that TV minimization can serve as a competitive benchmark method, at least for noiseless measurements.} data, and consider subsampled Fourier measurements as in Scenario~\refBone{} with 50 radial lines ($m = 17178$ and $N=320^2=102400$).
As before, the data processing is complex-valued, while the actual reconstructions are computed as real-valued magnitude images.
The model-based and learned inversion layers are realized as in Scenario~\refBone{}. 
As common in the fastMRI challenge, we have trained all networks with a cost function based on a combination of the $\l{1}$- and SSIM-distance, see also~\cite{zgfk17}. 
TV~minimization is solved in the unconstrained formulation, with the regularization parameter determined by a grid search over a subset of the validation set.

\begin{figure}
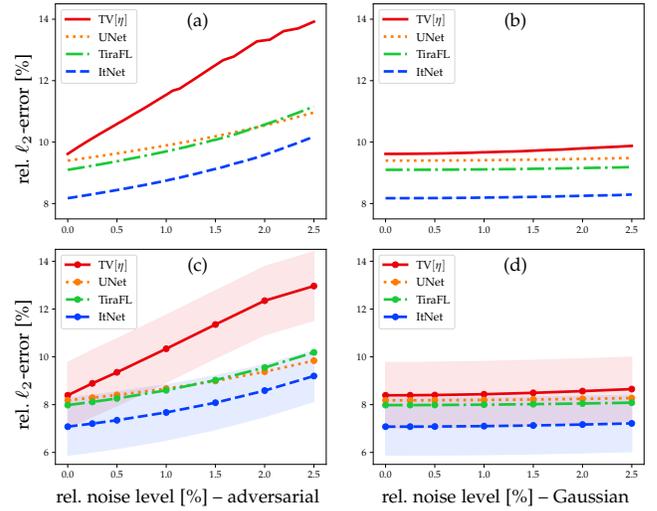

	\centering
	\scriptsize
	\setlength\tabcolsep{0pt}
	\begin{tabular}{cc@{\,\,}c}
		\rotatebox[origin=c]{90}{rel.~$\l{2}$-error [\%]} &
		\graphicwithlabel{fastmri_v4/results/attacks/fig_example_S8_adv_curve.pdf}{.22\textwidth}{(a)}{.9}{.55} &
		\graphicwithlabel{fastmri_v4/results/attacks/fig_example_S8_gauss_curve.pdf}{.22\textwidth}{(b)}{.9}{.55} \\
		\rotatebox[origin=c]{90}{rel.~$\l{2}$-error [\%]} &
		\graphicwithlabel{fastmri_v4/results/attacks/fig_table_adv.pdf}{.22\textwidth}{(c)}{.9}{.55} &
		\graphicwithlabel{fastmri_v4/results/attacks/fig_table_gauss.pdf}{.22\textwidth}{(d)}{.9}{.55} \\
		& rel.~noise level [\%] -- adversarial & rel.~noise level [\%] -- Gaussian
	\end{tabular}
	\caption{\textbf{Case Study C -- fastMRI.} (a)~shows the adversarial noise-to-error curve for the randomly selected image of Fig.~\ref{fig:fastmri:example_adv}. (b)~shows the corresponding Gaussian noise-to-error curve, where the mean and (almost imperceptible) standard deviation are computed over 50~draws of $\Noise$. (c)~and (d)~display the respective curves averaged over 30~images from the validation set. For the sake of clarity, we have omitted the standard deviations for $\UNet$ and $\TiraFL$, which behave similarly.}
	\label{fig:fastmri:table}
\end{figure}

\begin{figure*}
	\centering
    \scriptsize
   \begin{tabular}{l@{\,}c@{\,}c@{\,}c@{\,}c@{\enspace}c}
 & noiseless & 1\% rel.~noise -- adversarial & 1.5\% rel.~noise -- adversarial & 2.5\% rel.~noise -- adversarial & \\
\rotatebox[origin=c]{90}{$\TV[\noisebnd]$} &
\includegraphics[valign=c,width=0.2\textwidth]{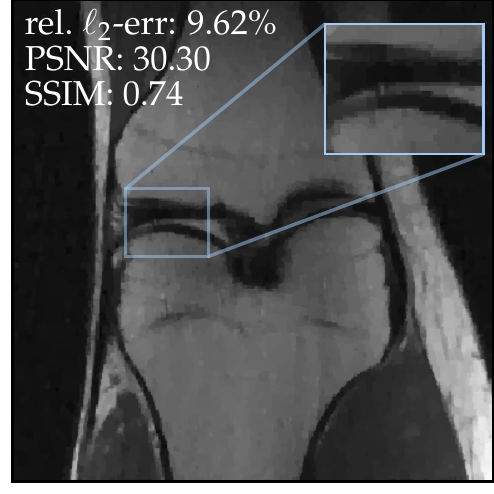} &
\includegraphics[valign=c,width=0.2\textwidth]{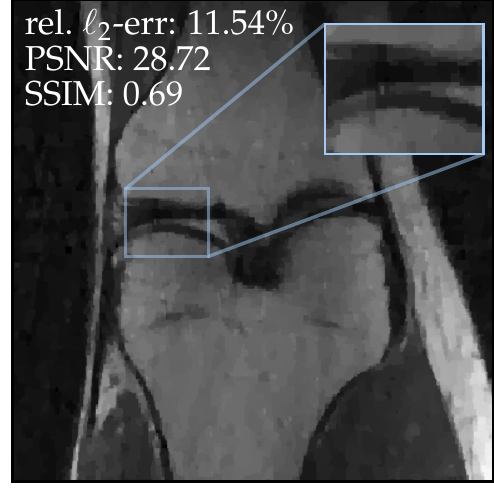} & \includegraphics[valign=c,width=0.2\textwidth]{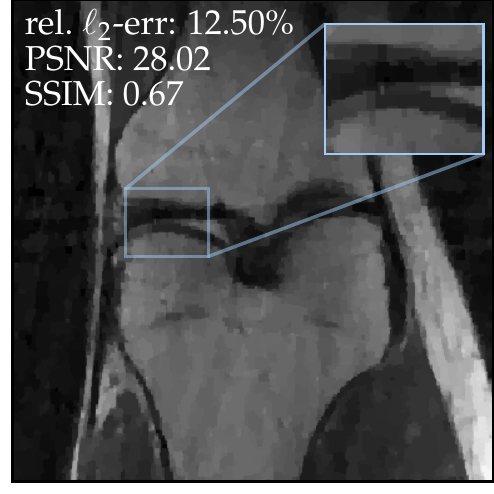} & \includegraphics[valign=c,width=0.2\textwidth]{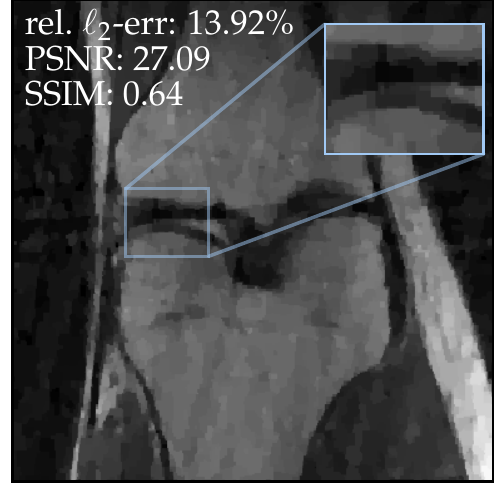} & \\
\rotatebox[origin=c]{90}{$\TiraFL$} &
\includegraphics[valign=c,width=0.2\textwidth]{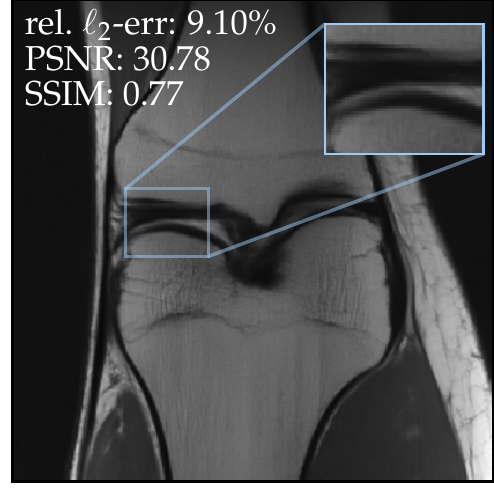} &
\includegraphics[valign=c,width=0.2\textwidth]{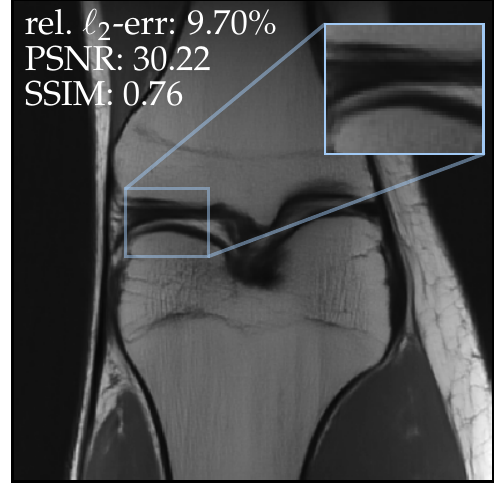} & \includegraphics[valign=c,width=0.2\textwidth]{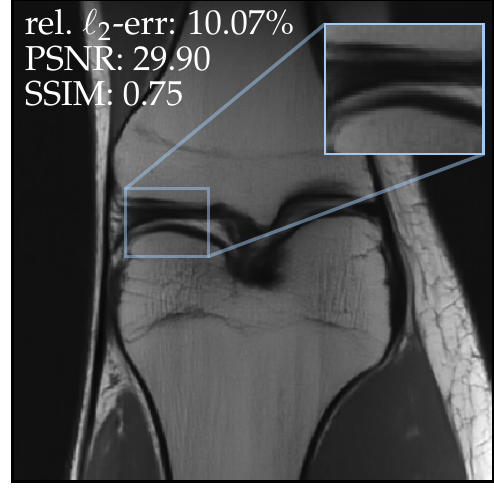} & \includegraphics[valign=c,width=0.2\textwidth]{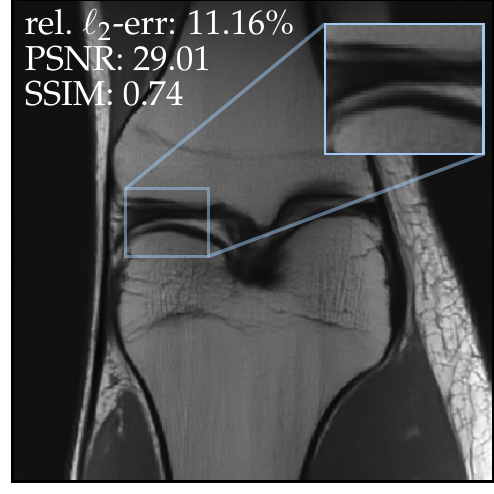} & \\
\multicolumn{4}{l}{\parbox{.55\textwidth}{\captionof{figure}{\label{fig:fastmri:example_adv}\textbf{Case Study C -- fastMRI.} Individual reconstructions of a central slice of a randomly selected volume from the validation set for different levels of adversarial noise. The reconstructed images are displayed in the window $[0.05,4.50]$, which is also used for the computation of the PSNR and SSIM. For error plots and the results of $\UNet$ and $\ItNet$, we refer to Fig.~\ref{fig:fastmri:example_adv_supp}. The ground truth image $\xgrtr$ is shown at the bottom right.}}} &
\includegraphics[valign=c,width=0.2\textwidth]{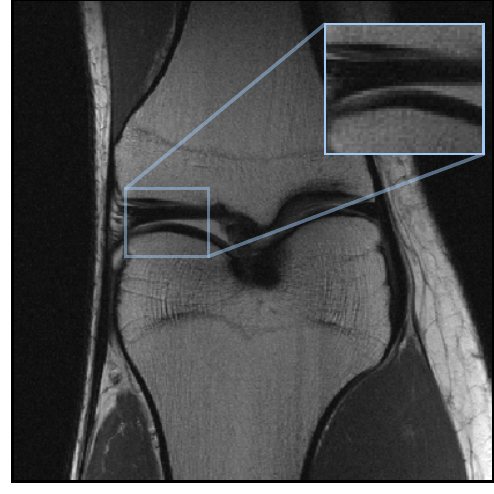} &
\rotatebox[origin=c]{90}{ground truth}
\end{tabular}
\end{figure*}

Fig.~\ref{fig:fastmri:table} shows the noise-to-error curves; see also Table~\ref{tab:fastmri:table_adv} and~\ref{tab:fastmri:table_ref}.
The associated individual reconstructions for $\TV[\noisebnd]$ and $\TiraFL$ with adversarial noise are displayed in Fig.~\ref{fig:fastmri:example_adv}; see Fig.~\ref{fig:fastmri:example_adv_supp} for the remaining networks and Fig.~\ref{fig:fastmri:example_ref} for the corresponding results with Gaussian noise.

\textsf{\textbf{Conclusions:}} Our experimental results show that the main findings of Case Study~A and~B carry over to real-world data.
The noise-to-error curves in Fig.~\ref{fig:fastmri:table} reveal a superior robustness of the learned reconstruction schemes over TV minimization, even for noiseless measurements (cf.~Scenario~\refAtwo{}).
Fig.~\ref{fig:fastmri:example_adv} underpins this observation from a qualitative viewpoint: the model-based prior of $\TV[\noisebnd]$ tends to blur fine details in the reconstructed images---this ``oil~painting'' effect becomes stronger with larger perturbations.
In contrast, the NN-based reconstructions always yield high resolution images.
Despite adversarial noise, the central image region---which is of main medical interest---remains largely unaffected, whereas tiny vessel structures appear in the outside (fat) region.
Such an amplification of existing patterns is comparable to the ellipsoidal artifacts in Case Study~B.
We emphasize that this phenomenon only occurs for large adversarial perturbations, where the benchmark of TV minimization already suffers from severe distortions.
In particular, the performance of the learned methods is not impaired by the same amount of Gaussian noise (see Fig.~\ref{fig:fastmri:example_ref}).

\section{Further Aspects of Robustness}
\label{sec:additional}

This section presents several additional experiments that allow for further insights into the robustness of learned methods.

\subsection{Training Without Noise -- An Inverse Crime?}
\label{sec:additional:crime}

\begin{figure}
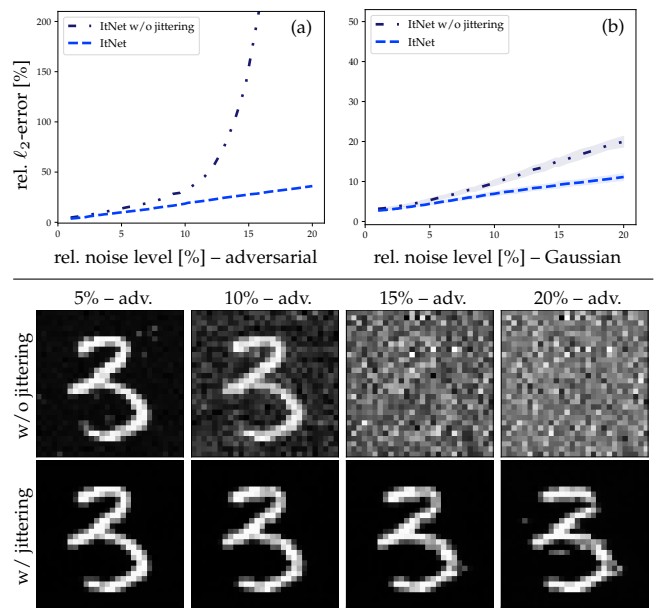

	\centering
	\scriptsize
	\setlength\tabcolsep{0pt}
	\begin{tabular}{ccc}
		\rotatebox[origin=c]{90}{rel.~$\l{2}$-error [\%]} &
		\graphicwithlabel{mnist/results/attacks/fig_example_S0_crime_curve.pdf}{.22\textwidth}{(a)}{.9}{.9} &
		\graphicwithlabel{mnist/results/attacks/fig_example_S0_crime_curve_gauss.pdf}{.22\textwidth}{(b)}{.9}{.9} \\
		& rel.~noise level [\%] -- adversarial & rel.~noise level [\%] -- Gaussian \\[.5em] \hline
	\end{tabular}
	\\[.5em]
	\begin{tabular}{cc@{\!\!}c@{\!\!}c@{\!\!}c@{\!\!}}
		& 5\% -- adv. & 10\% -- adv. & 15\% -- adv. & 20\% -- adv. \\
		\rotatebox[origin=c]{90}{w/o jittering} &
		\graphicwithlabel{{mnist/results/attacks/fig_example_S0_crime_unet_it_nojit_5.00e-02}.pdf}{.11\textwidth}{}{.95}{.95} &
		\graphicwithlabel{{mnist/results/attacks/fig_example_S0_crime_unet_it_nojit_1.00e-01}.pdf}{.11\textwidth}{}{.95}{.95} &
		\graphicwithlabel{{mnist/results/attacks/fig_example_S0_crime_unet_it_nojit_1.50e-01}.pdf}{.11\textwidth}{}{.95}{.95} &
		\graphicwithlabel{{mnist/results/attacks/fig_example_S0_crime_unet_it_nojit_2.00e-01}.pdf}{.11\textwidth}{}{.95}{.95} \\
		\rotatebox[origin=c]{90}{w/ jittering} &
		\graphicwithlabel{{mnist/results/attacks/fig_example_S0_crime_unet_it_jit_5.00e-02}.pdf}{.11\textwidth}{}{.95}{.95} &
		\graphicwithlabel{{mnist/results/attacks/fig_example_S0_crime_unet_it_jit_1.00e-01}.pdf}{.11\textwidth}{}{.95}{.95} &
		\graphicwithlabel{{mnist/results/attacks/fig_example_S0_crime_unet_it_jit_1.50e-01}.pdf}{.11\textwidth}{}{.95}{.95} &
		\graphicwithlabel{{mnist/results/attacks/fig_example_S0_crime_unet_it_jit_2.00e-01}.pdf}{.11\textwidth}{}{.95}{.95} \\
	\end{tabular}
	\caption{\textbf{An inverse crime?} A comparison between $\ItNet$ trained with and without jittering. The above noise-to-error curves are generated for the MNIST-digit \texttt{3} from Fig.~\ref{fig:mnist:example_adv} with (a) adversarial and (b) Gaussian noise. Individual reconstructions for adversarial noise are shown below (the intermediate steps performed by $\ItNet$ are visualized in Fig.~\ref{fig:crime:internal}).}
	\label{fig:mnist:crime}
\end{figure}

\begin{figure}
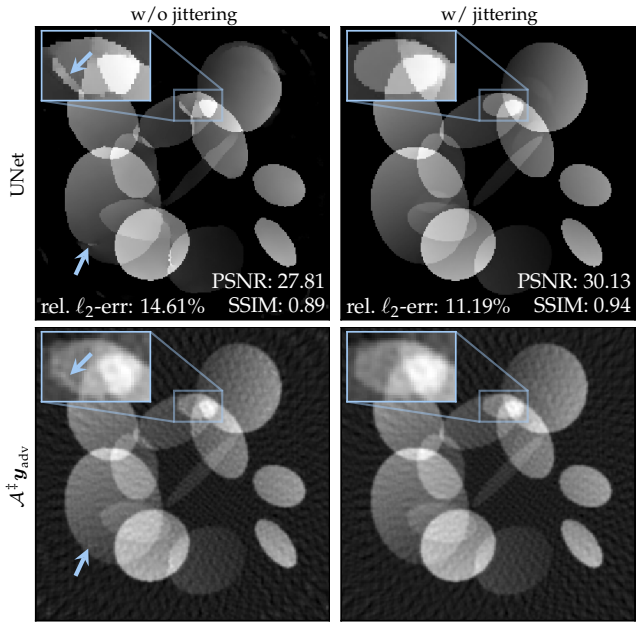

	\centering
	\scriptsize
	\setlength\tabcolsep{0pt}
	\begin{tabular}{cc@{\!\!}c}
		& w/o jittering & w/ jittering \\
		\rotatebox[origin=c]{90}{$\UNet$} &
		\graphicwitharrows{ellipses/results/attacks/fig_example_S48_radon_crime_unet_1e-02.pdf}{.22\textwidth}{.265}{.2}{245}{0.84}{0.14}{45} &
		\graphicwithlabel{ellipses/results/attacks/fig_example_S48_radon_crime_unet_jit_1e-02.pdf}{.22\textwidth}{\color{white}{}}{.95}{.95} \\
		\rotatebox[origin=c]{90}{$\inv{\A}\yadv$} &
		\graphicwitharrows{ellipses/results/attacks/fig_example_S48_radon_crime_inverter_unet_1e-02.pdf}{.22\textwidth}{.265}{.2}{245}{0.84}{0.14}{45} & 
		\graphicwithlabel{ellipses/results/attacks/fig_example_S48_radon_crime_inverter_unet_jit_1e-02.pdf}{.22\textwidth}{\color{white}{}}{.95}{.95} \\
	\end{tabular}
	\caption{\textbf{An inverse crime?} A comparison between $\UNet$ trained with and without jittering for image recovery from sparse-angle Radon measurements, see Fig.~\ref{fig:ellipses:example_adv_radon} in Scenario~\refBtwo{}. The reconstructions are obtained for 1\% adversarial noise. The bottom figures show the FBP inversions of the found perturbations, respectively. The blue arrows highlight tiny distortions that are amplified by the post-processing network.}
	\label{fig:ellipses:crime}
\end{figure}

In this section, the importance of \emph{jittering} for the stability of deep-learning-based reconstruction schemes is discussed (see Section~\ref{sec:methods:nntrain}). We have found that this technique can be beneficial for promoting adversarial robustness, in particular, for iterative architectures.
The previous claim is verified by an ablation study, comparing two versions of $\ItNet$ for Scenario~\refAtwo{}, one trained with jittering and the other without. The resulting noise-to-error curves in Fig.~\ref{fig:mnist:crime} reveal that noiseless training data can have drastic consequences. Indeed, the relative recovery error blows up at $\sim$15\% adversarial noise if jittering is not used. 
In a similar experiment, we analyze the adversarial robustness of image recovery from Radon measurements as in Scenario~\refBtwo{}. The results of Fig.~\ref{fig:ellipses:crime} show a clear superiority of the $\UNet$ that was subjected to noise during training (see also Fig.~\ref{fig:ellipses:example_poisson_radon} for the effect of Poisson noise). Without jittering, almost imperceptible distortions in the FBP inversions are intensified by the post-processing network (see blue arrows).

The above observations can be related to the notion of \emph{inverse crimes} in the literature on inverse problems, e.g., see \cite{ks06,ms12}.
This term is commonly used to explain the phenomenon of exact, but highly unstable, recovery from noiseless, simulated measurements.
In a similar way, networks seem to learn accurate, but unstable, reconstruction rules if they are trained with noiseless data.
We note that this does not only concern simulated phantom data but also real-world scenarios.
Indeed, in medical imaging applications, one often acquires fully sampled (noisy) reference scans $\{\tilde\y^i\}_{i = 1}^M$, which are used to generate the ground truth training images $\xgrtr^i = \A^{-1}_{\texttt{full}} \tilde\y^i$. The measurements are usually subsampled retrospectively by $\y^i = P \tilde\y^i$, where $P$ denotes an appropriate selection operator. NN-based solution methods for the limited data problem \eqref{eq:intro:problem} with $\A = P \A_{\texttt{full}}$ are then obtained by training on $\{(\y^i,\xgrtr^i)\}_{i = 1}^M$.
Importantly, such data pairs also ``commit'' an inverse crime, since they follow the noiseless forward model $\A \xgrtr^i =P \A_{\texttt{full}} \xgrtr^i = \y^i$.
Hence, we believe that simulating additional noise might be helpful in the situation of real-world measurements as well.
Jittering is a simple and natural remedy in that regard that can additionally reduce overfitting \cite{sd91}. The exploration of further regularization techniques or more sophisticated ways of injecting noise during training is left to future research.  

\subsection{Adversarial Examples for Classification From Compressed Measurements}
\label{sec:additional:cclass}

In medical healthcare, image recovery is merely one component of the entire data-processing chain.
Indeed, machine learning techniques are particularly suitable for automated diagnosis or personalized treatment recommendations.
As argued in the introduction of this article, the study of adversarial examples for such classification tasks differs from the robustness analysis of reconstruction methods.
In this section, we shed further light on this subject by analyzing classification from compressed measurements---think of detecting a tumor from a subsampled MRI scan.

To this end, we revisit the toy model of Scenario~\refAtwo{}, with the goal to predict MNIST digits from their Gaussian measurements.
This is realized by training a basic convolutional NN classifier $\ConvNet \colon \R^N \to [0, 1]^{10}$, mapping images to class probabilities for each of the 10 digits.
The concatenation with a reconstruction method $\Rec \colon \R^m \to \R^N$ then yields the following classification map:
\begin{equation}\label{eq:additional:compclass}
	\CompClass \colon \R^m \to [0,1]^{10}, \ \y \mapsto [\ConvNet \circ \Rec](\y).
\end{equation}
The approach of $\CompClass$ can be seen as a simplified model for the automated diagnosis from subsampled measurements; see also \cite{bmr17} and the references therein for the related problem of \emph{compressed classification}.

\begin{figure}
	\centering
	\scriptsize
	\setlength\tabcolsep{0pt}
	\begin{tabular}{cc}
		\rotatebox[origin=c]{90}{accuracy [\%]} &
		\includegraphics[valign=c,width=.30\textwidth]{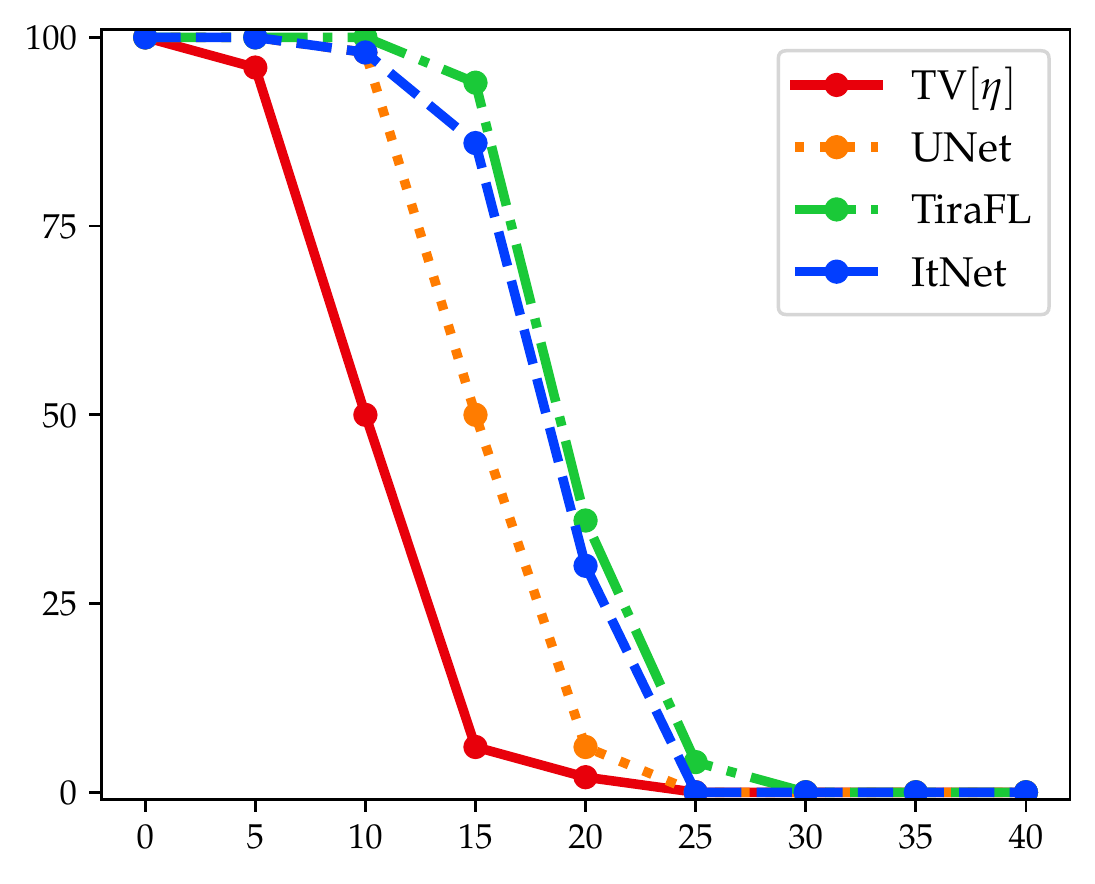} \\
		& rel.~noise level [\%] -- adversarial  \\[.5em] \hline
	\end{tabular} \\[.5em]
	\begin{tabular}{l@{\,}c@{\,}c@{\,}c@{\,}c}
		& 5\% & 10\% & 15\%  & 20\%  \\
		\rotatebox[origin=c]{90}{$\TV[\noisebnd]$} &
		\includegraphics[valign=c,width=0.06\textwidth]{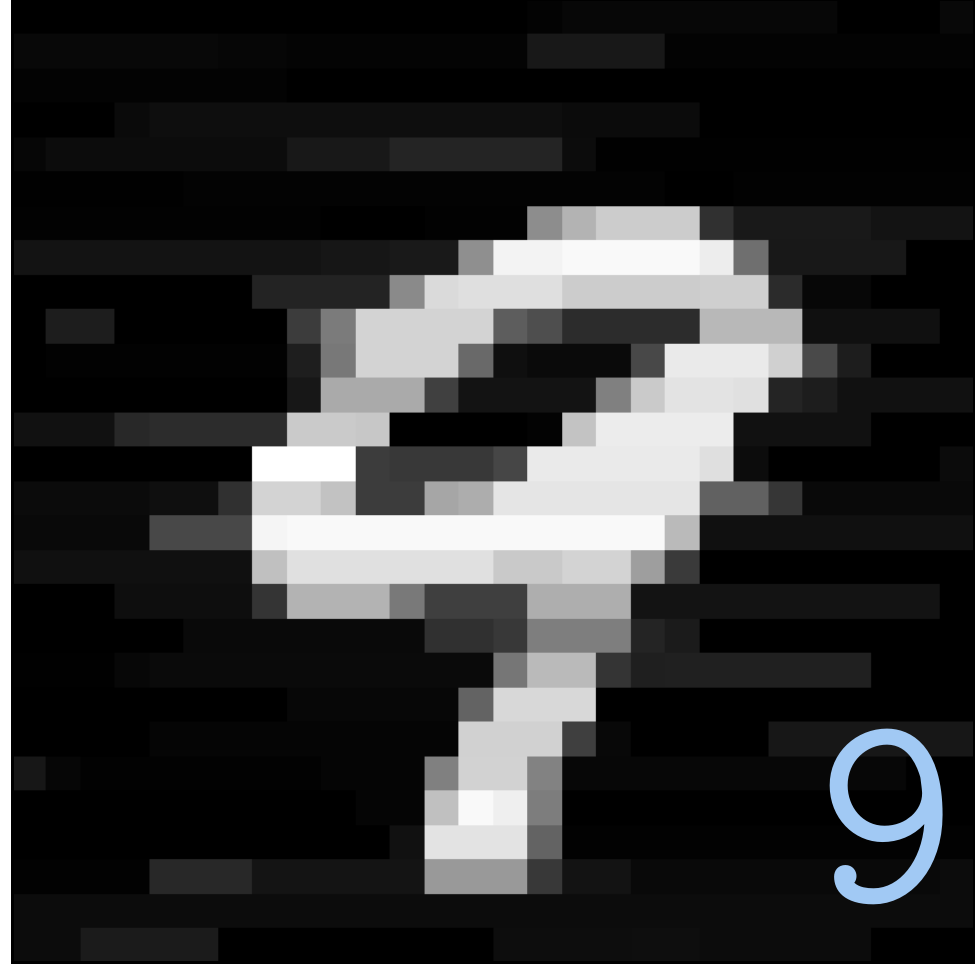} &
		\includegraphics[valign=c,width=0.06\textwidth]{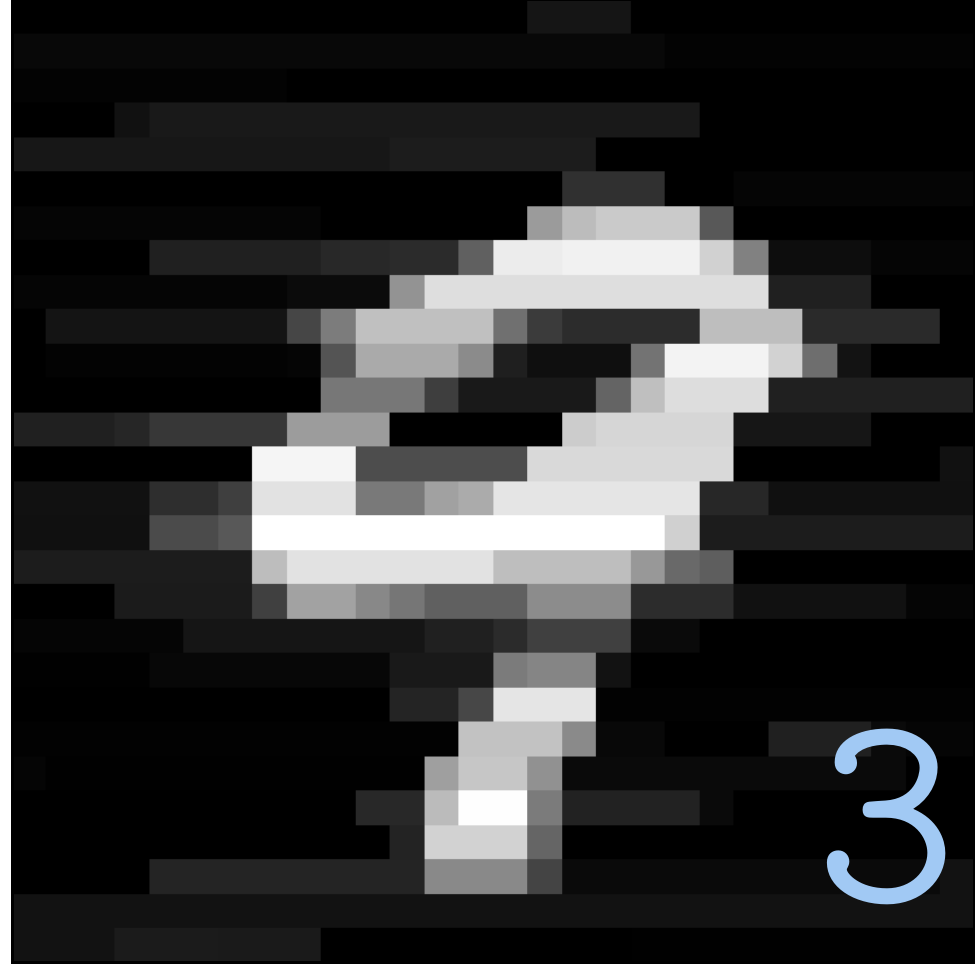} &
		\includegraphics[valign=c,width=0.06\textwidth]{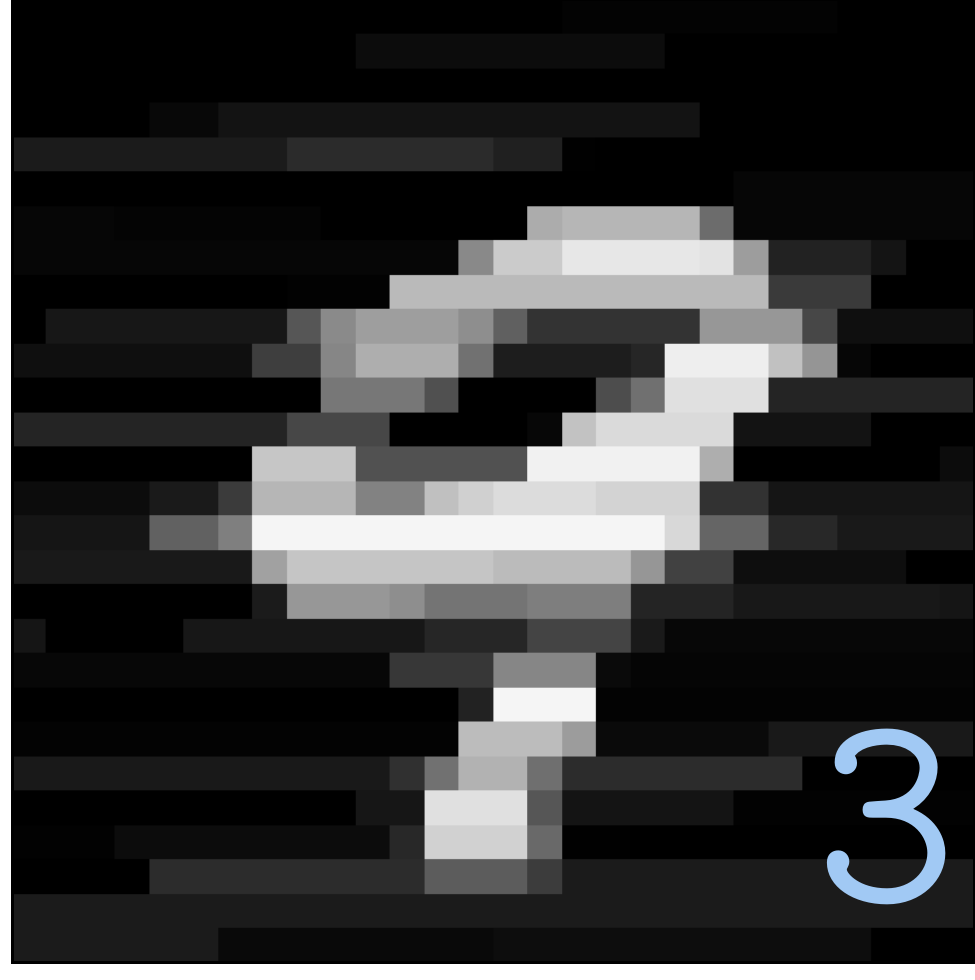} &
		\includegraphics[valign=c,width=0.06\textwidth]{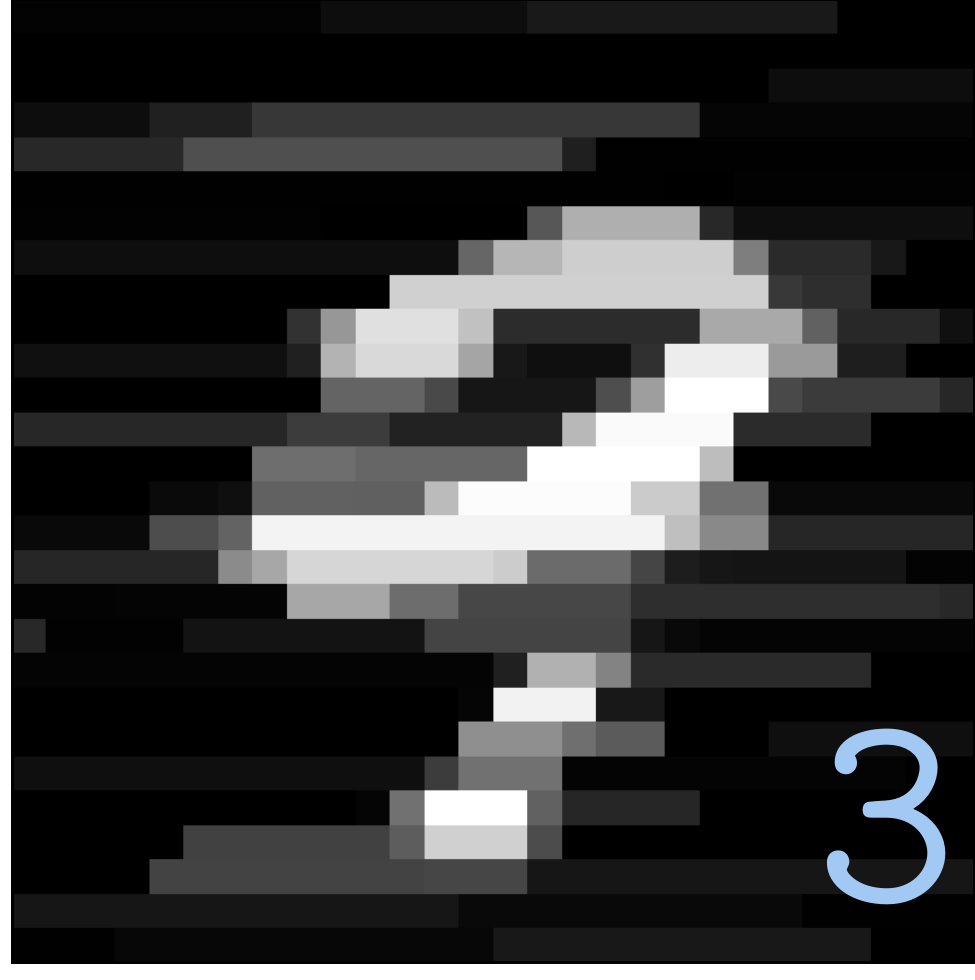} \\
		\rotatebox[origin=c]{90}{$\UNet$} &
		\includegraphics[valign=c,width=0.06\textwidth]{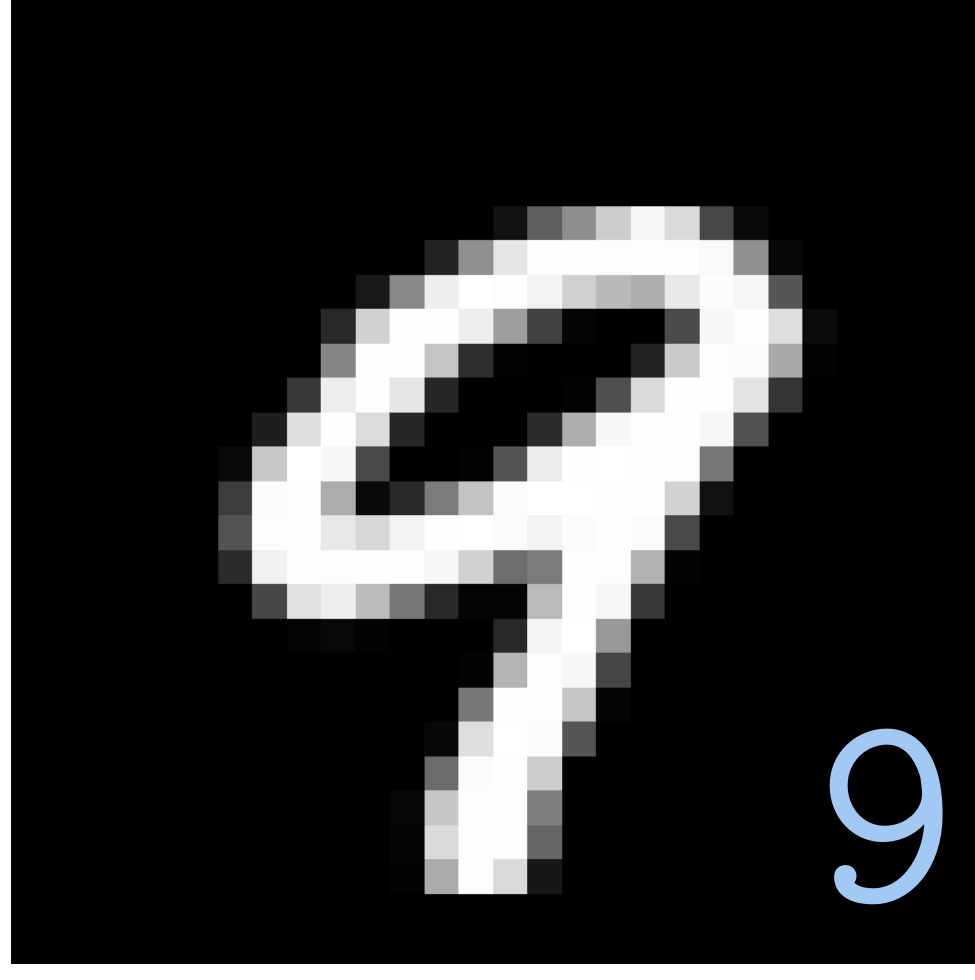}  &
		\includegraphics[valign=c,width=0.06\textwidth]{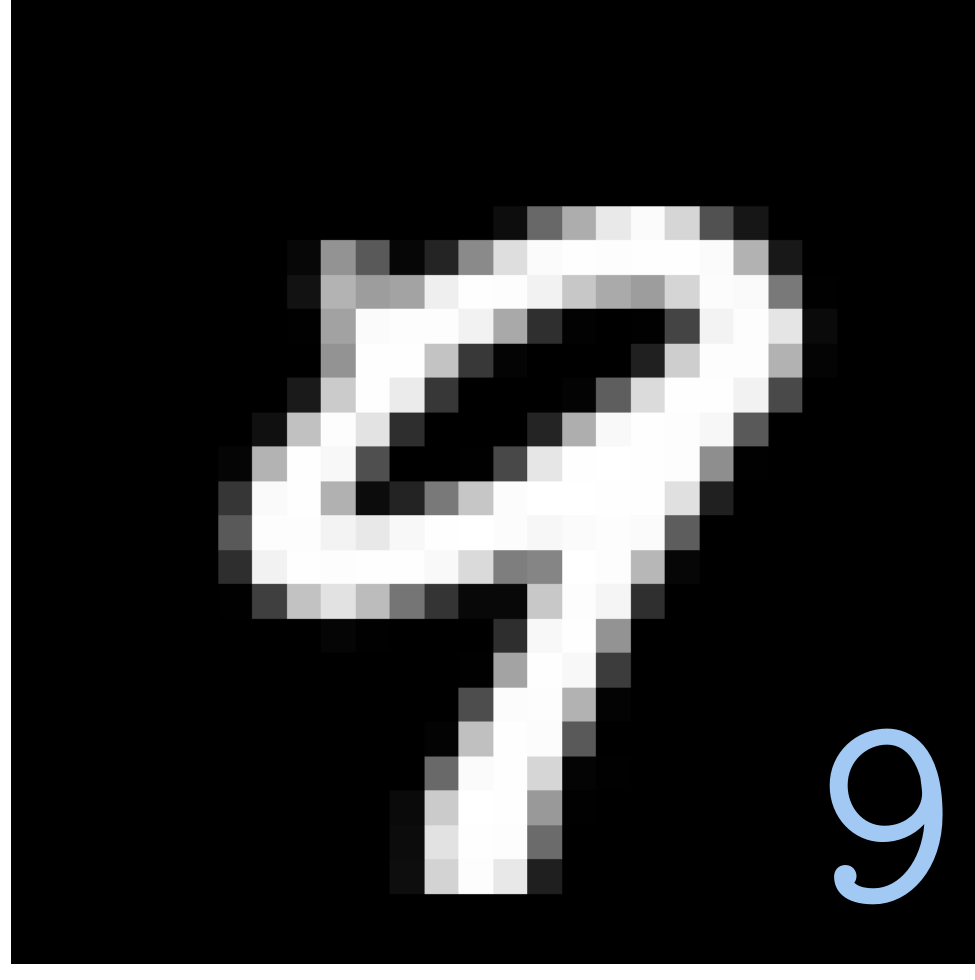}  &
		\includegraphics[valign=c,width=0.06\textwidth]{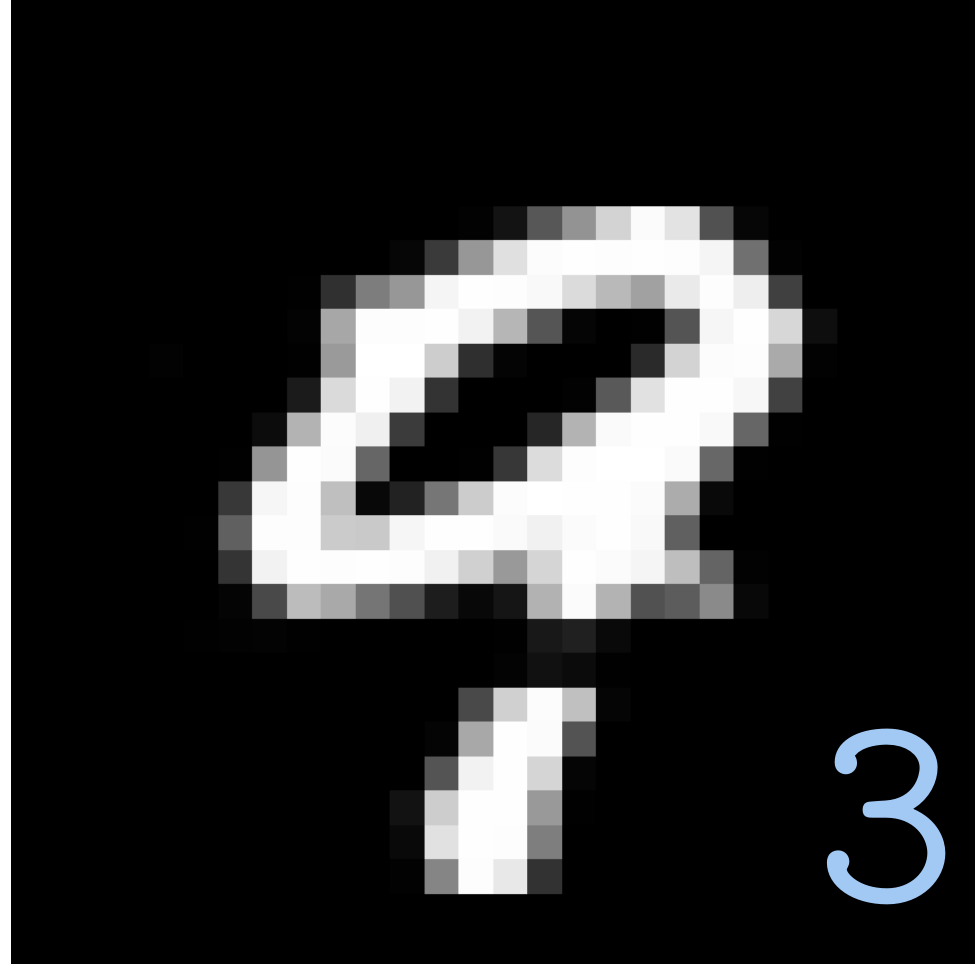}  &
		\includegraphics[valign=c,width=0.06\textwidth]{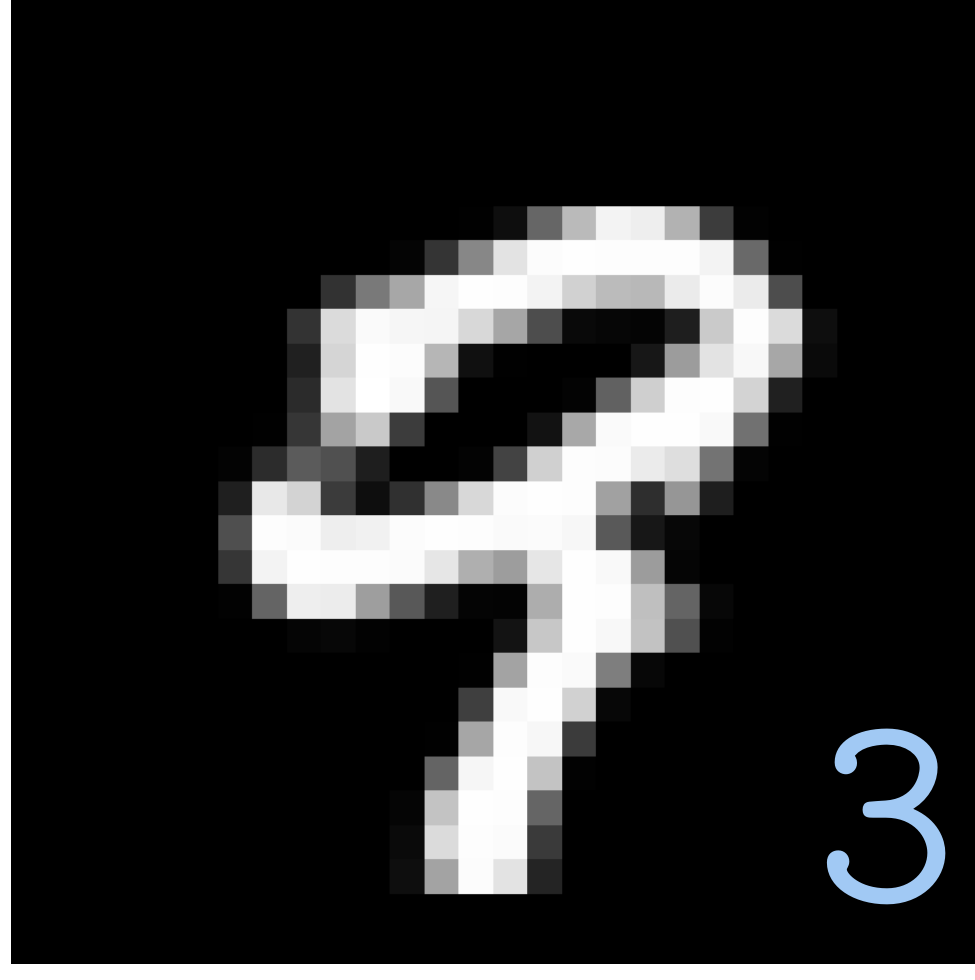}  \\
		\rotatebox[origin=c]{90}{$\TiraFL$} &
		\includegraphics[valign=c,width=0.06\textwidth]{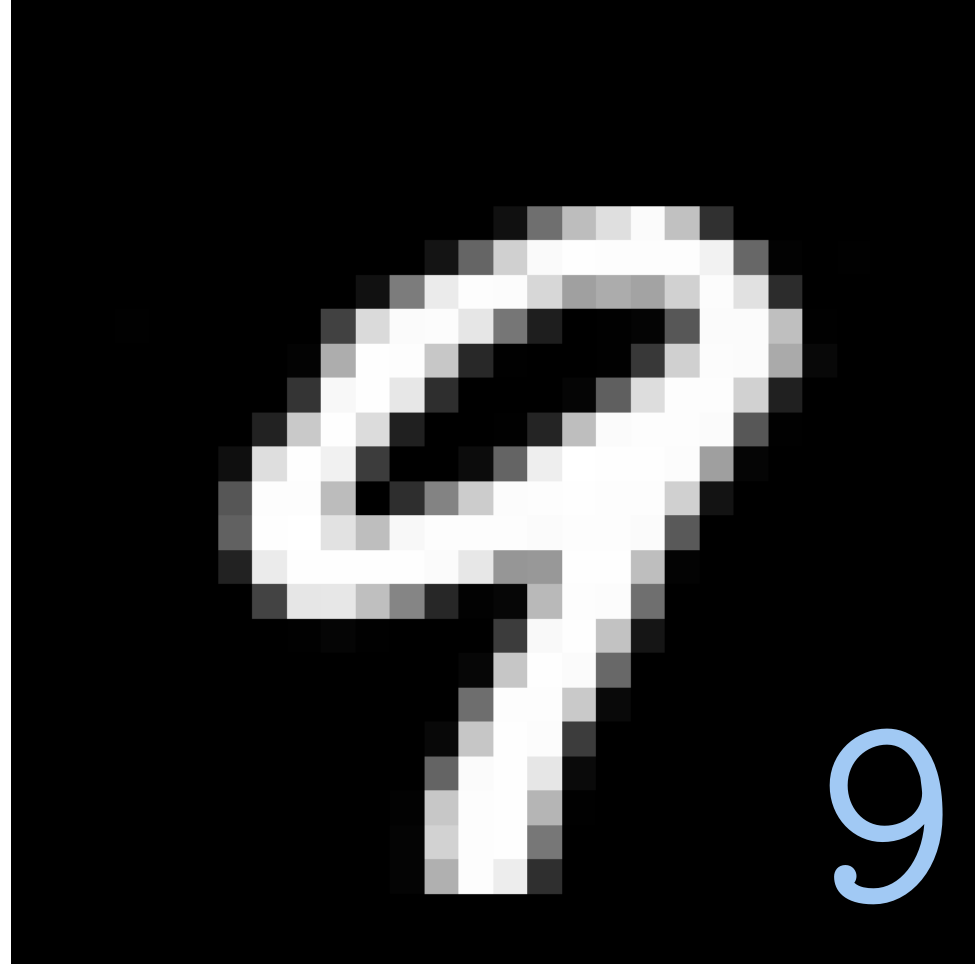}  &
		\includegraphics[valign=c,width=0.06\textwidth]{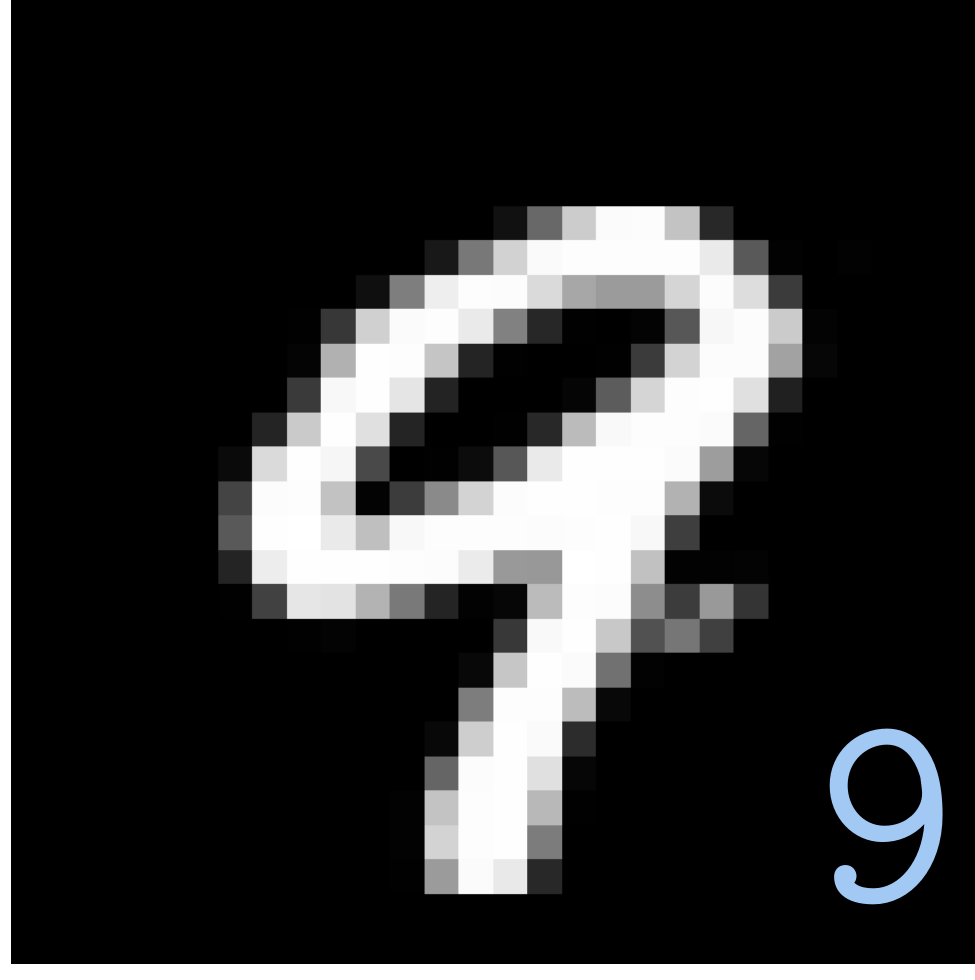}  &
		\includegraphics[valign=c,width=0.06\textwidth]{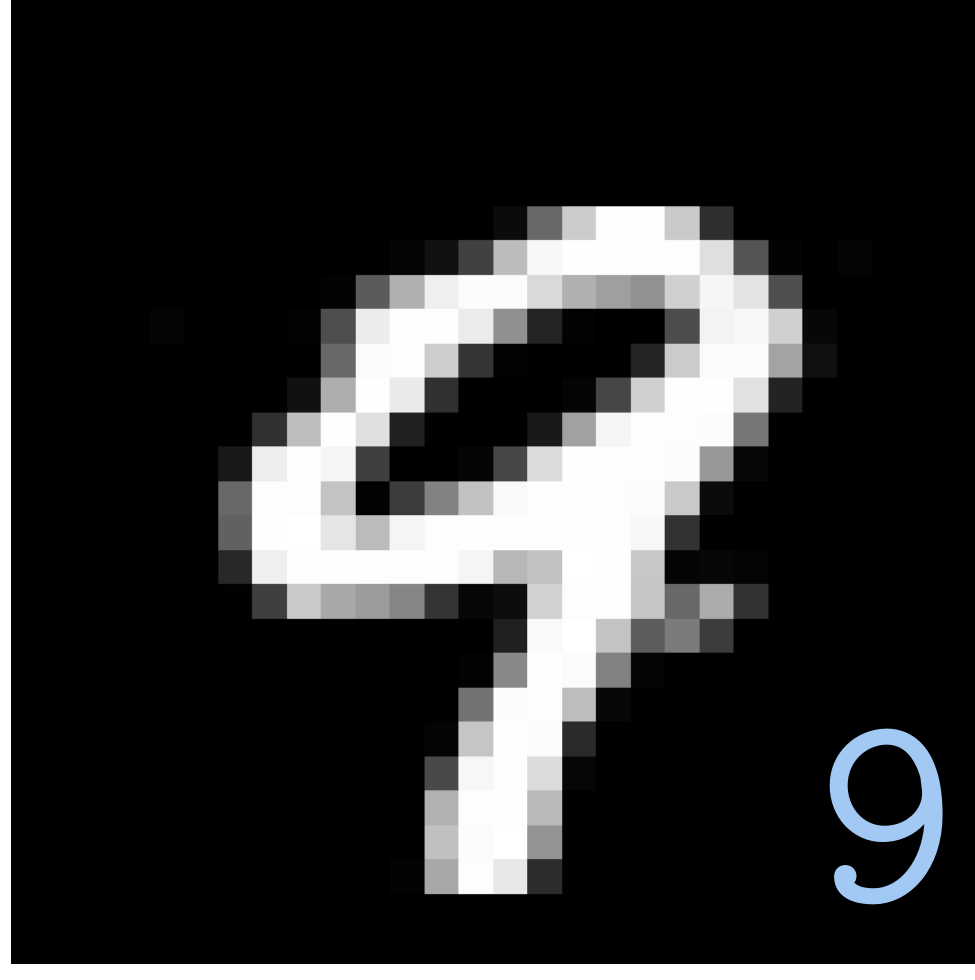}  &
		\includegraphics[valign=c,width=0.06\textwidth]{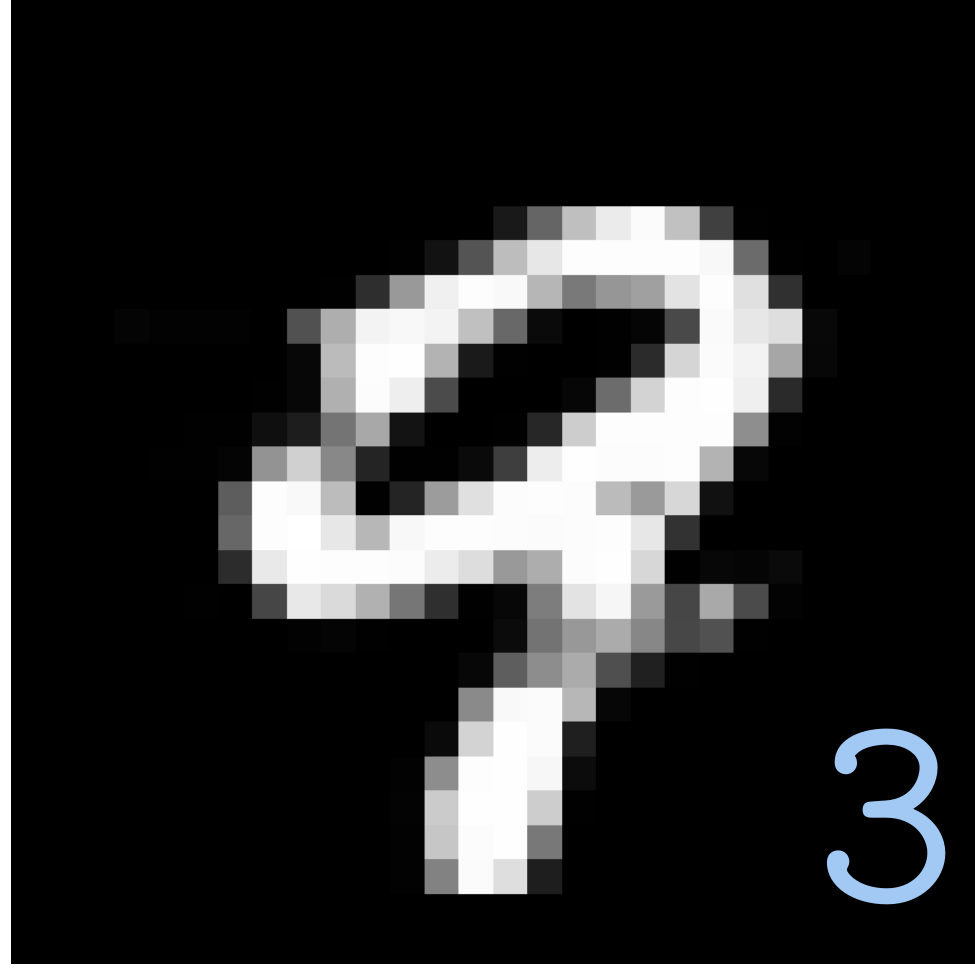}  \\
		\rotatebox[origin=c]{90}{$\ItNet$} &
		\includegraphics[valign=c,width=0.06\textwidth]{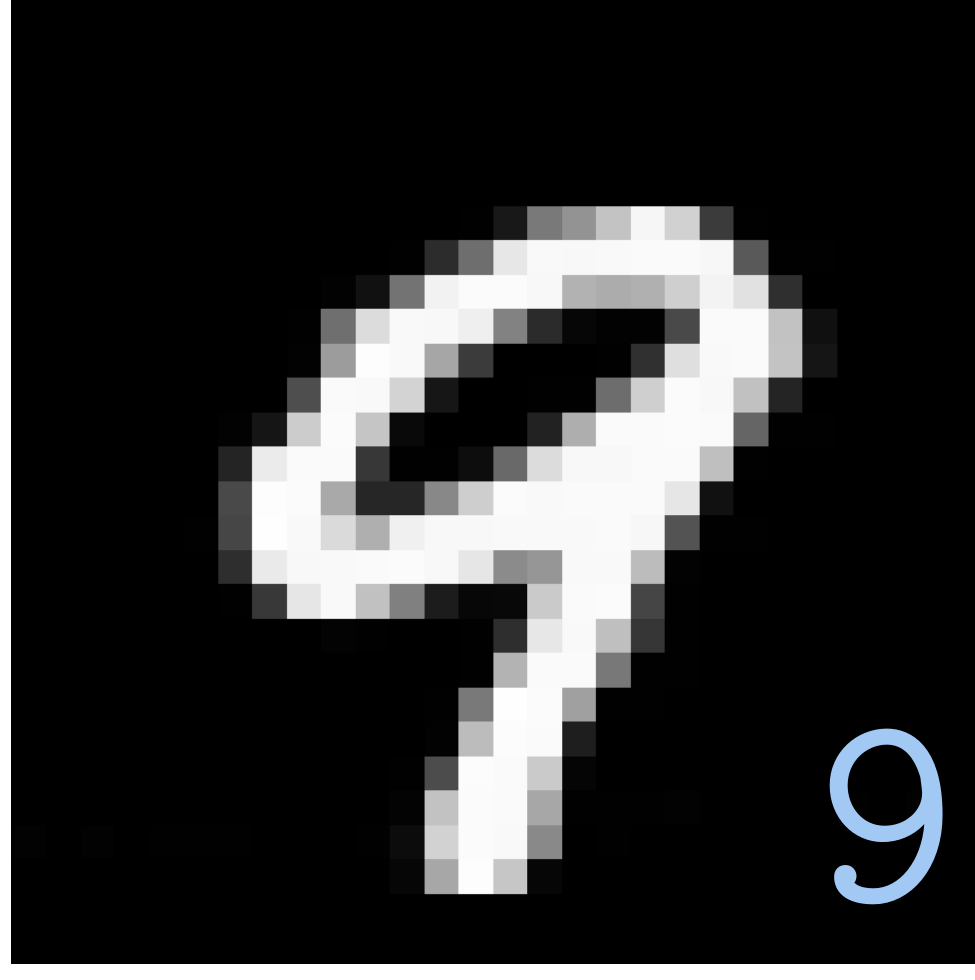} &
		\includegraphics[valign=c,width=0.06\textwidth]{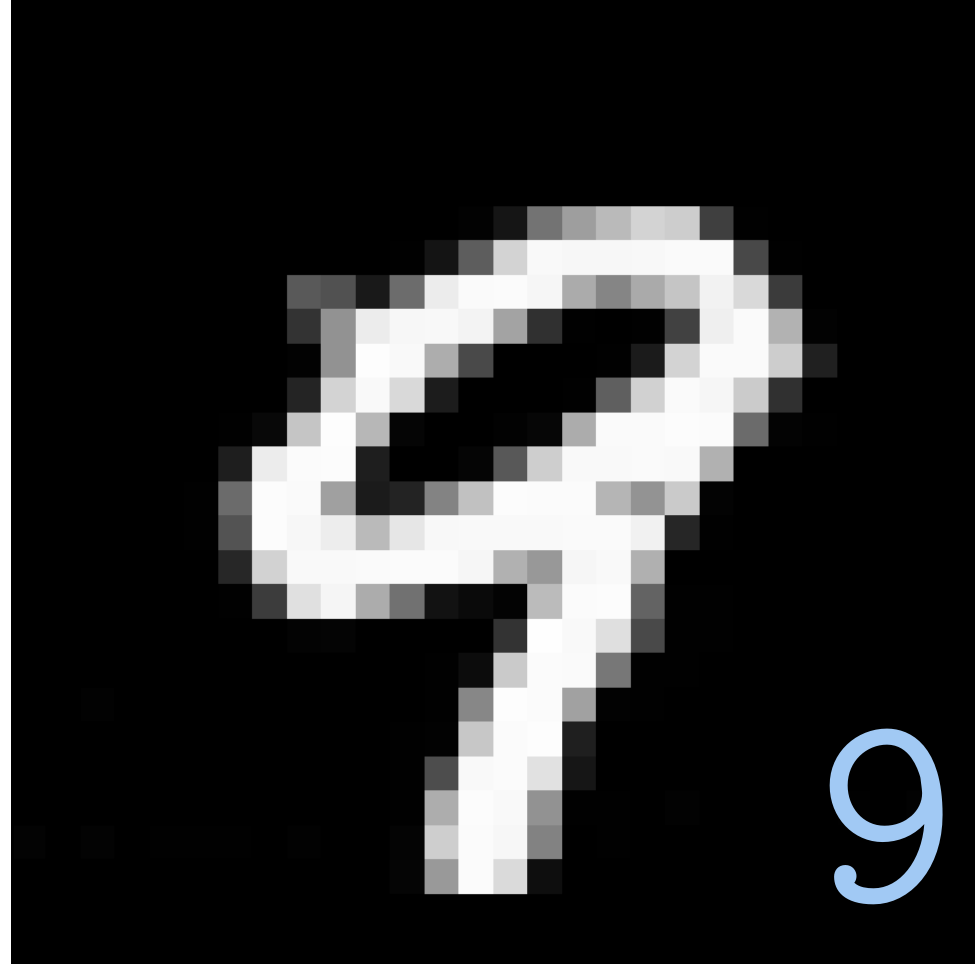} &
		\includegraphics[valign=c,width=0.06\textwidth]{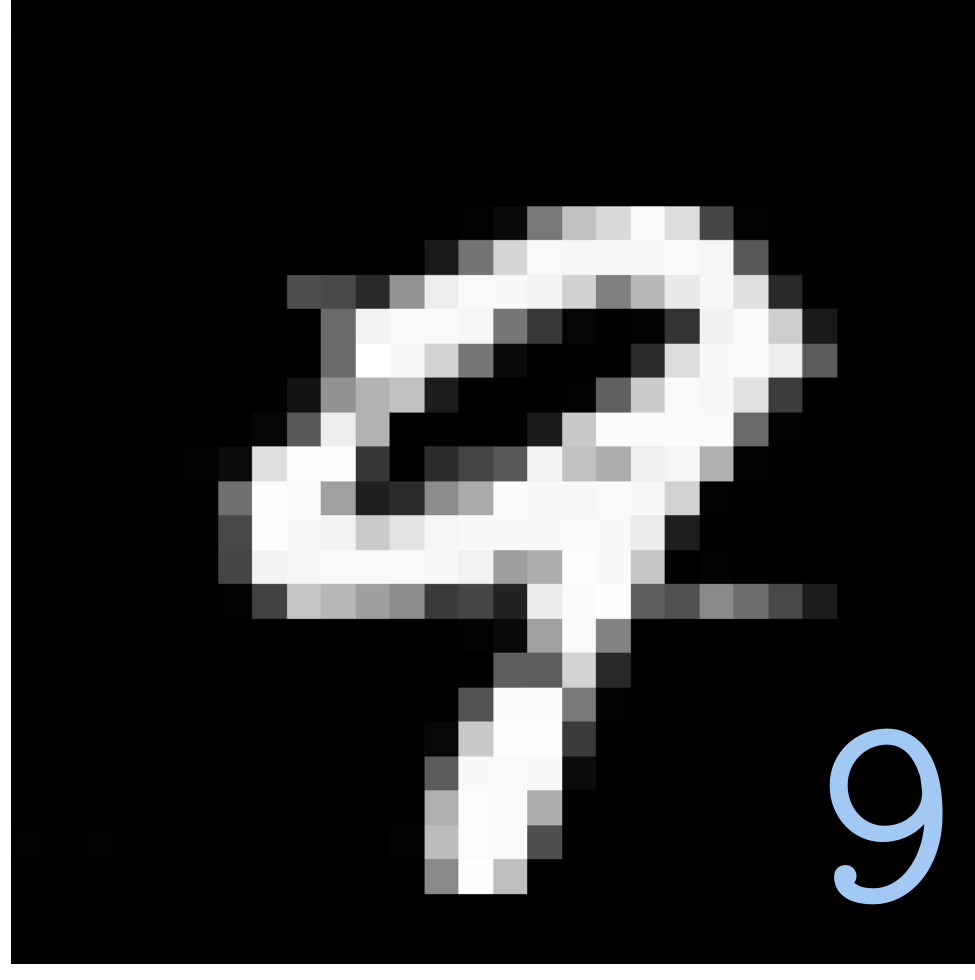} &
		\includegraphics[valign=c,width=0.06\textwidth]{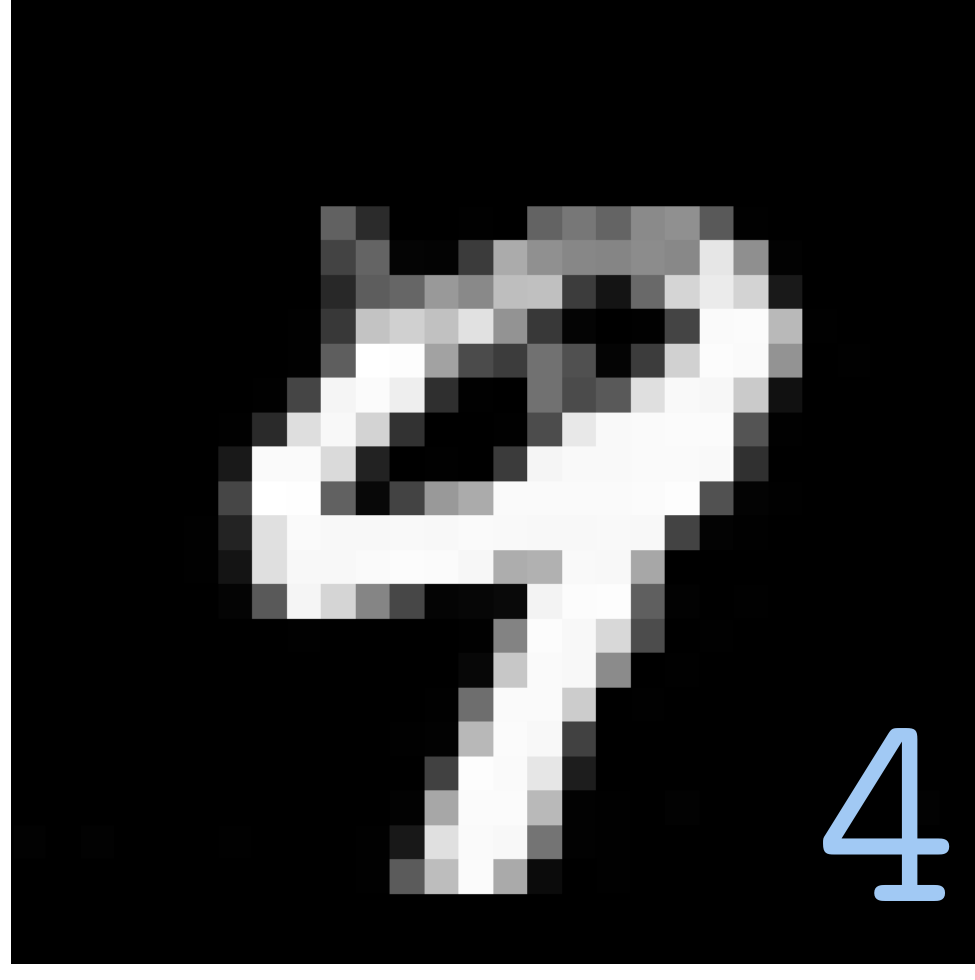}
	\end{tabular}
	\caption{\textbf{Classification from compressed measurements.} The above curve plots the relative adversarial noise level against the prediction accuracy of the classfier \eqref{eq:additional:compclass} for different recovery methods (averged over 50 digits from the test set). The intermediate reconstructions of a randomly selected digit are shown below for different noise levels. Their predicted class labels are displayed in the bottom right corner.}
	\label{fig:mnist:table_classification}
\end{figure}

Inspired by \cite{cw17}, we adapt the attack strategy \eqref{eq:methods:findadv} to the classification setting by (approximately) solving
\[
	\Noiseadv = \argmax_{\lnorm{\Noise} \leq \noisebnd} \ \max_{k\neq c} \ [\CompClass(\ygrtr + \Noise)]_k - [\CompClass(\ygrtr + \Noise)]_c
\]
where $c\in\{0,1,\dots,9\}$ is the true class label of $\xgrtr$.
Fig.~\ref{fig:mnist:table_classification} shows a noise-to-\emph{accuracy} curve visualizing the relative amount of correct classifications for different choices of $\Rec$. The corresponding image reconstructions $\Rec(\ygrtr + \Noiseadv)$ as well as the predicted classes $\argmax_k [\CompClass(\ygrtr + \Noiseadv)]_k$ for an example digit are presented below.

All classifiers exhibit a transition behavior: the success rate is almost perfect for small perturbations and then drops to zero at some point.
The associated images show that we have found adversarial examples in the ordinary sense of machine learning.
Indeed, every visualized reconstruction is still recognizable as the digit \texttt{9}.
In other words, although being stable, each of the recovery methods is capable of producing slightly perturbed images that fool the $\ConvNet$-part.
Remarkably, this phenomenon occurs independently of using a model-based or learned solver for \eqref{eq:intro:problem}.
We conclude that deep-learning-based data-processing pipelines (as in medical healthcare) remain vulnerable to adversarial attacks, even if provably robust reconstruction schemes are employed.

\subsection{The Original fastMRI Challenge Setup}
\label{sec:additional:badmeas}

\begin{figure}
	\centering
	\scriptsize
	\begin{tabular}{l@{\,}c@{\,}c@{\,}c@{\,}c}
		& noiseless & 2.5\% rel.~noise -- adv. \\
		\rotatebox[origin=c]{90}{$\TV[\noisebnd]$} &
		\includegraphics[valign=c,width=0.2\textwidth]{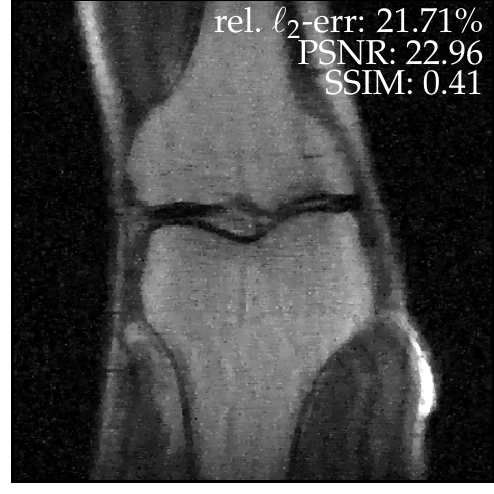} & \graphicwithbox{{fastmri/results/attacks/fig_example_challenge_V0S15_adv_tv_3e-02}.pdf}{0.2\textwidth}{0.0275}{0.15}{0.4}{0.425} \\
		\rotatebox[origin=c]{90}{$\Tira$} &
		\includegraphics[valign=c,width=0.2\textwidth]{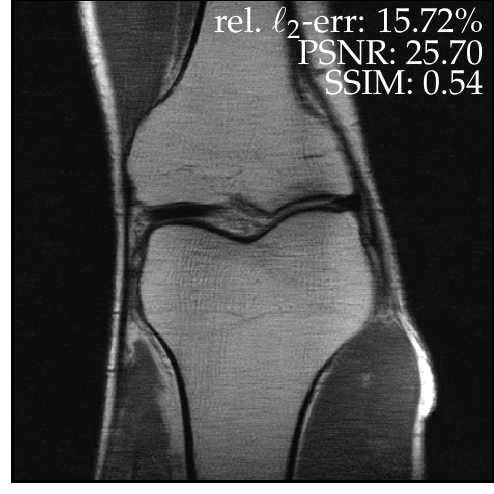} & \graphicwithbox{{fastmri/results/attacks/fig_example_challenge_V0S15_adv_tiramisu_jit_3e-02}.pdf}{0.2\textwidth}{0.0275}{0.15}{0.4}{0.425}
	\end{tabular}
	\caption{\textbf{The original fastMRI challenge setup.} Reconstructions of a randomly selected image from the validation set. Compared to the analogous experiment in Fig.~\ref{fig:fastmri:example_adv}, the Fourier subsampling operator is based on vertical lines in the k-space instead of a radial mask. The reconstructed images are displayed in the window $[0.05,4.50]$, which is also used for the computation of the PSNR and SSIM. Note that the data are given as emulated single-coil (ESC) measurements, whereas the reconstructions in Fig.~\ref{fig:fastmri:example_adv} are based on multi-coil images. Hence, the signal-to-noise ratios are not directly comparable.}
	\label{fig:additional:badmeas}
\end{figure}

This section demonstrates that the original fastMRI challenge data for single-coil MRI is more susceptible to adversarial noise.
In contrast to Case Study~C, the challenge measurement setup is based on omitting k-space lines in the phase-encoding direction (corresponding to 4-fold acceleration), i.e., the subsampling mask is defined by vertical lines. The resulting undersampling ratio of $\sim$23\% is higher than in Case Study~C ($\sim$17\%).
Fig.~\ref{fig:additional:badmeas} shows individual image reconstructions for $\TV[\noisebnd]$ and $\Tira$.\footnote{Since the fastMRI challenge setup does not rely on a fixed subsampling mask, the fully-learned approach for Tiramisu is not available here. Our $\Tira$-net performs competitively in the fastMRI public leaderboard: We have achieved an SSIM of $0.765$, whereas the leading method has $0.783$ (\url{https://fastmri.org/leaderboards/}, teamname \texttt{AnItalianDessert}, accessed on 2020-11-08).}
Compared to Fig.~\ref{fig:fastmri:example_adv}, the outcomes indicate a loss of adversarial robustness, as the reconstructed images exhibit undesired line-shaped artifacts (see blue box in Fig.~\ref{fig:additional:badmeas}).
This phenomenon occurs regardless of using a model-based ($\TV[\noisebnd]$) or learned method ($\Tira$).
In fact, the observed artifacts are a consequence of the underlying measurement system: the anisotropic mask pattern implies that vertical image features become more ``aligned''  with the kernel of the forward operator.
Hence, clearly visible distortions may be caused by relatively small perturbations of the measurements (cf.~\cite{gaah20}).
This confirms that the design of sampling patterns does not only influence the accuracy of a reconstruction method (e.g., see \cite{bcckw16}), but also its adversarial robustness.

\section{Discussion}
\label{sec:discussion}

In an extensive series of experiments, this work has analyzed the robustness of deep-learning-based solution methods for inverse problems.
Central to our approach was to study the effect of adversarial noise, i.e., worst-case perturbations of the measurements that maximize the reconstruction error. A systematic comparison with a model-based reference method has shown that standard deep NN schemes are remarkably resilient against statistical and adversarial distortions.
On the other hand, we have demonstrated that instabilities might be caused by the ``inverse crime'' of training with noiseless data.
A simple remedy in that regard is jittering---a standard regularization and robustification technique in deep learning \cite{gbc16}.
However, it is well known that this does not cure the adversarial vulnerability of deep NN classifiers, which requires more sophisticated defense strategies \cite{msg20}.
While such defenses may also improve the robustness in the context of image recovery \cite{rbk20}, our results allow for a surprising conclusion:
Injecting Gaussian random noise in the training phase seems sufficient to obtain solution methods for inverse problems that are resistant to other types of noise, including adversarial perturbations.

Admittedly, there are several aspects that go beyond the scope of our study:
(i)~We are restricted to a selection of end-to-end NN architectures, excluding other approaches, such as generative models \cite{bjpd17}, the deep image prior \cite{uvl18}, or learned regularizers \cite{lsah20}.
However, since these algorithms typically involve more model-based components, we expect their robustness to be comparable to the schemes considered in the present work.
(ii)~Due to the non-convexity of \eqref{eq:methods:findadv}, a theoretical optimality certificate for our attack strategy is lacking.
Nevertheless, our results provide empirical evidence that we have solved the problem adequately: The gap between worst-case and statistical perturbations appears consistent across all considered scenarios.
More importantly, we have verified the ability to detect an error blowup caused by adversarial noise (see Fig.~\ref{fig:mnist:crime}).
(iii)~Our analysis takes a mathematical perspective on robustness, thereby relying on standard similarity measures, in particular, the Euclidean norm.
It is well known that such quantitative metrics are insensitive to several types of visual distortions.
For instance, even the winning networks of the fastMRI challenge were unable to capture certain tiny pathological features that rarely appear in the data \cite{kno+20b}.
While some of these details are possibly lost in the subsampling process, this failure could also be due to instabilities of deep learning techniques; see \cite{ccshmp20} for recent progress in that direction.

The relevance of artificial intelligence for future healthcare is undeniable.
In this field, reliable reconstruction methods are indispensable, since errors caused by instabilities can be fatal.
Although there is typically no ``adversary'' in medical imaging (i.e., an agent that intentionally manipulates the measurements), it is reassuring to know the limits of what could go wrong in principle.
Of more practical interest is the robustness against random perturbations, which is a more realistic noise model for common imaging modalities.
We believe that our work makes progress in both regards, by showing optimistic results on the use of deep NNs for inverse problems in imaging.

%
%
%
%
%
%
 \ifCLASSOPTIONcompsoc
   \section*{Acknowledgments}
	M.G.~and M.M.~acknowledge support by the DFG Priority Programme DFG-SPP 1798.
	We express our gratitude to the Institute of Mathematics of the Technical University of Berlin for providing us hardware resources to realize the numerical experiments presented in this work.

 \else
   \section*{Acknowledgment}
 \fi
%
%
%
%
%
%
%
%
%
 \bibliographystyle{IEEEtran}
 \bibliography{references}

\begin{thebibliography}{10}
\providecommand{\url}[1]{#1}
\csname url@samestyle\endcsname
\providecommand{\newblock}{\relax}
\providecommand{\bibinfo}[2]{#2}
\providecommand{\BIBentrySTDinterwordspacing}{\spaceskip=0pt\relax}
\providecommand{\BIBentryALTinterwordstretchfactor}{4}
\providecommand{\BIBentryALTinterwordspacing}{\spaceskip=\fontdimen2\font plus
\BIBentryALTinterwordstretchfactor\fontdimen3\font minus
  \fontdimen4\font\relax}
\providecommand{\BIBforeignlanguage}[2]{{%
\expandafter\ifx\csname l@#1\endcsname\relax
\typeout{** WARNING: IEEEtran.bst: No hyphenation pattern has been}%
\typeout{** loaded for the language `#1'. Using the pattern for}%
\typeout{** the default language instead.}%
\else
\language=\csname l@#1\endcsname
\fi
#2}}
\providecommand{\BIBdecl}{\relax}
\BIBdecl

\bibitem{ldsp08}
M.~Lustig, D.~L. Donoho, J.~M. Santos, and J.~M. Pauly, ``Compressed sensing
  {MRI},'' \emph{IEEE Signal Process. Mag.}, vol.~25, no.~2, pp. 72--82, 2008.

\bibitem{hbrn08}
J.~{Haupt}, W.~U. {Bajwa}, M.~{Rabbat}, and R.~{Nowak}, ``Compressed sensing
  for networked data,'' \emph{IEEE Signal Process. Mag.}, vol.~25, no.~2, pp.
  92--101, 2008.

\bibitem{spm02}
J.~L. Starck, E.~Pantin, and F.~Murtagh, ``Deconvolution in astronomy: A
  review,'' \emph{Publ. Astron. Soc. Pac.}, vol. 114, no. 800, pp. 1051--1069,
  2002.

\bibitem{tar81}
A.~Tarantola and B.~Valetta, ``Inverse problems = quest for information,''
  \emph{J. Geophys.}, vol.~50, no.~1, pp. 159--170, 1981.

\bibitem{fh13}
S.~Foucart and H.~Rauhut, \emph{A Mathematical Introduction to Compressive
  Sensing}, ser. Applied and Numerical Harmonic Analysis.\hskip 1em plus 0.5em
  minus 0.4em\relax Birkh{\"a}user Basel, 2013.

\bibitem{ksh12}
A.~Krizhevsky, I.~Sutskever, and G.~E. Hinton, ``{ImageNet} classification with
  deep convolutional neural networks,'' in \emph{Advances in Neural Information
  Processing Systems 25}, F.~Pereira, C.~J.~C. Burges, L.~Bottou, and K.~Q.
  Weinberger, Eds.\hskip 1em plus 0.5em minus 0.4em\relax Curran Associates,
  Inc., 2012, pp. 1097--1105.

\bibitem{lbh15}
Y.~LeCun, Y.~Bengio, and G.~Hinton, ``Deep learning,'' \emph{Nature}, vol. 521,
  no. 7553, pp. 436--444, 2015.

\bibitem{gbc16}
I.~Goodfellow, Y.~Bengio, and A.~Courville, \emph{Deep Learning}.\hskip 1em
  plus 0.5em minus 0.4em\relax MIT Press, 2016.

\bibitem{kl10}
K.~Gregor and Y.~LeCun, ``Learning fast approximations of sparse coding,'' in
  \emph{Proceedings of the 27th International Conference on International
  Conference on Machine Learning (ICML)}, J.~F{\"u}rnkranz and T.~Joachims,
  Eds., 2010, pp. 399--406.

\bibitem{yslx16}
Y.~Yang, J.~Sun, H.~Li, and Z.~Xu, ``Deep {ADMM-Net} for compressive sensing
  {MRI},'' in \emph{Advances in Neural Information Processing Systems 29},
  D.~D. Lee, M.~Sugiyama, U.~V. Luxburg, I.~Guyon, and R.~Garnett, Eds.\hskip
  1em plus 0.5em minus 0.4em\relax Curran Associates, Inc., 2016, pp. 10--18.

\bibitem{dlht16}
C.~{Dong}, C.~C. {Loy}, K.~{He}, and X.~{Tang}, ``Image super-resolution using
  deep convolutional networks,'' \emph{IEEE Trans. Pattern Anal. Mach.
  Intell.}, vol.~38, no.~2, pp. 295--307, 2016.

\bibitem{kmy17}
E.~Kang, J.~Min, and J.~C. Ye, ``A deep convolutional neural network using
  directional wavelets for low-dose {X-ray CT} reconstruction,'' \emph{Med.
  Phys.}, vol.~44, no.~10, pp. e360--e375, 2017.

\bibitem{jmfu17}
K.~H. Jin, M.~T. McCann, E.~Froustey, and M.~Unser, ``Deep convolutional neural
  network for inverse problems in imaging,'' \emph{IEEE Trans. Image Process.},
  vol.~26, no.~9, pp. 4509--4522, 2017.

\bibitem{ham+17}
K.~Hammernik, T.~Klatzer, E.~Kobler, M.~P. Recht, D.~K. Sodickson, T.~Pock, and
  F.~Knoll, ``Learning a variational network for reconstruction of accelerated
  {MRI} data,'' \emph{Magn. Reson. Med.}, vol.~79, no.~6, pp. 3055--3071, 2018.

\bibitem{che+17b}
H.~Chen, Y.~Zhang, W.~Zhang, P.~Liao, K.~Li, J.~Zhou, and G.~Wang, ``Low-dose
  {CT} via convolutional neural network,'' \emph{Biomed. Opt. Express}, vol.~8,
  no.~2, pp. 679--694, 2017.

\bibitem{bjpd17}
A.~Bora, A.~Jalal, E.~Price, and A.~G. Dimakis, ``Compressed sensing using
  generative models,'' in \emph{Proceedings of the 34th International
  Conference on Machine Learning (ICML)}, D.~Precup and Y.~W. Teh, Eds.,
  vol.~70, 2017, pp. 537--546.

\bibitem{zlcrr18}
B.~Zhu, J.~Z. Liu, S.~F. Cauley, B.~R. Rosen, and M.~S. Rosen, ``Image
  reconstruction by domain-transform manifold learning,'' \emph{Nature}, vol.
  555, no. 7697, pp. 487--492, 2018.

\bibitem{bub+19}
T.~A. Bubba, G.~Kutyniok, M.~Lassas, M.~M{\"a}rz, W.~Samek, S.~Siltanen, and
  V.~Srinivasan, ``Learning the invisible: A hybrid deep learning-shearlet
  framework for limited angle computed tomography,'' \emph{Inverse Probl.},
  vol.~35, no.~6, p. 064002, 2019.

\bibitem{amos19}
S.~Arridge, P.~Maass, O.~{\"O}ktem, and C.-B. Sch{\"o}nlieb, ``Solving inverse
  problems using data-driven models,'' \emph{Acta Numer.}, vol.~28, pp. 1--174,
  2019.

\bibitem{ela17}
M.~Elad, ``{Deep, Deep Trouble: Deep Learning's Impact on Image Processing,
  Mathematics, and Humanity},'' SIAM News, available online:
  \url{https://sinews.siam.org/Details-Page/deep-deep-trouble}, 2017.

\bibitem{che+17}
H.~Chen, Y.~Zhang, M.~K. Kalra, F.~Lin, Y.~Chen, P.~Liao, J.~Zhou, and G.~Wang,
  ``Low-dose {CT} with a residual encoder-decoder convolutional neural
  network,'' \emph{IEEE Trans. Med. Imag.}, vol.~36, no.~12, pp. 2524--2535,
  2017.

\bibitem{haao20}
A.~{Hauptmann}, J.~{Adler}, S.~R. {Arridge}, and O.~{{\"O}ktem}, ``Multi-scale
  learned iterative reconstruction,'' \emph{IEEE Trans. Comput. Imag.}, 2020,
  available online: \url{https://doi.org/10.1109/TCI.2020.2990299}.

\bibitem{hua+18}
Y.~Huang, T.~W{\"u}rfl, K.~Breininger, L.~Liu, G.~Lauritsch, and A.~Maier,
  ``Some investigations on robustness of deep learning in limited angle
  tomography,'' in \emph{Medical Image Computing and Computer Assisted
  Intervention -- MICCAI 2018}, A.~F. Frangi, J.~A. Schnabel, C.~Davatzikos,
  C.~Alberola-L{\'o}pez, and G.~Fichtinger, Eds.\hskip 1em plus 0.5em minus
  0.4em\relax Springer Cham, 2018, pp. 145--153.

\bibitem{arpah20}
V.~Antun, F.~Renna, C.~Poon, B.~Adcock, and A.~C. Hansen, ``On instabilities of
  deep learning in image reconstruction and the potential costs of {AI},''
  \emph{Proc. Natl. Acad. Sci.}, 2020, available online:
  \url{https://doi.org/10.1073/pnas.1907377117}.

\bibitem{gaah20}
N.~M. Gottschling, V.~Antun, B.~Adcock, and A.~C. Hansen, ``The troublesome
  kernel: why deep learning for inverse problems is typically unstable,'' 2020,
  preprint arXiv:2001.01258.

\bibitem{rbk20}
A.~Raj, Y.~Bresler, and B.~Li, ``Improving robustness of deep-learning-based
  image reconstruction,'' in \emph{Proceedings of the 37th International
  Conference on Machine Learning (ICML)}, H.~Daum{\'e} and A.~Singh, Eds.,
  2017.

\bibitem{sze+14}
C.~Szegedy, W.~Zaremba, I.~Sutskever, J.~Bruna, D.~Erhan, I.~Goodfellow, and
  R.~Fergus, ``Intriguing properties of neural networks,'' 2014, preprint
  arXiv:1312.6199.

\bibitem{kgb17}
A.~Kurakin, I.~Goodfellow, and S.~Bengio, ``Adversarial examples in the
  physical world,'' 2017, preprint arXiv:1607.02533.

\bibitem{eyk18}
K.~Eykholt, I.~Evtimov, E.~Fernandes, B.~Li, A.~Rahmati, C.~Xiao, A.~Prakash,
  T.~Kohno, and D.~Song, ``Robust physical-world attacks on deep learning
  visual classification,'' in \emph{Proceedings of the IEEE Conference on
  Computer Vision and Pattern Recognition (CVPR)}, 2018, pp. 1625--1634.

\bibitem{zbo+19}
J.~Zbontar, F.~Knoll, A.~Sriram, T.~Murrell, Z.~Huang, M.~J. Muckley,
  A.~Defazio, R.~Stern, P.~Johnson, M.~Bruno, M.~Parente, K.~J. Geras,
  J.~Katsnelson, H.~Chandarana, Z.~Zhang, M.~Drozdzal, A.~Romero, M.~Rabbat,
  P.~Vincent, N.~Yakubova, J.~Pinkerton, D.~Wang, E.~Owens, C.~L. Zitnick,
  M.~P. Recht, D.~K. Sodickson, and Y.~W. Lui, ``{fastMRI:} an open dataset and
  benchmarks for accelerated {MRI},'' 2018, preprint arXiv:1811.08839.

\bibitem{kno+20}
F.~Knoll, J.~Zbontar, A.~Sriram, M.~J. Muckley, M.~Bruno, A.~Defazio,
  M.~Parente, K.~J. Geras, J.~Katsnelson, H.~Chandarana, Z.~Zhang,
  M.~Drozdzalv, A.~Romero, M.~Rabbat, P.~Vincent, J.~Pinkerton, D.~Wang,
  N.~Yakubova, E.~Owens, C.~L. Zitnick, M.~P. Recht, D.~K. Sodickson, and Y.~W.
  Lui, ``{fastMRI:} a publicly available raw k-space and {DICOM} dataset of
  knee images for accelerated {MR} image reconstruction using machine
  learning,'' \emph{Radiology Artif. Intell.}, vol.~2, no.~1, p. e190007, 2020.

\bibitem{pas+17}
A.~Paszke, S.~Gross, S.~Chintala, G.~Chanan, E.~Yang, Z.~DeVito, Z.~Lin,
  A.~Desmaison, L.~Antiga, and A.~Lerer, ``Automatic differentiation in
  {PyTorch},'' Contribution to the NIPS 2017 Autodiff Workshop, available
  online: \url{https://openreview.net/forum?id=BJJsrmfCZ}, 2017.

\bibitem{car20}
N.~Carlini, ``{A Complete List of All (arXiv) Adversarial Example Papers},''
  available online:
  \url{https://nicholas.carlini.com/writing/2019/all-adversarial-example-papers.html},
  accessed on 2020-11-02, 2020.

\bibitem{yhzl19}
X.~Yuan, P.~He, Q.~Zhu, and X.~Li, ``Adversarial examples: Attacks and defenses
  for deep learning,'' \emph{IEEE Trans. Neural Netw. Learn. Syst.}, vol.~30,
  no.~9, pp. 2805--2824, 2019.

\bibitem{ommf20}
G.~Ortiz-Jimenez, A.~Modas, S.-M. Moosavi-Dezfooli, and P.~Frossard, ``Optimism
  in the face of adversity: Understanding and improving deep learning through
  adversarial robustness,'' 2020, preprint arXiv:2010.09624.

\bibitem{amt20}
A.~Arnab, O.~Miksik, and P.~H. Torr, ``On the robustness of semantic
  segmentation models to adversarial attacks,'' \emph{IEEE Trans. Pattern Anal.
  Mach. Intell.}, vol.~42, no.~12, pp. 3040--3053, 2020.

\bibitem{qui93}
E.~T. Quinto, ``{Singularities of the X-Ray Transform and Limited Data
  Tomography in $\mathbb{R}^2 $ and $\mathbb{R}^3$},'' \emph{SIAM J. Math.
  Anal.}, vol.~24, no.~5, pp. 1215--1225, 1993.

\bibitem{schpr17}
J.~Schlemper, J.~Caballero, J.~V. Hajnal, A.~N. Price, and D.~Rueckert, ``A
  deep cascade of convolutional neural networks for dynamic mr image
  reconstruction,'' \emph{IEEE Trans. Med. Imag.}, vol.~37, no.~2, pp.
  491--503, 2017.

\bibitem{kekp20}
E.~Kobler, A.~Effland, K.~Kunisch, and T.~Pock, ``Total deep variation: A
  stable regularizer for inverse problems,'' 2020, arXiv: 2006.08789.

\bibitem{ojbmdw20}
G.~Ongie, A.~Jalal, R.~G. Baraniuk, C.~A. Metzler, A.~G. Dimakis, and
  R.~Willett, ``Deep learning techniques for inverse problems in imaging,''
  \emph{IEEE J. Sel. Areas Inf. Theory}, vol.~1, no.~1, pp. 39--56, 2020.

\bibitem{rfb15}
O.~Ronneberger, P.~Fischer, and T.~Brox, ``{U-Net:} convolutional networks for
  biomedical image segmentation,'' in \emph{Medical Image Computing and
  Computer Assisted Intervention -- MICCAI 2015}, N.~Navab, J.~Hornegger, W.~M.
  Wells, and A.~F. Frangi, Eds.\hskip 1em plus 0.5em minus 0.4em\relax Springer
  Cham, 2015, pp. 234--241.

\bibitem{hzrs16}
K.~He, X.~Zhang, S.~Ren, and J.~Sun, ``Deep residual learning for image
  recognition,'' in \emph{Proceedings of the IEEE Conference on Computer Vision
  and Pattern Recognition (CVPR)}, 2016, pp. 770--778.

\bibitem{wlvi17}
J.~M. Wolterink, T.~Leiner, M.~A. Viergever, and I.~I{\v{s}}gum, ``Generative
  adversarial networks for noise reduction in low-dose {CT},'' \emph{IEEE
  Trans. Med. Imag.}, vol.~36, no.~12, pp. 2536--2545, 2017.

\bibitem{yan+18}
Q.~Yang, P.~Yan, Y.~Zhang, H.~Yu, Y.~Shi, X.~Mou, M.~K. Kalra, Y.~Zhang,
  L.~Sun, and G.~Wang, ``Low-dose {CT} image denoising using a generative
  adversarial network with {W}asserstein distance and perceptual loss,''
  \emph{IEEE Trans. Med. Imag.}, vol.~37, no.~6, pp. 1348--1357, 2018.

\bibitem{jdvrb17}
S.~J{\'e}gou, M.~Drozdzal, D.~Vazquez, A.~Romero, and Y.~Bengio, ``The one
  hundred layers tiramisu: Fully convolutional densenets for semantic
  segmentation,'' in \emph{Proceedings of the IEEE Conference on Computer
  Vision and Pattern Recognition (CVPR)}, 2017, pp. 11--19.

\bibitem{hlvw17}
G.~Huang, Z.~Liu, L.~Van Der~Maaten, and K.~Q. Weinberger, ``Densely connected
  convolutional networks,'' in \emph{Proceedings of the IEEE Conference on
  Computer Vision and Pattern Recognition (CVPR)}, 2017, pp. 4700--4708.

\bibitem{sch+19}
J.~Schlemper, I.~Oksuz, J.~R. Clough, J.~Duan, A.~P. King, J.~A. Schnabel,
  J.~V. Hajnal, and D.~Rueckert, ``{dAUTOMAP:} decomposing {AUTOMAP} to achieve
  scalability and enhance performance,'' 2019, preprint arXiv:1909.10995.

\bibitem{amj18}
H.~K. Aggarwal, M.~P. Mani, and M.~Jacob, ``{MoDL:} model-based deep learning
  architecture for inverse problems,'' \emph{IEEE Trans. Med. Imag.}, vol.~38,
  no.~2, pp. 394--405, 2018.

\bibitem{ao18}
J.~Adler and O.~{\"O}ktem, ``Learned primal-dual reconstruction,'' \emph{IEEE
  Trans. Med. Imag.}, vol.~37, no.~6, pp. 1322--1332, 2018.

\bibitem{hsqdsr19}
K.~Hammernik, J.~Schlemper, C.~Qin, J.~Duan, R.~M. Summers, and D.~Rueckert,
  ``$\sigma$-net: Systematic evaluation of iterative deep neural networks for
  fast parallel {MR Image} reconstruction,'' 2019, preprint arXiv:1912.09278.

\bibitem{chlf+20}
I.~Y. {Chun}, Z.~{Huang}, H.~{Lim}, and J.~{Fessler}, ``{Momentum-Net:} fast
  and convergent iterative neural network for inverse problems,'' \emph{IEEE
  Trans. Pattern Anal. Mach. Intell.}, 2020, available online:
  \url{https://doi.org/10.1109/TPAMI.2020.3012955}.

\bibitem{kb14}
D.~P. Kingma and J.~Ba, ``{Adam:} a method for stochastic optimization,'' 2014,
  preprint arXiv:1412.6980.

\bibitem{sd91}
J.~Sietsma and R.~J. Dow, ``Creating artificial neural networks that
  generalize,'' \emph{Neural Netw.}, vol.~4, no.~1, pp. 67--79, 1991.

\bibitem{hk92}
L.~Holmstrom and P.~Koistinen, ``Using additive noise in back-propagation
  training,'' \emph{IEEE Trans. Neural Netw.}, vol.~3, no.~1, pp. 24--38, 1992.

\bibitem{bis95}
C.~M. Bishop, ``Training with noise is equivalent to {Tikhonov}
  regularization,'' \emph{Neural Comput.}, vol.~7, no.~1, pp. 108--116, 1995.

\bibitem{vllbm10}
P.~Vincent, H.~Larochelle, I.~Lajoie, Y.~Bengio, and P.-A. Manzagol, ``Stacked
  denoising autoencoders: Learning useful representations in a deep network
  with a local denoising criterion,'' \emph{J. Mach. Learn. Res.}, vol.~11, pp.
  3371--3408, 2010.

\bibitem{rof92}
L.~I. Rudin, S.~Osher, and E.~Fatemi, ``{Nonlinear total variation based noise
  removal algorithms},'' \emph{Physica D: Nonlinear Phenomena}, vol.~60, no.
  1--4, pp. 259--268, 1992.

\bibitem{cl97}
A.~Chambolle and P.-L. Lions, ``Image recovery via total variation minimization
  and related problems,'' \emph{Numer. Math.}, vol.~76, no.~2, pp. 167--188,
  1997.

\bibitem{bb18}
M.~Benning and M.~Burger, ``Modern regularization methods for inverse
  problems,'' \emph{Acta Numer.}, vol.~27, pp. 1--111, 2018.

\bibitem{crt06a}
E.~J. Cand{\`e}s, J.~K. Romberg, and T.~Tao, ``Robust uncertainty principles:
  exact signal reconstruction from highly incomplete frequency information,''
  \emph{IEEE Trans. Inf. Theory}, vol.~52, no.~2, pp. 489--509, 2006.

\bibitem{nw13}
D.~{Needell} and R.~{Ward}, ``Near-optimal compressed sensing guarantees for
  total variation minimization,'' \emph{IEEE Trans. Imag. Proc.}, vol.~22,
  no.~10, pp. 3941--3949, 2013.

\bibitem{poo15}
C.~Poon, ``On the role of total variation in compressed sensing,'' \emph{SIAM
  J. Imag. Sci.}, vol.~8, no.~1, pp. 682--720, 2015.

\bibitem{gms20}
M.~Genzel, M.~M{\"a}rz, and R.~Seidel, ``Compressed sensing with {1D} total
  variation: Breaking sample complexity barriers via non-uniform recovery,''
  2020, preprint arXiv:2001.09952.

\bibitem{gm75}
R.~Glowinski and A.~Marroco, ``Sur l'approximation, par {\'e}l{\'e}ments finis
  d'ordre un, et la r{\'e}solution, par p{\'e}nalisation-dualit{\'e} d'une
  classe de probl{\`e}mes de {D}irichlet non lin{\'e}aires,'' \emph{RAIRO Anal.
  Numer.}, vol.~9, no.~R2, pp. 41--76, 1975.

\bibitem{gm76}
D.~Gabay and B.~Mercier, ``A dual algorithm for the solution of nonlinear
  variational problems via finite element approximation,'' \emph{Comput. Math.
  Appl.}, vol.~2, no.~1, pp. 17--40, 1976.

\bibitem{cw17}
N.~Carlini and D.~Wagner, ``Towards evaluating the robustness of neural
  networks,'' in \emph{IEEE Symposium on Security and Privacy}, 2017, pp.
  39--57.

\bibitem{aabbdk19}
A.~Agrawal, B.~Amos, S.~Barratt, S.~Boyd, S.~Diamond, and J.~Z. Kolter,
  ``Differentiable convex optimization layers,'' in \emph{Advances in Neural
  Information Processing Systems 32}, H.~Wallach, H.~Larochelle,
  A.~Beygelzimer, F.~d'Alch\'{e} Buc, E.~Fox, and R.~Garnett, Eds.\hskip 1em
  plus 0.5em minus 0.4em\relax Curran Associates, Inc., 2019, pp. 9562--9574.

\bibitem{lbbh98}
Y.~LeCun, L.~Bottou, Y.~Bengio, and P.~Haffner, ``Gradient-based learning
  applied to document recognition,'' \emph{Proc. IEEE}, vol.~86, no.~11, pp.
  2278--2324, 1998.

\bibitem{almt14}
D.~Amelunxen, M.~Lotz, M.~B. McCoy, and J.~A. Tropp, ``Living on the edge:
  phase transitions in convex programs with random data,'' \emph{Inf.
  Inference}, vol.~3, no.~3, pp. 224--294, 2014.

\bibitem{ao17}
J.~Adler and O.~{\"O}ktem, ``Solving ill-posed inverse problems using iterative
  deep neural networks,'' \emph{Inverse Probl.}, vol.~33, no.~12, p. 124007,
  2017.

\bibitem{sp08}
E.~Y. Sidky and X.~Pan, ``Image reconstruction in circular cone-beam computed
  tomography by constrained, total-variation minimization,'' \emph{Phys. Med.
  Biol.}, vol.~53, no.~17, pp. 4777--4807, 2008.

\bibitem{wbss04}
Z.~Wang, A.~C. Bovik, H.~R. Sheikh, and E.~P. Simoncelli, ``Image quality
  assessment: from error visibility to structural similarity,'' \emph{IEEE
  Trans. Image Process.}, vol.~13, no.~4, pp. 600--612, 2004.

\bibitem{ern20}
P.~Ernst, ``Pytorch implementation of scikit-image's radon function, {version
  0.1.4},'' available online: \url{https://github.com/phernst/pytorch_radon},
  2020.

\bibitem{pmg16}
N.~Papernot, P.~McDaniel, and I.~Goodfellow, ``Transferability in machine
  learning: from phenomena to black-box attacks using adversarial samples,''
  2016, preprint arXiv:1605.07277.

\bibitem{zgfk17}
H.~{Zhao}, O.~{Gallo}, I.~{Frosio}, and J.~{Kautz}, ``Loss functions for image
  restoration with neural networks,'' \emph{IEEE Trans. Comput. Imag.}, vol.~3,
  no.~1, pp. 47--57, 2017.

\bibitem{ks06}
J.~Kaipio and E.~Somersalo, \emph{Statistical and computational inverse
  problems}, ser. Applied Mathematical Sciences.\hskip 1em plus 0.5em minus
  0.4em\relax Springer New York, 2006, vol. 160.

\bibitem{ms12}
J.~L. Mueller and S.~Siltanen, \emph{Linear and nonlinear inverse problems with
  practical applications}.\hskip 1em plus 0.5em minus 0.4em\relax SIAM, 2012.

\bibitem{bmr17}
A.~S. Bandeira, D.~G. Mixon, and B.~Recht, ``Compressive classification and the
  rare eclipse problem,'' in \emph{Compressed Sensing and its Applications:
  Second International MATHEON Conference 2015}, ser. Applied and Numerical
  Harmonic Analysis, H.~Boche, G.~Caire, R.~Calderbank, M.~M{\"a}rz,
  G.~Kutyniok, and R.~Mathar, Eds.\hskip 1em plus 0.5em minus 0.4em\relax
  Springer Cham, 2017, pp. 197--220.

\bibitem{bcckw16}
C.~Boyer, N.~Chauffert, P.~Ciuciu, J.~Kahn, and P.~Weiss, ``On the generation
  of sampling schemes for magnetic resonance imaging,'' \emph{SIAM J. Imaging
  Sci.}, vol.~9, no.~4, pp. 2039--2072, 2016.

\bibitem{msg20}
G.~R. Machado, E.~Silva, and R.~R. Goldschmidt, ``Adversarial machine learning
  in image classification: A survey towards the defender's perspective,'' 2020,
  preprint arXiv:2009.03728.

\bibitem{uvl18}
D.~Ulyanov, A.~Vedaldi, and V.~Lempitsky, ``Deep image prior,'' in
  \emph{Proceedings of the IEEE Conference on Computer Vision and Pattern
  Recognition (CVPR)}, 2018, pp. 9446--9454.

\bibitem{lsah20}
H.~Li, J.~Schwab, S.~Antholzer, and M.~Haltmeier, ``{NETT:} solving inverse
  problems with deep neural networks,'' \emph{Inverse Probl.}, vol.~36, no.~6,
  p. 065005, 2020.

\bibitem{kno+20b}
F.~Knoll, T.~Murrell, A.~Sriram, N.~Yakubova, J.~Zbontar, M.~Rabbat,
  A.~Defazio, M.~J. Muckley, D.~K. Sodickson, C.~L. Zitnick, and M.~P. Recht,
  ``Advancing machine learning for {MR} image reconstruction with an open
  competition: Overview of the 2019 {fastMRI} challenge,'' \emph{Magn. Reson.
  Med.}, vol.~84, no.~6, pp. 3054--3070, 2020.

\bibitem{ccshmp20}
K.~Cheng, F.~Caliv\'a, R.~Shah, M.~Han, S.~Majumdar, and V.~Pedoia,
  ``Addressing the false negative problem of deep learning {MRI} reconstruction
  models by adversarial attacks and robust training,'' in \emph{Proceedings of
  the 3rd Conference on Medical Imaging with Deep Learning (MIDL)}, T.~Arbel,
  I.~B. Ayed, M.~de~Bruijne, M.~Descoteaux, H.~Lombaert, and C.~Pal, Eds., vol.
  121, 2020, pp. 121--135.

\end{thebibliography}
\clearpage

\onecolumn
\setcounter{section}{0}
\renewcommand*{\thesection}{S\arabic{section}}
\setcounter{table}{0}
\renewcommand*{\thetable}{S\arabic{table}}
\setcounter{figure}{0}
\renewcommand*{\thefigure}{S\arabic{figure}}

\raggedbottom

\begin{center}
	\Huge\sffamily Supplementary Material
\end{center}

\par\noindent\hfill\IEEEcompsocdiamondline\hfill\hbox{}\par

\vspace{\baselineskip}

\noindent
The supplementary material is organized as follows:
\begin{itemize}
	\item
	Section~\ref{sec:supp:csA}--\ref{sec:supp:csC} contain supplementary results for Case Study~A--C, respectively.
	\item
	Section~\ref{sec:supp:additional} contains supplementary results for Section~\ref{sec:additional}.
	\item
	Section~\ref{sec:supp:hyperparam} provides an overview of all empirically selected hyper-parameters for the considered network architectures, training processes, and adversarial attacks.
\end{itemize}

\section{Supplementary Results for Case Study~A (CS With Gaussian Measurements)}
\label{sec:supp:csA}

\begin{table}[H]
	\robustify\bfseries
	\begin{tabular}{llS[separate-uncertainty]S[separate-uncertainty]S[separate-uncertainty]S[separate-uncertainty]S[separate-uncertainty]S[separate-uncertainty]S[separate-uncertainty]}
\toprule
\multicolumn{2}{l}{rel.~noise -- adversarial} &                  {0.0\%} &                  {0.1\%} &                  {0.5\%} &                  {1.0\%} &                  {2.0\%} &                   {4.0\%} &                   {6.0\%} \\
\midrule
TV$[\eta]$ & rel.~$\l{2}$-err. [\%] &  \bfseries 0.00 \pm 0.00 &  \bfseries 0.32 \pm 0.08 &            1.66 \pm 0.42 &            3.36 \pm 0.86 &            6.63 \pm 1.67 &            12.26 \pm 2.57 &            17.21 \pm 2.98 \\
UNet & rel.~$\l{2}$-err. [\%] &            2.53 \pm 1.97 &            2.67 \pm 2.01 &            3.36 \pm 2.11 &            4.49 \pm 2.31 &            7.29 \pm 2.52 &            13.15 \pm 3.24 &            18.27 \pm 3.58 \\
UNetFL & rel.~$\l{2}$-err. [\%] &            2.01 \pm 1.70 &            2.14 \pm 1.73 &            2.82 \pm 1.84 &            3.95 \pm 2.01 &            6.46 \pm 2.21 &            11.91 \pm 2.54 &            16.98 \pm 2.72 \\
Tira & rel.~$\l{2}$-err. [\%] &            1.22 \pm 1.15 &            1.33 \pm 1.18 &            1.95 \pm 1.33 &            3.05 \pm 1.64 &            5.90 \pm 2.23 &            11.97 \pm 3.13 &            17.18 \pm 3.27 \\
TiraFL & rel.~$\l{2}$-err. [\%] &            0.98 \pm 0.88 &            1.10 \pm 0.92 &            1.73 \pm 1.15 &            2.74 \pm 1.42 &            5.32 \pm 1.89 &  \bfseries 11.07 \pm 2.82 &            16.43 \pm 3.42 \\
ItNet & rel.~$\l{2}$-err. [\%] &            0.45 \pm 0.18 &            0.52 \pm 0.18 &  \bfseries 0.93 \pm 0.17 &  \bfseries 1.80 \pm 0.80 &  \bfseries 4.50 \pm 1.90 &            11.62 \pm 3.67 &  \bfseries 16.42 \pm 3.70 \\
\bottomrule
\end{tabular}

	\caption{\textbf{Scenario~\refAone{} -- CS with 1D signals.} A numerical representation of the results of Fig.~\ref{fig:tvsynth:table}(c), including the additional methods $\UNetFL$ and $\Tira$. The smallest relative error per noise level is highlighted in bold.}
	\label{tab:tvsynth:table_adv}
\end{table}

\begin{table}[H]
	\robustify\bfseries
	\begin{tabular}{llS[separate-uncertainty]S[separate-uncertainty]S[separate-uncertainty]S[separate-uncertainty]S[separate-uncertainty]S[separate-uncertainty]S[separate-uncertainty]}
\toprule
\multicolumn{2}{l}{rel.~noise -- Gaussian}  &                  {0.0\%} &                  {0.1\%} &                  {0.5\%} &                  {1.0\%} &                  {2.0\%} &                  {4.0\%} &                  {6.0\%} \\
\midrule
TV$[\eta]$ & rel.~$\l{2}$-err. [\%] &  \bfseries 0.00 \pm 0.00 &  \bfseries 0.18 \pm 0.06 &            0.90 \pm 0.30 &            1.79 \pm 0.60 &            3.61 \pm 1.18 &            6.84 \pm 2.15 &            9.74 \pm 2.64 \\
UNet & rel.~$\l{2}$-err. [\%] &            2.53 \pm 1.97 &            2.58 \pm 1.99 &            2.79 \pm 2.03 &            3.09 \pm 2.07 &            3.82 \pm 2.11 &            5.67 \pm 2.25 &            7.70 \pm 2.49 \\
UNetFL & rel.~$\l{2}$-err. [\%] &            2.01 \pm 1.70 &            2.05 \pm 1.71 &            2.24 \pm 1.75 &            2.53 \pm 1.79 &            3.23 \pm 1.91 &            4.99 \pm 2.05 &            7.03 \pm 2.21 \\
Tira & rel.~$\l{2}$-err. [\%] &            1.22 \pm 1.15 &            1.26 \pm 1.16 &            1.43 \pm 1.21 &            1.71 \pm 1.25 &            2.35 \pm 1.43 &            4.15 \pm 1.79 &            6.42 \pm 2.23 \\
TiraFL & rel.~$\l{2}$-err. [\%] &            0.98 \pm 0.88 &            1.02 \pm 0.90 &            1.20 \pm 0.95 &            1.48 \pm 1.05 &            2.10 \pm 1.24 &            3.86 \pm 1.63 &            5.98 \pm 2.19 \\
ItNet & rel.~$\l{2}$-err. [\%] &            0.45 \pm 0.18 &            0.47 \pm 0.18 &  \bfseries 0.59 \pm 0.17 &  \bfseries 0.80 \pm 0.17 &  \bfseries 1.38 \pm 0.50 &  \bfseries 2.91 \pm 0.99 &  \bfseries 5.33 \pm 2.15 \\
\bottomrule
\end{tabular}

	\caption{\textbf{Scenario~\refAone{} -- CS with 1D signals.} A numerical representation of the results of Fig.~\ref{fig:tvsynth:table}(d), including the additional methods $\UNetFL$ and $\Tira$. The smallest relative error per noise level is highlighted in bold.}
	\label{tab:tvsynth:table_ref}
\end{table}

\begin{figure}[H]
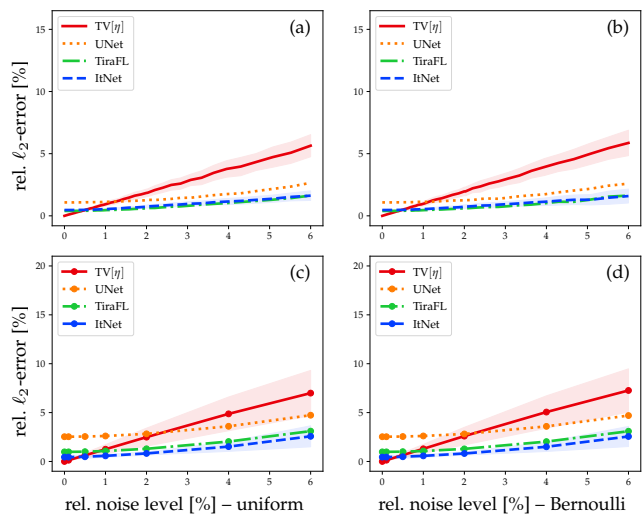

	\centering
	\scriptsize
	\setlength\tabcolsep{0pt}
	\begin{tabular}{cc@{\,\,}c}
		\rotatebox[origin=c]{90}{rel.~$\l{2}$-error [\%]} &
		\graphicwithlabel{tvsynth/results/attacks/fig_example_S6_uniform_curve.pdf}{.22\textwidth}{(a)}{.9}{.9} &
		\graphicwithlabel{tvsynth/results/attacks/fig_example_S6_bernoulli_curve.pdf}{.22\textwidth}{(b)}{.9}{.9} \\
		\rotatebox[origin=c]{90}{rel.~$\l{2}$-error [\%]} &
		\graphicwithlabel{tvsynth/results/attacks/fig_table_uniform.pdf}{.22\textwidth}{(c)}{.9}{.9} &
		\graphicwithlabel{tvsynth/results/attacks/fig_table_bernoulli.pdf}{.22\textwidth}{(d)}{.9}{.9} \\
		& rel.~noise level [\%] -- uniform & rel.~noise level [\%] -- Bernoulli
	\end{tabular}
	\caption{\textbf{Scenario~\refAone{} -- CS with 1D signals.} (a) and (b)~show uniform and Bernoulli noise-to-error curves, respectively, for the signal of Fig.~\ref{fig:tvsynth:example_adv}. In the latter case, we have generated symmetrized Bernoulli noise with $p = 0.025$. The mean and standard deviation are computed over 200~draws of~$\Noise$. (c)~and (d)~display the respective curves averaged over 50~signals from the test set. For the sake of clarity, we have omitted the standard deviations for $\UNet$ and $\TiraFL$, which behave similarly.}
	\label{fig:tvsynth:table_2}
\end{figure}

\begin{figure}[H]
	\centering
	\scriptsize
	\begin{tabular}{l@{\,}c@{\,}c@{\,}c@{\,}c}
	 & 0.5\% rel.~noise -- Gaussian & 2\% rel.~noise -- Gaussian & 6\% rel.~noise -- Gaussian & 12\% rel.~noise -- Gaussian \\
	\rotatebox[origin=c]{90}{$\TV[\noisebnd]$} &
	\includegraphics[valign=c,width=0.24\textwidth]{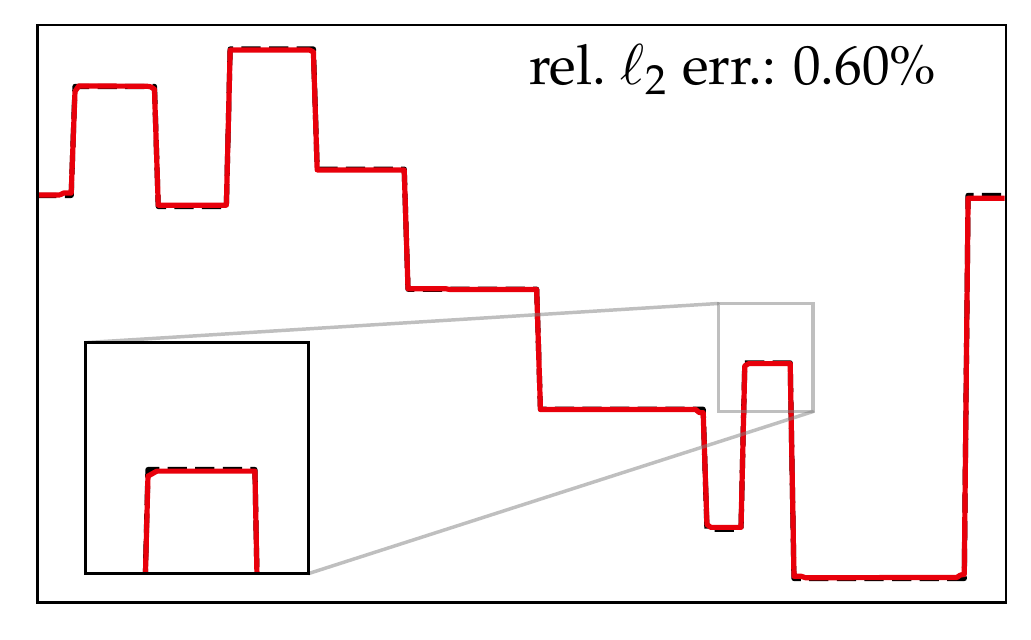} &
	\includegraphics[valign=c,width=0.24\textwidth]{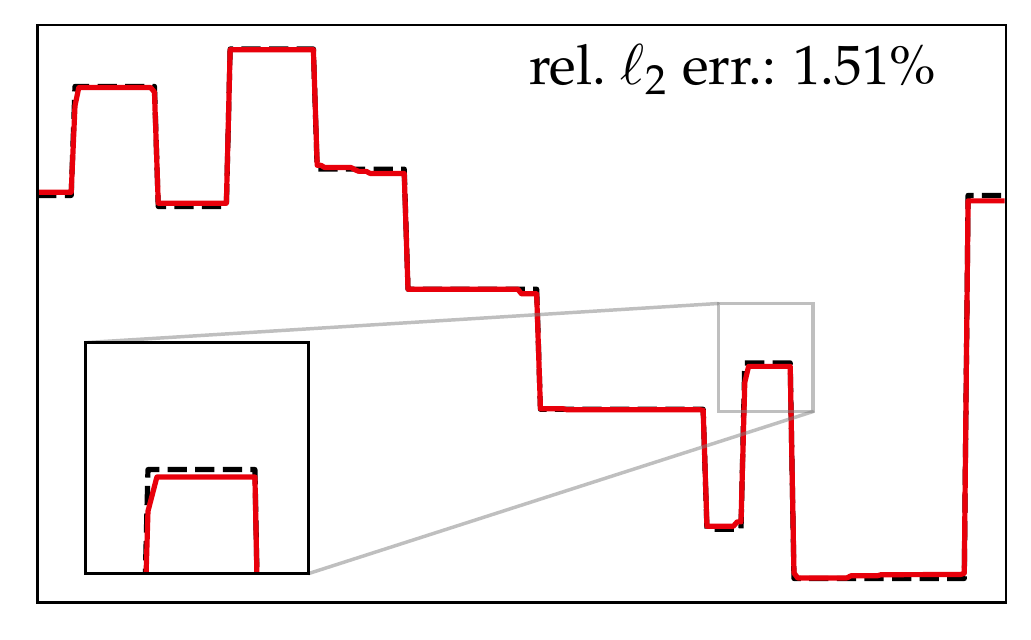} & \includegraphics[valign=c,width=0.24\textwidth]{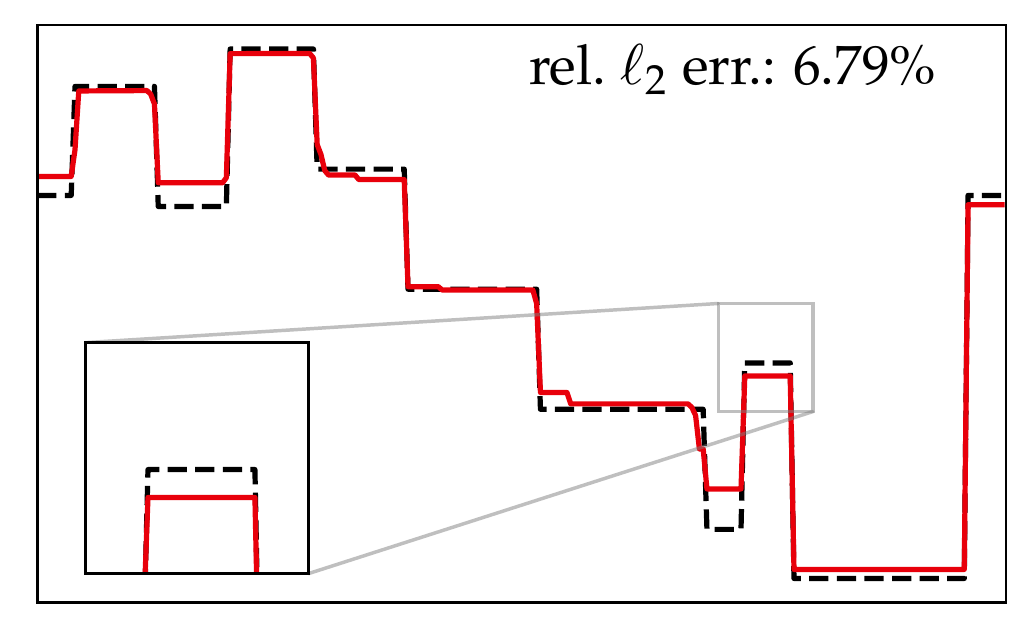} & \includegraphics[valign=c,width=0.24\textwidth]{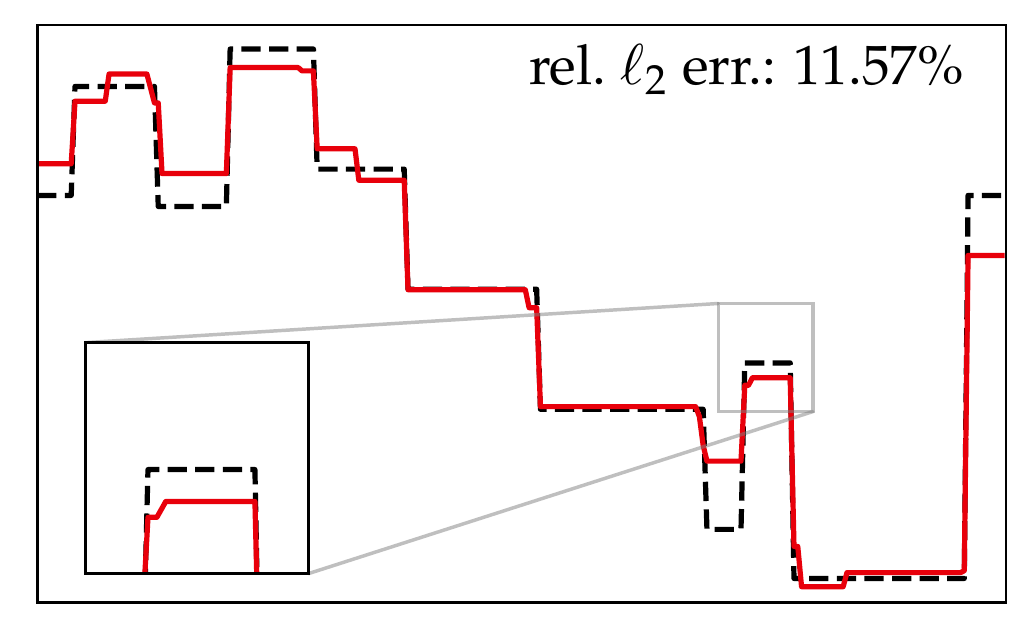} \\
	\rotatebox[origin=c]{90}{$\UNet$} &
	\includegraphics[valign=c,width=0.24\textwidth]{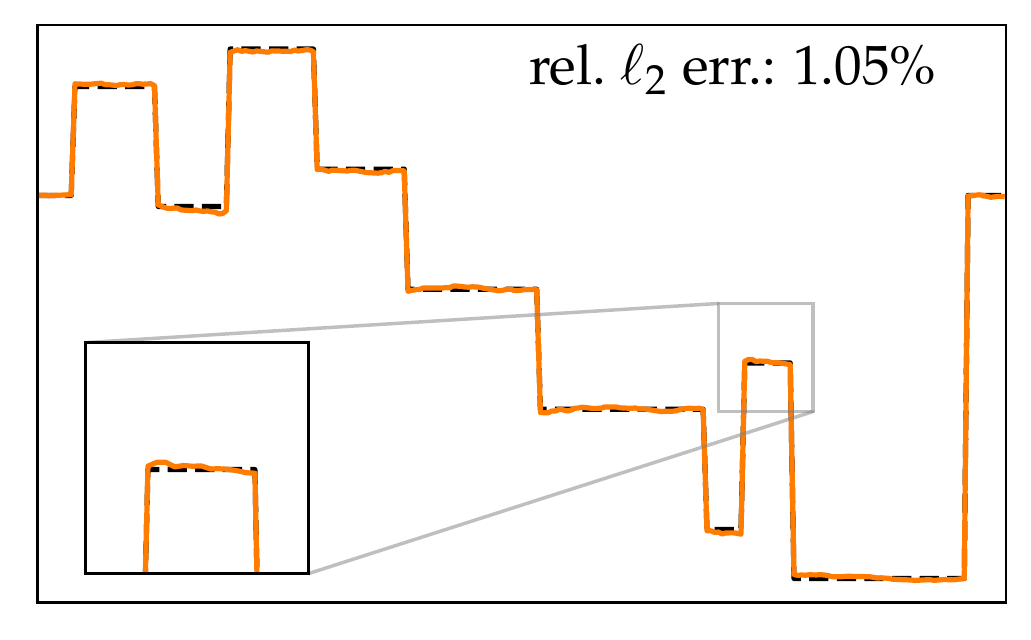} &
	\includegraphics[valign=c,width=0.24\textwidth]{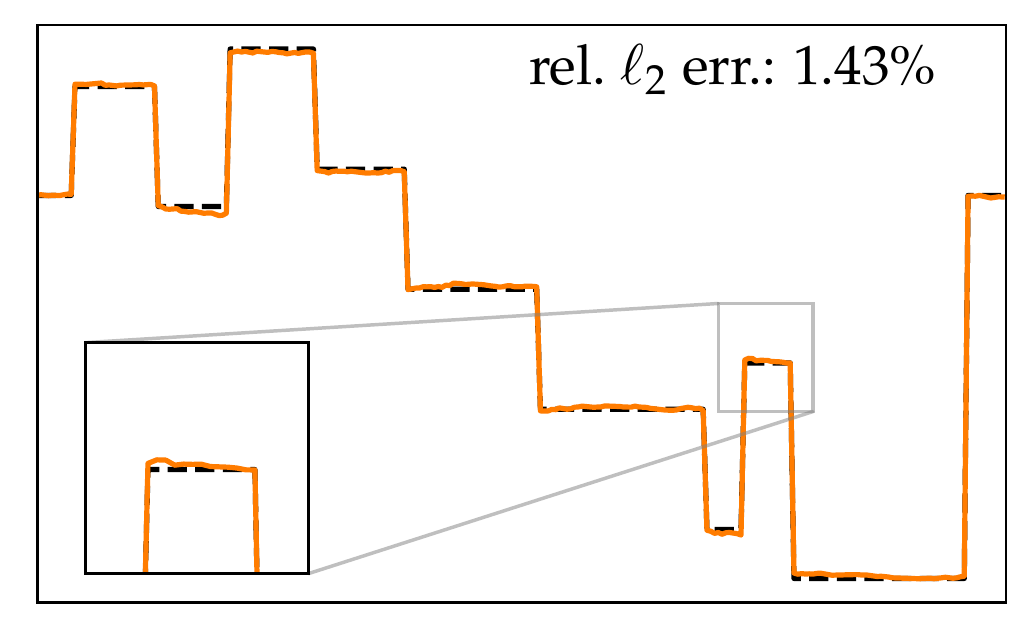} & \includegraphics[valign=c,width=0.24\textwidth]{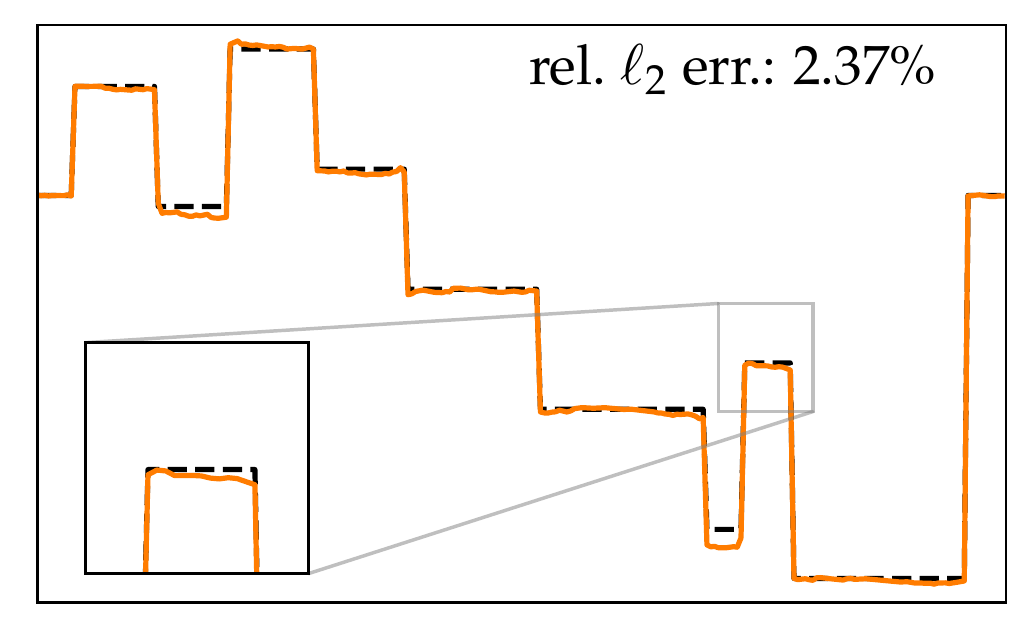} & \includegraphics[valign=c,width=0.24\textwidth]{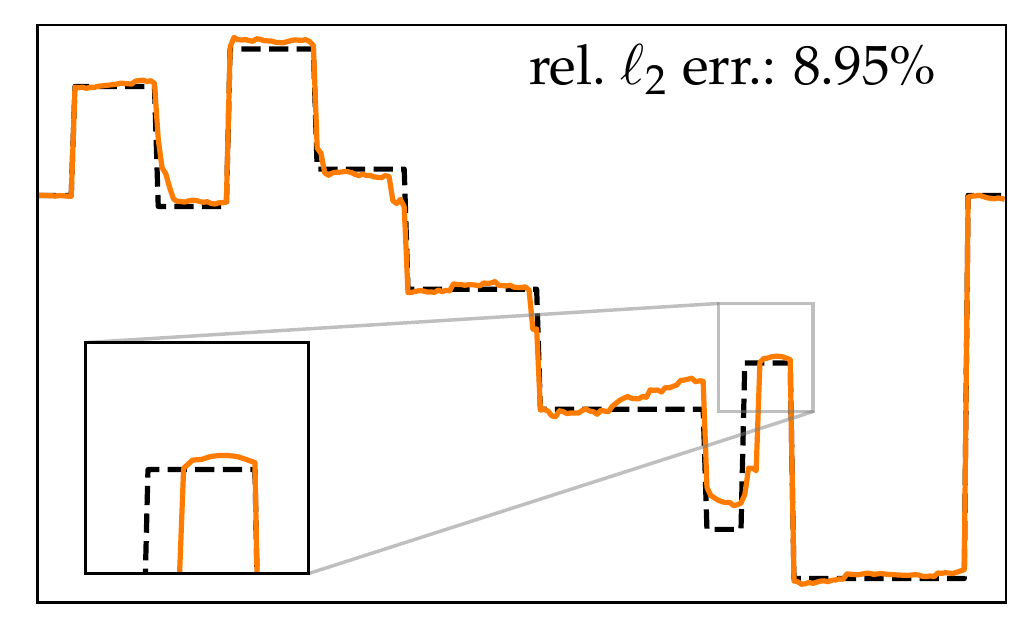} \\
	\rotatebox[origin=c]{90}{$\TiraFL$} &
	\includegraphics[valign=c,width=0.24\textwidth]{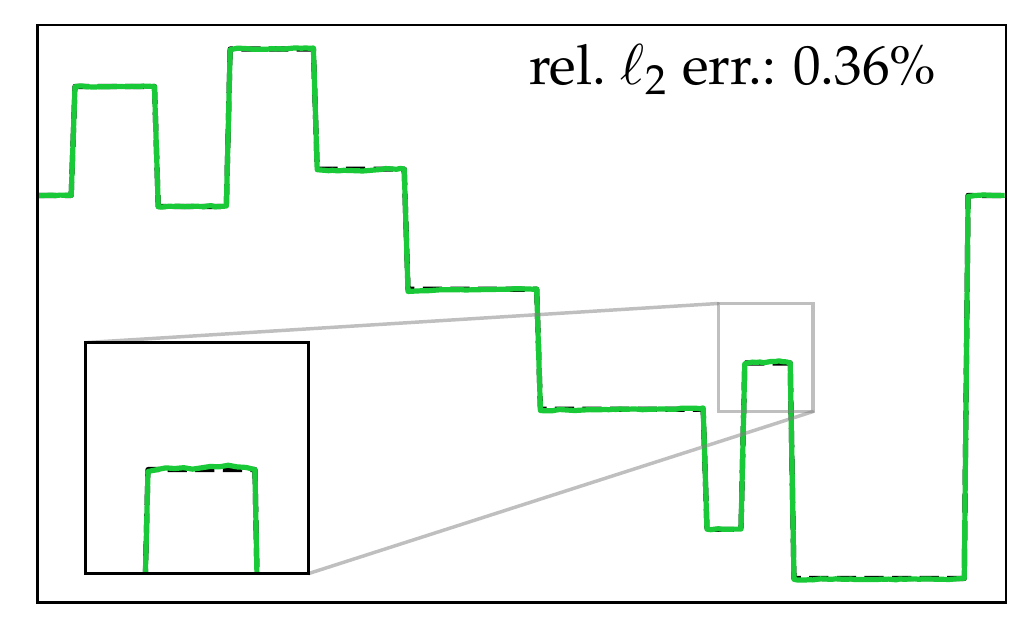} &
	\includegraphics[valign=c,width=0.24\textwidth]{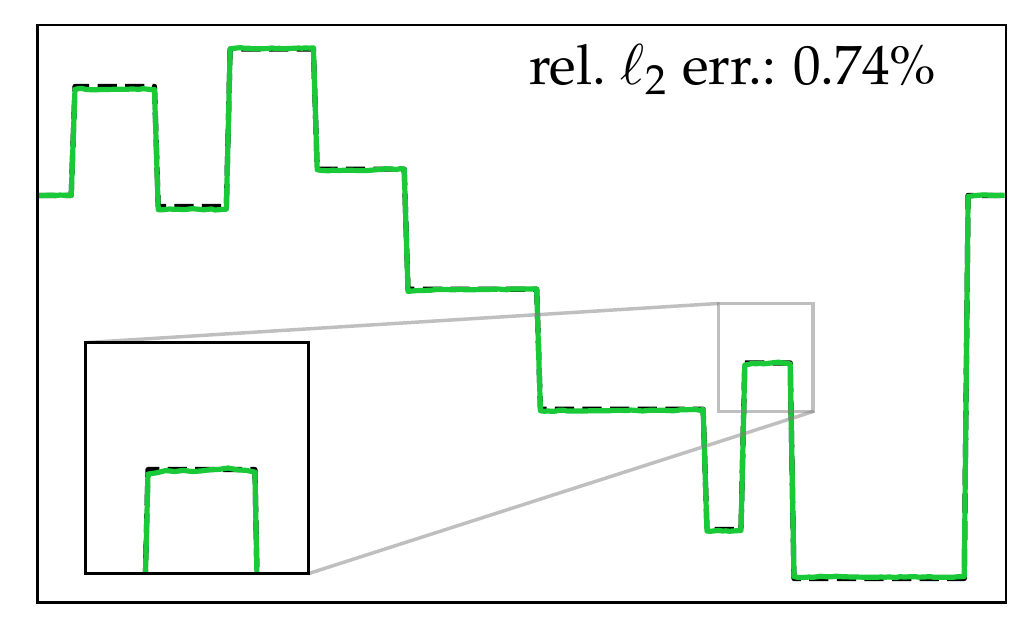} & \includegraphics[valign=c,width=0.24\textwidth]{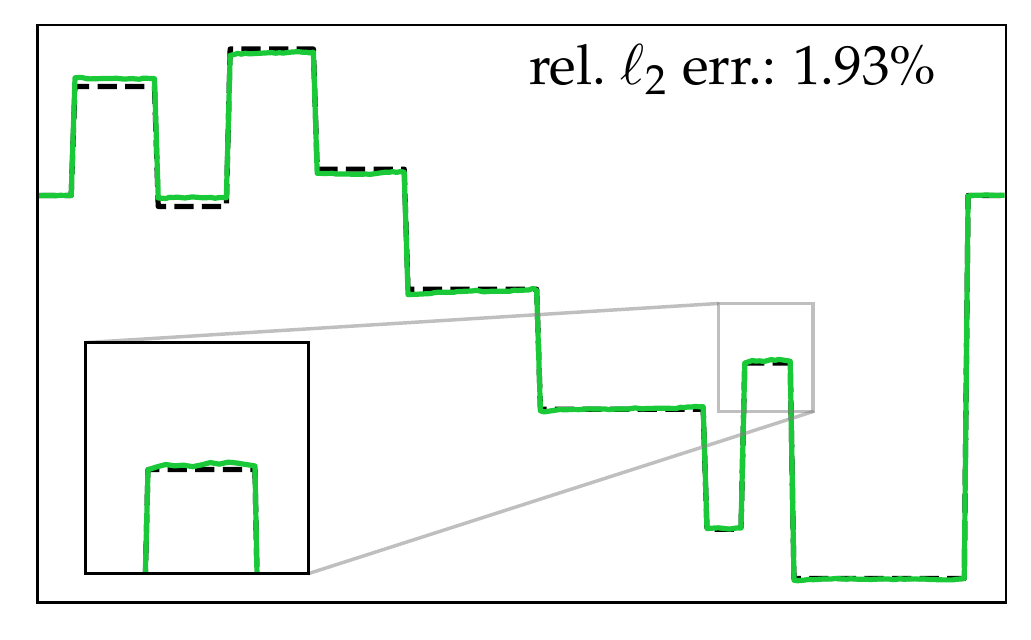} & \includegraphics[valign=c,width=0.24\textwidth]{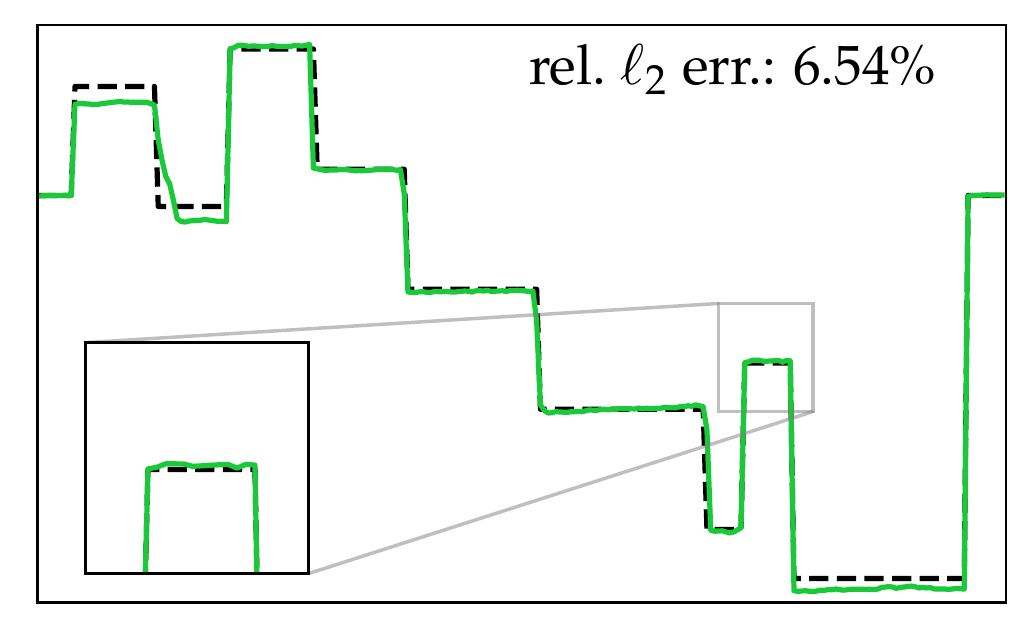} \\
	\rotatebox[origin=c]{90}{$\ItNet$} &
	\includegraphics[valign=c,width=0.24\textwidth]{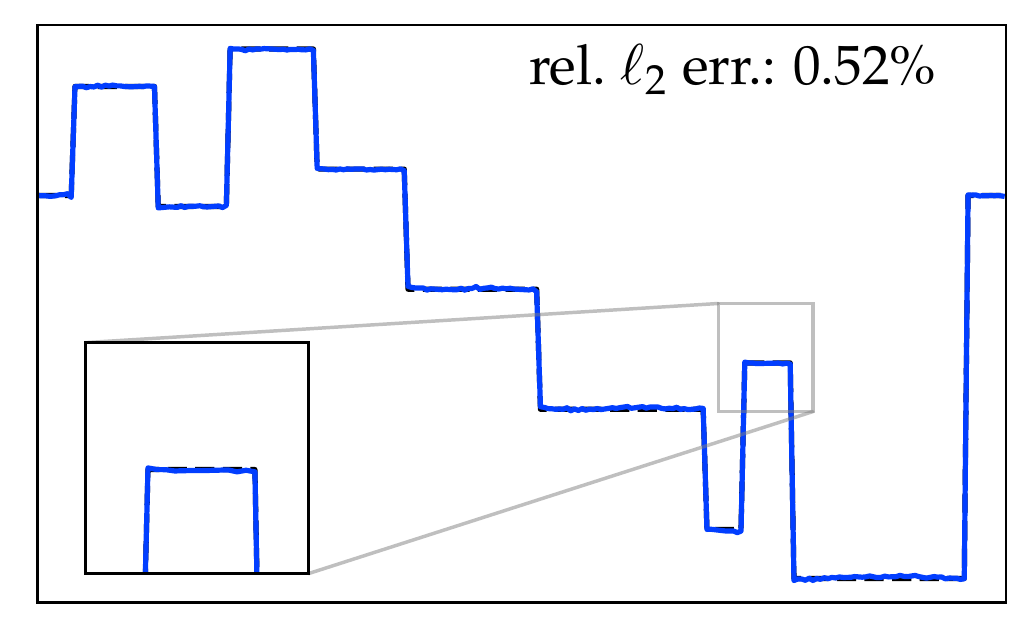} &
	\includegraphics[valign=c,width=0.24\textwidth]{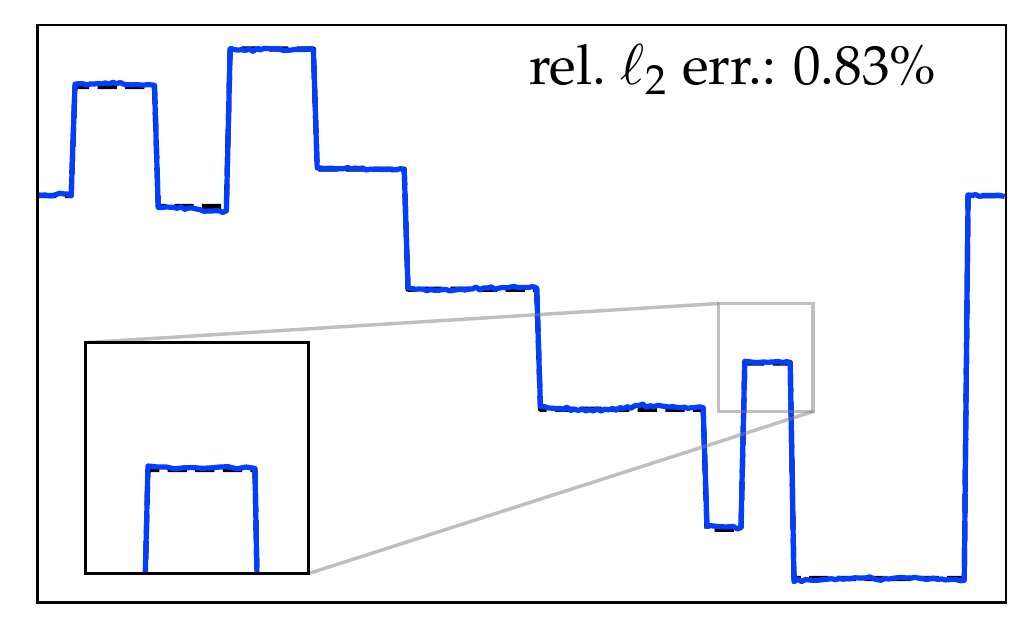} & \includegraphics[valign=c,width=0.24\textwidth]{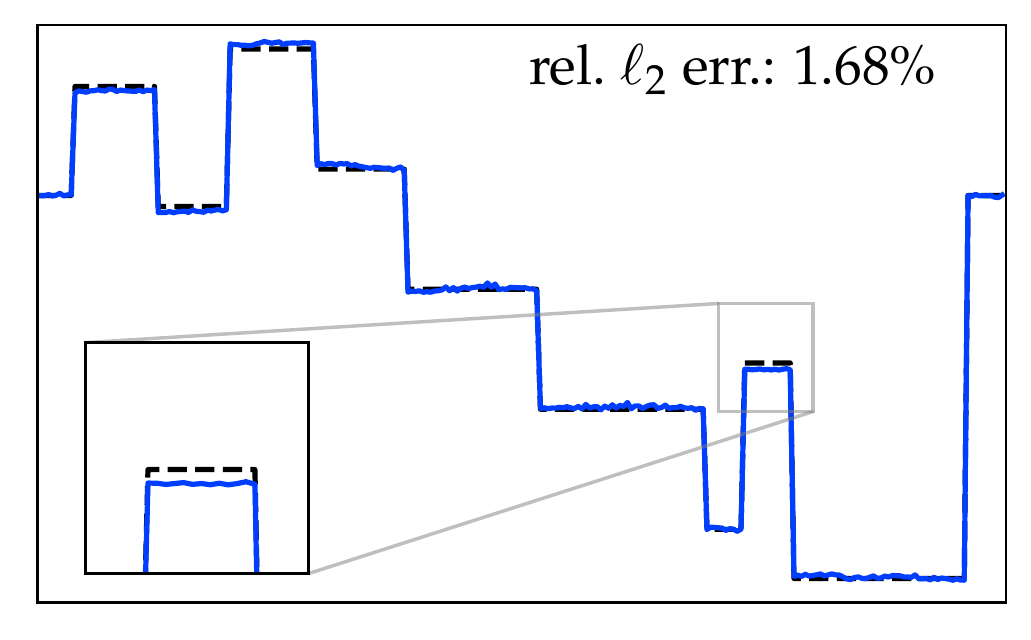} & \includegraphics[valign=c,width=0.24\textwidth]{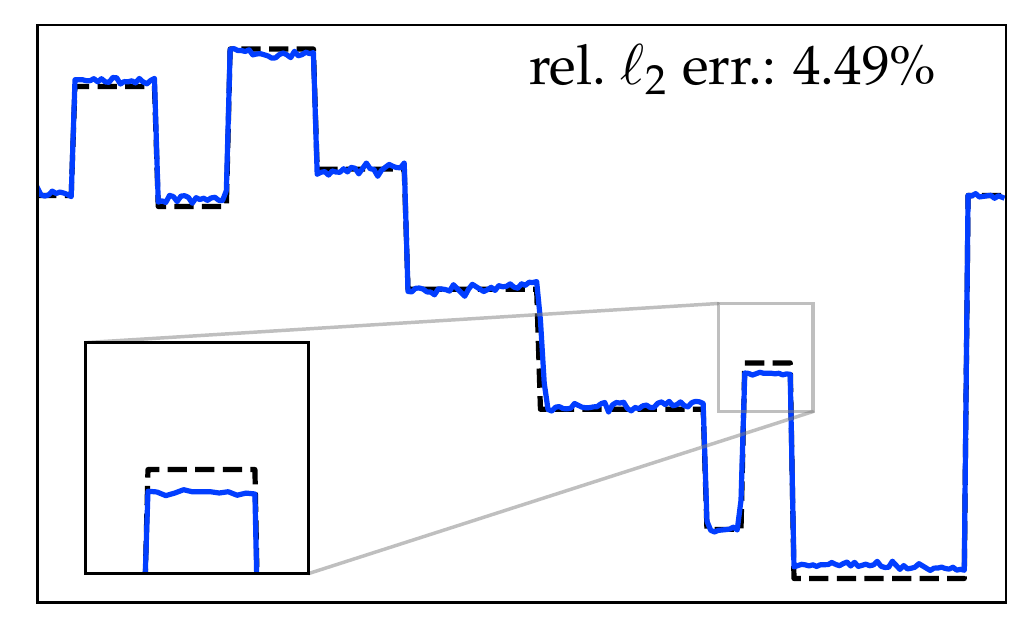}
	\end{tabular}
	\caption{\textbf{Scenario~\refAone{} -- CS with 1D signals.} Individual reconstructions of the signal from Fig.~\ref{fig:tvsynth:example_adv} under Gaussian noise. The ground truth signal is visualized by a dashed line. In favor of the more insightful noise level 12\%, we have omitted the noiseless case.}
	\label{fig:tvsynth:example_gauss}
\end{figure}

\begin{table}[H]
	\robustify\bfseries
	\begin{tabular}{llS[separate-uncertainty]S[separate-uncertainty]S[separate-uncertainty]S[separate-uncertainty]S[separate-uncertainty]S[separate-uncertainty]S[separate-uncertainty]}
\toprule
\multicolumn{2}{l}{rel.~noise -- adversarial}    &                  {0.0\%} &                  {0.5\%} &                  {1.0\%} &                  {3.0\%} &                  {5.0\%} &                   {7.5\%} &                  {10.0\%} \\
\midrule
TV$[\eta]$ & rel.~$\l{2}$-err. [\%] &          15.32 \pm 10.13 &           18.18 \pm 9.83 &           20.50 \pm 9.60 &           28.68 \pm 8.77 &           35.92 \pm 8.45 &            43.87 \pm 7.95 &            50.85 \pm 7.35 \\
UNet & rel.~$\l{2}$-err. [\%] &            9.79 \pm 2.14 &           10.24 \pm 2.17 &           10.71 \pm 2.19 &           12.96 \pm 2.37 &           15.71 \pm 2.58 &            20.23 \pm 2.91 &            25.08 \pm 3.15 \\
UNetFL & rel.~$\l{2}$-err. [\%] &            7.88 \pm 1.42 &            8.23 \pm 1.42 &            8.60 \pm 1.42 &           10.23 \pm 1.42 &           12.13 \pm 1.45 &            14.97 \pm 1.47 &            18.28 \pm 1.51 \\
Tira & rel.~$\l{2}$-err. [\%] &            8.56 \pm 1.77 &            8.95 \pm 1.78 &            9.37 \pm 1.79 &           11.21 \pm 1.81 &           13.50 \pm 1.80 &            16.87 \pm 1.95 &            20.66 \pm 2.11 \\
TiraFL & rel.~$\l{2}$-err. [\%] &            7.64 \pm 1.38 &            7.99 \pm 1.37 &            8.36 \pm 1.36 &            9.99 \pm 1.34 &           11.94 \pm 1.34 &            14.91 \pm 1.28 &            18.25 \pm 1.30 \\
ItNet & rel.~$\l{2}$-err. [\%] &  \bfseries 2.47 \pm 0.58 &  \bfseries 2.96 \pm 0.60 &  \bfseries 3.53 \pm 0.60 &  \bfseries 6.26 \pm 0.59 &  \bfseries 9.35 \pm 0.72 &  \bfseries 13.62 \pm 1.31 &  \bfseries 18.06 \pm 1.77 \\
\bottomrule
\end{tabular}

	\caption{\textbf{Scenario~\refAtwo{} -- CS with MNIST.} A numerical representation of the results of Fig.~\ref{fig:mnist:table}(c), including the additional methods $\UNetFL$ and $\Tira$. The smallest relative error per noise level is highlighted in bold.}
	\label{tab:mnist:table_adv}
\end{table}

\begin{table}[H]
	\robustify\bfseries
	\begin{tabular}{llS[separate-uncertainty]S[separate-uncertainty]S[separate-uncertainty]S[separate-uncertainty]S[separate-uncertainty]S[separate-uncertainty]S[separate-uncertainty]}
\toprule
\multicolumn{2}{l}{rel.~noise -- Gaussian} &            {0.0\%} &                  {0.5\%} &                  {1.0\%} &                  {3.0\%} &                  {5.0\%} &                  {7.5\%} &                 {10.0\%} \\
\midrule
TV$[\eta]$ & rel.~$\l{2}$-err. [\%] &          15.32 \pm 10.13 &           16.55 \pm 9.71 &           17.48 \pm 9.31 &           21.52 \pm 8.19 &           25.53 \pm 7.75 &           30.00 \pm 7.48 &           34.20 \pm 7.18 \\
UNet & rel.~$\l{2}$-err. [\%] &            9.79 \pm 2.14 &            9.87 \pm 2.15 &            9.96 \pm 2.14 &           10.36 \pm 2.13 &           10.86 \pm 2.18 &           11.54 \pm 2.16 &           12.37 \pm 2.11 \\
UNetFL & rel.~$\l{2}$-err. [\%] &            7.88 \pm 1.42 &            7.88 \pm 1.42 &            7.89 \pm 1.42 &            7.99 \pm 1.41 &            8.16 \pm 1.40 &            8.49 \pm 1.38 &            8.92 \pm 1.36 \\
Tira & rel.~$\l{2}$-err. [\%] &            8.56 \pm 1.77 &            8.56 \pm 1.77 &            8.57 \pm 1.76 &            8.67 \pm 1.75 &            8.85 \pm 1.72 &            9.17 \pm 1.69 &            9.59 \pm 1.65 \\
TiraFL & rel.~$\l{2}$-err. [\%] &            7.64 \pm 1.38 &            7.70 \pm 1.38 &            7.77 \pm 1.37 &            8.12 \pm 1.35 &            8.52 \pm 1.36 &            9.18 \pm 1.35 &            9.88 \pm 1.35 \\
ItNet & rel.~$\l{2}$-err. [\%] &  \bfseries 2.47 \pm 0.58 &  \bfseries 2.58 \pm 0.59 &  \bfseries 2.72 \pm 0.58 &  \bfseries 3.60 \pm 0.58 &  \bfseries 4.65 \pm 0.66 &  \bfseries 6.00 \pm 0.73 &  \bfseries 7.32 \pm 0.80 \\
\bottomrule
\end{tabular}

	\caption{\textbf{Scenario~\refAtwo{} -- CS with MNIST.} A numerical representation of the results of Fig.~\ref{fig:mnist:table}(d), including the additional methods $\UNetFL$ and $\Tira$. The smallest relative error per noise level is highlighted in bold.}
	\label{tab:mnist:table_ref}
\end{table}

\begin{figure}[H]
	\centering
	\begin{tabular}{cc}
		\scriptsize
		\setlength\tabcolsep{1pt}
		\begin{tabular}{lccc}
			& 2\% rel.~noise -- adv. & 5\% rel.~noise -- adv. & 10\% rel.~noise -- adv. \\
			\rotatebox[origin=c]{90}{$\TV[\noisebnd]$} &
			\includegraphics[valign=c,width=0.14\textwidth]{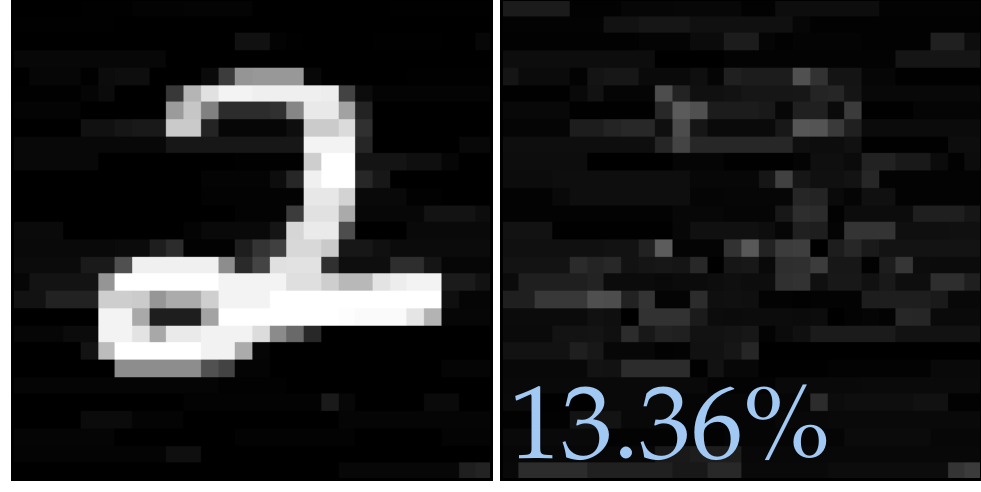} & \includegraphics[valign=c,width=0.14\textwidth]{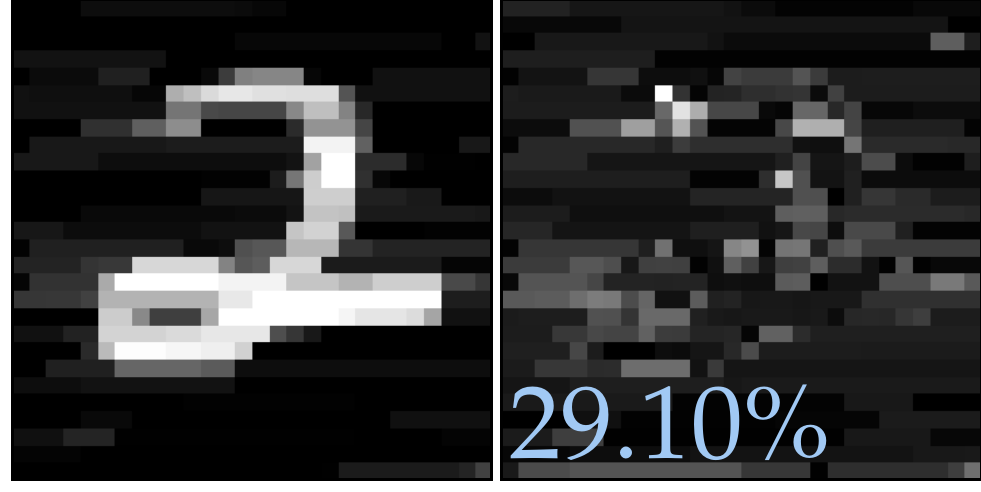} & \includegraphics[valign=c,width=0.14\textwidth]{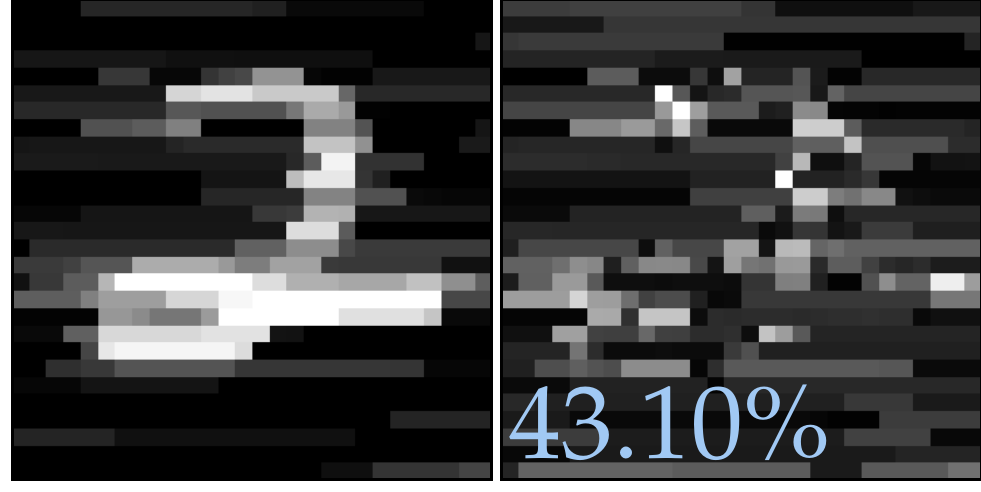} \\
			\rotatebox[origin=c]{90}{$\UNet$} &
			\includegraphics[valign=c,width=0.14\textwidth]{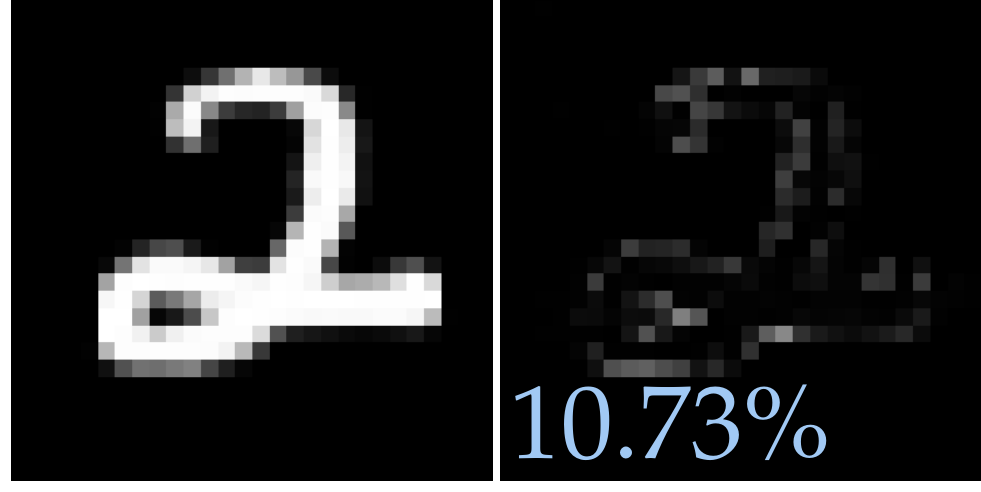} & \includegraphics[valign=c,width=0.14\textwidth]{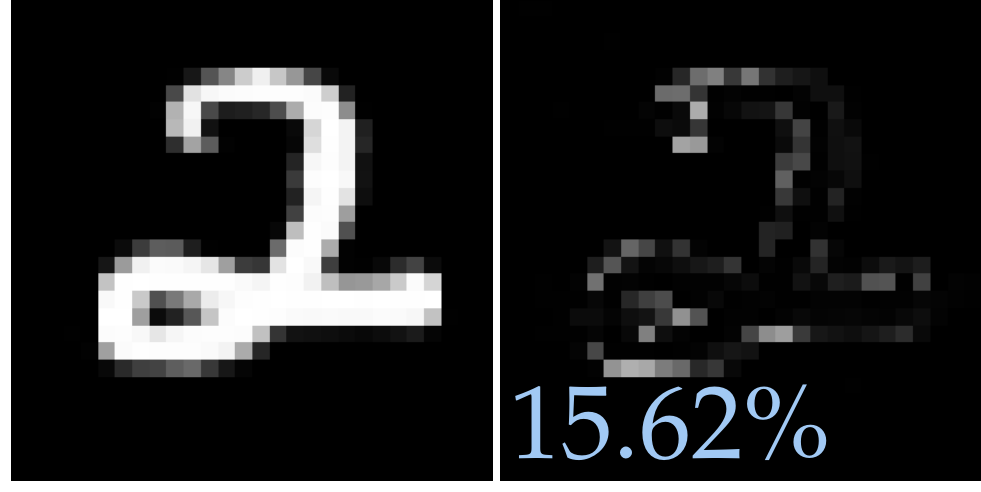} & \includegraphics[valign=c,width=0.14\textwidth]{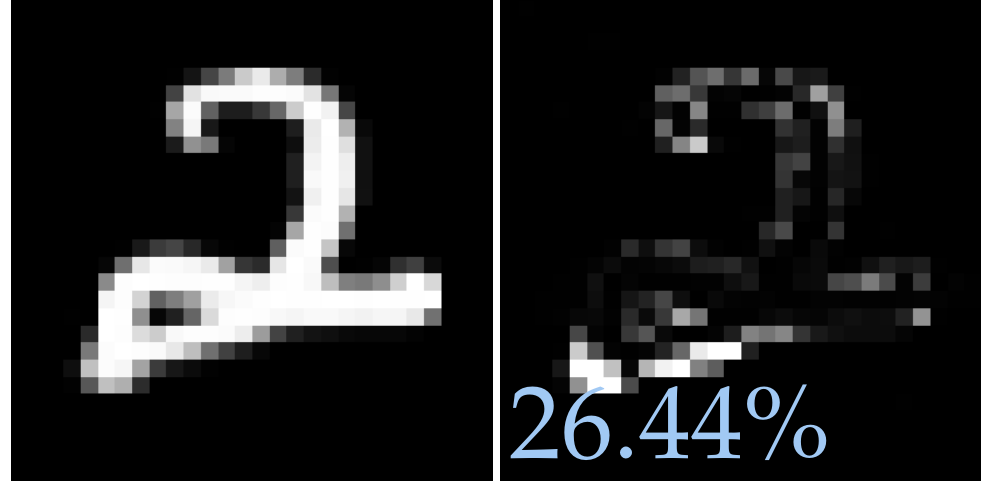} \\
			\rotatebox[origin=c]{90}{$\TiraFL$} &
			\includegraphics[valign=c,width=0.14\textwidth]{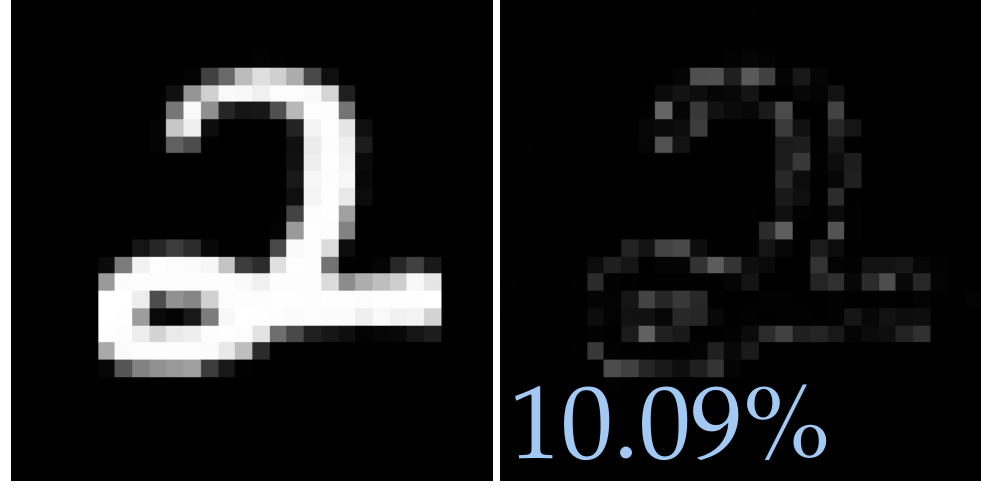} & \includegraphics[valign=c,width=0.14\textwidth]{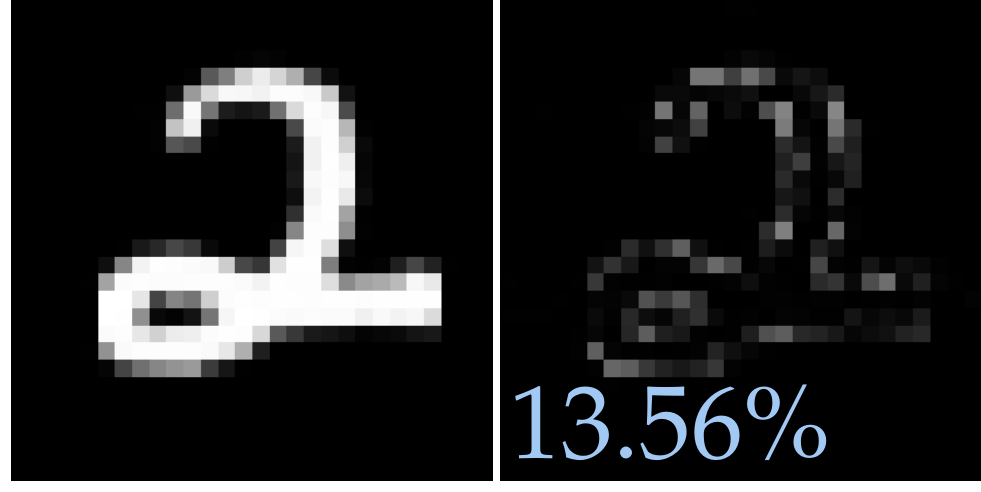} & \includegraphics[valign=c,width=0.14\textwidth]{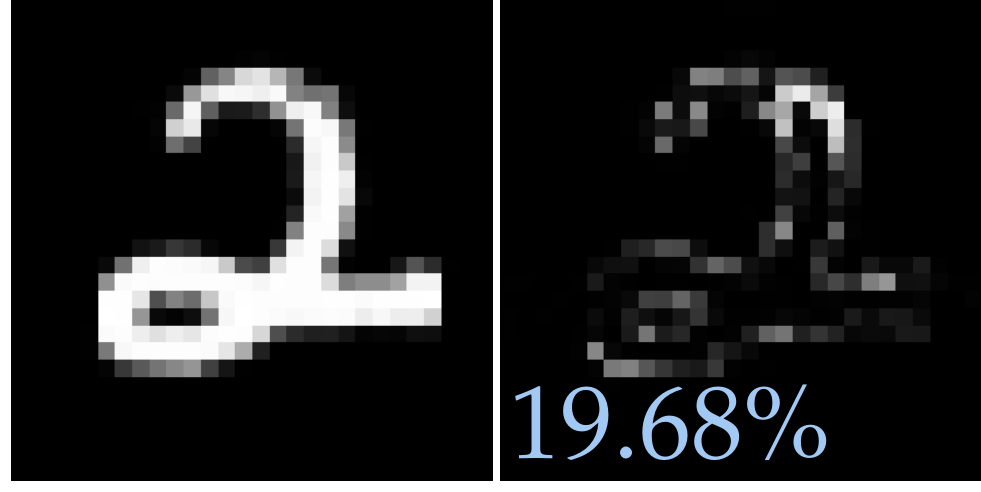} \\
			\rotatebox[origin=c]{90}{$\ItNet$} &
			\includegraphics[valign=c,width=0.14\textwidth]{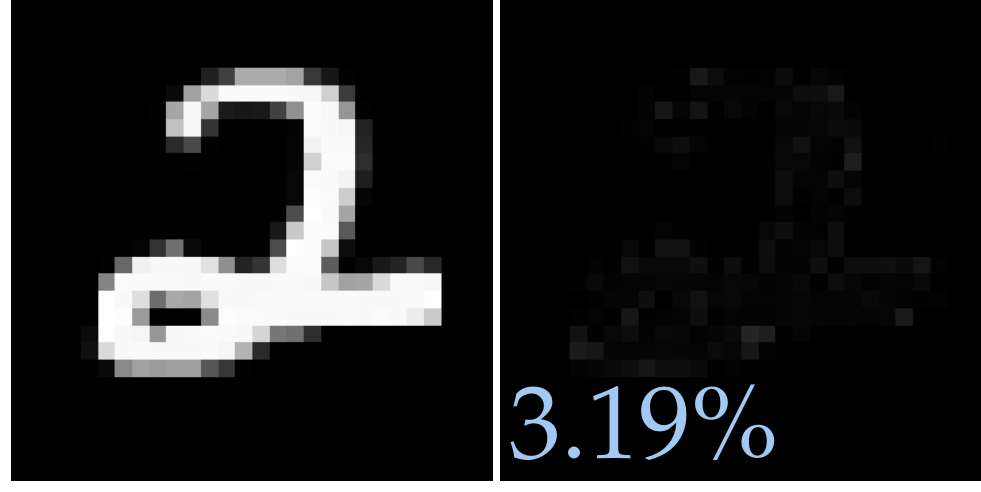} & \includegraphics[valign=c,width=0.14\textwidth]{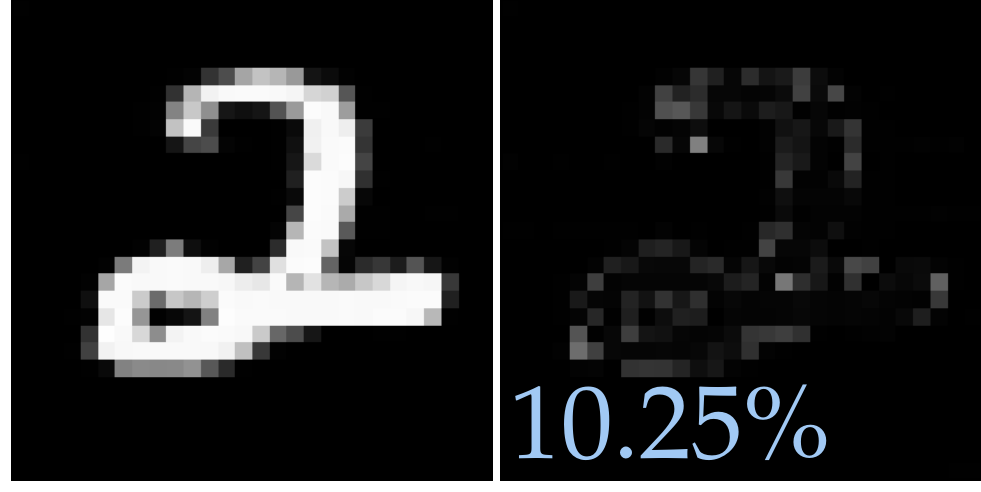} & \includegraphics[valign=c,width=0.14\textwidth]{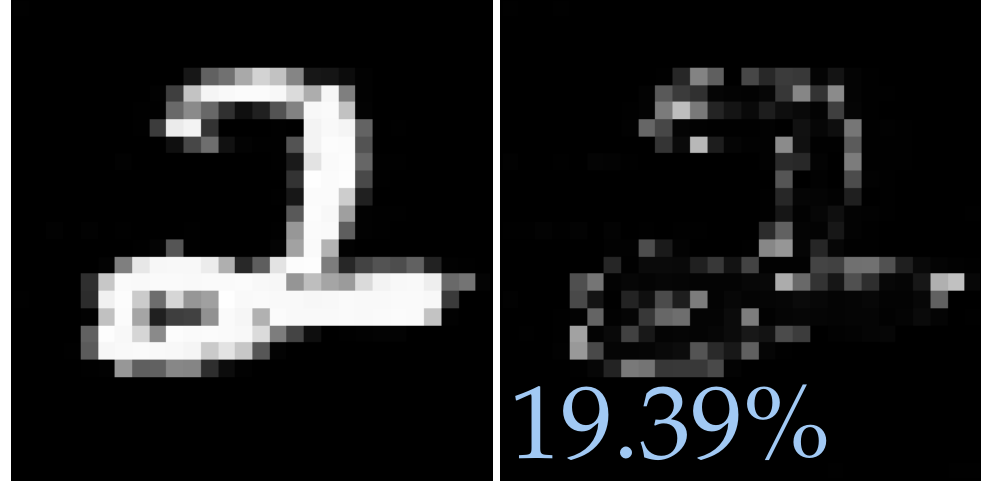} \\
		\end{tabular}

		&
		\scriptsize
		\setlength\tabcolsep{1pt}
		\begin{tabular}{lccc}
			& 2\% rel.~noise -- adv. & 5\% rel.~noise -- adv. & 10\% rel.~noise -- adv. \\
			&
			\includegraphics[valign=c,width=0.14\textwidth]{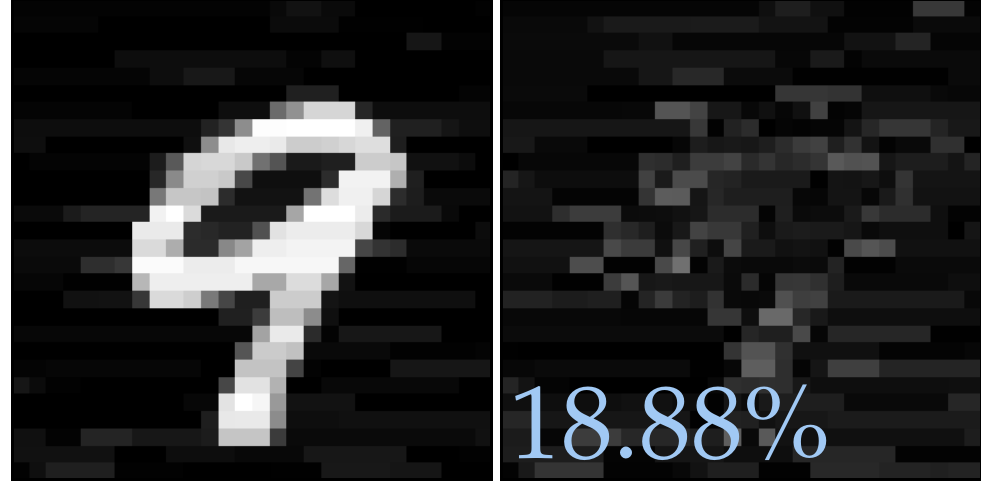} & \includegraphics[valign=c,width=0.14\textwidth]{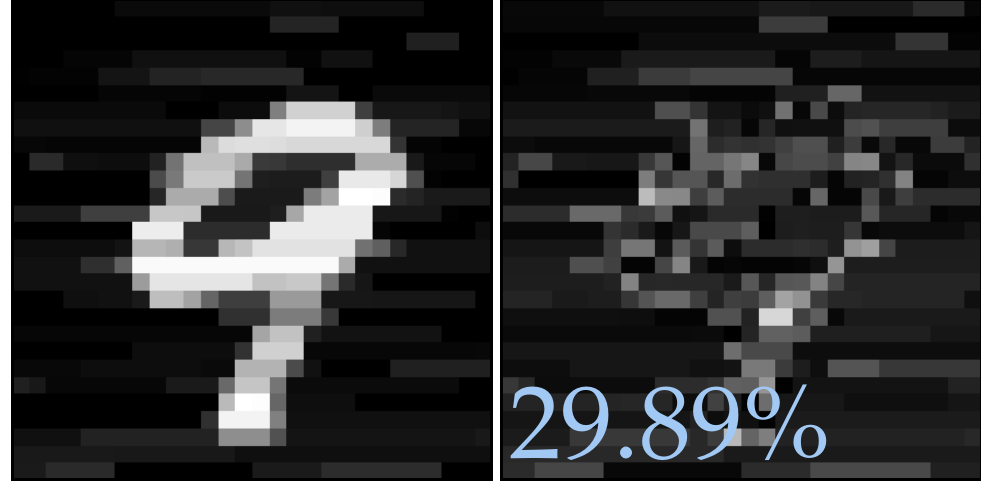} & \includegraphics[valign=c,width=0.14\textwidth]{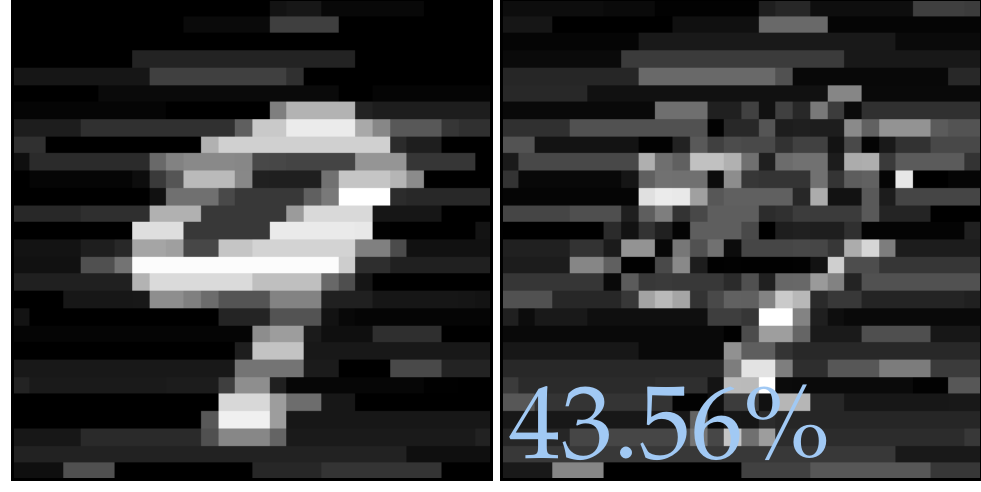} \\
			&
			\includegraphics[valign=c,width=0.14\textwidth]{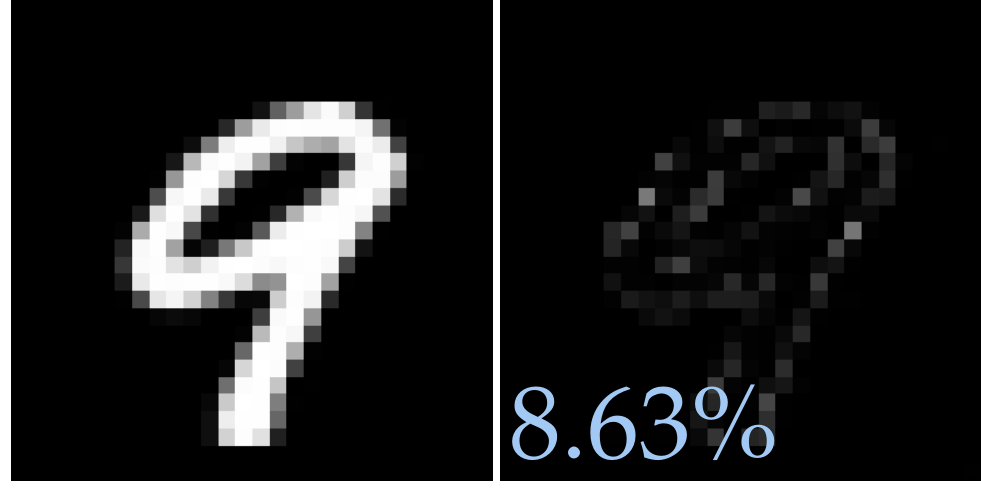} & \includegraphics[valign=c,width=0.14\textwidth]{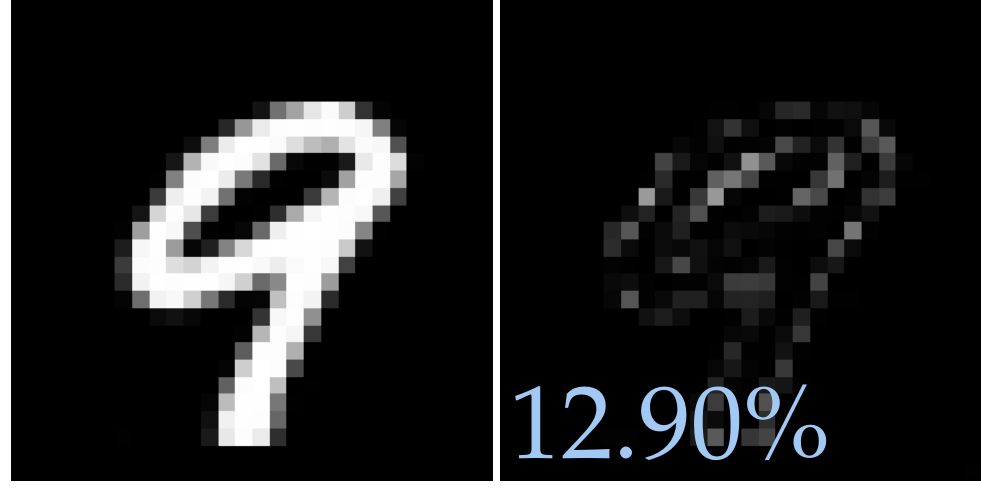} & \includegraphics[valign=c,width=0.14\textwidth]{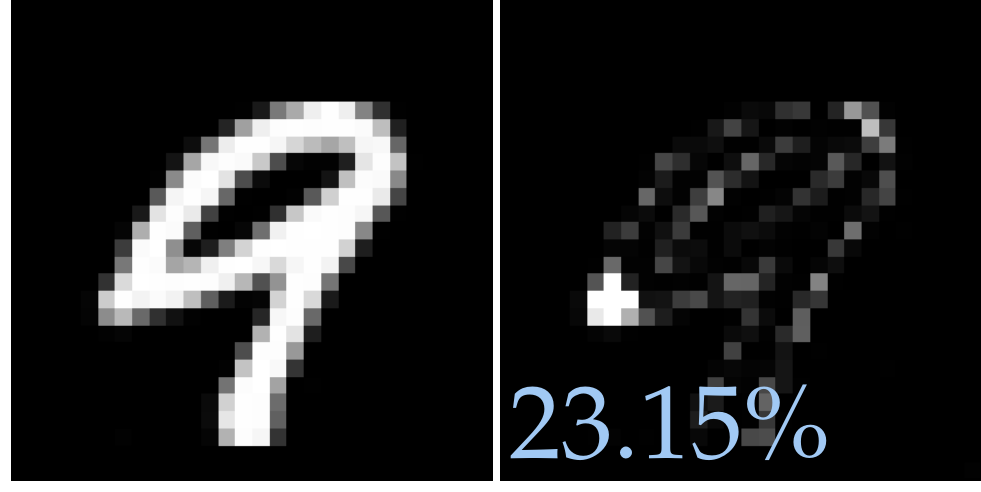} \\
			&
			\includegraphics[valign=c,width=0.14\textwidth]{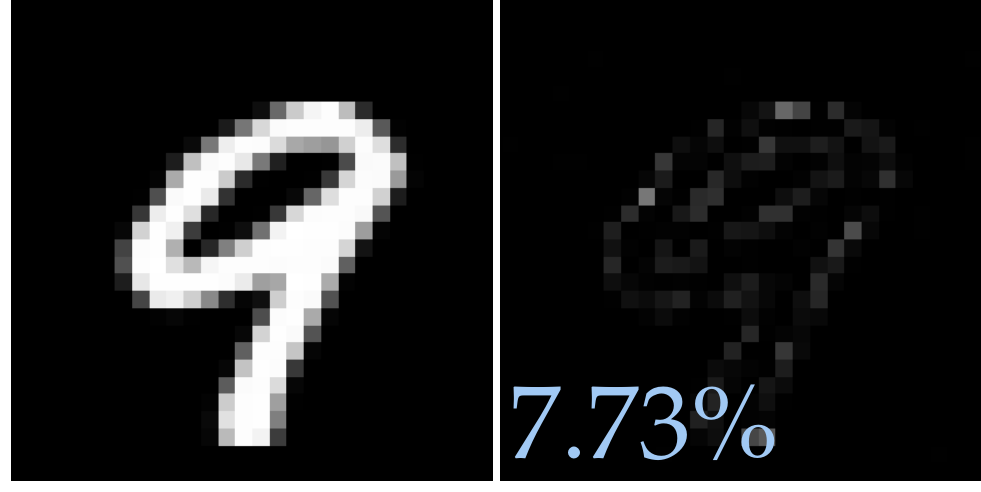} & \includegraphics[valign=c,width=0.14\textwidth]{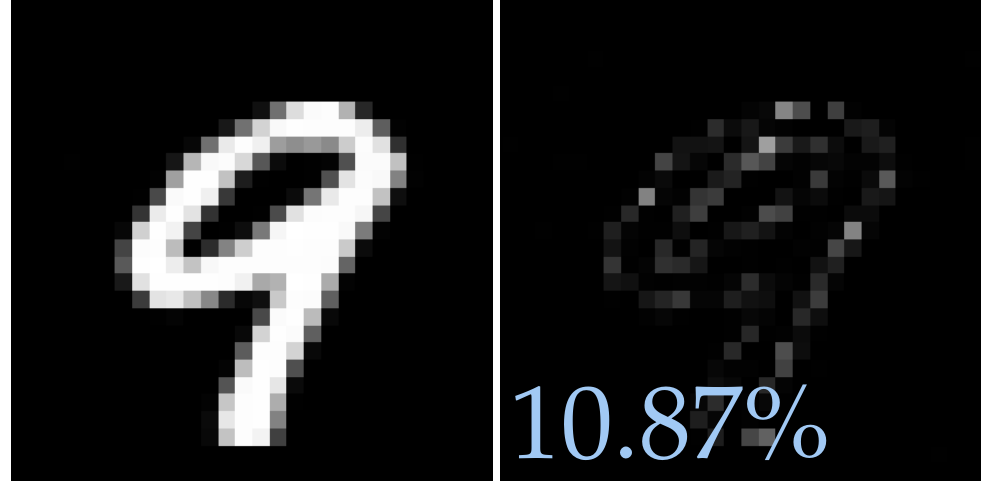} & \includegraphics[valign=c,width=0.14\textwidth]{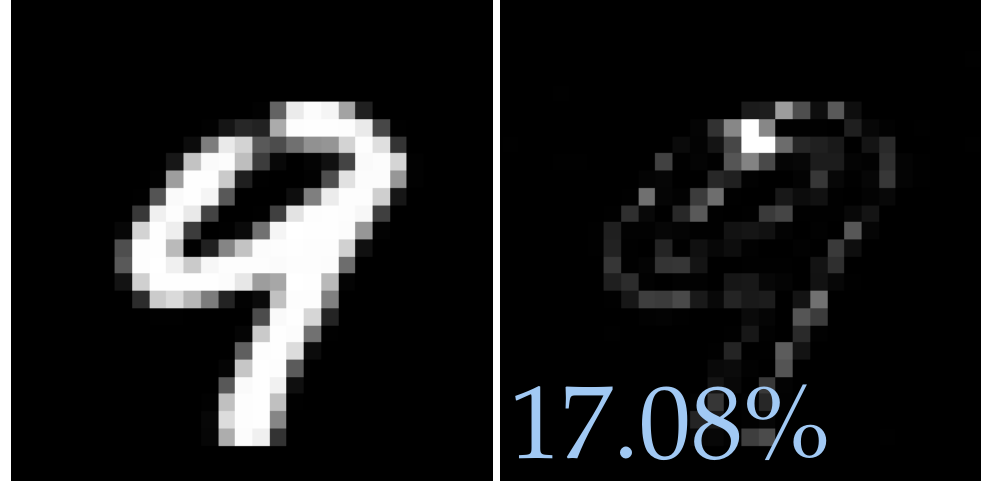} \\
			&
			\includegraphics[valign=c,width=0.14\textwidth]{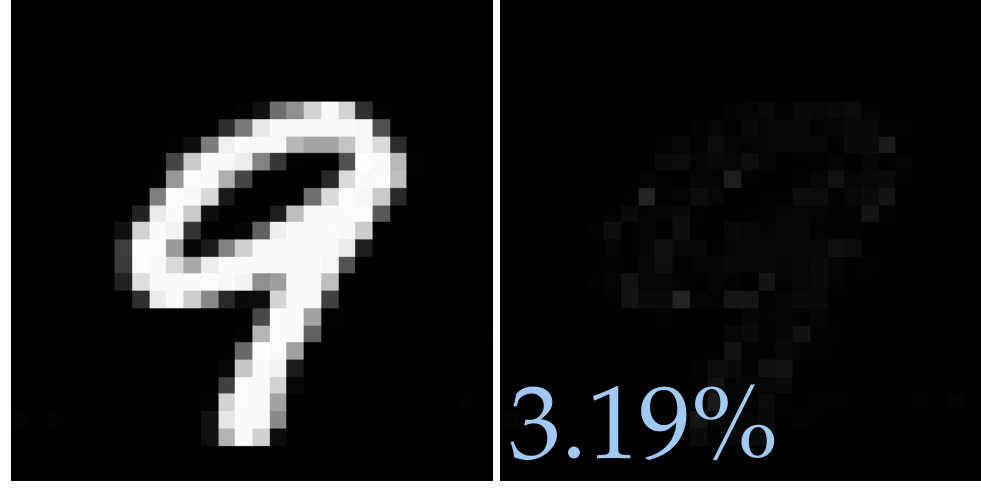} & \includegraphics[valign=c,width=0.14\textwidth]{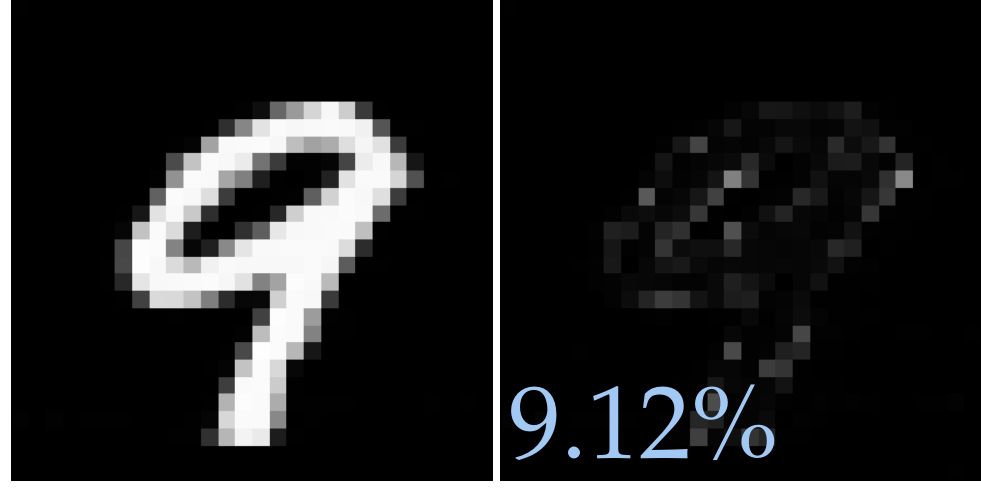} & \includegraphics[valign=c,width=0.14\textwidth]{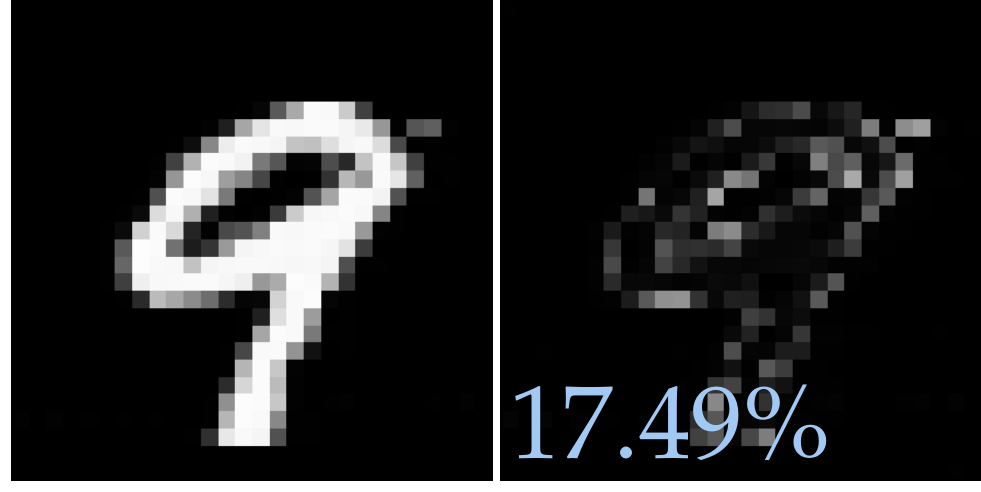} \\
		\end{tabular}
	\end{tabular}
	\caption{\textbf{Scenario~\refAtwo{} -- CS with MNIST.} Individual reconstructions of two additional digits from the test set for different levels of adversarial noise (see Fig.~\ref{fig:mnist:example_adv}). The reconstructed digits and their error plots (with relative $\l{2}$-error) are displayed in the windows $[0,1]$ and $[0, 0.6]$, respectively.}
	\label{fig:mnist:example_adv_supp}
\end{figure}

\begin{figure}[H]
\centering
\begin{tabular}{cc}
 \scriptsize
  \setlength\tabcolsep{1pt}
  \begin{tabular}{lccc}
     & 5\% rel.~noise -- Gauss. & 10\% rel.~noise -- Gauss. & 25\% rel.~noise -- Gauss. \\
     \rotatebox[origin=c]{90}{$\TV[\noisebnd]$} &
    \includegraphics[valign=c,width=0.14\textwidth]{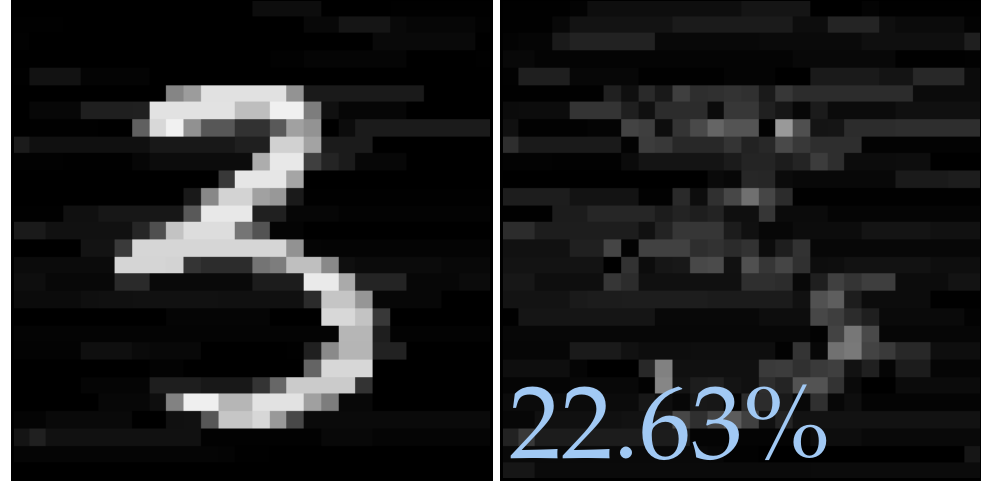} & \includegraphics[valign=c,width=0.14\textwidth]{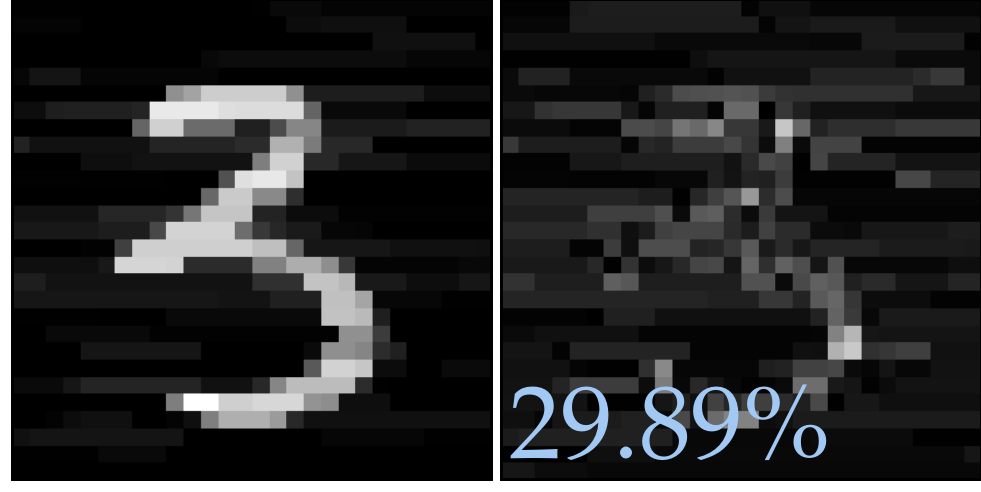} & \includegraphics[valign=c,width=0.14\textwidth]{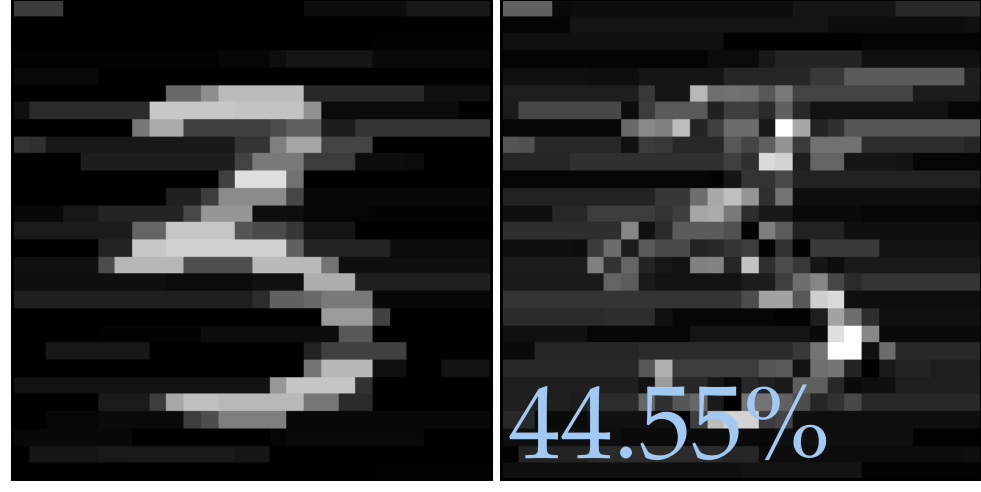} \\
    \rotatebox[origin=c]{90}{$\UNet$} &
    \includegraphics[valign=c,width=0.14\textwidth]{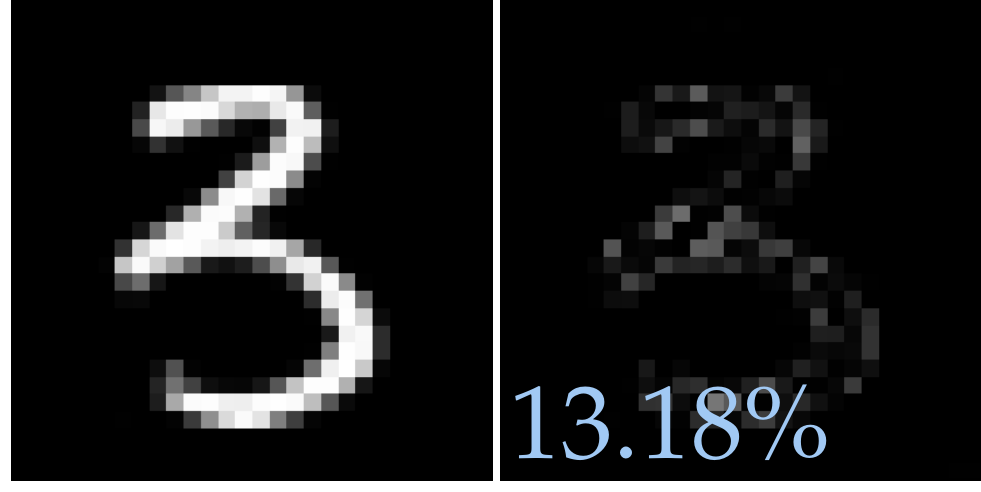} & \includegraphics[valign=c,width=0.14\textwidth]{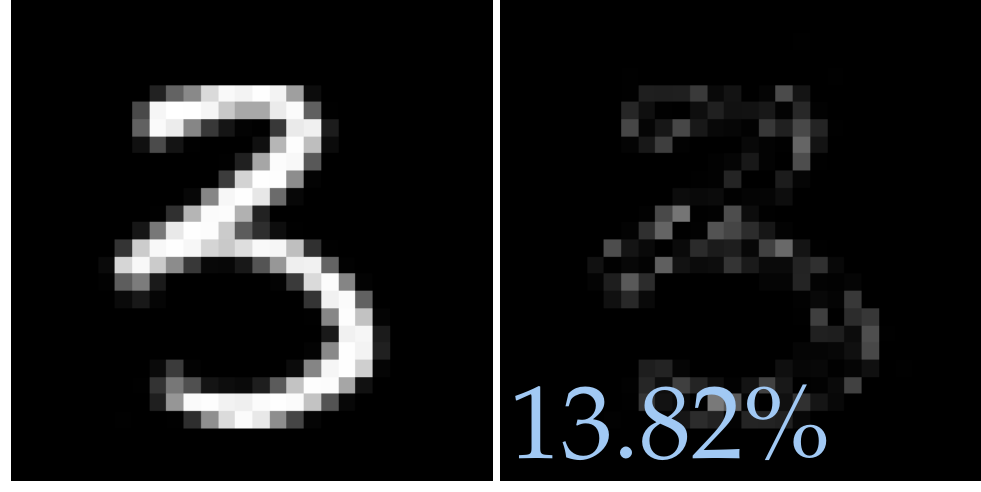} & \includegraphics[valign=c,width=0.14\textwidth]{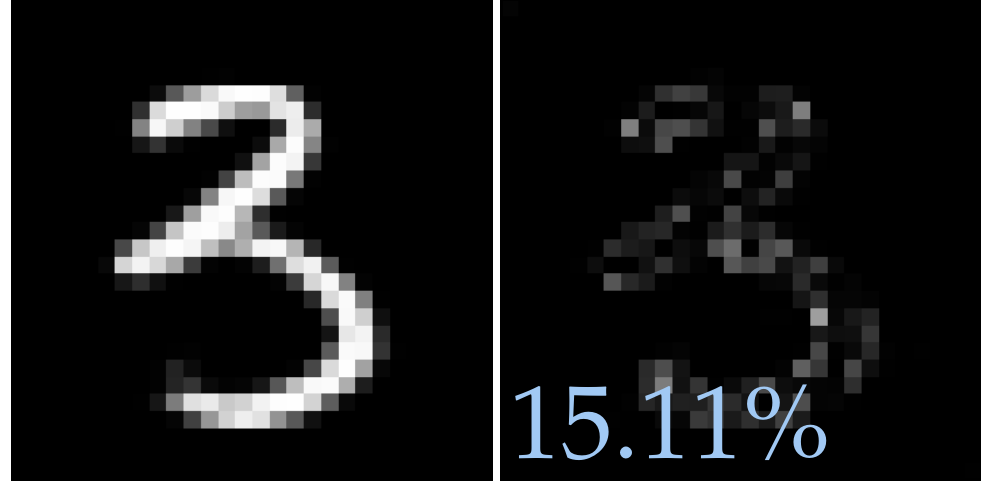} \\
    \rotatebox[origin=c]{90}{$\TiraFL$} &
    \includegraphics[valign=c,width=0.14\textwidth]{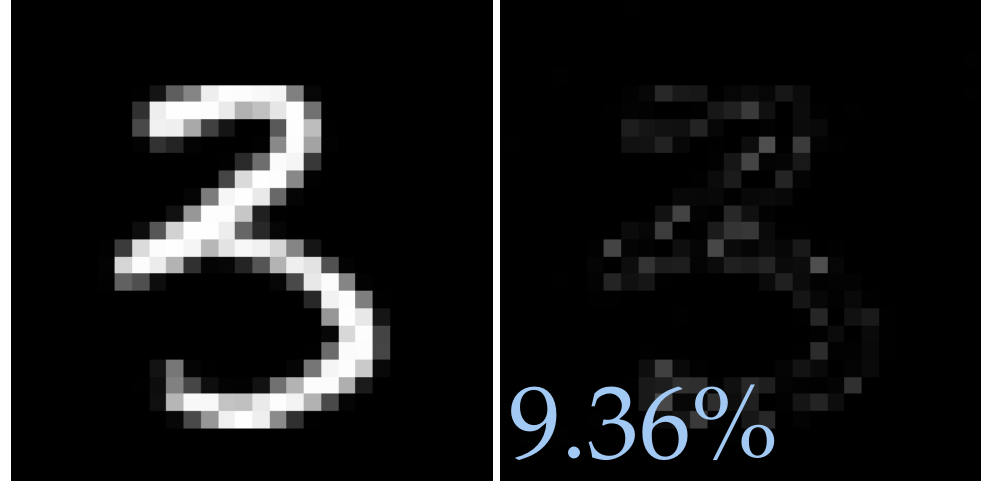} & \includegraphics[valign=c,width=0.14\textwidth]{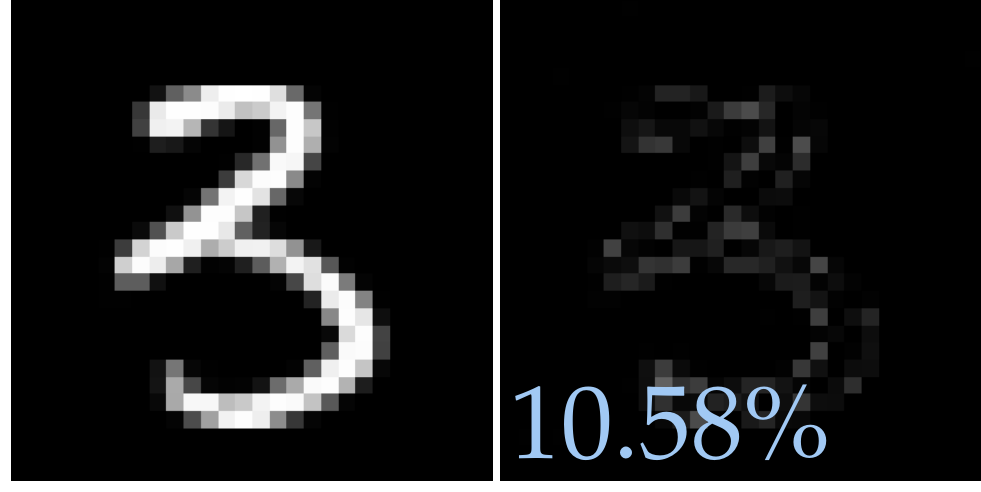} & \includegraphics[valign=c,width=0.14\textwidth]{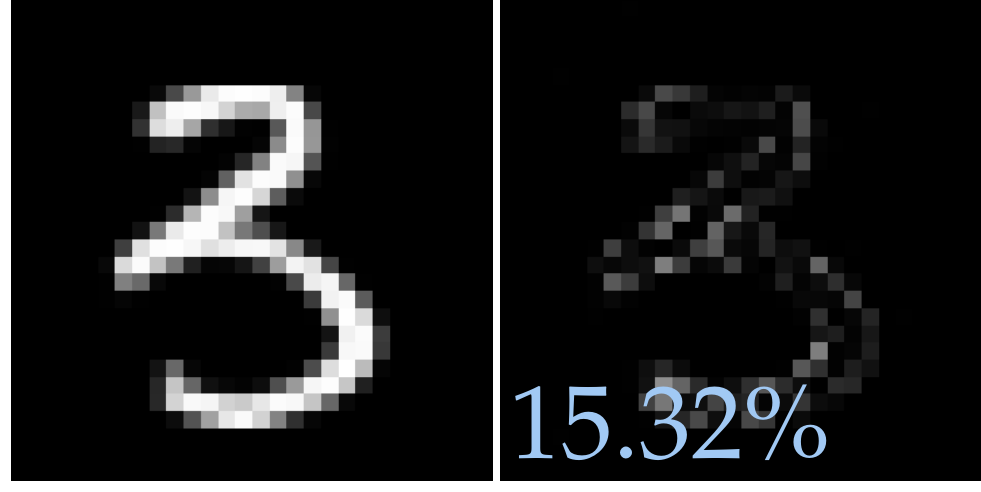} \\
    \rotatebox[origin=c]{90}{$\ItNet$} &
    \includegraphics[valign=c,width=0.14\textwidth]{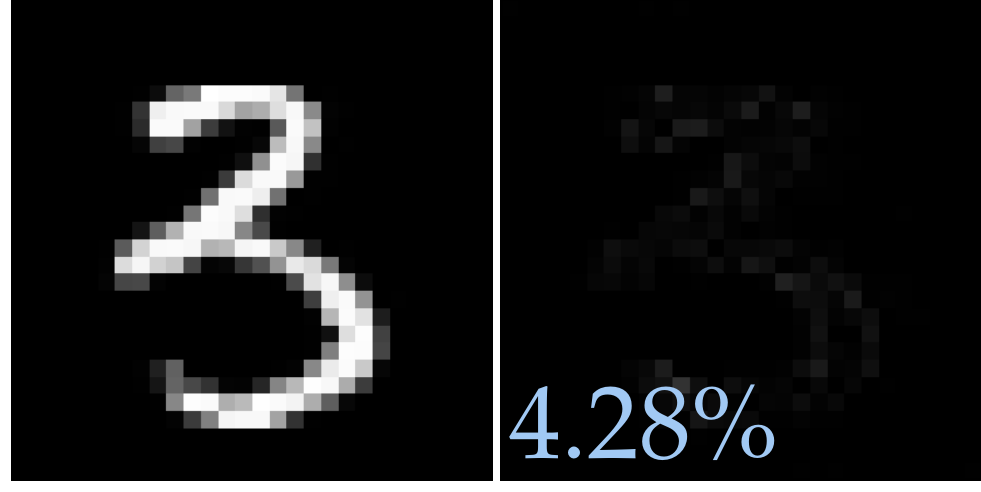} & \includegraphics[valign=c,width=0.14\textwidth]{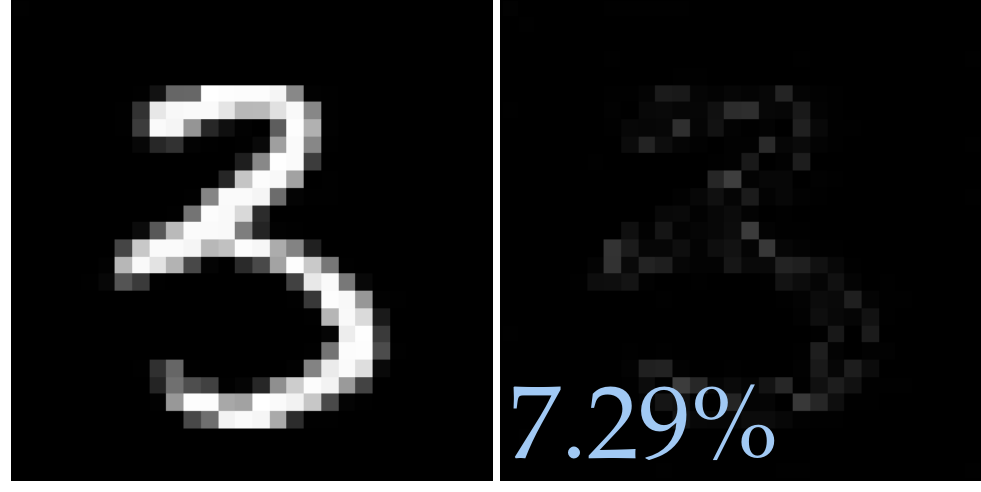} & \includegraphics[valign=c,width=0.14\textwidth]{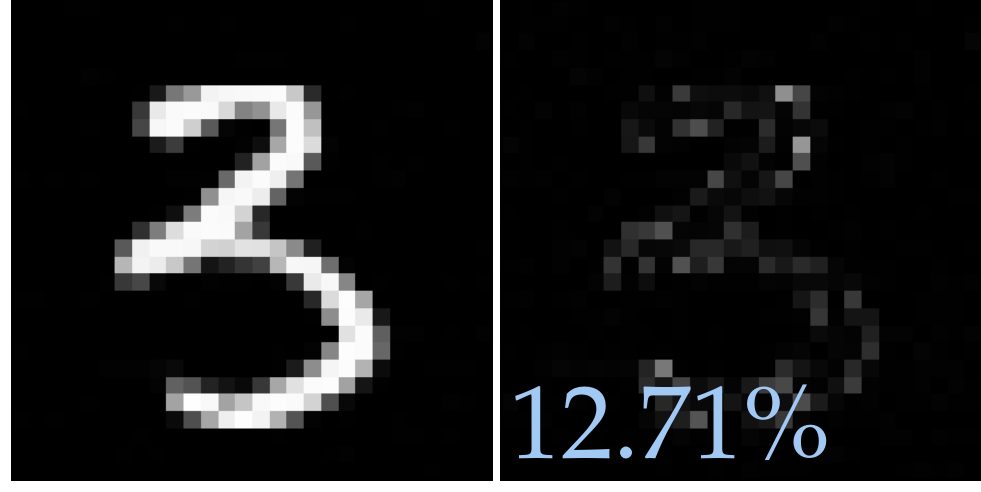} \\
 \end{tabular}

 &
 \scriptsize
  \setlength\tabcolsep{1pt}
   \begin{tabular}{lccc}
     & 5\% rel.~noise -- Gauss. & 10\% rel.~noise -- Gauss. & 25\% rel.~noise -- Gauss. \\
      &
    \includegraphics[valign=c,width=0.14\textwidth]{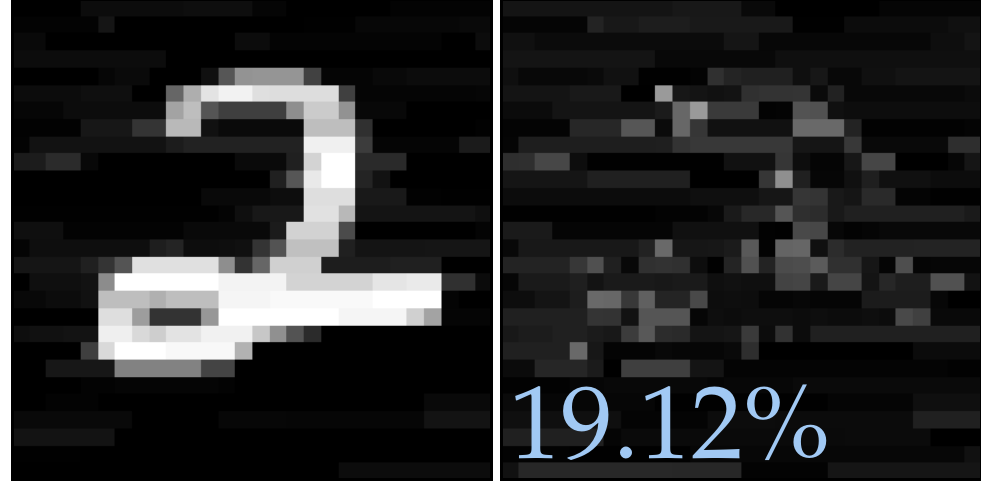} & \includegraphics[valign=c,width=0.14\textwidth]{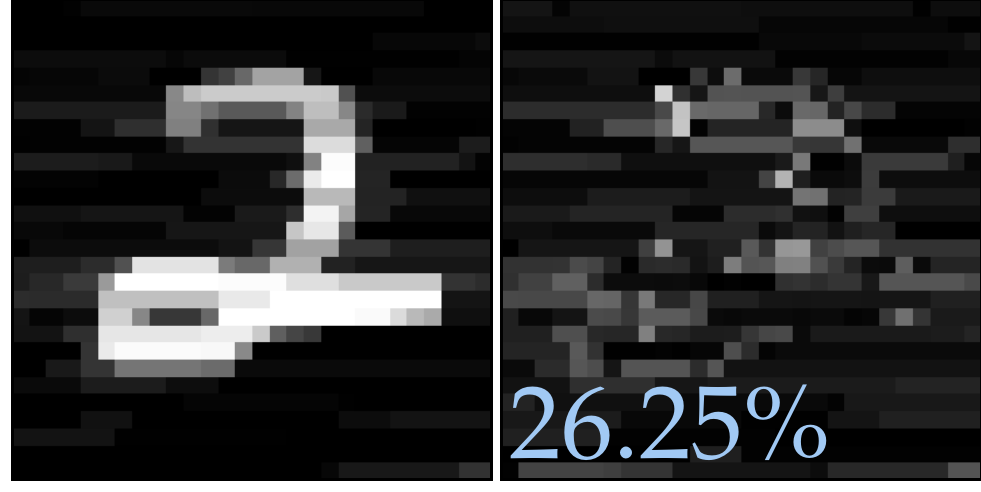} & \includegraphics[valign=c,width=0.14\textwidth]{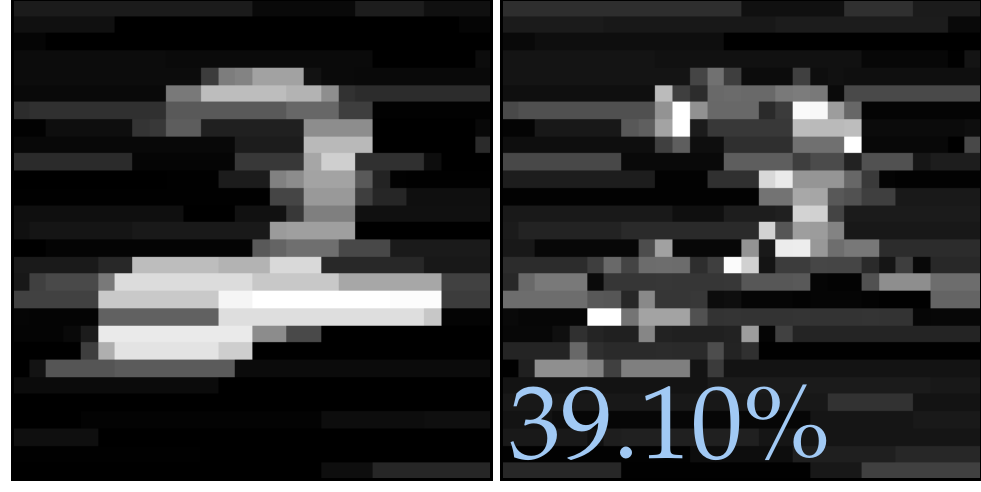} \\
     &
    \includegraphics[valign=c,width=0.14\textwidth]{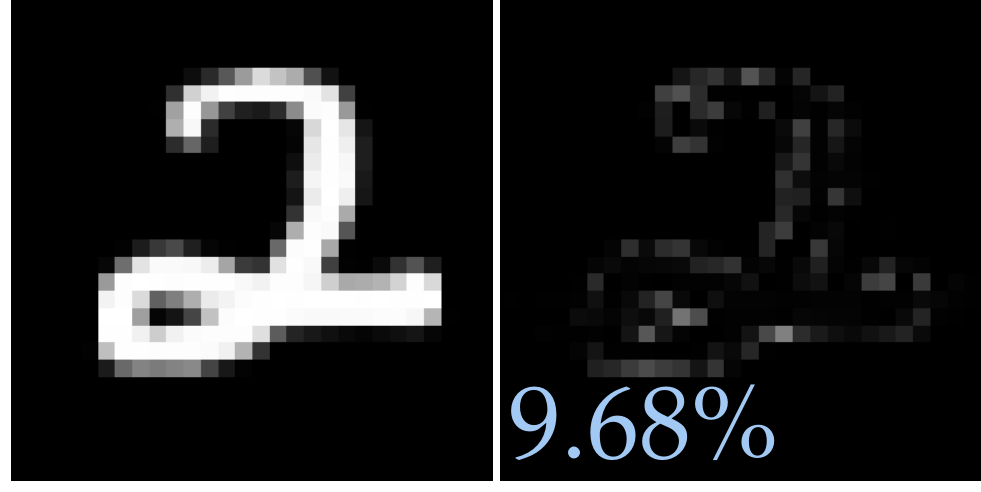} & \includegraphics[valign=c,width=0.14\textwidth]{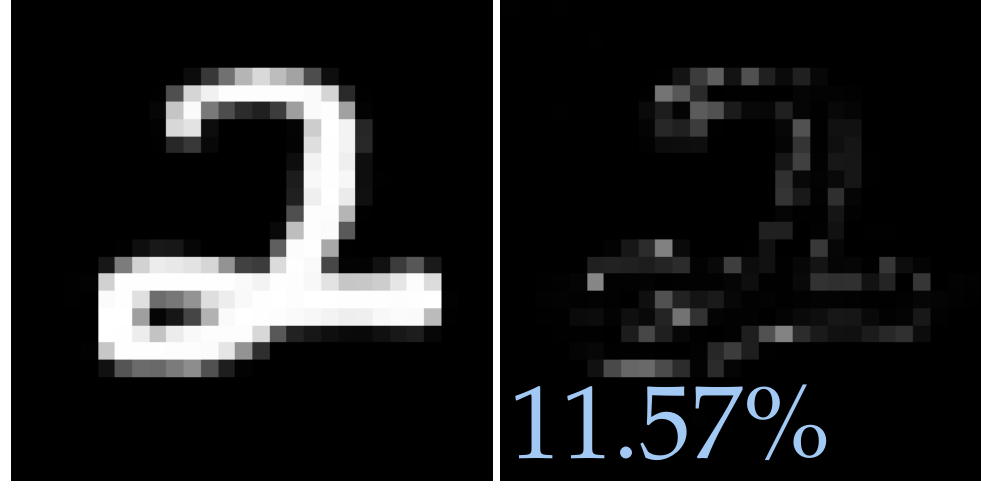} & \includegraphics[valign=c,width=0.14\textwidth]{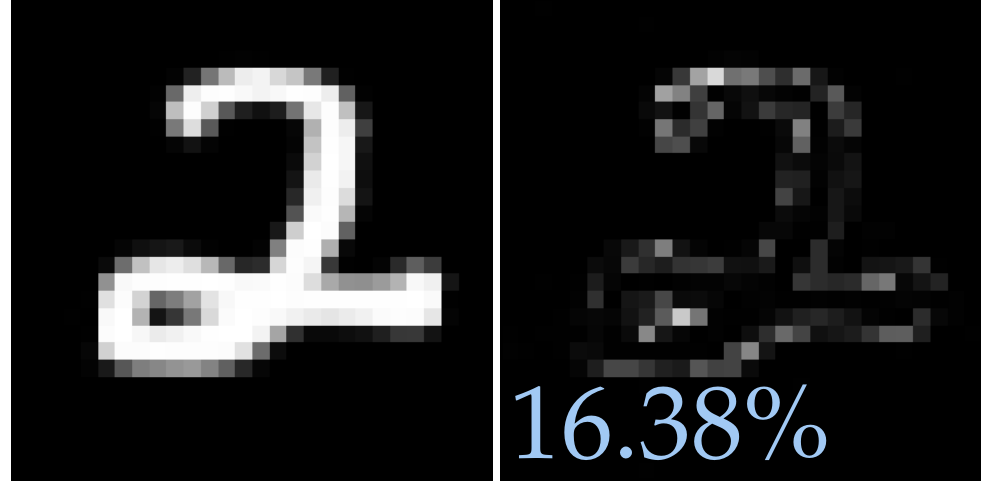} \\
     &
    \includegraphics[valign=c,width=0.14\textwidth]{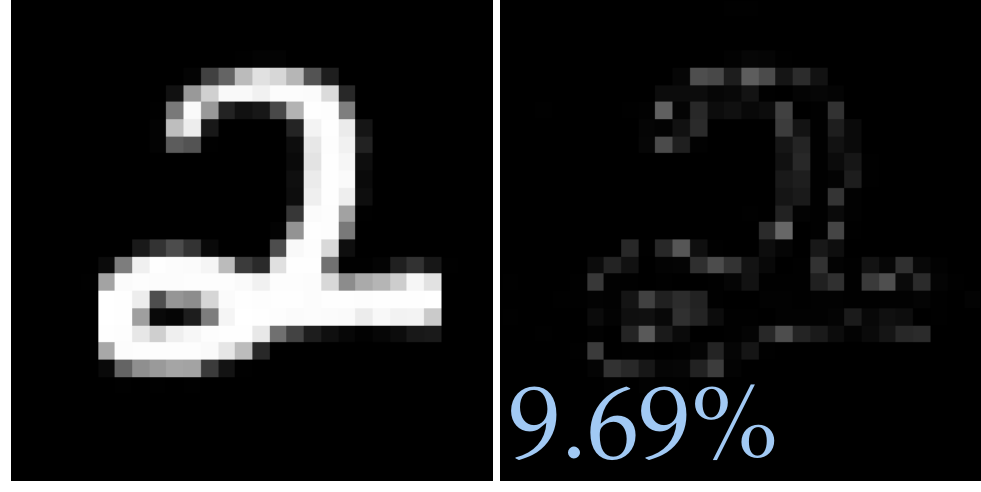} & \includegraphics[valign=c,width=0.14\textwidth]{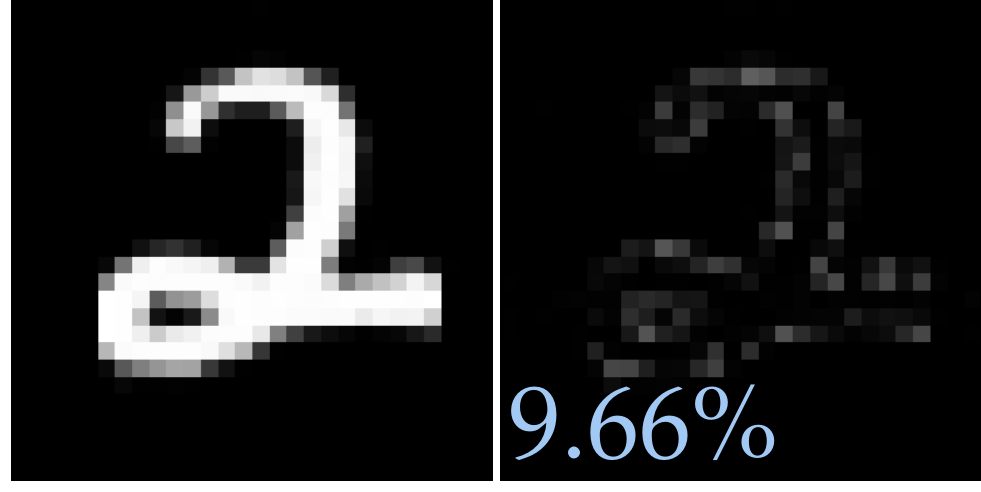} & \includegraphics[valign=c,width=0.14\textwidth]{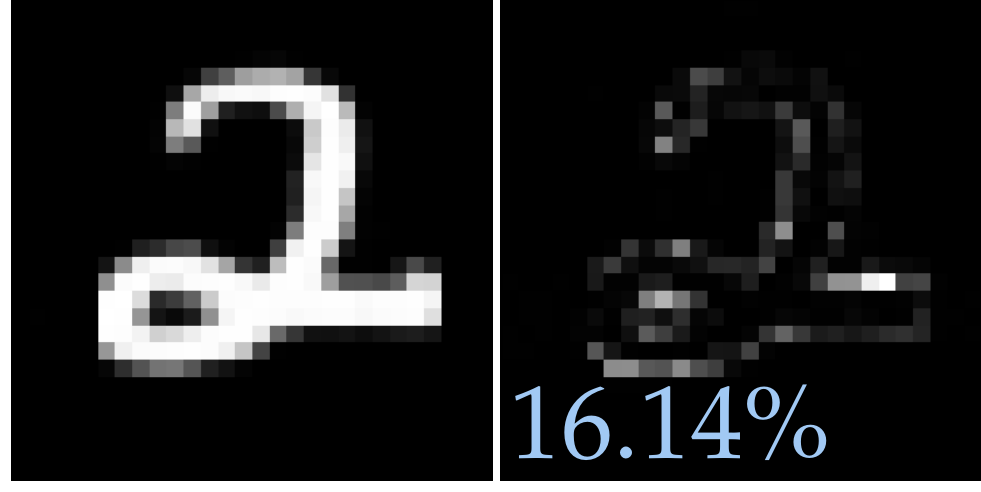} \\
     &
    \includegraphics[valign=c,width=0.14\textwidth]{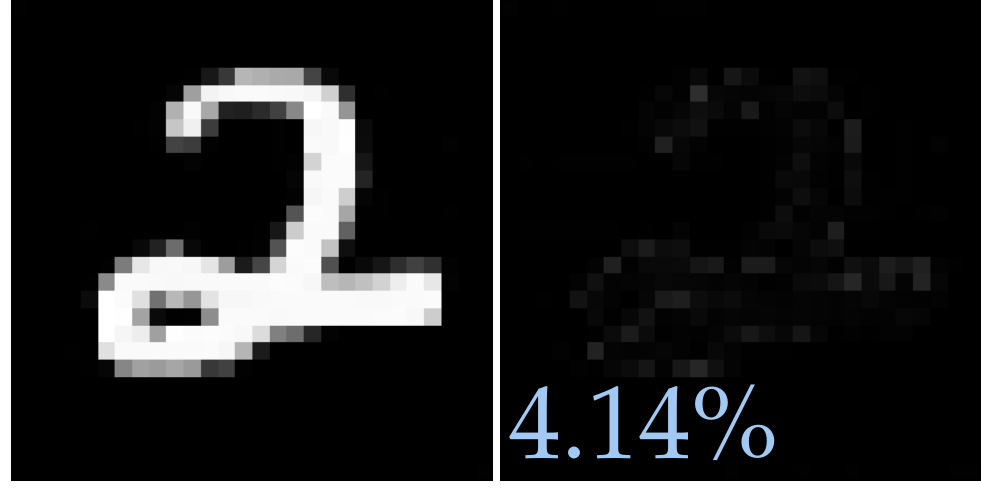} & \includegraphics[valign=c,width=0.14\textwidth]{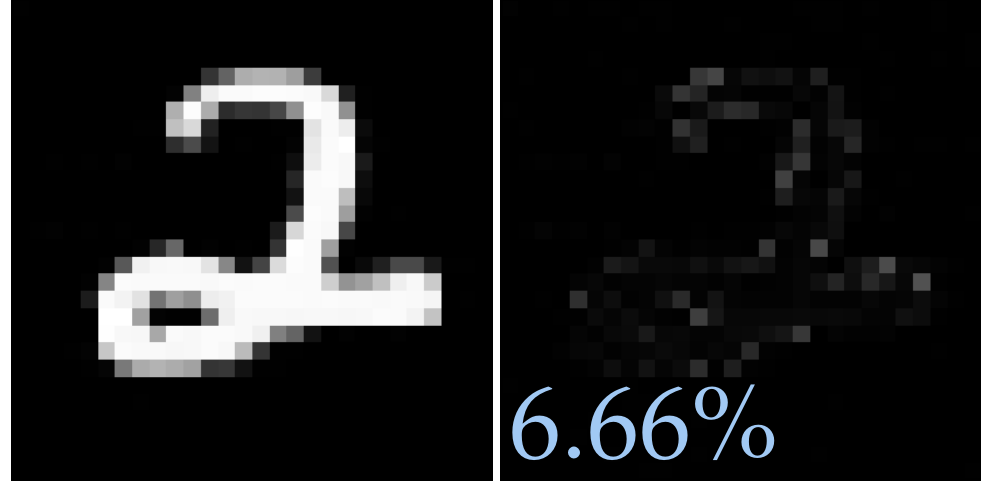} & \includegraphics[valign=c,width=0.14\textwidth]{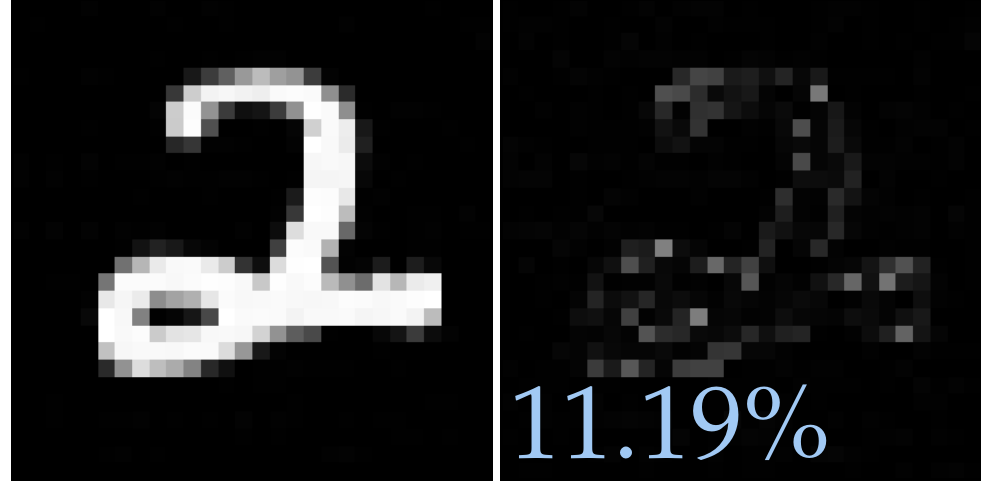} \\
 \end{tabular}

 \\
 \\
  \scriptsize
  \setlength\tabcolsep{1pt}
   \begin{tabular}{lccc}
     \rotatebox[origin=c]{90}{$\TV[\noisebnd]$} &
    \includegraphics[valign=c,width=0.14\textwidth]{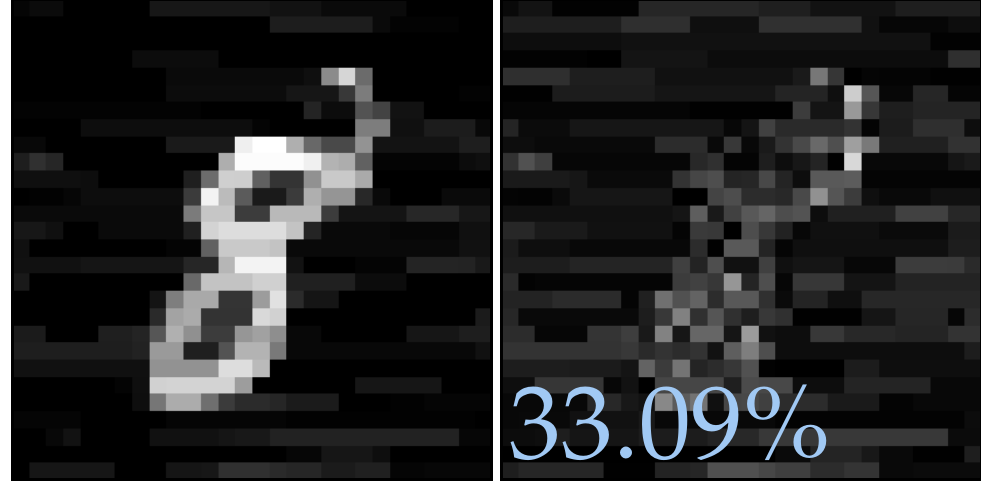} & \includegraphics[valign=c,width=0.14\textwidth]{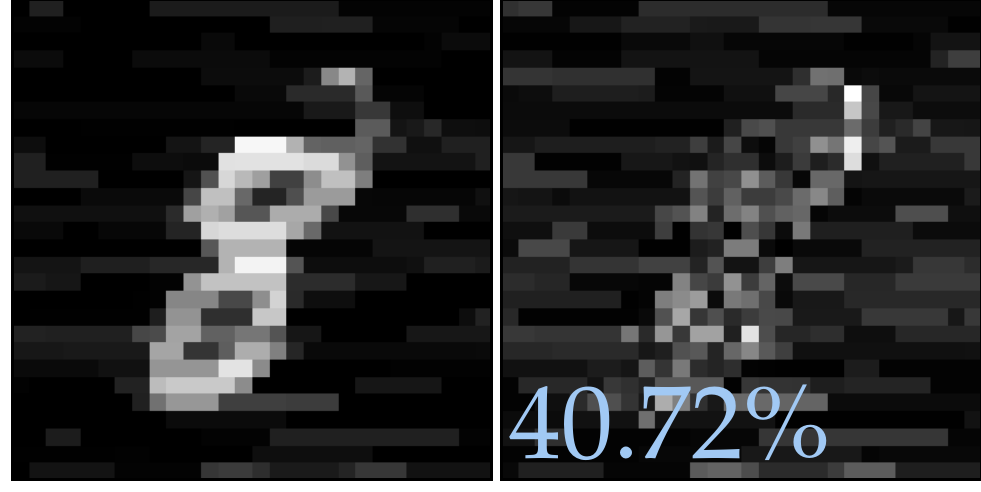} & \includegraphics[valign=c,width=0.14\textwidth]{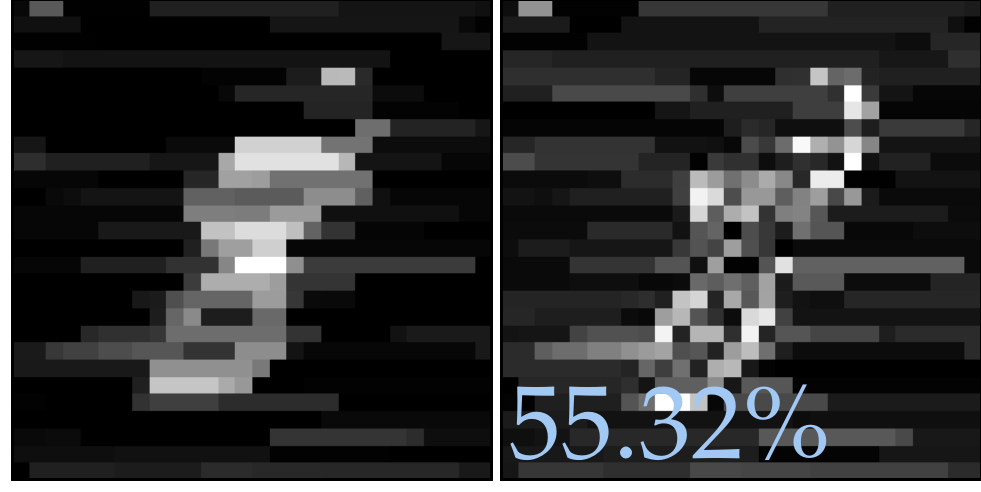} \\
    \rotatebox[origin=c]{90}{$\UNet$} &
    \includegraphics[valign=c,width=0.14\textwidth]{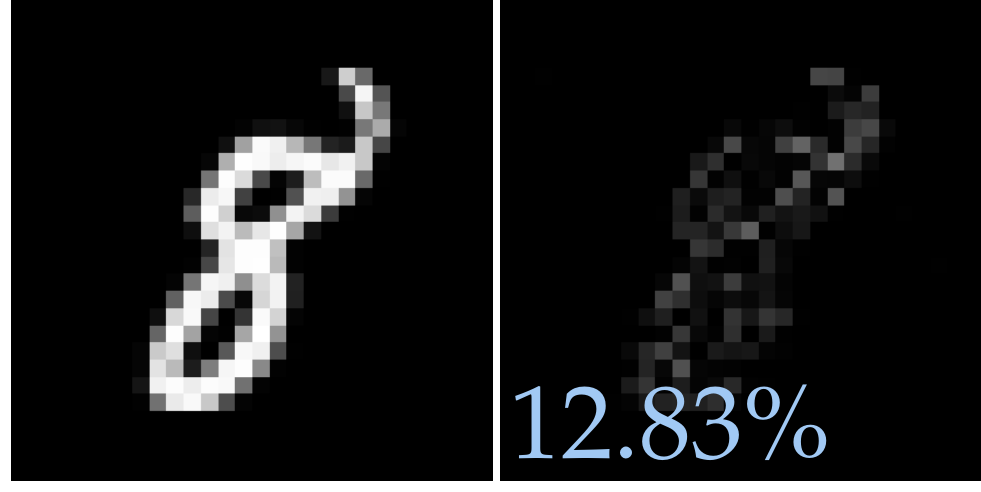} & \includegraphics[valign=c,width=0.14\textwidth]{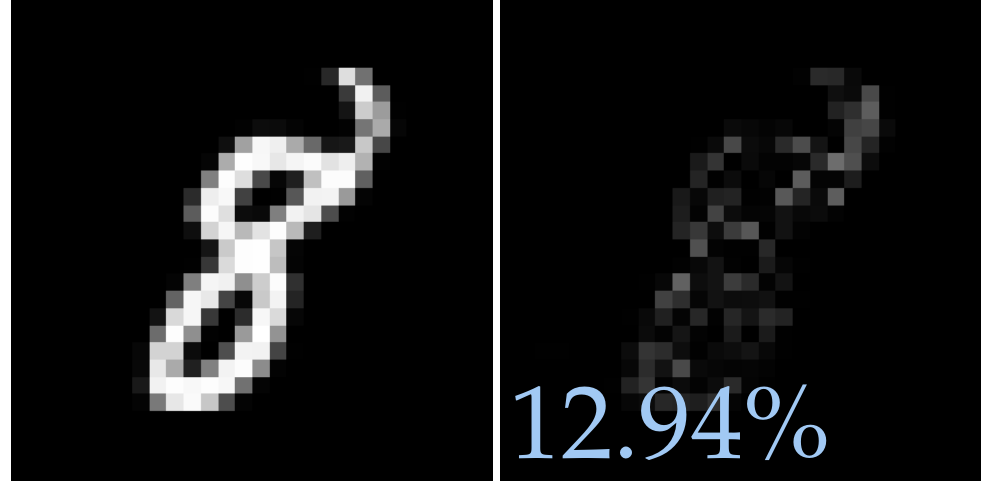} & \includegraphics[valign=c,width=0.14\textwidth]{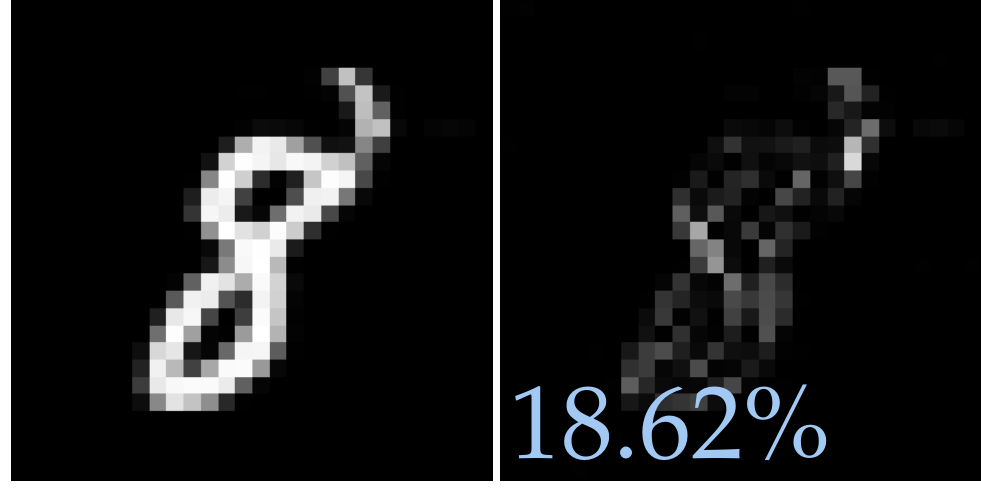} \\
    \rotatebox[origin=c]{90}{$\TiraFL$} &
    \includegraphics[valign=c,width=0.14\textwidth]{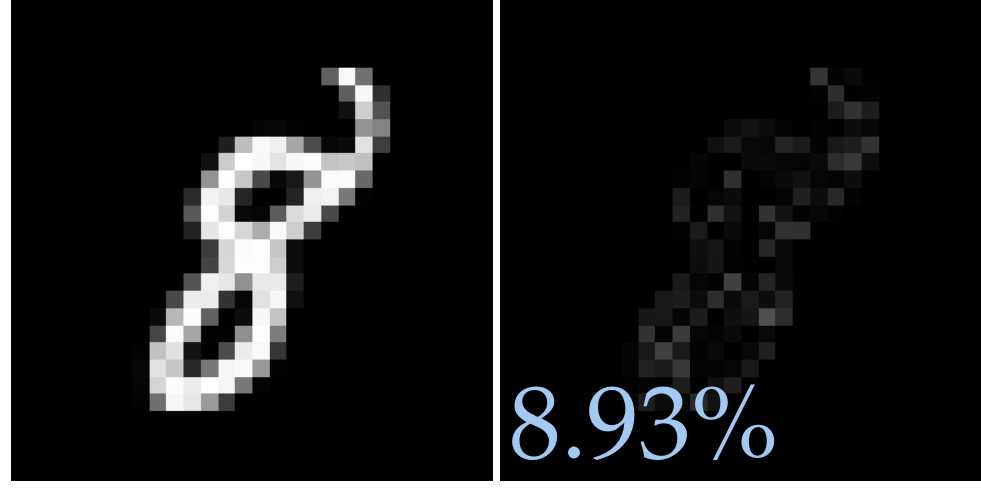} & \includegraphics[valign=c,width=0.14\textwidth]{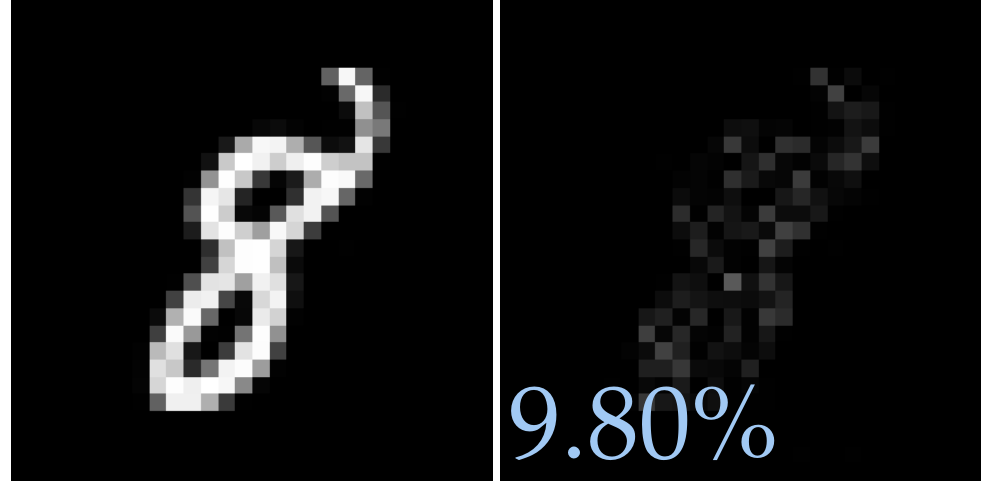} & \includegraphics[valign=c,width=0.14\textwidth]{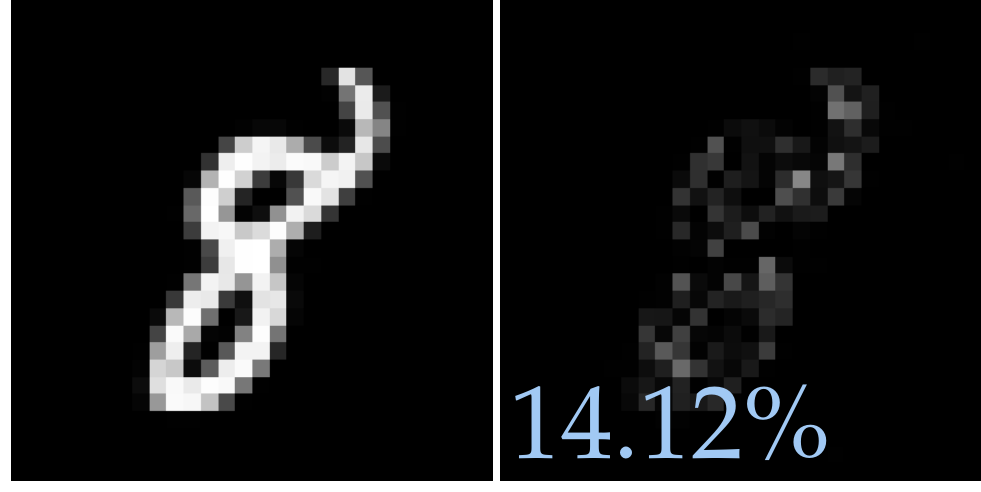} \\
    \rotatebox[origin=c]{90}{$\ItNet$} &
    \includegraphics[valign=c,width=0.14\textwidth]{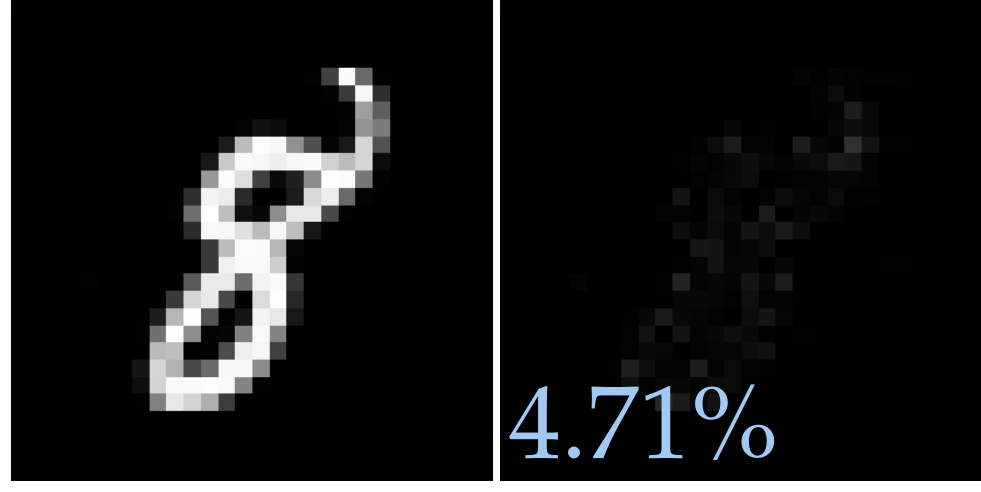} & \includegraphics[valign=c,width=0.14\textwidth]{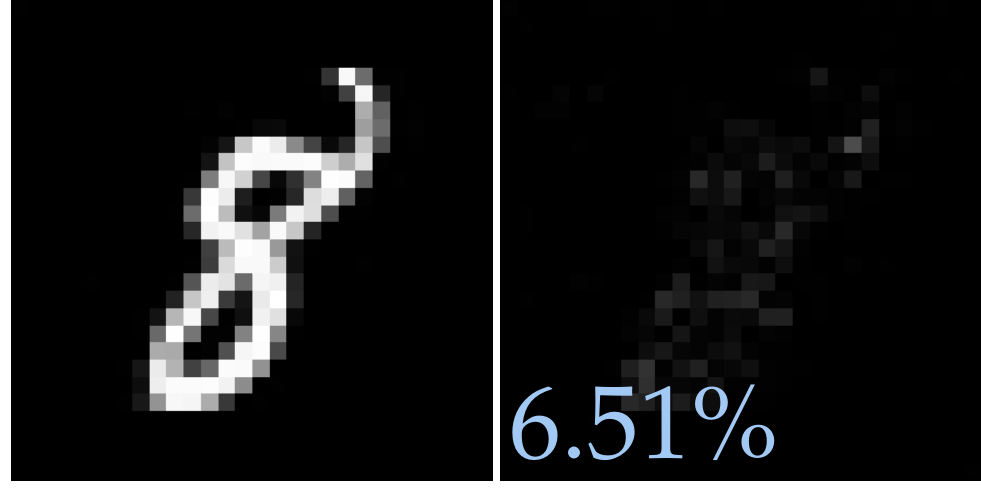} & \includegraphics[valign=c,width=0.14\textwidth]{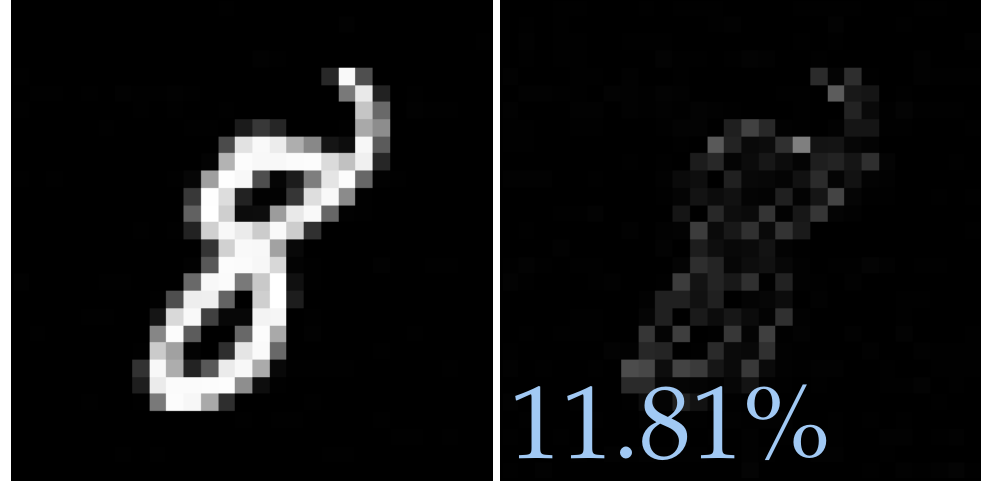} \\
 \end{tabular}

 &
 \scriptsize
  \setlength\tabcolsep{1pt}
   \begin{tabular}{lccc}
      &
    \includegraphics[valign=c,width=0.14\textwidth]{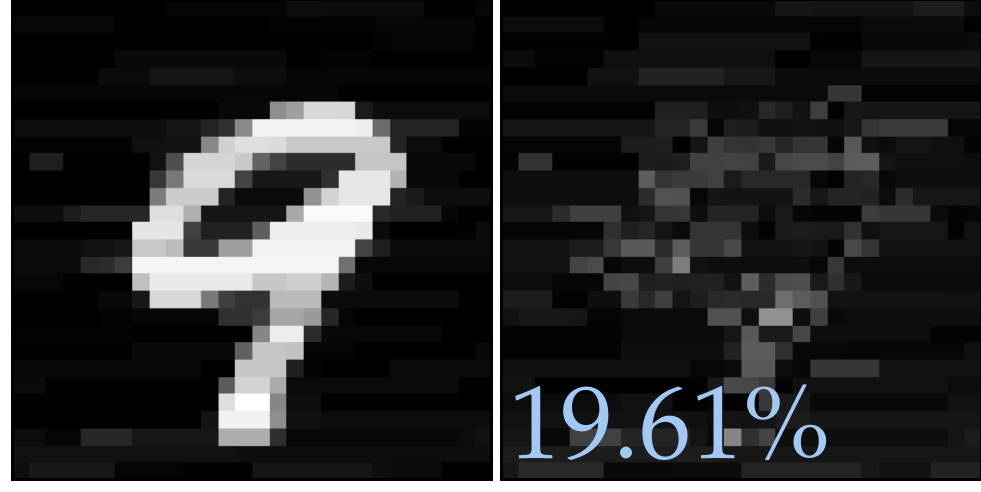} & \includegraphics[valign=c,width=0.14\textwidth]{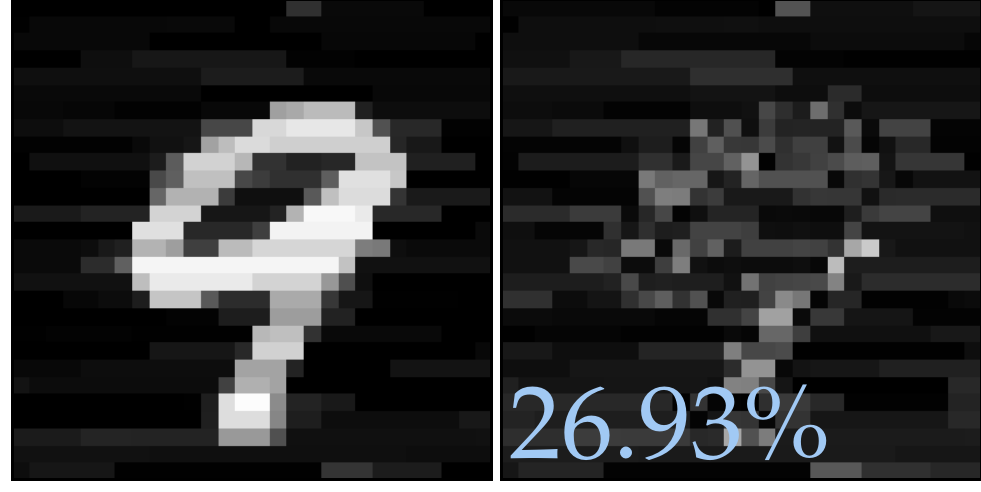} & \includegraphics[valign=c,width=0.14\textwidth]{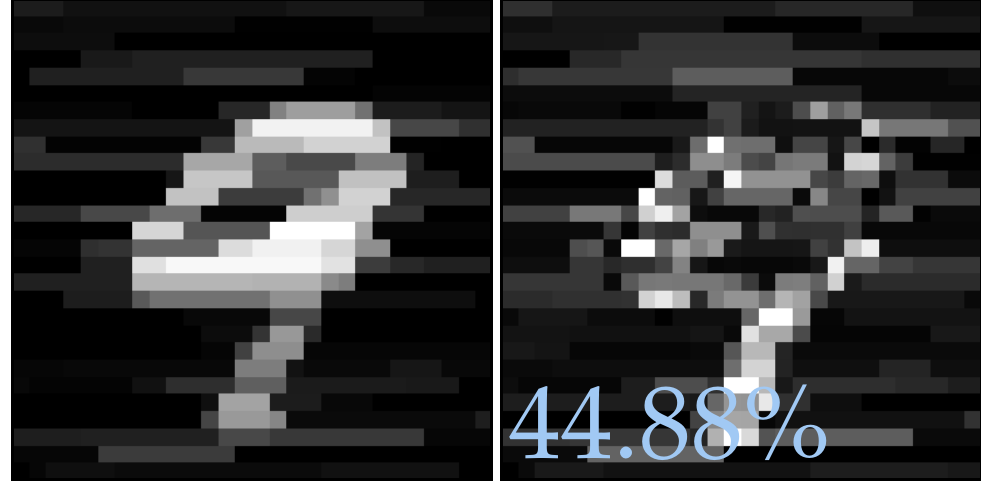} \\
     &
    \includegraphics[valign=c,width=0.14\textwidth]{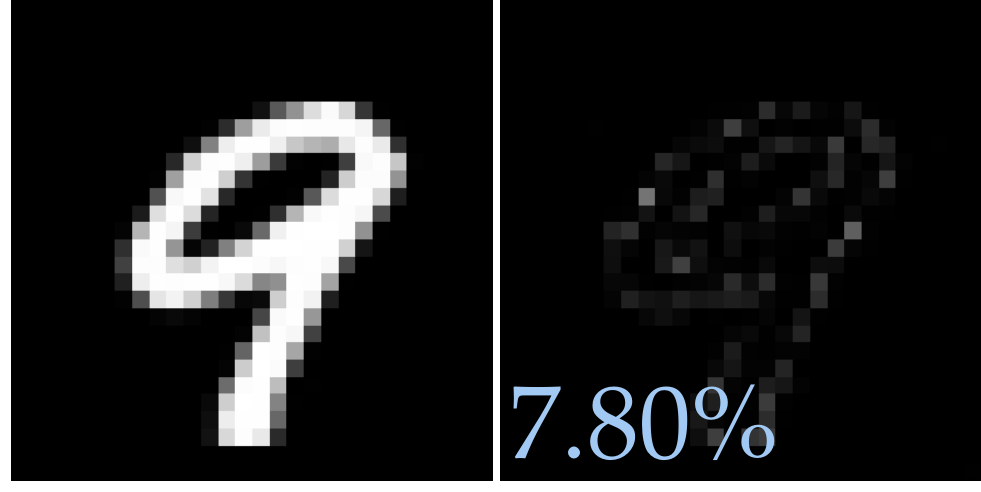} & \includegraphics[valign=c,width=0.14\textwidth]{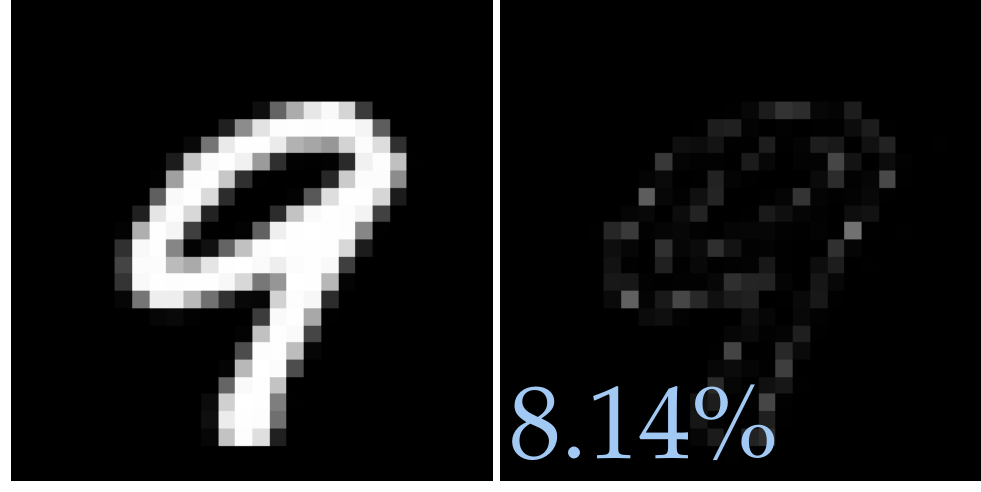} & \includegraphics[valign=c,width=0.14\textwidth]{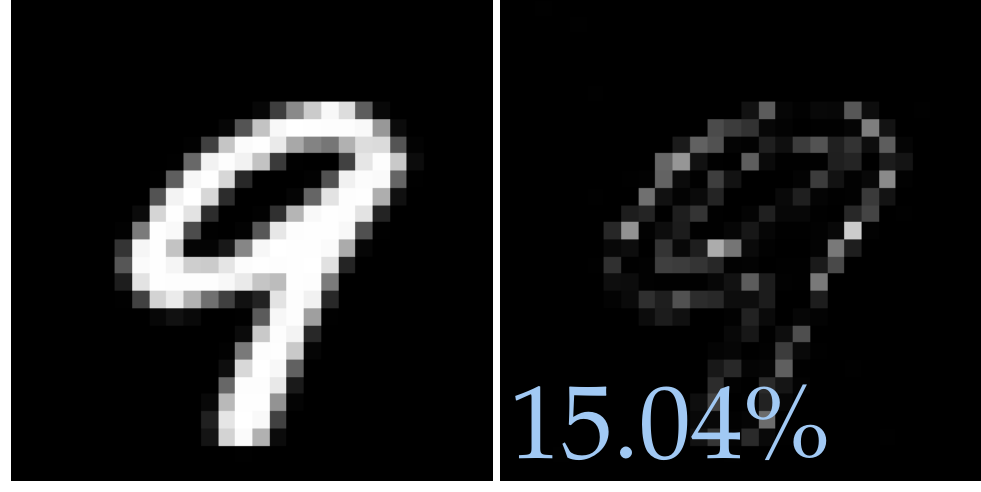} \\
     &
    \includegraphics[valign=c,width=0.14\textwidth]{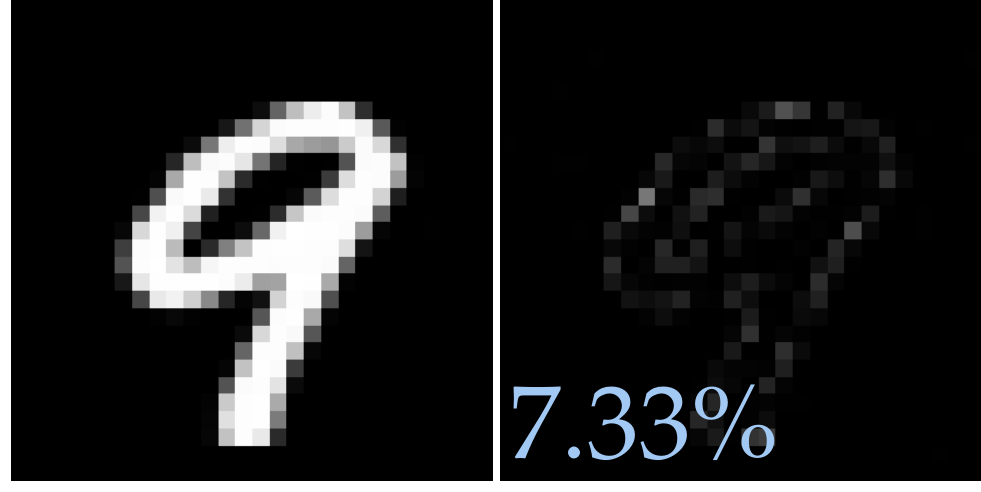} & \includegraphics[valign=c,width=0.14\textwidth]{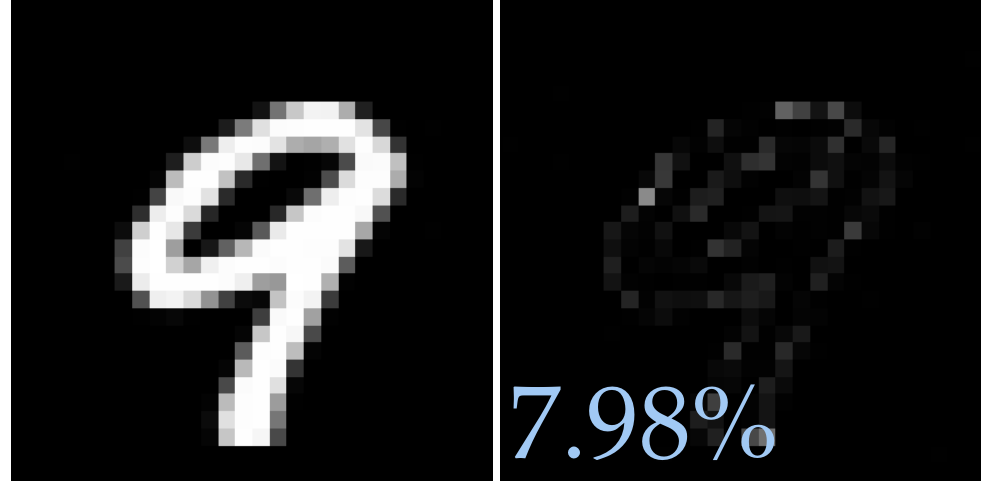} & \includegraphics[valign=c,width=0.14\textwidth]{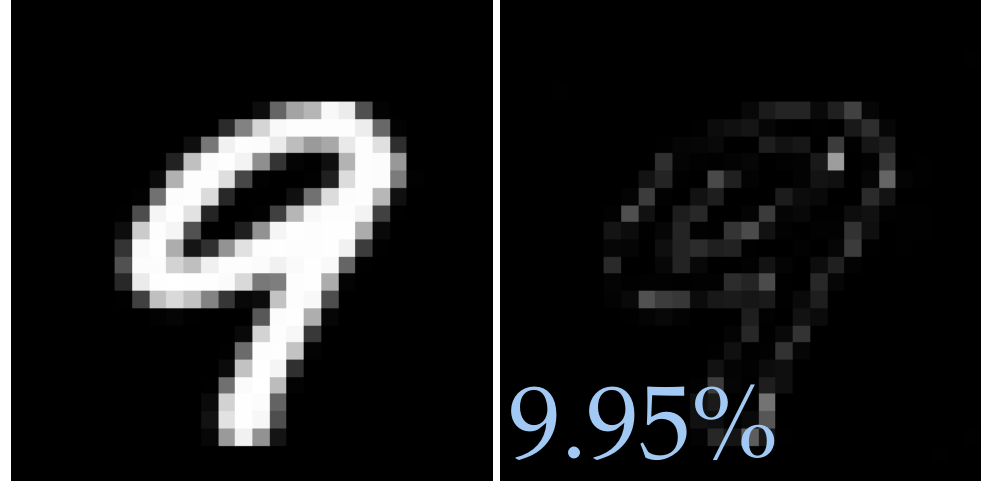} \\
     &
    \includegraphics[valign=c,width=0.14\textwidth]{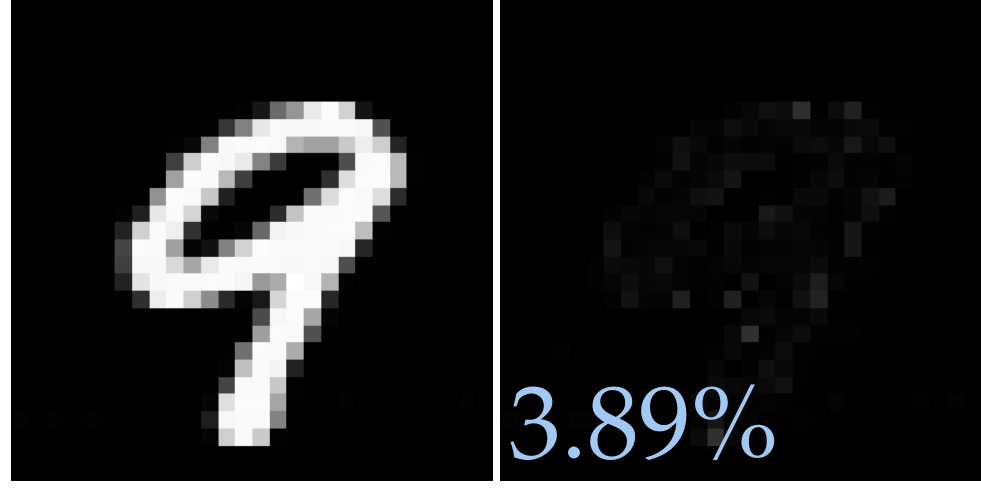} & \includegraphics[valign=c,width=0.14\textwidth]{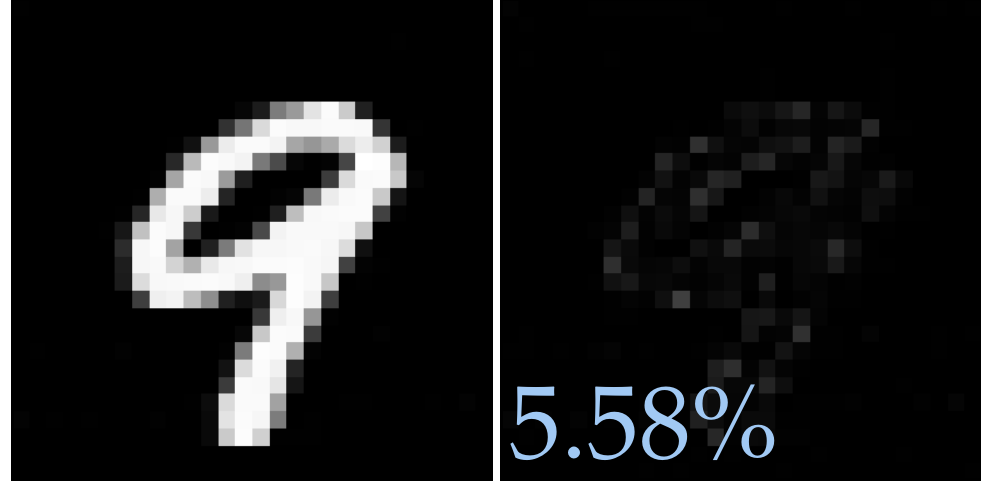} & \includegraphics[valign=c,width=0.14\textwidth]{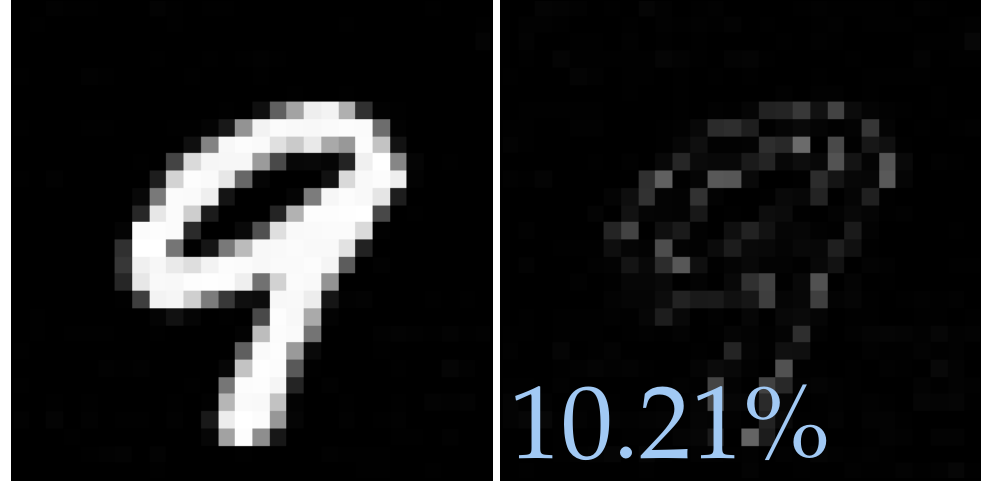} \\
 \end{tabular}
\end{tabular}
\caption{\textbf{Scenario~\refAtwo{} -- CS with MNIST.} Individual reconstructions of the digits from Fig.~\ref{fig:mnist:example_adv} and~\ref{fig:mnist:example_adv_supp} under Gaussian noise. The reconstructed digits and their error plots (with relative $\l{2}$-error) are displayed in the windows $[0,1]$ and $[0, 0.6]$, respectively. In favor of the more insightful noise level 25\%, we have omitted 2\%.}
\label{fig:mnist:example_gauss}
\end{figure}

\clearpage
\section{Supplementary Results for Case Study~B (Image Recovery of Phantom Ellipses)}
\label{sec:supp:csB}

\begin{table}[H]
	\robustify\bfseries
	\begin{tabular}{llS[separate-uncertainty]S[separate-uncertainty]S[separate-uncertainty]S[separate-uncertainty]S[separate-uncertainty]S[separate-uncertainty]S[separate-uncertainty]}
\toprule
\multicolumn{2}{l}{rel.~noise -- adversarial}  &                   {0.0\%} &                   {0.5\%} &                   {1.0\%} &                   {2.0\%} &                   {3.0\%} &                   {5.0\%} &                   {8.0\%} \\
\midrule
\multirow{3}{*}{TV$[\eta]$} & rel.~$\l{2}$-err. [\%] &   \bfseries 0.44 \pm 0.11 &   \bfseries 2.51 \pm 0.24 &             4.34 \pm 0.35 &             7.35 \pm 0.45 &             9.96 \pm 0.46 &            14.19 \pm 0.44 &            20.72 \pm 0.63 \\
      & PSNR &  \bfseries 60.00 \pm 3.26 &  \bfseries 44.73 \pm 2.20 &            39.98 \pm 2.16 &            35.38 \pm 2.00 &            32.74 \pm 1.91 &            29.66 \pm 1.82 &            26.37 \pm 1.93 \\
      & SSIM &   \bfseries 1.00 \pm 0.00 &             0.99 \pm 0.00 &             0.98 \pm 0.01 &             0.96 \pm 0.02 &             0.94 \pm 0.03 &   \bfseries 0.93 \pm 0.03 &   \bfseries 0.93 \pm 0.02 \\
\cline{1-9}
\multirow{3}{*}{UNet} & rel.~$\l{2}$-err. [\%] &             2.94 \pm 0.63 &             4.27 \pm 0.55 &             5.70 \pm 0.53 &             8.38 \pm 0.52 &            10.88 \pm 0.53 &            15.20 \pm 0.70 &            20.41 \pm 0.96 \\
      & PSNR &            43.52 \pm 2.06 &            40.15 \pm 1.79 &            37.61 \pm 1.72 &            34.24 \pm 1.67 &            31.97 \pm 1.66 &            29.07 \pm 1.69 &            26.51 \pm 1.78 \\
      & SSIM &             0.99 \pm 0.01 &             0.98 \pm 0.01 &             0.97 \pm 0.01 &             0.96 \pm 0.01 &             0.94 \pm 0.02 &             0.91 \pm 0.02 &             0.85 \pm 0.03 \\
\cline{1-9}
\multirow{3}{*}{UNetFL} & rel.~$\l{2}$-err. [\%] &             2.72 \pm 0.50 &             4.12 \pm 0.44 &             5.57 \pm 0.43 &             8.35 \pm 0.43 &            10.97 \pm 0.43 &            15.58 \pm 0.67 &            21.03 \pm 1.02 \\
      & PSNR &            44.13 \pm 2.29 &            40.46 \pm 1.96 &            37.80 \pm 1.87 &            34.28 \pm 1.76 &            31.90 \pm 1.73 &            28.85 \pm 1.76 &            26.25 \pm 1.82 \\
      & SSIM &             0.99 \pm 0.00 &             0.99 \pm 0.00 &             0.98 \pm 0.01 &             0.97 \pm 0.01 &             0.95 \pm 0.02 &             0.91 \pm 0.03 &             0.85 \pm 0.04 \\
\cline{1-9}
\multirow{3}{*}{Tira} & rel.~$\l{2}$-err. [\%] &             1.74 \pm 0.37 &             3.33 \pm 0.33 &             4.85 \pm 0.37 &             7.73 \pm 0.42 &            10.42 \pm 0.53 &            15.01 \pm 0.74 &            20.39 \pm 0.99 \\
      & PSNR &            48.05 \pm 2.50 &            42.27 \pm 1.93 &            39.01 \pm 1.85 &            34.94 \pm 1.80 &            32.35 \pm 1.76 &            29.18 \pm 1.76 &            26.52 \pm 1.83 \\
      & SSIM &             1.00 \pm 0.00 &             0.99 \pm 0.00 &   \bfseries 0.99 \pm 0.00 &             0.97 \pm 0.01 &             0.95 \pm 0.02 &             0.91 \pm 0.03 &             0.87 \pm 0.04 \\
\cline{1-9}
\multirow{3}{*}{TiraFL} & rel.~$\l{2}$-err. [\%] &             1.75 \pm 0.39 &             3.42 \pm 0.34 &             4.94 \pm 0.41 &             7.82 \pm 0.44 &            10.54 \pm 0.51 &            15.14 \pm 0.69 &            20.55 \pm 0.95 \\
      & PSNR &            48.05 \pm 2.58 &            42.05 \pm 1.93 &            38.85 \pm 1.86 &            34.85 \pm 1.83 &            32.24 \pm 1.80 &            29.10 \pm 1.75 &            26.45 \pm 1.81 \\
      & SSIM &             1.00 \pm 0.00 &   \bfseries 0.99 \pm 0.00 &             0.99 \pm 0.01 &             0.97 \pm 0.01 &             0.95 \pm 0.02 &             0.91 \pm 0.03 &             0.87 \pm 0.04 \\
\cline{1-9}
\multirow{3}{*}{ItNet} & rel.~$\l{2}$-err. [\%] &             1.45 \pm 0.29 &             2.81 \pm 0.28 &   \bfseries 4.21 \pm 0.32 &   \bfseries 6.87 \pm 0.37 &   \bfseries 9.37 \pm 0.40 &  \bfseries 13.65 \pm 0.49 &  \bfseries 18.98 \pm 0.65 \\
      & PSNR &            49.63 \pm 1.80 &            43.76 \pm 1.62 &  \bfseries 40.23 \pm 1.64 &  \bfseries 35.97 \pm 1.63 &  \bfseries 33.27 \pm 1.67 &  \bfseries 30.00 \pm 1.64 &  \bfseries 27.13 \pm 1.65 \\
      & SSIM &             0.99 \pm 0.00 &             0.99 \pm 0.00 &             0.98 \pm 0.00 &   \bfseries 0.97 \pm 0.01 &   \bfseries 0.96 \pm 0.01 &             0.92 \pm 0.02 &             0.88 \pm 0.03 \\
\bottomrule
\end{tabular}

	\caption{\textbf{Scenario~\refBone{} -- Fourier meas.~with ellipses.} A numerical representation of the results of Fig.~\ref{fig:ellipses:table}(c), including the additional methods $\UNetFL$ and $\Tira$. The best relative error/PSNR/SSIM per noise level is highlighted in bold. Note that the high SSIM values for $\TV[\noisebnd]$ for 5\% and 8\% can be explained by the fact that adversarial perturbations for $\TV[\noisebnd]$ cause point-like artifacts, see the zoomed region in Fig.~\ref{fig:ellipses:example_adv}. In contrast to the PSNR, the SSIM seems to be less sensitive to such types of errors.}
	\label{tab:ellipses:table_adv}
\end{table}

\begin{table}[H]
	\robustify\bfseries
	\begin{tabular}{llS[separate-uncertainty]S[separate-uncertainty]S[separate-uncertainty]S[separate-uncertainty]S[separate-uncertainty]S[separate-uncertainty]S[separate-uncertainty]}
\toprule
\multicolumn{2}{l}{rel.~noise -- Gaussian} & {0.0\%} &                   {0.5\%} &                   {1.0\%} &                   {2.0\%} &                   {3.0\%} &                   {5.0\%} &                   {8.0\%} \\
\midrule
\multirow{3}{*}{TV$[\eta]$} & rel.~$\l{2}$-err. [\%] &   \bfseries 0.44 \pm 0.11 &   \bfseries 0.80 \pm 0.12 &   \bfseries 1.18 \pm 0.15 &             1.93 \pm 0.25 &             2.64 \pm 0.35 &             3.90 \pm 0.50 &             5.52 \pm 0.58 \\
      & PSNR &  \bfseries 60.00 \pm 3.26 &  \bfseries 54.73 \pm 2.39 &  \bfseries 51.29 \pm 2.44 &            47.08 \pm 2.46 &            44.33 \pm 2.47 &            40.94 \pm 2.25 &            37.91 \pm 2.17 \\
      & SSIM &   \bfseries 1.00 \pm 0.00 &   \bfseries 1.00 \pm 0.00 &             0.99 \pm 0.00 &             0.98 \pm 0.01 &             0.96 \pm 0.02 &             0.96 \pm 0.02 &             0.92 \pm 0.04 \\
\cline{1-9}
\multirow{3}{*}{UNet} & rel.~$\l{2}$-err. [\%] &             2.94 \pm 0.63 &             2.94 \pm 0.63 &             2.95 \pm 0.63 &             3.00 \pm 0.63 &             3.07 \pm 0.62 &             3.28 \pm 0.61 &             3.72 \pm 0.61 \\
      & PSNR &            43.52 \pm 2.06 &            43.50 \pm 2.06 &            43.47 \pm 2.06 &            43.33 \pm 2.07 &            43.12 \pm 2.07 &            42.51 \pm 2.09 &            41.41 \pm 2.12 \\
      & SSIM &             0.99 \pm 0.01 &             0.99 \pm 0.01 &             0.99 \pm 0.01 &             0.99 \pm 0.01 &             0.99 \pm 0.01 &             0.99 \pm 0.00 &             0.98 \pm 0.00 \\
\cline{1-9}
\multirow{3}{*}{UNetFL} & rel.~$\l{2}$-err. [\%] &             2.72 \pm 0.50 &             2.73 \pm 0.50 &             2.74 \pm 0.50 &             2.80 \pm 0.51 &             2.88 \pm 0.51 &             3.13 \pm 0.53 &             3.61 \pm 0.55 \\
      & PSNR &            44.13 \pm 2.29 &            44.12 \pm 2.29 &            44.07 \pm 2.29 &            43.90 \pm 2.28 &            43.64 \pm 2.28 &            42.92 \pm 2.26 &            41.65 \pm 2.20 \\
      & SSIM &             0.99 \pm 0.00 &             0.99 \pm 0.00 &             0.99 \pm 0.00 &             0.99 \pm 0.00 &             0.99 \pm 0.00 &             0.99 \pm 0.00 &             0.99 \pm 0.00 \\
\cline{1-9}
\multirow{3}{*}{Tira} & rel.~$\l{2}$-err. [\%] &             1.74 \pm 0.37 &             1.75 \pm 0.37 &             1.77 \pm 0.37 &             1.83 \pm 0.38 &             1.92 \pm 0.38 &             2.19 \pm 0.40 &             2.70 \pm 0.45 \\
      & PSNR &            48.05 \pm 2.50 &            48.02 \pm 2.50 &            47.95 \pm 2.49 &            47.65 \pm 2.48 &            47.20 \pm 2.46 &            46.03 \pm 2.41 &            44.19 \pm 2.36 \\
      & SSIM &             1.00 \pm 0.00 &             1.00 \pm 0.00 &             1.00 \pm 0.00 &             1.00 \pm 0.00 &             1.00 \pm 0.00 &             0.99 \pm 0.00 &             0.99 \pm 0.00 \\
\cline{1-9}
\multirow{3}{*}{TiraFL} & rel.~$\l{2}$-err. [\%] &             1.75 \pm 0.39 &             1.76 \pm 0.39 &             1.77 \pm 0.39 &             1.83 \pm 0.40 &             1.92 \pm 0.40 &             2.19 \pm 0.42 &             2.70 \pm 0.46 \\
      & PSNR &            48.05 \pm 2.58 &            48.02 \pm 2.57 &            47.94 \pm 2.57 &            47.66 \pm 2.55 &            47.20 \pm 2.54 &            46.04 \pm 2.47 &            44.21 \pm 2.39 \\
      & SSIM &             1.00 \pm 0.00 &             1.00 \pm 0.00 &   \bfseries 1.00 \pm 0.00 &   \bfseries 1.00 \pm 0.00 &   \bfseries 1.00 \pm 0.00 &   \bfseries 1.00 \pm 0.00 &   \bfseries 0.99 \pm 0.00 \\
\cline{1-9}
\multirow{3}{*}{ItNet} & rel.~$\l{2}$-err. [\%] &             1.45 \pm 0.29 &             1.46 \pm 0.29 &             1.47 \pm 0.29 &   \bfseries 1.53 \pm 0.30 &   \bfseries 1.62 \pm 0.32 &   \bfseries 1.92 \pm 0.37 &   \bfseries 2.50 \pm 0.45 \\
      & PSNR &            49.63 \pm 1.80 &            49.60 \pm 1.80 &            49.52 \pm 1.82 &  \bfseries 49.19 \pm 1.89 &  \bfseries 48.65 \pm 1.98 &  \bfseries 47.18 \pm 2.18 &  \bfseries 44.89 \pm 2.27 \\
      & SSIM &             0.99 \pm 0.00 &             0.99 \pm 0.00 &             0.99 \pm 0.00 &             0.99 \pm 0.00 &             0.99 \pm 0.00 &             0.99 \pm 0.00 &             0.99 \pm 0.00 \\
\bottomrule
\end{tabular}

	\caption{\textbf{Scenario~\refBone{} -- Fourier meas.~with ellipses.} A numerical representation of the results of Fig.~\ref{fig:ellipses:table}(d), including the additional methods $\UNetFL$ and $\Tira$. The best relative error/PSNR/SSIM per noise level is highlighted in bold.}
	\label{tab:ellipses:table_ref}
\end{table}

\begin{figure}[H]
	\centering
	\scriptsize
	\begin{tabular}{l@{\,}c@{\,}c@{\,}c@{\,}c}
		& noiseless & 1\% rel.~noise -- adv. & 3\% rel.~noise -- adv. & 8\% rel.~noise -- adv. \\
		\rotatebox[origin=c]{90}{$\TV[\noisebnd]$} &
		\includegraphics[valign=c,width=0.15\textwidth]{{ellipses/results/attacks/fig_example_S66_adv_tv_0.00e+00}.pdf} &
		\includegraphics[valign=c,width=0.15\textwidth]{{ellipses/results/attacks/fig_example_S66_adv_tv_1.00e-02}.pdf} & \includegraphics[valign=c,width=0.15\textwidth]{{ellipses/results/attacks/fig_example_S66_adv_tv_3.00e-02}.pdf} & \includegraphics[valign=c,width=0.15\textwidth]{{ellipses/results/attacks/fig_example_S66_adv_tv_8.00e-02}.pdf} \\
		&
		\includegraphics[valign=c,width=0.15\textwidth]{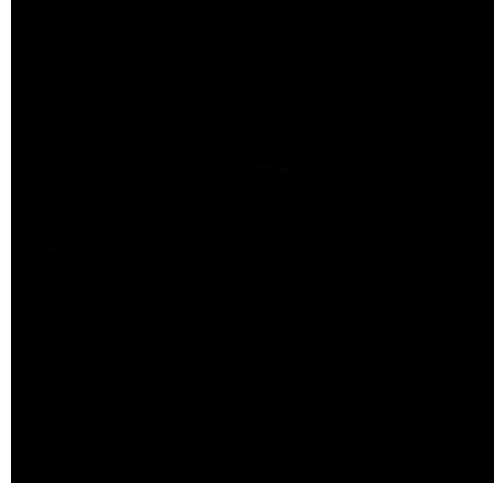} &
		\includegraphics[valign=c,width=0.15\textwidth]{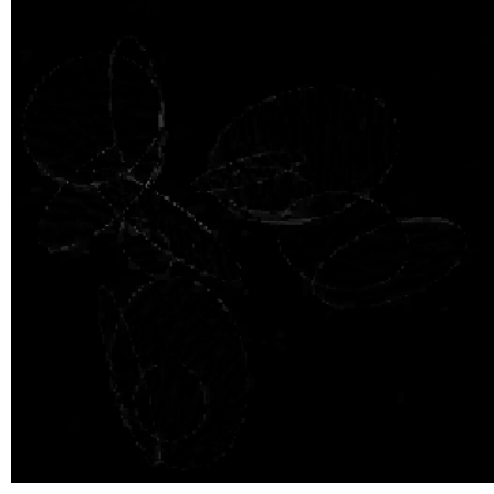} & \includegraphics[valign=c,width=0.15\textwidth]{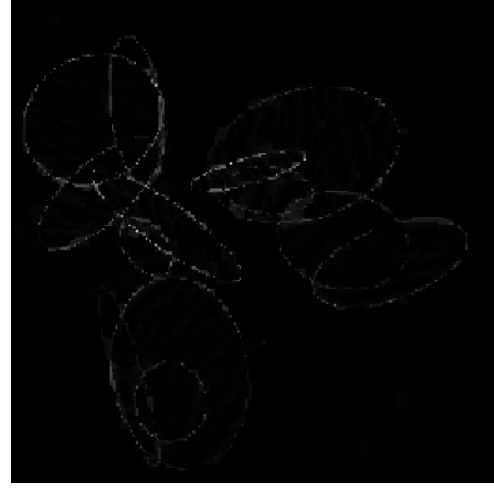} & \includegraphics[valign=c,width=0.15\textwidth]{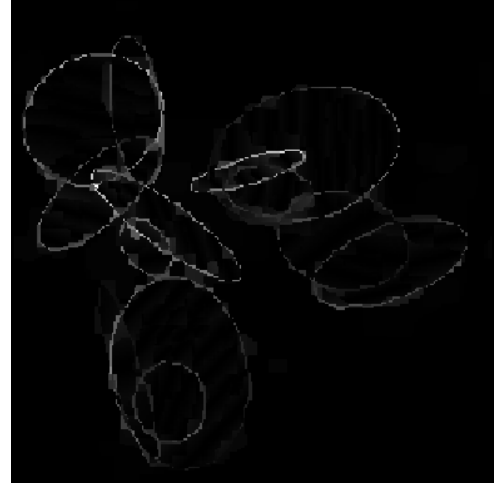} \\ \\

		\rotatebox[origin=c]{90}{$\UNet$} &
		\includegraphics[valign=c,width=0.15\textwidth]{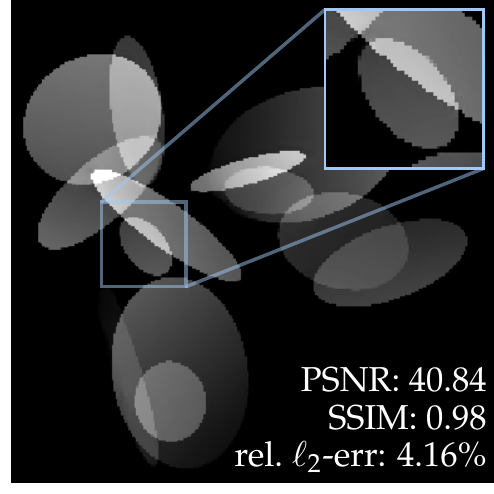} &
		\includegraphics[valign=c,width=0.15\textwidth]{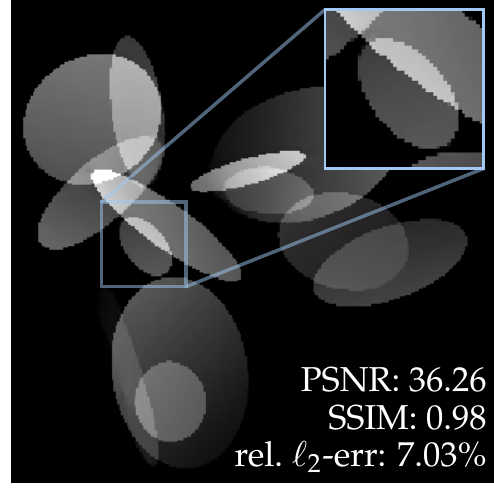} & \includegraphics[valign=c,width=0.15\textwidth]{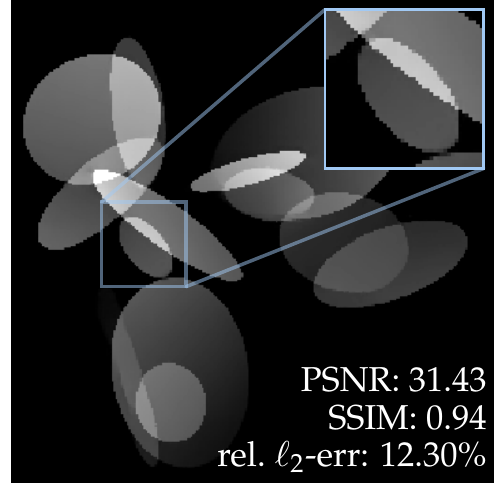} & \includegraphics[valign=c,width=0.15\textwidth]{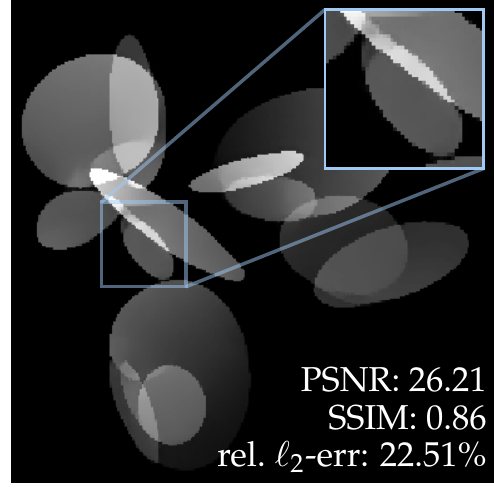} \\
		&
		\includegraphics[valign=c,width=0.15\textwidth]{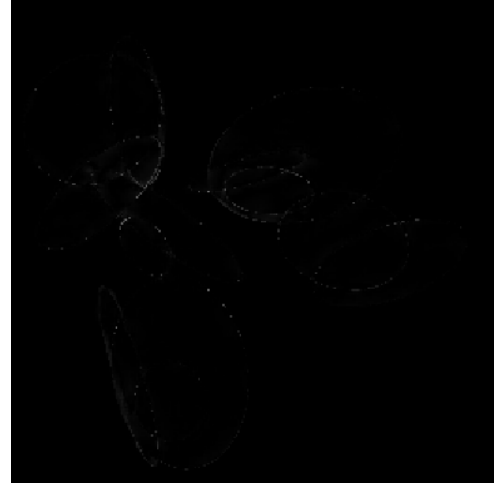} &
		\includegraphics[valign=c,width=0.15\textwidth]{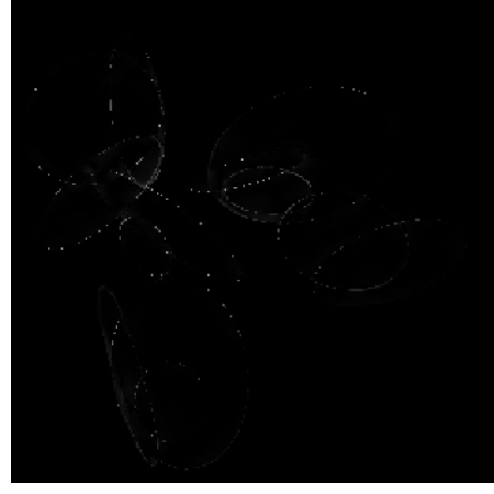} & \includegraphics[valign=c,width=0.15\textwidth]{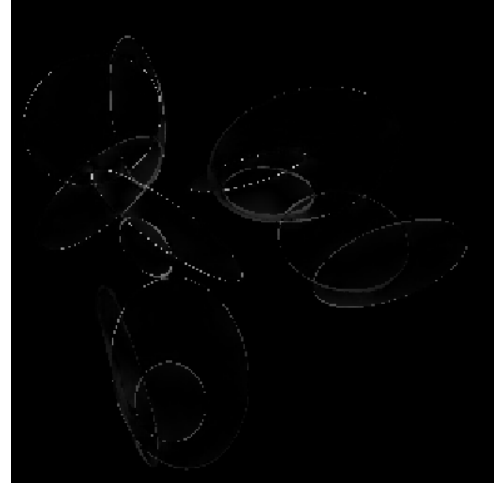} & \includegraphics[valign=c,width=0.15\textwidth]{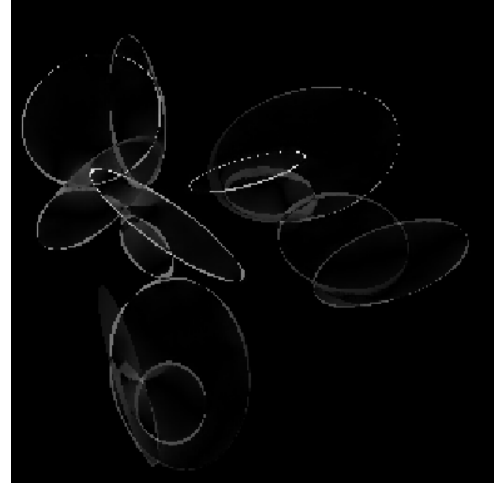} \\ \\

		\rotatebox[origin=c]{90}{$\TiraFL$} &
		\includegraphics[valign=c,width=0.15\textwidth]{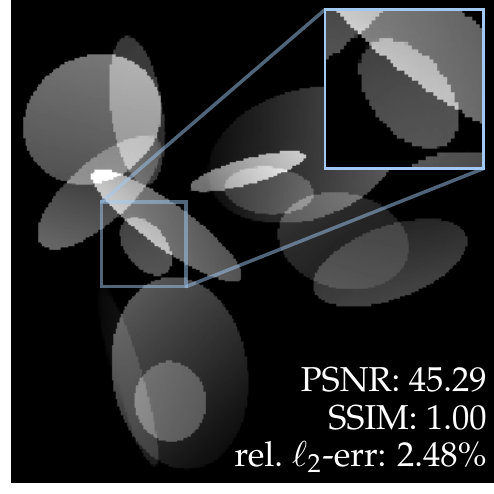} &
		\includegraphics[valign=c,width=0.15\textwidth]{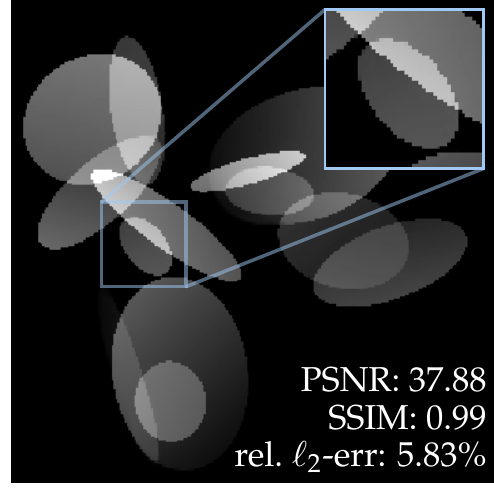} & \includegraphics[valign=c,width=0.15\textwidth]{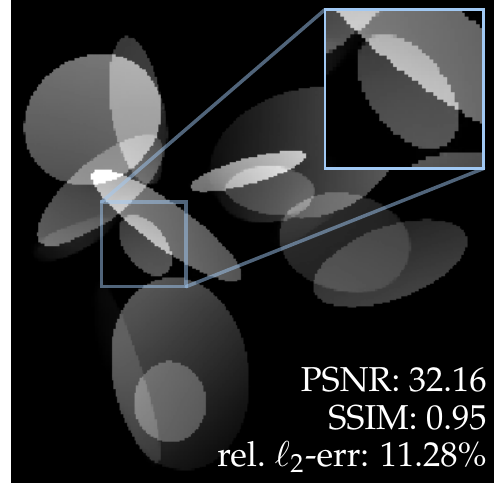} & \includegraphics[valign=c,width=0.15\textwidth]{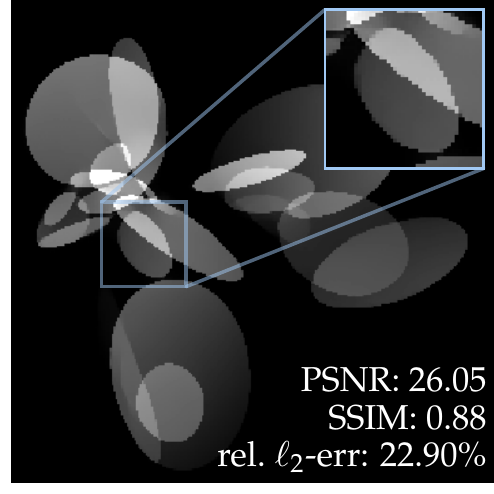} \\
		&
		\includegraphics[valign=c,width=0.15\textwidth]{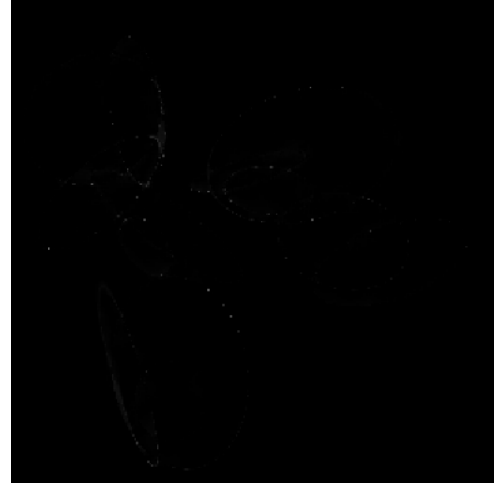} &
		\includegraphics[valign=c,width=0.15\textwidth]{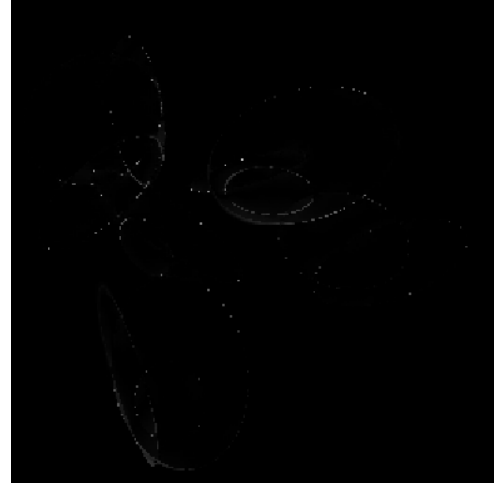} & \includegraphics[valign=c,width=0.15\textwidth]{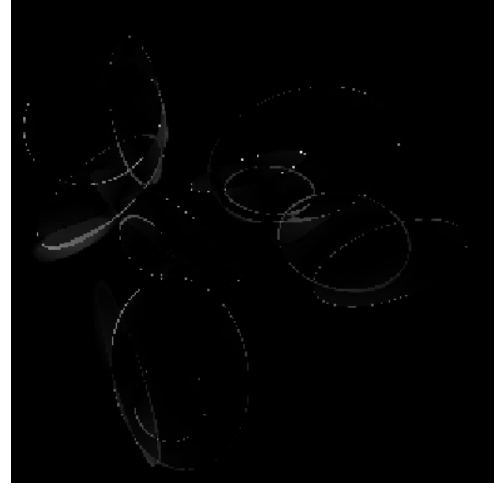} & \includegraphics[valign=c,width=0.15\textwidth]{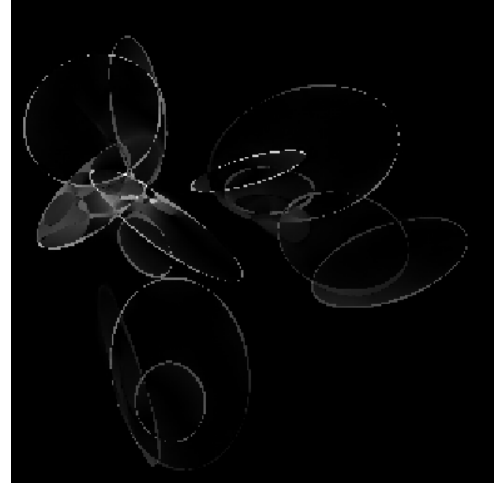} \\ \\

		\rotatebox[origin=c]{90}{$\ItNet$} &
		\includegraphics[valign=c,width=0.15\textwidth]{{ellipses/results/attacks/fig_example_S66_adv_unet_it_jit_0.00e+00}.pdf} &
		\includegraphics[valign=c,width=0.15\textwidth]{{ellipses/results/attacks/fig_example_S66_adv_unet_it_jit_1.00e-02}.pdf} & \includegraphics[valign=c,width=0.15\textwidth]{{ellipses/results/attacks/fig_example_S66_adv_unet_it_jit_3.00e-02}.pdf} & \includegraphics[valign=c,width=0.15\textwidth]{{ellipses/results/attacks/fig_example_S66_adv_unet_it_jit_8.00e-02}.pdf} \\
		&
		\includegraphics[valign=c,width=0.15\textwidth]{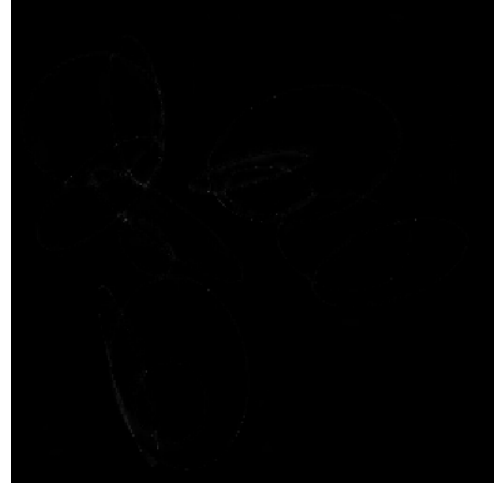} &
		\includegraphics[valign=c,width=0.15\textwidth]{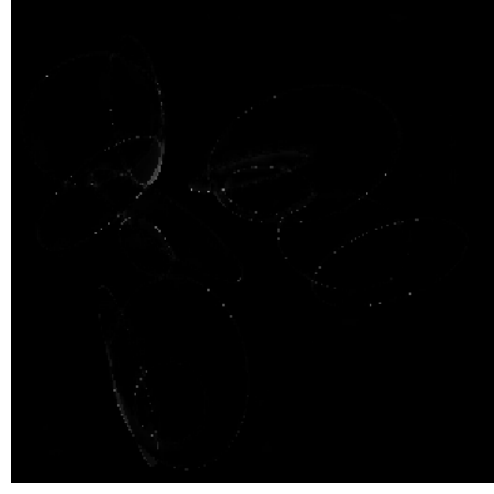} & \includegraphics[valign=c,width=0.15\textwidth]{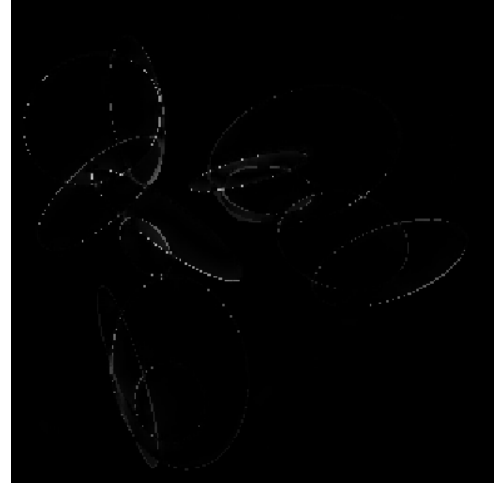} & \includegraphics[valign=c,width=0.15\textwidth]{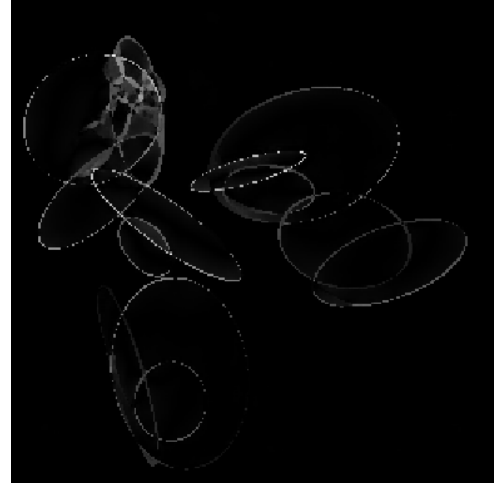}
	\end{tabular}
	\caption{\textbf{Scenario~\refBone{} -- Fourier meas.~with ellipses.} Individual reconstructions of the image from Fig.~\ref{fig:ellipses:example_adv} for different levels of adversarial noise. The reconstructed images are displayed in the window $[0,0.9]$, which is also used for the computation of the PSNR and SSIM. The error plots shown below each reconstruction are displayed in the window $[0, 0.6]$.}
	\label{fig:ellipses:example_adv_supp}
\end{figure}

\begin{figure}[H]
 \centering
 \scriptsize
\begin{tabular}{l@{\,}c@{\,}c@{\,}c@{\,}c}
 & 1\% rel.~noise -- Gauss. & 3\% rel.~noise -- Gauss. & 8\% rel.~noise -- Gauss. & 16\% rel.~noise -- Gauss. \\
\rotatebox[origin=c]{90}{$\TV[\noisebnd]$} &
\includegraphics[valign=c,width=0.15\textwidth]{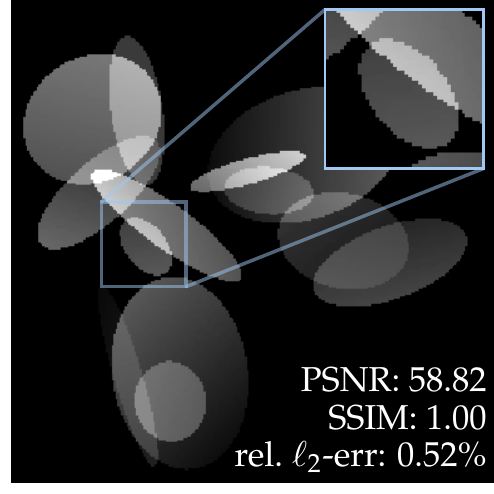} &
\includegraphics[valign=c,width=0.15\textwidth]{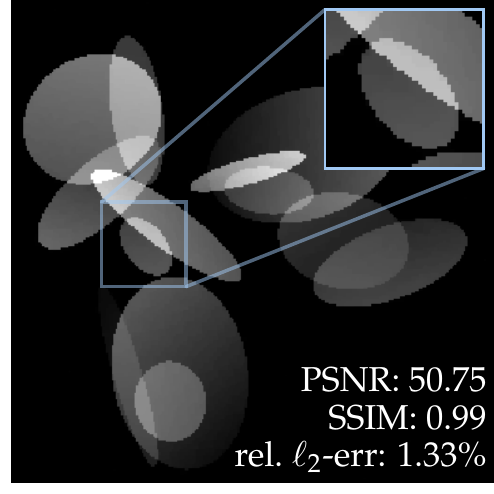} & \includegraphics[valign=c,width=0.15\textwidth]{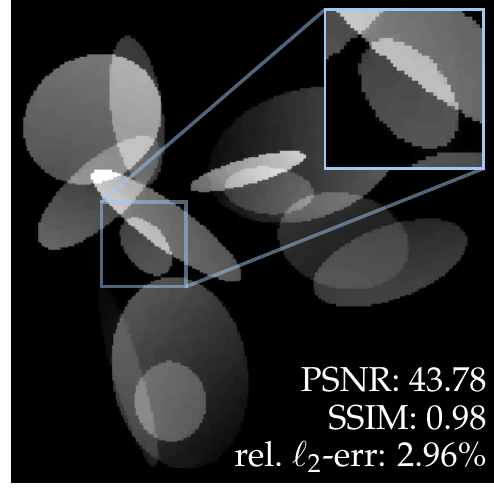} & \includegraphics[valign=c,width=0.15\textwidth]{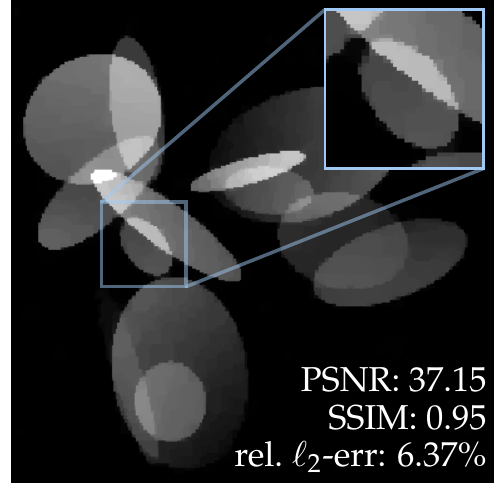} \\
&
\includegraphics[valign=c,width=0.15\textwidth]{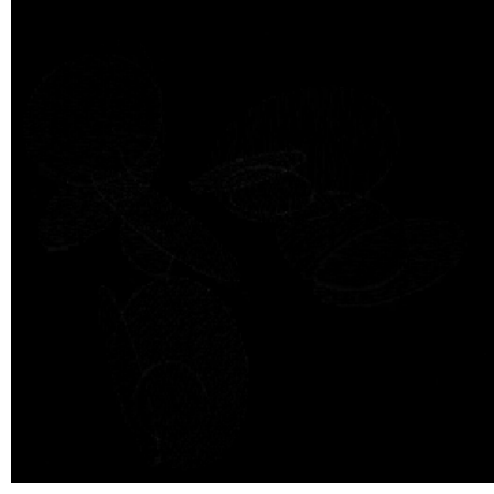} &
\includegraphics[valign=c,width=0.15\textwidth]{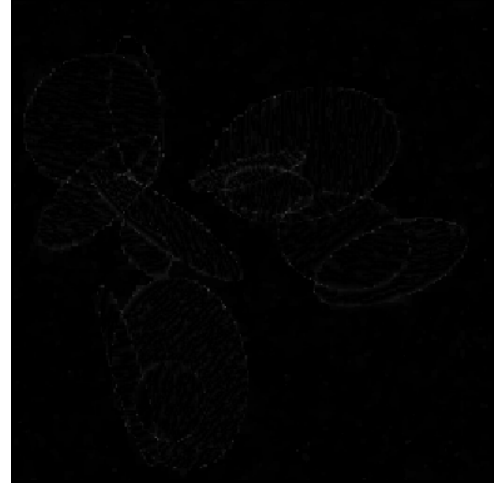} & \includegraphics[valign=c,width=0.15\textwidth]{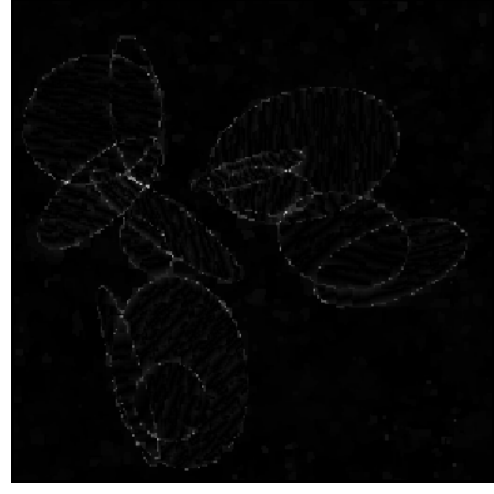} & \includegraphics[valign=c,width=0.15\textwidth]{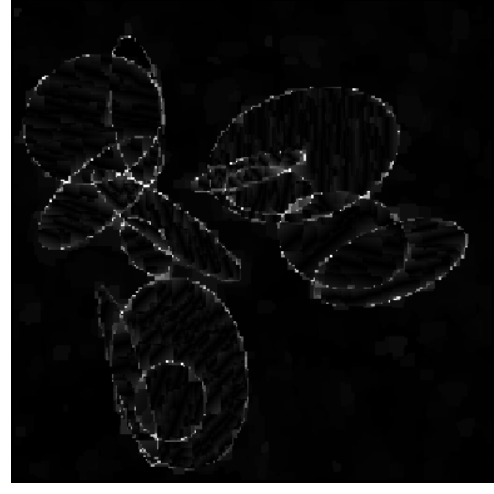} \\ \\

\rotatebox[origin=c]{90}{$\UNet$} &
\includegraphics[valign=c,width=0.15\textwidth]{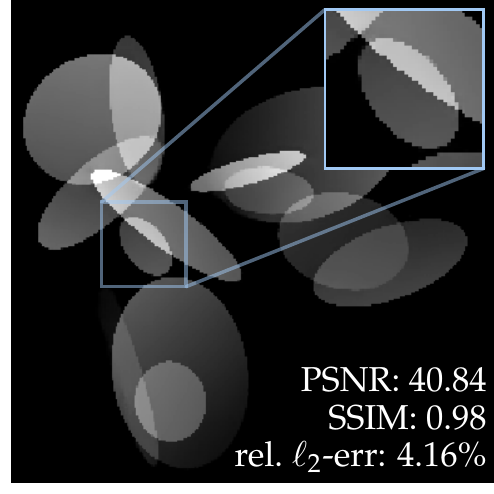} &
\includegraphics[valign=c,width=0.15\textwidth]{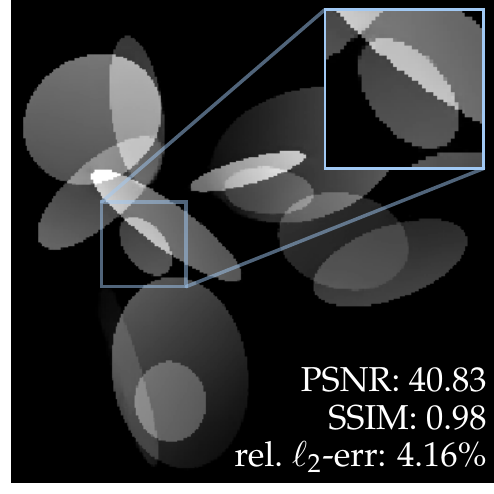} & \includegraphics[valign=c,width=0.15\textwidth]{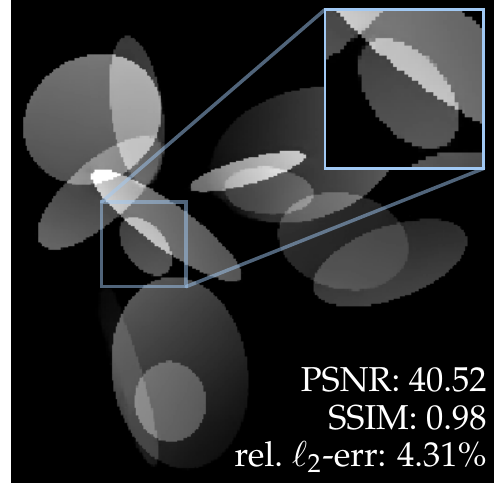} & \includegraphics[valign=c,width=0.15\textwidth]{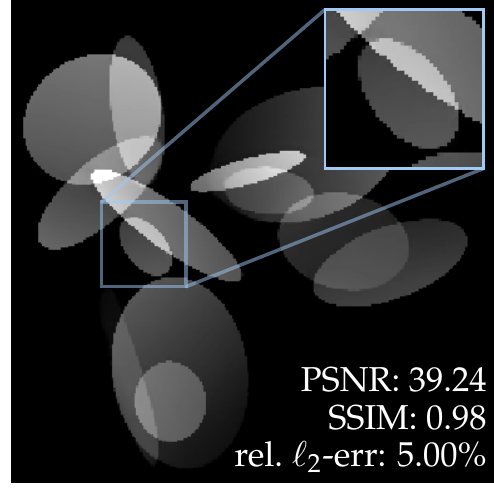} \\
&
\includegraphics[valign=c,width=0.15\textwidth]{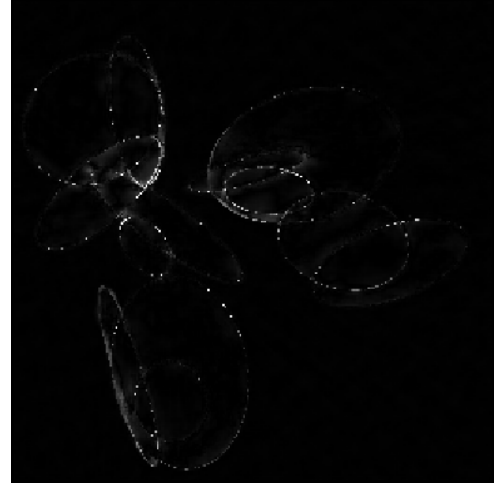} &
\includegraphics[valign=c,width=0.15\textwidth]{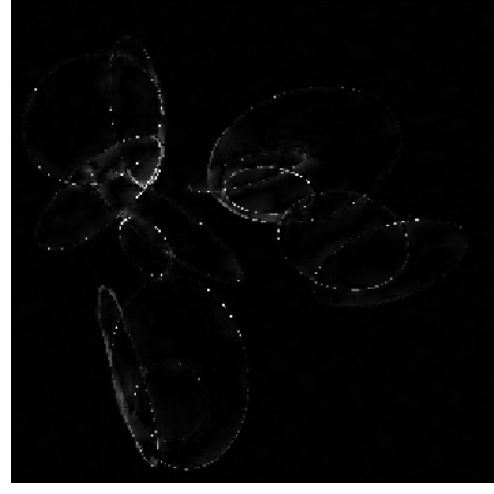} & \includegraphics[valign=c,width=0.15\textwidth]{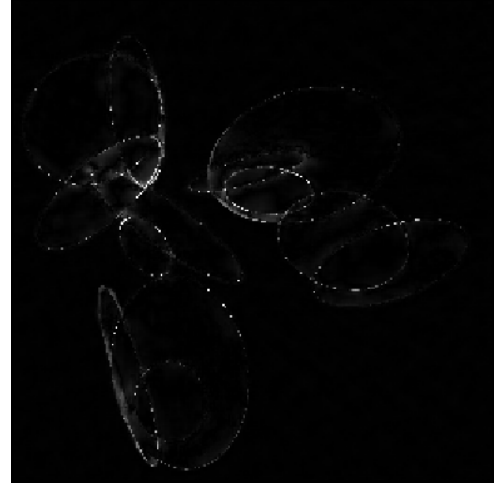} & \includegraphics[valign=c,width=0.15\textwidth]{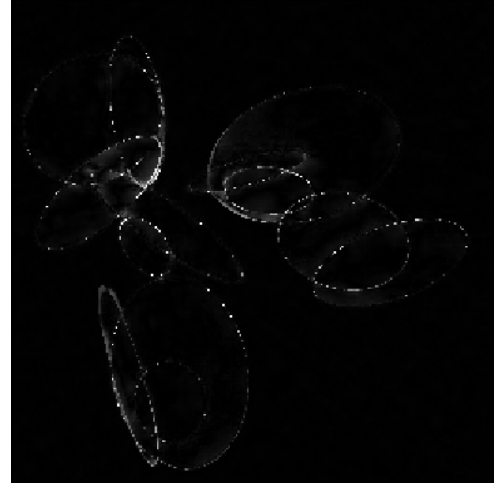} \\ \\

\rotatebox[origin=c]{90}{$\TiraFL$} &
\includegraphics[valign=c,width=0.15\textwidth]{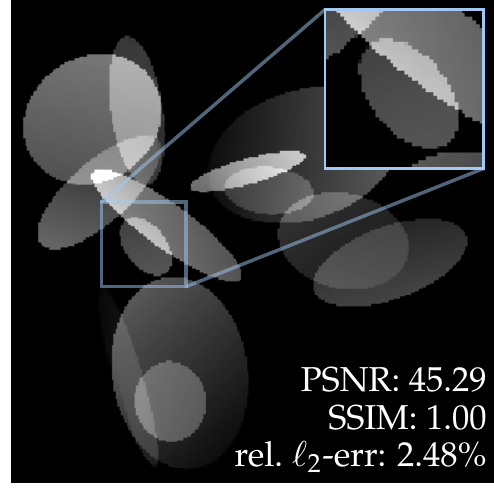} &
\includegraphics[valign=c,width=0.15\textwidth]{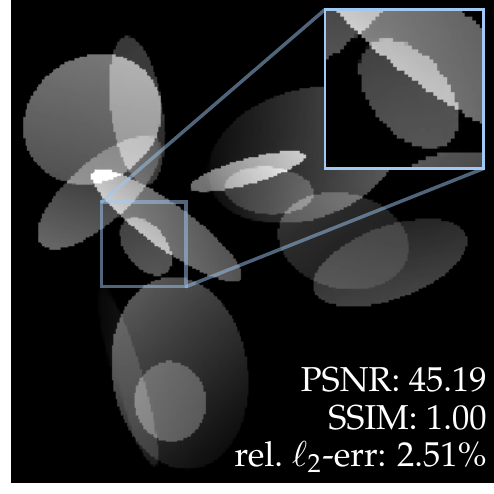} & \includegraphics[valign=c,width=0.15\textwidth]{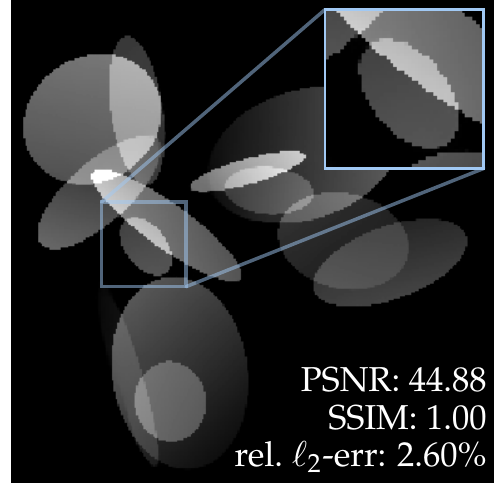} & \includegraphics[valign=c,width=0.15\textwidth]{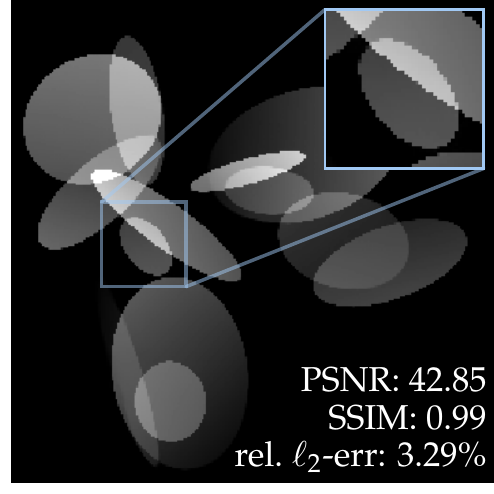} \\
&
\includegraphics[valign=c,width=0.15\textwidth]{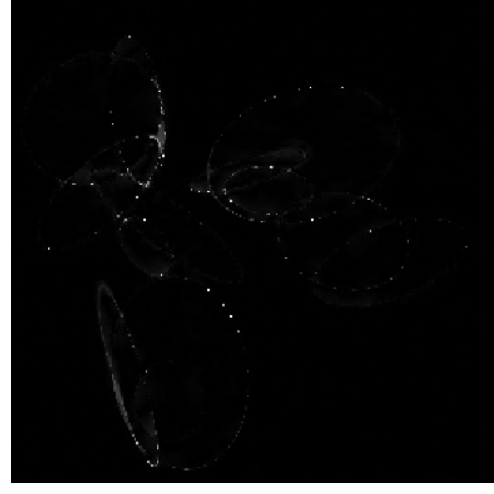} &
\includegraphics[valign=c,width=0.15\textwidth]{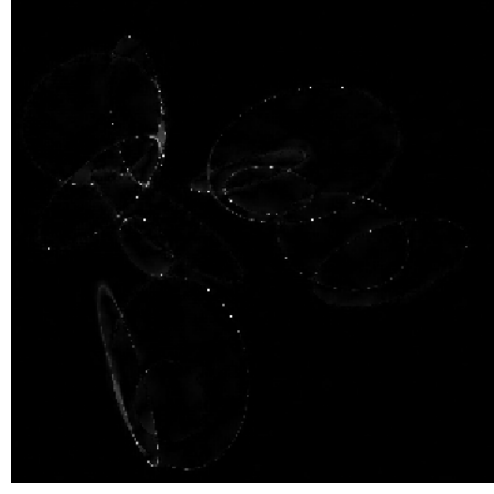} & \includegraphics[valign=c,width=0.15\textwidth]{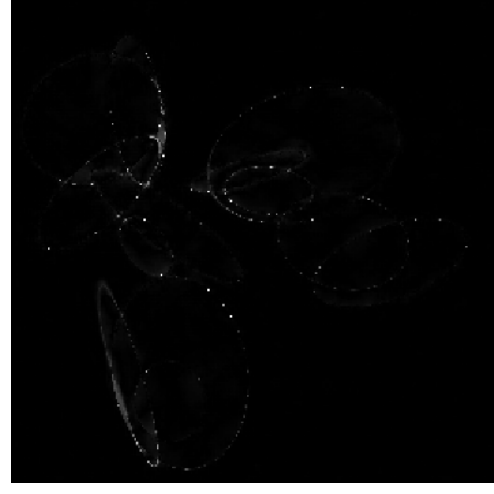} & \includegraphics[valign=c,width=0.15\textwidth]{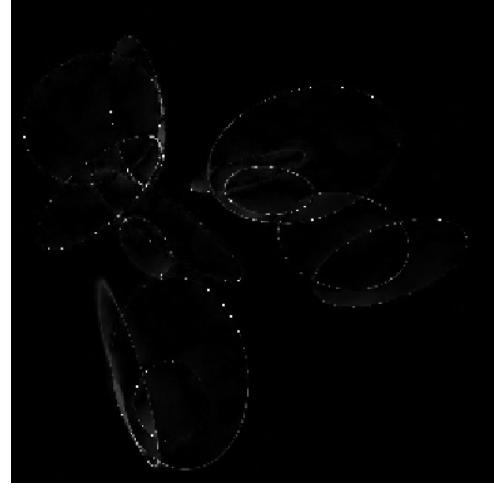} \\ \\

\rotatebox[origin=c]{90}{$\ItNet$} &
\includegraphics[valign=c,width=0.15\textwidth]{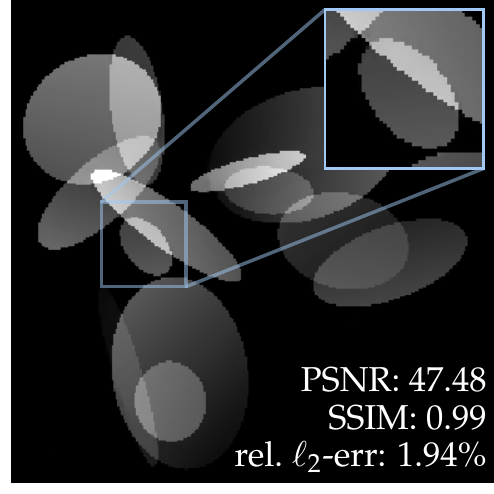} &
\includegraphics[valign=c,width=0.15\textwidth]{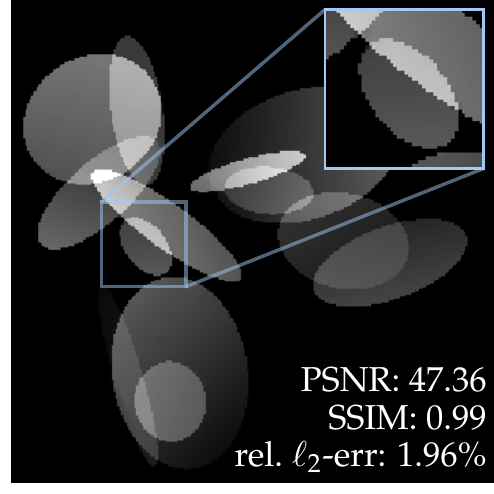} & \includegraphics[valign=c,width=0.15\textwidth]{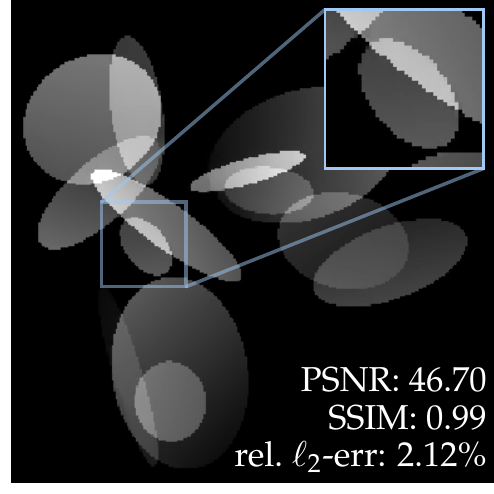} & \includegraphics[valign=c,width=0.15\textwidth]{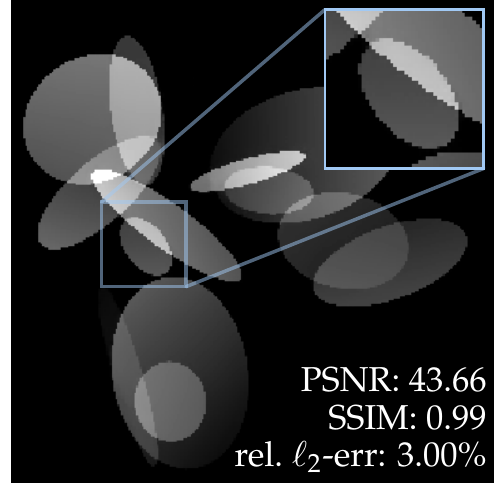} \\
&
\includegraphics[valign=c,width=0.15\textwidth]{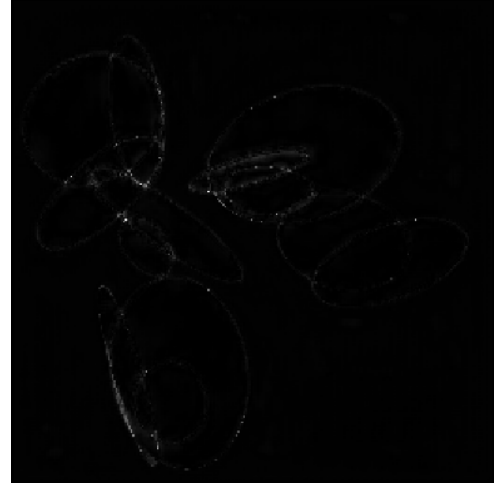} &
\includegraphics[valign=c,width=0.15\textwidth]{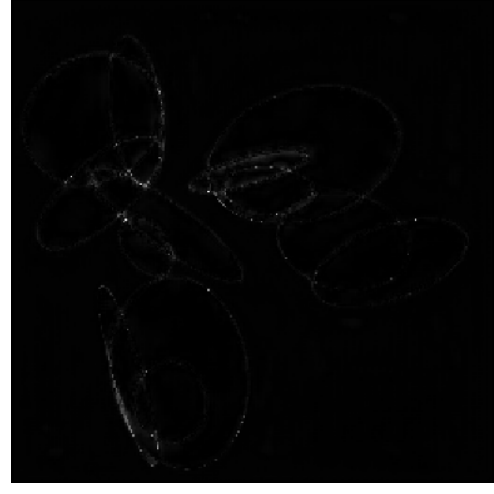} & \includegraphics[valign=c,width=0.15\textwidth]{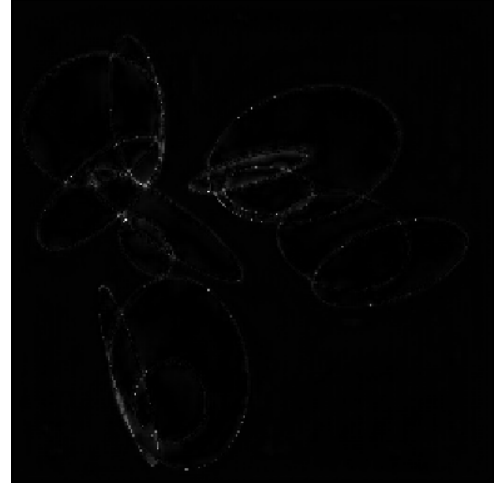} & \includegraphics[valign=c,width=0.15\textwidth]{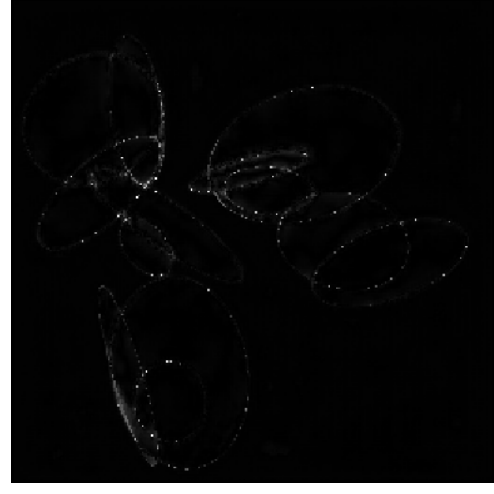}
\end{tabular}
\caption{\textbf{Scenario~\refBone{} -- Fourier meas.~with ellipses.} Individual reconstructions of the image from Fig.~\ref{fig:ellipses:example_adv} under Gaussian noise. The reconstructed images are displayed in the window $[0,0.9]$, which is also used for the computation of the PSNR and SSIM. The error plots shown below each reconstruction are displayed in the window $[0, 0.15]$. In favor of the more insightful noise level 16\%, we have omitted the noiseless case.}
\label{fig:ellipses:example_gauss}
\end{figure}


\begin{figure}[H]
 \centering
 \scriptsize
\begin{tabular}{l@{\,}c@{\,}c@{\,}c@{\,}c}
 & 0.5\% rel.~noise -- Poisson & 1\% rel.~noise -- Poisson & 2\% rel.~noise -- Poisson & 3\% rel.~noise -- Poisson \\
\rotatebox[origin=c]{90}{$\TV[\noisebnd]$} &
\includegraphics[valign=c,width=0.2\textwidth]{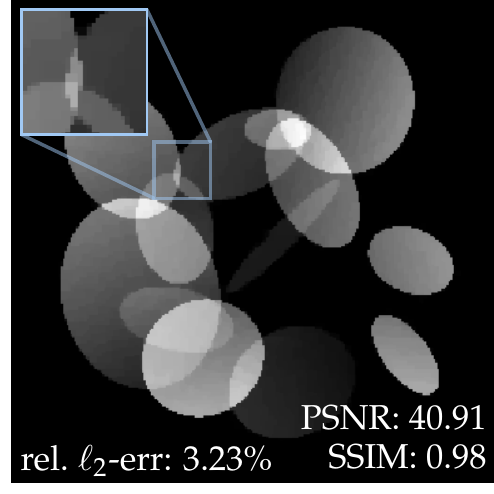} & \includegraphics[valign=c,width=0.2\textwidth]{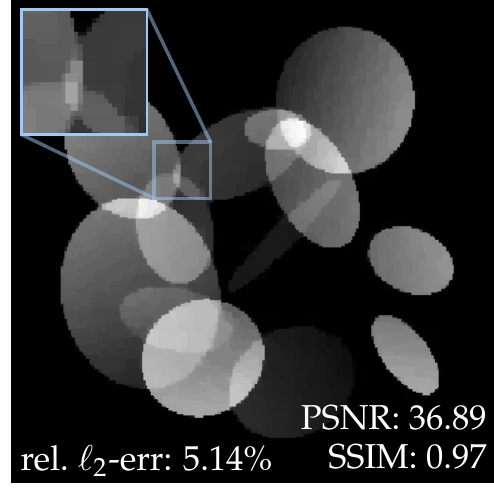} & \includegraphics[valign=c,width=0.2\textwidth]{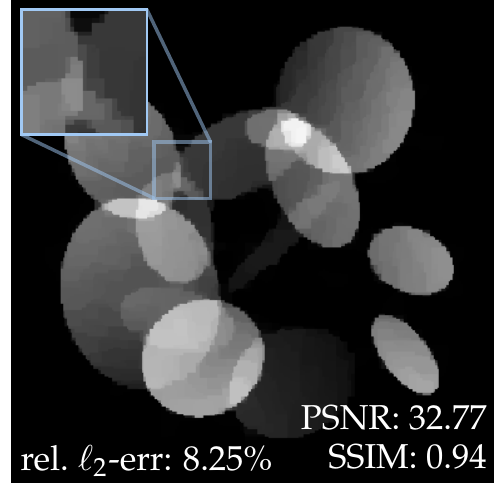} &
\includegraphics[valign=c,width=0.2\textwidth]{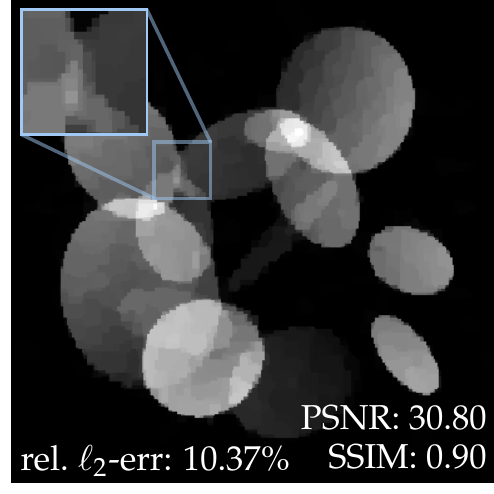} \\
\rotatebox[origin=c]{90}{$\UNet$} &
\includegraphics[valign=c,width=0.2\textwidth]{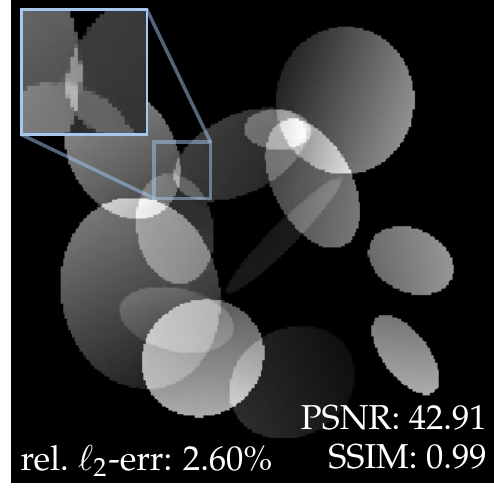} & \includegraphics[valign=c,width=0.2\textwidth]{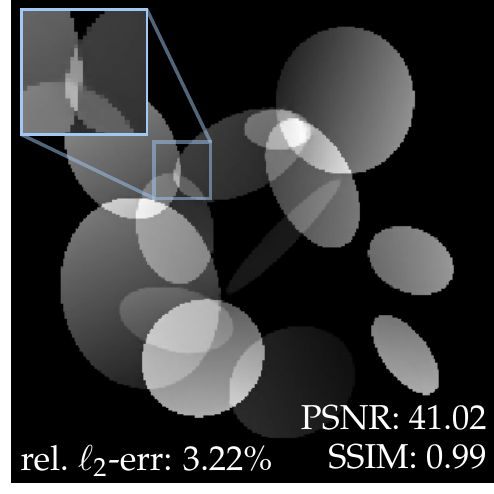} & \includegraphics[valign=c,width=0.2\textwidth]{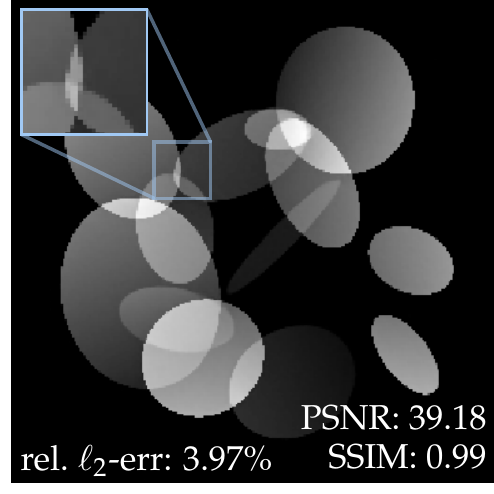} &
\includegraphics[valign=c,width=0.2\textwidth]{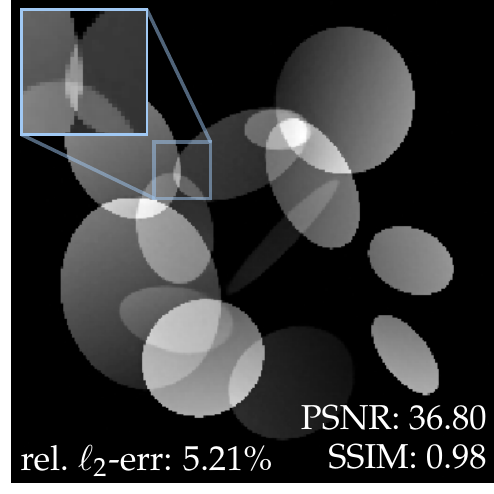} \\
\rotatebox[origin=c]{90}{$\UNet$ w/o Jitter} &
\includegraphics[valign=c,width=0.2\textwidth]{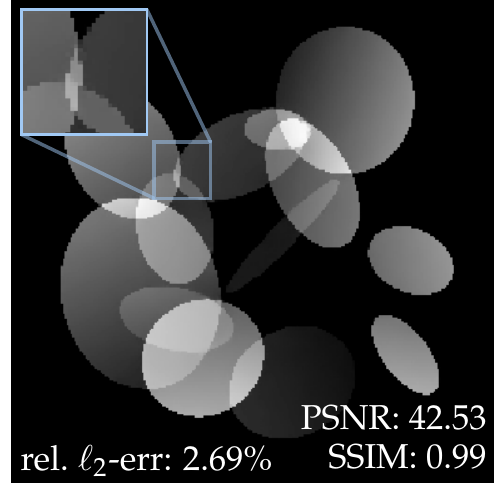} & \includegraphics[valign=c,width=0.2\textwidth]{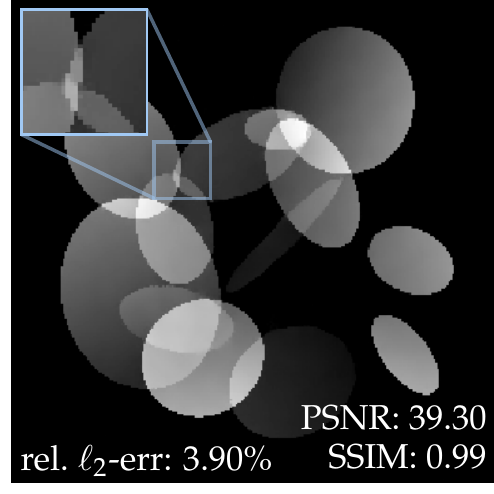} & \includegraphics[valign=c,width=0.2\textwidth]{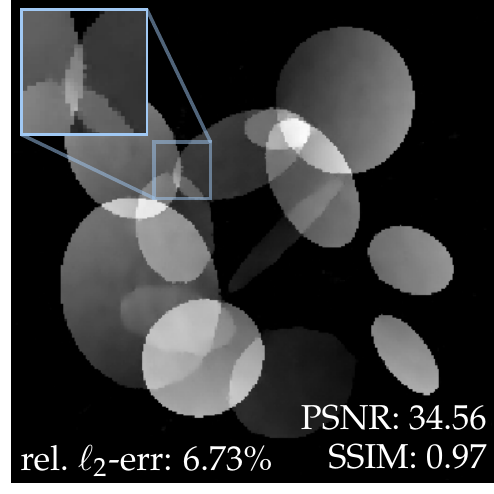} &
\includegraphics[valign=c,width=0.2\textwidth]{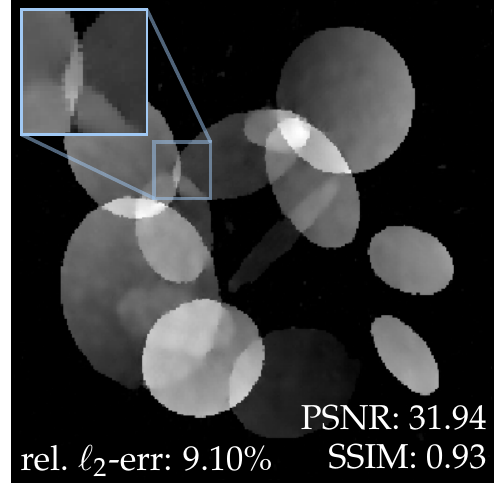}
\end{tabular}
\caption{\textbf{Scenario~\refBtwo{} -- Radon meas.~with ellipses.} Individual reconstructions of the image from Fig.~\ref{fig:ellipses:example_adv_radon} under Poisson noise. The reconstructed images are displayed in the window $[0,1]$, which is also used for the computation of the PSNR and SSIM. In favor of the more insightful noise level 3\%, we have omitted the noiseless case. The bottom row shows the corresponding reconstructions for a $\UNet$ that is trained without jittering; see also Section~\ref{sec:additional:crime} on the inverse crime.}
\label{fig:ellipses:example_poisson_radon}
\end{figure}

\begin{figure}[H]
	\centering
	\scriptsize
	\begin{tabular}{l@{\,}c@{\,}c@{\,}c@{\,}c@{\,}c}
		& $\ItNet(\yadv)$ -- 8\% rel.~noise & $\TV[\noisebnd \cong \text{0\%}](\yadv)$ & $\TV[\noisebnd \cong \text{1\%}](\yadv)$ & $\TV[\noisebnd \cong \text{3\%}](\yadv)$ & $\TV[\noisebnd \cong \text{8\%}](\yadv)$ \\[.25em]
		\rotatebox[origin=c]{90}{Fourier: $\ItNet \longrightarrow \TV[\noisebnd]$} &
		\includegraphics[valign=c,width=0.19\textwidth]{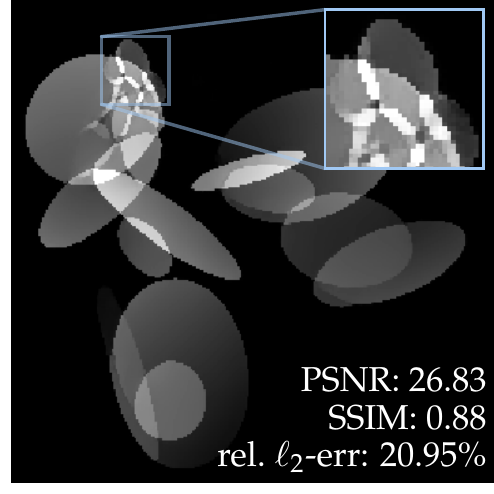} &
		\includegraphics[valign=c,width=0.19\textwidth]{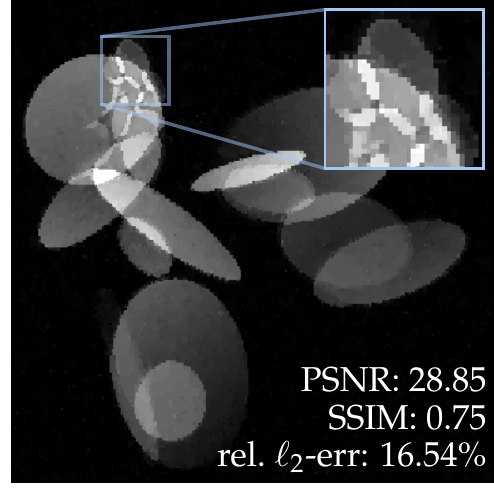} &
		\includegraphics[valign=c,width=0.19\textwidth]{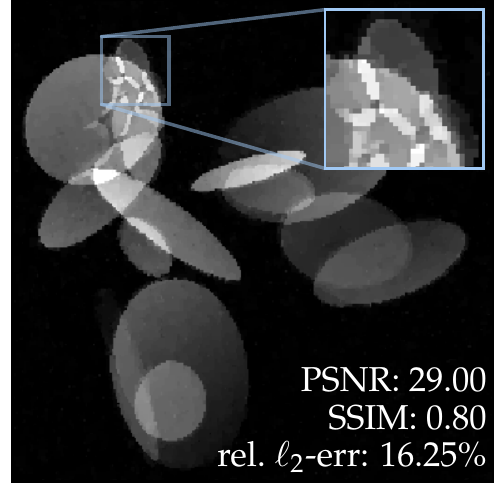} & \includegraphics[valign=c,width=0.19\textwidth]{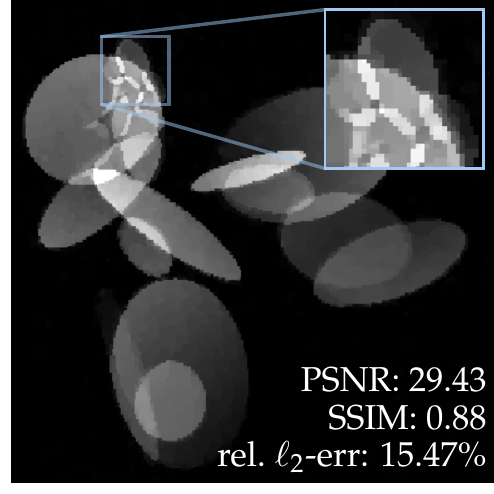} & \includegraphics[valign=c,width=0.19\textwidth]{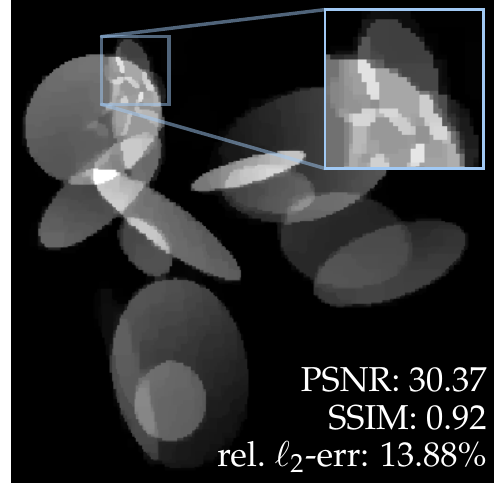} \\ \\
		& $\TV[\noisebnd](\yadv)$ -- 8\% rel.~noise & $\UNet(\yadv)$ & $\TiraFL(\yadv)$ & $\ItNet(\yadv)$ & \\[.25em]
		\rotatebox[origin=c]{90}{Fourier: $\TV[\noisebnd] \longrightarrow \text{NNs}$} &
		\includegraphics[valign=c,width=0.19\textwidth]{{ellipses/results/attacks/fig_example_S66_adv_tv_8.00e-02}.pdf} &
		\includegraphics[valign=c,width=0.19\textwidth]{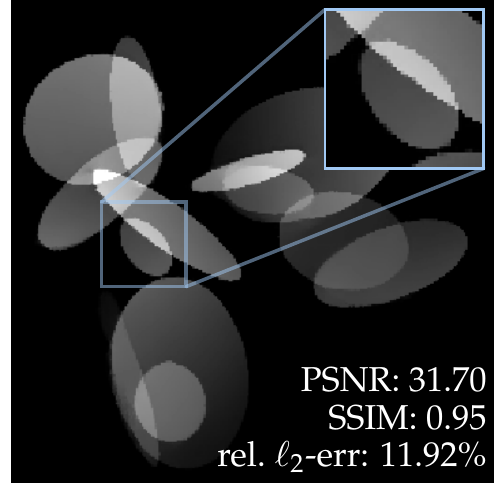} &
		\includegraphics[valign=c,width=0.19\textwidth]{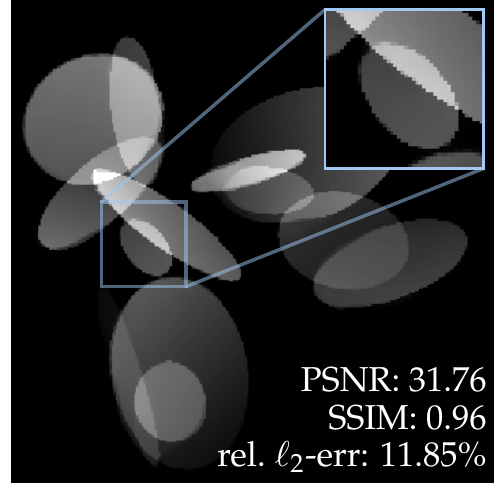} & \includegraphics[valign=c,width=0.19\textwidth]{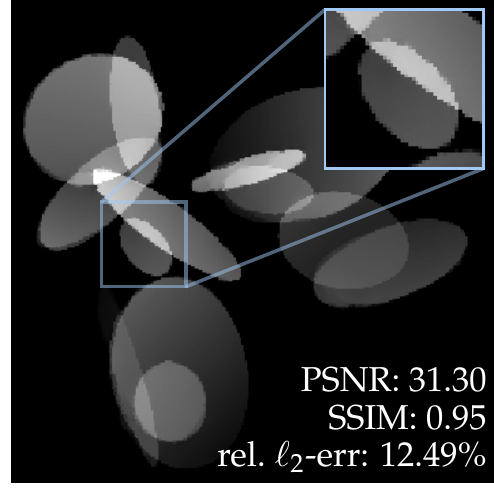} \\ \\
		& $\UNet(\yadv)$ -- 2\% rel.~noise & $\TV[\noisebnd \cong \text{0\%}](\yadv)$ & $\TV[\noisebnd \cong \text{0.5\%}](\yadv)$ & $\TV[\noisebnd \cong \text{1\%}](\yadv)$ & $\TV[\noisebnd \cong \text{2\%}](\yadv)$ \\[.25em]
		\rotatebox[origin=c]{90}{Radon: $\UNet \longrightarrow \TV[\noisebnd]$} &
		\includegraphics[valign=c,width=0.19\textwidth]{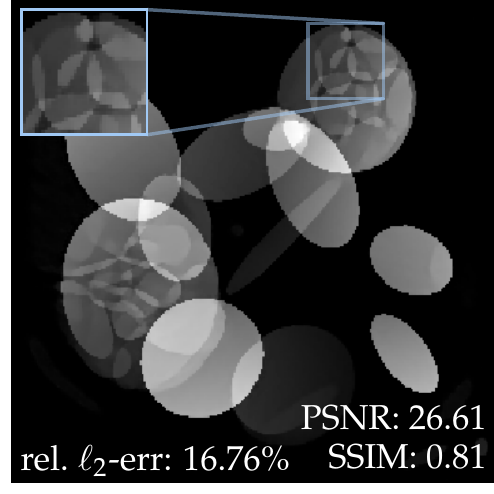} &
		\includegraphics[valign=c,width=0.19\textwidth]{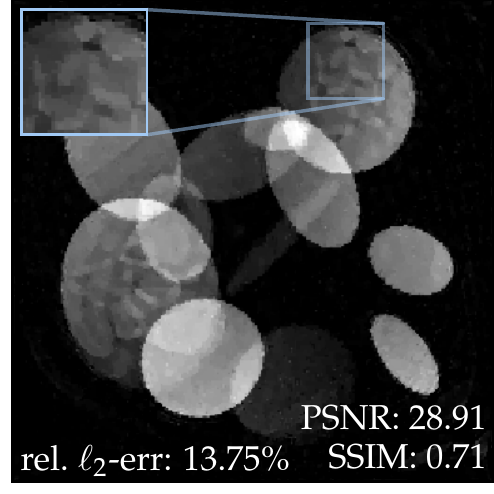} &
		\includegraphics[valign=c,width=0.19\textwidth]{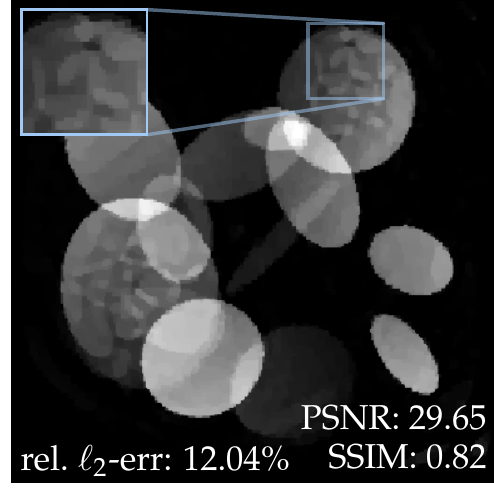} & \includegraphics[valign=c,width=0.19\textwidth]{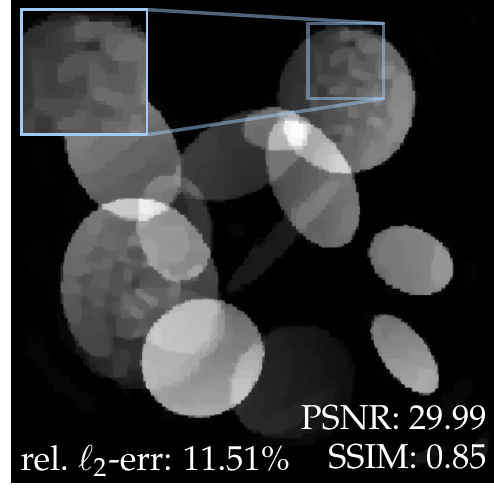} & \includegraphics[valign=c,width=0.19\textwidth]{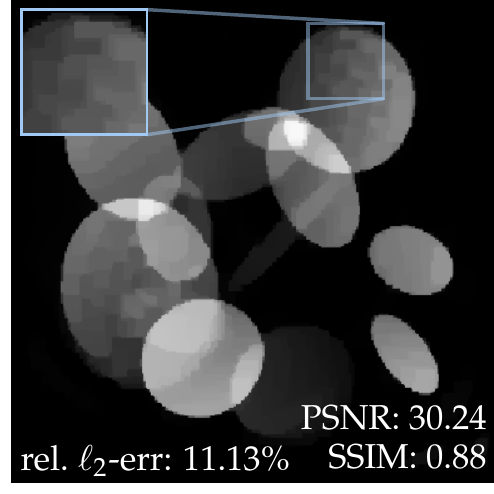}
	\end{tabular}
	\caption{\textbf{Case Study~B -- Transferability of perturbations.} This figure analyzes how adversarial noise transfers between TV minimization and NN-based solvers. The top row shows the recovery behavior of $\TV[\noisebnd]$ in the case of Fourier measurements when an adversarial perturbation $\yadv$ found for $\ItNet$ is used as input (cf.~Fig.~\ref{fig:ellipses:example_adv}). Here, we also demonstrate the impact of the noise tuning parameter $\noisebnd$, which controls the degree of regularization for TV minimization. The middle row presents the reverse experiment: an adversarial perturbation $\yadv$ found for $\TV[\noisebnd]$ is plugged into each considered NN. The bottom row is the analog of the top row in the case of Radon measurements (cf.~Fig.~\ref{fig:ellipses:example_adv_radon}).}
	\label{fig:ellipses:transfer}
\end{figure}

\clearpage
\section{Supplementary Results for Case Study~C (MRI on Real-World Data)}
\label{sec:supp:csC}

\begin{table}[H]
\robustify\bfseries
\begin{tabular}{llS[separate-uncertainty]S[separate-uncertainty]S[separate-uncertainty]S[separate-uncertainty]S[separate-uncertainty]S[separate-uncertainty]S[separate-uncertainty]}
\toprule
\multicolumn{2}{l}{rel.~noise -- adversarial}  &                   {0.0\%} &                   {0.2\%} &                   {0.5\%} &                   {1.0\%} &                   {1.5\%} &                   {2.0\%} &                   {2.5\%} \\
\midrule
\multirow{3}{*}{TV} & rel.~$\l{2}$-err. [\%] &             8.39 \pm 1.38 &             8.89 \pm 1.40 &             9.35 \pm 1.41 &            10.34 \pm 1.41 &            11.35 \pm 1.42 &            12.35 \pm 1.43 &            12.96 \pm 1.44 \\
      & PSNR &            31.70 \pm 1.47 &            31.18 \pm 1.40 &            30.73 \pm 1.34 &            29.85 \pm 1.22 &            29.02 \pm 1.10 &            28.28 \pm 1.02 &            27.85 \pm 0.98 \\
      & SSIM &             0.78 \pm 0.04 &             0.77 \pm 0.04 &             0.76 \pm 0.04 &             0.74 \pm 0.04 &             0.72 \pm 0.04 &             0.70 \pm 0.04 &             0.68 \pm 0.04 \\
\cline{1-9}
\multirow{3}{*}{UNet} & rel.~$\l{2}$-err. [\%] &             8.18 \pm 1.27 &             8.29 \pm 1.27 &             8.41 \pm 1.27 &             8.67 \pm 1.26 &             8.99 \pm 1.25 &             9.38 \pm 1.22 &             9.84 \pm 1.18 \\
      & PSNR &            31.90 \pm 1.38 &            31.79 \pm 1.37 &            31.66 \pm 1.35 &            31.38 \pm 1.30 &            31.06 \pm 1.23 &            30.69 \pm 1.15 &            30.26 \pm 1.06 \\
      & SSIM &             0.80 \pm 0.04 &             0.80 \pm 0.03 &             0.79 \pm 0.03 &             0.79 \pm 0.03 &             0.78 \pm 0.03 &             0.78 \pm 0.03 &             0.77 \pm 0.03 \\
\cline{1-9}
\multirow{3}{*}{UNetFL} & rel.~$\l{2}$-err. [\%] &             8.23 \pm 1.28 &             8.35 \pm 1.28 &             8.47 \pm 1.28 &             8.75 \pm 1.27 &             9.10 \pm 1.25 &             9.51 \pm 1.21 &            10.01 \pm 1.17 \\
      & PSNR &            31.85 \pm 1.39 &            31.72 \pm 1.37 &            31.59 \pm 1.34 &            31.30 \pm 1.29 &            30.96 \pm 1.22 &            30.56 \pm 1.13 &            30.10 \pm 1.04 \\
      & SSIM &             0.79 \pm 0.04 &             0.79 \pm 0.04 &             0.79 \pm 0.04 &             0.78 \pm 0.03 &             0.78 \pm 0.03 &             0.77 \pm 0.03 &             0.77 \pm 0.03 \\
\cline{1-9}
\multirow{3}{*}{Tira} & rel.~$\l{2}$-err. [\%] &             7.97 \pm 1.26 &             8.10 \pm 1.26 &             8.24 \pm 1.26 &             8.58 \pm 1.25 &             9.00 \pm 1.21 &             9.52 \pm 1.17 &            10.16 \pm 1.12 \\
      & PSNR &            32.13 \pm 1.41 &            31.99 \pm 1.39 &            31.84 \pm 1.36 &            31.48 \pm 1.30 &            31.05 \pm 1.19 &            30.54 \pm 1.09 &            29.97 \pm 0.97 \\
      & SSIM &             0.80 \pm 0.03 &             0.80 \pm 0.03 &             0.80 \pm 0.03 &             0.79 \pm 0.03 &             0.79 \pm 0.03 &             0.78 \pm 0.03 &             0.77 \pm 0.03 \\
\cline{1-9}
\multirow{3}{*}{TiraFL} & rel.~$\l{2}$-err. [\%] &             7.98 \pm 1.27 &             8.11 \pm 1.27 &             8.26 \pm 1.27 &             8.60 \pm 1.27 &             9.03 \pm 1.24 &             9.55 \pm 1.20 &            10.19 \pm 1.15 \\
      & PSNR &            32.12 \pm 1.42 &            31.98 \pm 1.40 &            31.82 \pm 1.37 &            31.46 \pm 1.31 &            31.03 \pm 1.22 &            30.52 \pm 1.11 &            29.95 \pm 1.00 \\
      & SSIM &             0.80 \pm 0.03 &             0.80 \pm 0.03 &             0.80 \pm 0.03 &             0.79 \pm 0.03 &             0.78 \pm 0.03 &             0.78 \pm 0.03 &             0.77 \pm 0.03 \\
\cline{1-9}
\multirow{3}{*}{ItNet} & rel.~$\l{2}$-err. [\%] &   \bfseries 7.08 \pm 1.20 &   \bfseries 7.21 \pm 1.20 &   \bfseries 7.35 \pm 1.19 &   \bfseries 7.67 \pm 1.17 &   \bfseries 8.08 \pm 1.13 &   \bfseries 8.59 \pm 1.10 &   \bfseries 9.20 \pm 1.07 \\
      & PSNR &  \bfseries 33.18 \pm 1.52 &  \bfseries 33.02 \pm 1.49 &  \bfseries 32.85 \pm 1.45 &  \bfseries 32.45 \pm 1.35 &  \bfseries 31.99 \pm 1.23 &  \bfseries 31.45 \pm 1.12 &  \bfseries 30.84 \pm 1.02 \\
      & SSIM &   \bfseries 0.82 \pm 0.04 &   \bfseries 0.82 \pm 0.04 &   \bfseries 0.81 \pm 0.04 &   \bfseries 0.81 \pm 0.03 &   \bfseries 0.80 \pm 0.03 &   \bfseries 0.79 \pm 0.03 &   \bfseries 0.78 \pm 0.03 \\
\bottomrule
\end{tabular}

\caption{\textbf{Case Study C -- fastMRI.} A numerical representation of the results of Fig.~\ref{fig:fastmri:table}(c), including the additional methods $\UNetFL$ and $\Tira$. The best relative error/PSNR/SSIM per noise level is highlighted in bold.}
\label{tab:fastmri:table_adv}
\end{table}

\begin{table}[H]
\robustify\bfseries
\begin{tabular}{llS[separate-uncertainty]S[separate-uncertainty]S[separate-uncertainty]S[separate-uncertainty]S[separate-uncertainty]S[separate-uncertainty]S[separate-uncertainty]}
\toprule
\multicolumn{2}{l}{rel.~noise -- Gaussian} &                   {0.0\%} &                   {0.2\%} &                   {0.5\%} &                   {1.0\%} &                   {1.5\%} &                   {2.0\%} &                   {2.5\%} \\
\midrule
\multirow{3}{*}{TV} & rel.~$\l{2}$-err. [\%] &             8.39 \pm 1.38 &             8.39 \pm 1.38 &             8.40 \pm 1.38 &             8.44 \pm 1.37 &             8.49 \pm 1.36 &             8.57 \pm 1.35 &             8.65 \pm 1.34 \\
      & PSNR &            31.70 \pm 1.47 &            31.69 \pm 1.47 &            31.68 \pm 1.47 &            31.65 \pm 1.46 &            31.58 \pm 1.44 &            31.51 \pm 1.41 &            31.42 \pm 1.38 \\
      & SSIM &             0.78 \pm 0.04 &             0.78 \pm 0.04 &             0.78 \pm 0.04 &             0.78 \pm 0.04 &             0.78 \pm 0.04 &             0.77 \pm 0.04 &             0.77 \pm 0.04 \\
\cline{1-9}
\multirow{3}{*}{UNet} & rel.~$\l{2}$-err. [\%] &             8.18 \pm 1.27 &             8.18 \pm 1.27 &             8.18 \pm 1.27 &             8.20 \pm 1.26 &             8.22 \pm 1.26 &             8.24 \pm 1.26 &             8.27 \pm 1.26 \\
      & PSNR &            31.90 \pm 1.38 &            31.90 \pm 1.38 &            31.90 \pm 1.38 &            31.89 \pm 1.38 &            31.86 \pm 1.37 &            31.84 \pm 1.37 &            31.80 \pm 1.36 \\
      & SSIM &             0.80 \pm 0.04 &             0.80 \pm 0.04 &             0.80 \pm 0.04 &             0.80 \pm 0.04 &             0.80 \pm 0.04 &             0.80 \pm 0.04 &             0.79 \pm 0.04 \\
\cline{1-9}
\multirow{3}{*}{UNetFL} & rel.~$\l{2}$-err. [\%] &             8.23 \pm 1.28 &             8.24 \pm 1.28 &             8.24 \pm 1.28 &             8.25 \pm 1.28 &             8.26 \pm 1.28 &             8.29 \pm 1.27 &             8.31 \pm 1.27 \\
      & PSNR &            31.85 \pm 1.39 &            31.84 \pm 1.39 &            31.84 \pm 1.38 &            31.83 \pm 1.38 &            31.81 \pm 1.38 &            31.79 \pm 1.37 &            31.76 \pm 1.37 \\
      & SSIM &             0.79 \pm 0.04 &             0.79 \pm 0.04 &             0.79 \pm 0.04 &             0.79 \pm 0.04 &             0.79 \pm 0.04 &             0.79 \pm 0.04 &             0.79 \pm 0.04 \\
\cline{1-9}
\multirow{3}{*}{Tira} & rel.~$\l{2}$-err. [\%] &             7.97 \pm 1.26 &             7.97 \pm 1.26 &             7.98 \pm 1.26 &             7.99 \pm 1.26 &             8.01 \pm 1.26 &             8.04 \pm 1.25 &             8.07 \pm 1.25 \\
      & PSNR &            32.13 \pm 1.41 &            32.13 \pm 1.41 &            32.13 \pm 1.41 &            32.11 \pm 1.40 &            32.09 \pm 1.40 &            32.05 \pm 1.39 &            32.02 \pm 1.38 \\
      & SSIM &             0.80 \pm 0.03 &             0.80 \pm 0.03 &             0.80 \pm 0.03 &             0.80 \pm 0.03 &             0.80 \pm 0.03 &             0.80 \pm 0.03 &             0.80 \pm 0.03 \\
\cline{1-9}
\multirow{3}{*}{TiraFL} & rel.~$\l{2}$-err. [\%] &             7.98 \pm 1.27 &             7.98 \pm 1.27 &             7.99 \pm 1.27 &             8.00 \pm 1.27 &             8.02 \pm 1.27 &             8.05 \pm 1.26 &             8.08 \pm 1.26 \\
      & PSNR &            32.12 \pm 1.42 &            32.12 \pm 1.42 &            32.12 \pm 1.42 &            32.10 \pm 1.41 &            32.08 \pm 1.41 &            32.05 \pm 1.40 &            32.01 \pm 1.40 \\
      & SSIM &             0.80 \pm 0.03 &             0.80 \pm 0.03 &             0.80 \pm 0.03 &             0.80 \pm 0.03 &             0.80 \pm 0.03 &             0.80 \pm 0.03 &             0.80 \pm 0.04 \\
\cline{1-9}
\multirow{3}{*}{ItNet} & rel.~$\l{2}$-err. [\%] &   \bfseries 7.08 \pm 1.20 &   \bfseries 7.08 \pm 1.20 &   \bfseries 7.08 \pm 1.20 &   \bfseries 7.10 \pm 1.20 &   \bfseries 7.13 \pm 1.19 &   \bfseries 7.17 \pm 1.19 &   \bfseries 7.22 \pm 1.18 \\
      & PSNR &  \bfseries 33.18 \pm 1.52 &  \bfseries 33.18 \pm 1.52 &  \bfseries 33.17 \pm 1.52 &  \bfseries 33.15 \pm 1.51 &  \bfseries 33.12 \pm 1.50 &  \bfseries 33.07 \pm 1.48 &  \bfseries 33.01 \pm 1.47 \\
      & SSIM &   \bfseries 0.82 \pm 0.04 &   \bfseries 0.82 \pm 0.04 &   \bfseries 0.82 \pm 0.04 &   \bfseries 0.82 \pm 0.04 &   \bfseries 0.82 \pm 0.04 &   \bfseries 0.81 \pm 0.04 &   \bfseries 0.81 \pm 0.04 \\
\bottomrule
\end{tabular}

\caption{\textbf{Case Study C -- fastMRI.} A numerical representation of the results of Fig.~\ref{fig:fastmri:table}(d), including the additional methods $\UNetFL$ and $\Tira$. The best relative error/PSNR/SSIM per noise level is highlighted in bold.}
\label{tab:fastmri:table_ref}
\end{table}

\begin{figure}[H]
	\centering
	\scriptsize
	\begin{tabular}{l@{\,}c@{\,}c@{\,}c@{\,}c}
 		& noiseless & 1\% rel.~noise -- adv. & 1.5\% rel.~noise -- adv. & 2.5\% rel.~noise -- adv. \\
		\rotatebox[origin=c]{90}{$\TV[\noisebnd]$} &
		\includegraphics[valign=c,width=0.15\textwidth]{{fastmri_v4/results/attacks/fig_example_S8_adv_tv_0.00e+00}.pdf} &
		\includegraphics[valign=c,width=0.15\textwidth]{{fastmri_v4/results/attacks/fig_example_S8_adv_tv_1.00e-02}.pdf} & \includegraphics[valign=c,width=0.15\textwidth]{{fastmri_v4/results/attacks/fig_example_S8_adv_tv_1.50e-02}.pdf} & \includegraphics[valign=c,width=0.15\textwidth]{{fastmri_v4/results/attacks/fig_example_S8_adv_tv_2.50e-02}.pdf} \\
		&
		\includegraphics[valign=c,width=0.15\textwidth]{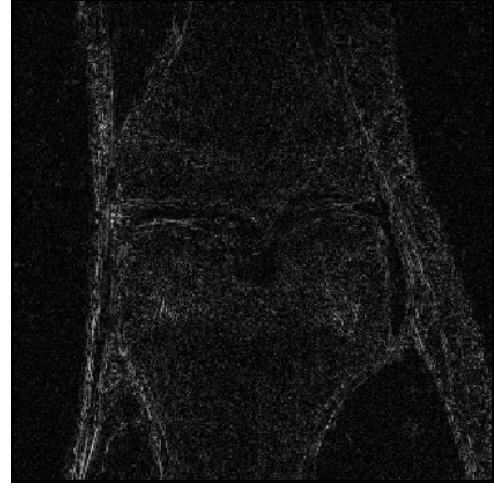} &
		\includegraphics[valign=c,width=0.15\textwidth]{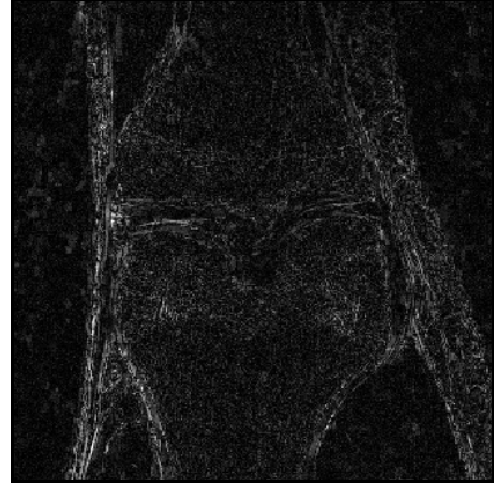} & \includegraphics[valign=c,width=0.15\textwidth]{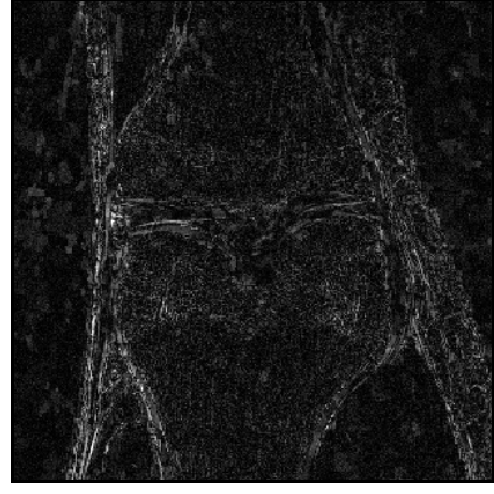} & \includegraphics[valign=c,width=0.15\textwidth]{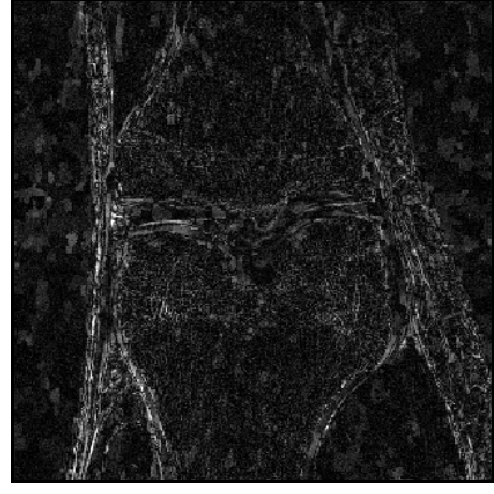} \\ \\

		\rotatebox[origin=c]{90}{$\UNet$} &
		\includegraphics[valign=c,width=0.15\textwidth]{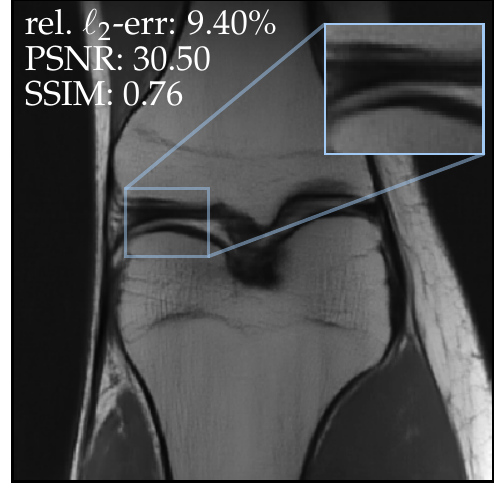} &
		\includegraphics[valign=c,width=0.15\textwidth]{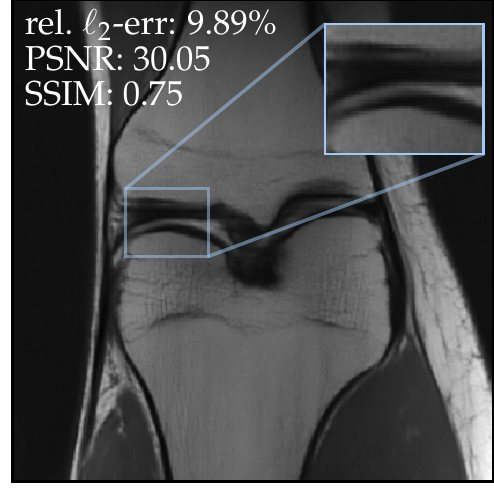} & \includegraphics[valign=c,width=0.15\textwidth]{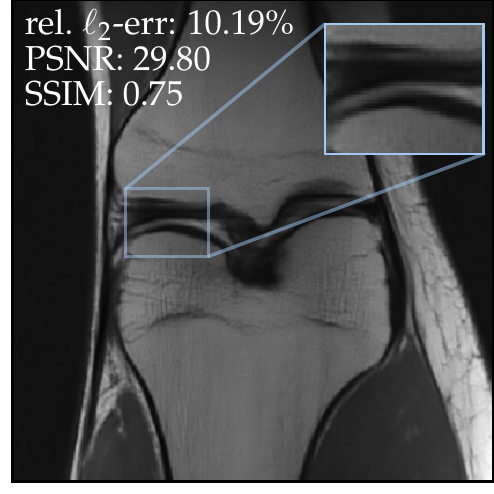} & \includegraphics[valign=c,width=0.15\textwidth]{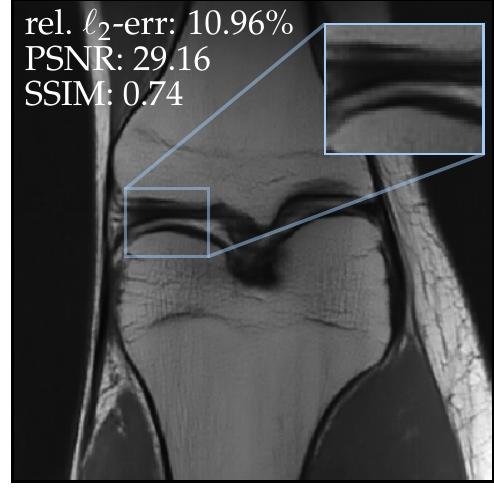} \\
		&
		\includegraphics[valign=c,width=0.15\textwidth]{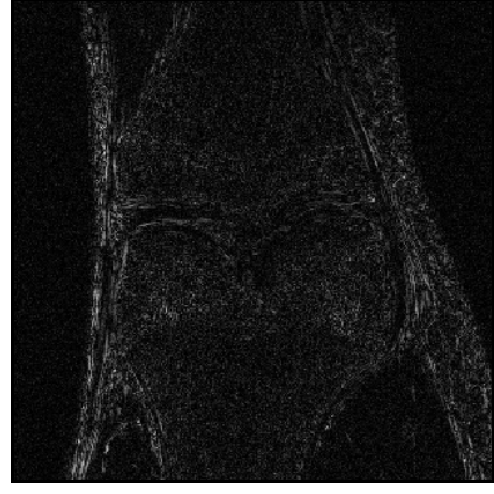} &
		\includegraphics[valign=c,width=0.15\textwidth]{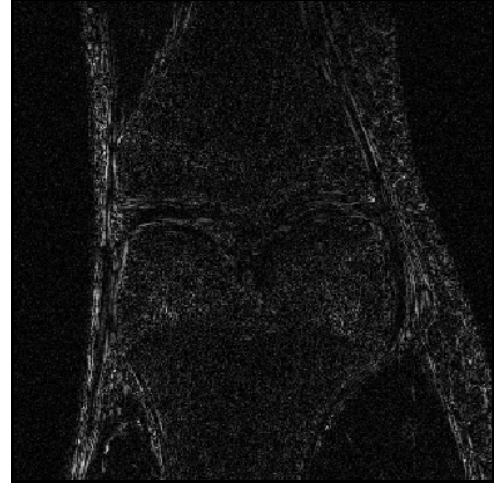} & \includegraphics[valign=c,width=0.15\textwidth]{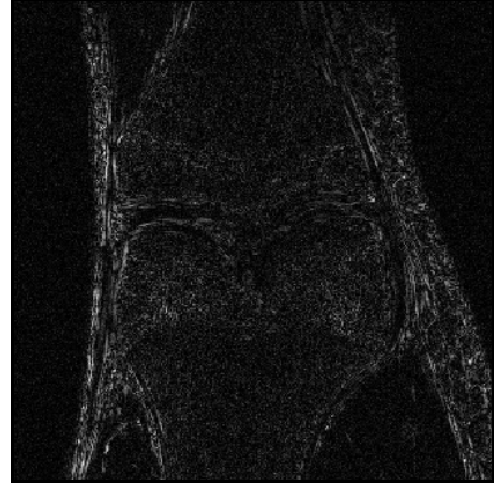} & \includegraphics[valign=c,width=0.15\textwidth]{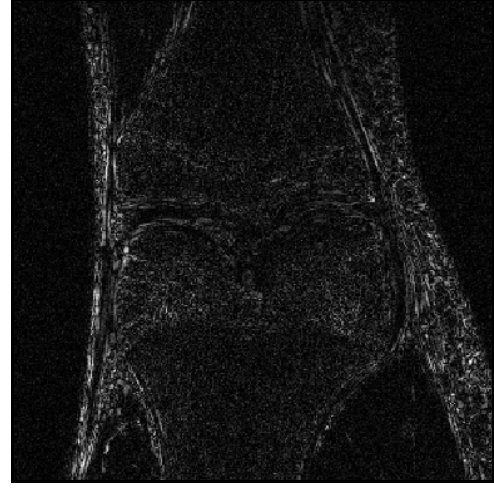} \\ \\

		\rotatebox[origin=c]{90}{$\TiraFL$} &
		\includegraphics[valign=c,width=0.15\textwidth]{{fastmri_v4/results/attacks/fig_example_S8_adv_tiramisu_ee_jit_0.00e+00}.pdf} &
		\includegraphics[valign=c,width=0.15\textwidth]{{fastmri_v4/results/attacks/fig_example_S8_adv_tiramisu_ee_jit_1.00e-02}.pdf} & \includegraphics[valign=c,width=0.15\textwidth]{{fastmri_v4/results/attacks/fig_example_S8_adv_tiramisu_ee_jit_1.50e-02}.pdf} & \includegraphics[valign=c,width=0.15\textwidth]{{fastmri_v4/results/attacks/fig_example_S8_adv_tiramisu_ee_jit_2.50e-02}.pdf} \\
		&
		\includegraphics[valign=c,width=0.15\textwidth]{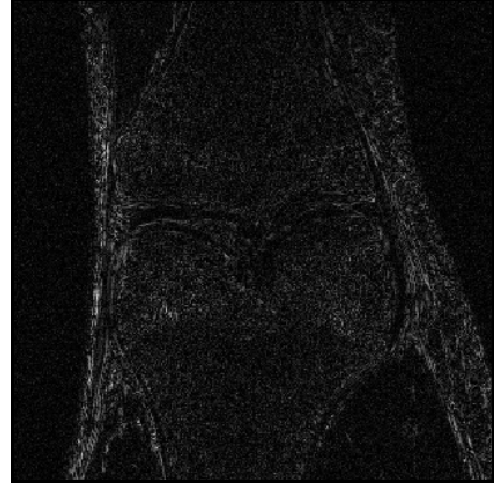} &
		\includegraphics[valign=c,width=0.15\textwidth]{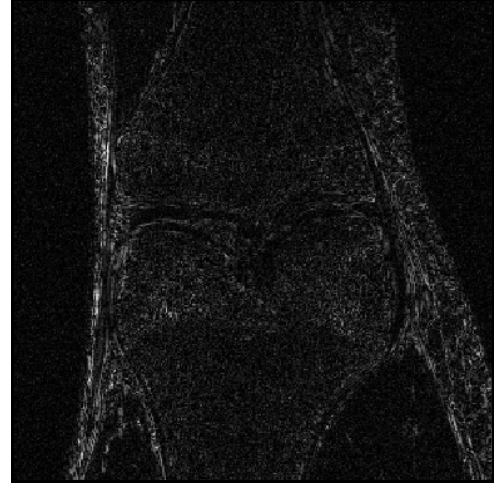} & \includegraphics[valign=c,width=0.15\textwidth]{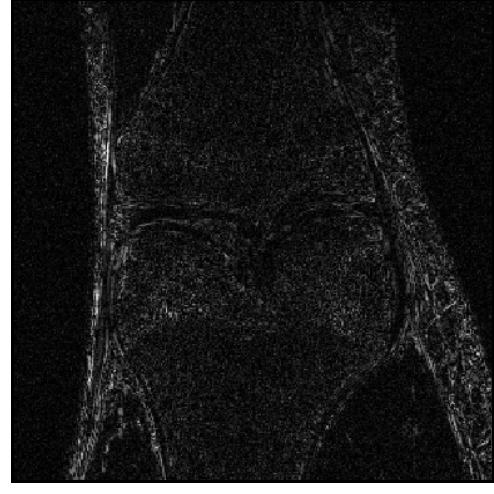} & \includegraphics[valign=c,width=0.15\textwidth]{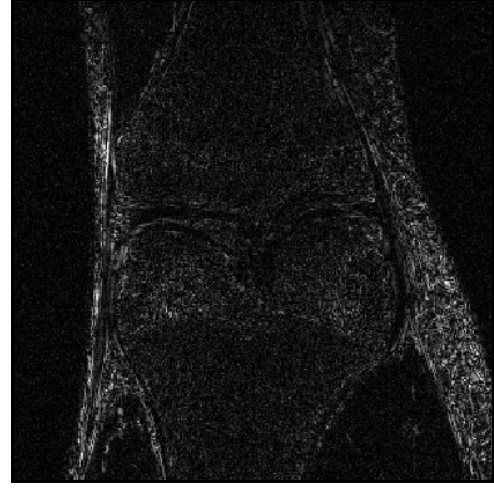} \\ \\

		\rotatebox[origin=c]{90}{$\ItNet$} &
		\includegraphics[valign=c,width=0.15\textwidth]{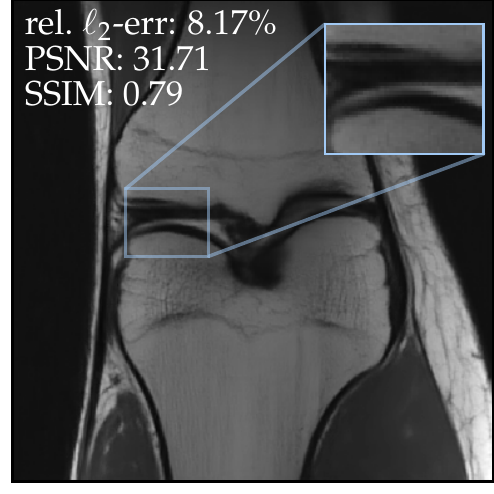} &
		\includegraphics[valign=c,width=0.15\textwidth]{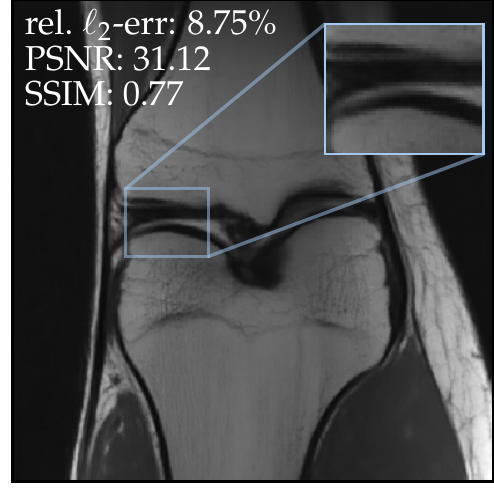} & \includegraphics[valign=c,width=0.15\textwidth]{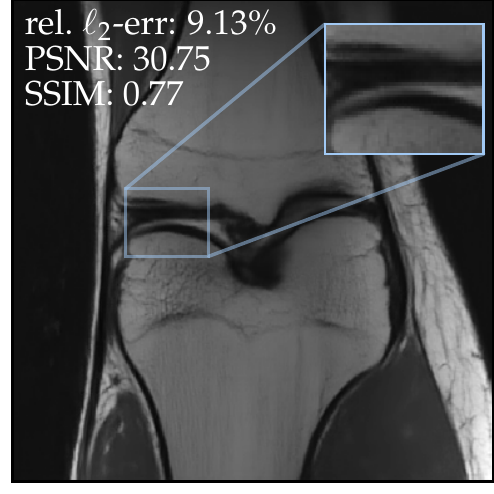} & \includegraphics[valign=c,width=0.15\textwidth]{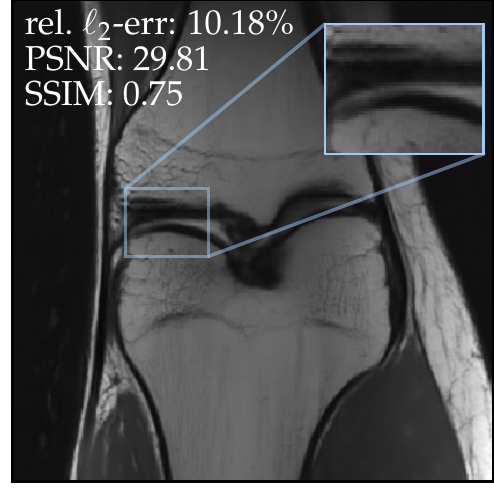} \\
		&
		\includegraphics[valign=c,width=0.15\textwidth]{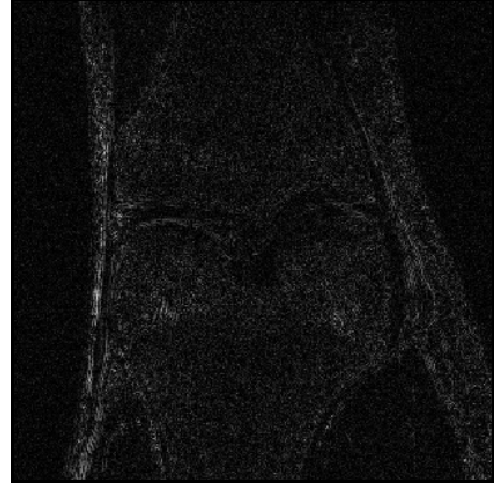} &
		\includegraphics[valign=c,width=0.15\textwidth]{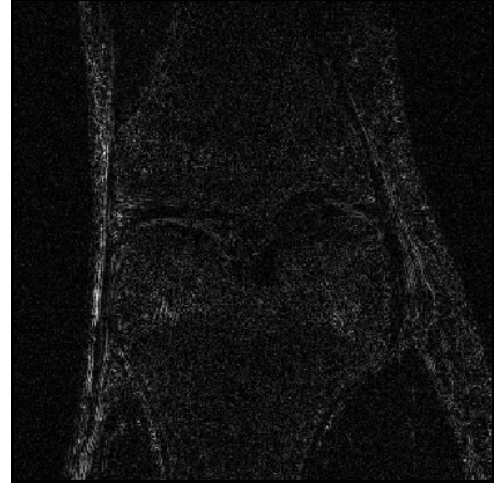} & \includegraphics[valign=c,width=0.15\textwidth]{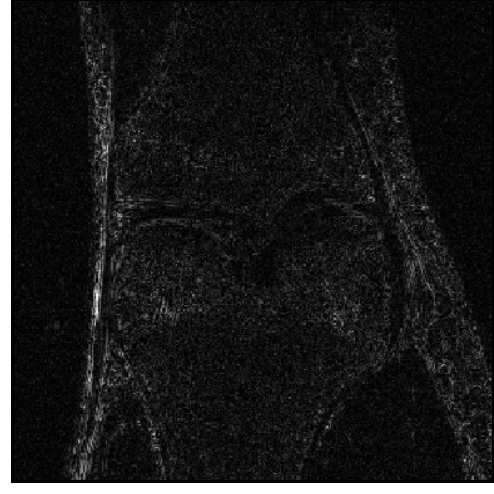} & \includegraphics[valign=c,width=0.15\textwidth]{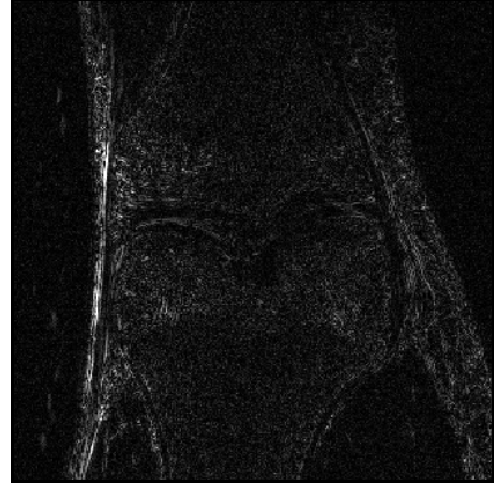}
	\end{tabular}
	\caption{\textbf{Case Study C -- fastMRI.} Individual reconstructions of the image from Fig.~\ref{fig:fastmri:example_adv} for different levels of adversarial noise. The reconstructed images are displayed in the window $[0.05,4.50]$, which is also used for the computation of the PSNR and SSIM. The error plots shown below each reconstruction are displayed in the window $[0, 1.25]$.}
	\label{fig:fastmri:example_adv_supp}
\end{figure}

\begin{figure}[H]
 \centering
 \scriptsize
\begin{tabular}{l@{\,}c@{\,}c@{\,}c@{\,}c}
& 1\% rel.~noise -- Gauss. & 1.5\% rel.~noise -- Gauss. & 2.5\% rel.~noise -- Gauss. & 10\% rel.~noise -- Gauss. \\
\rotatebox[origin=c]{90}{$\TV[\noisebnd]$} &
\includegraphics[valign=c,width=0.15\textwidth]{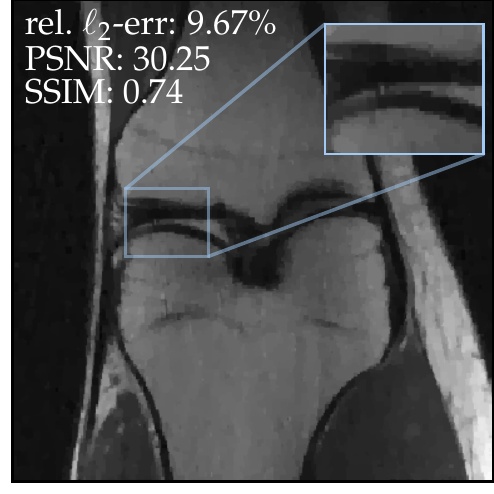} &
\includegraphics[valign=c,width=0.15\textwidth]{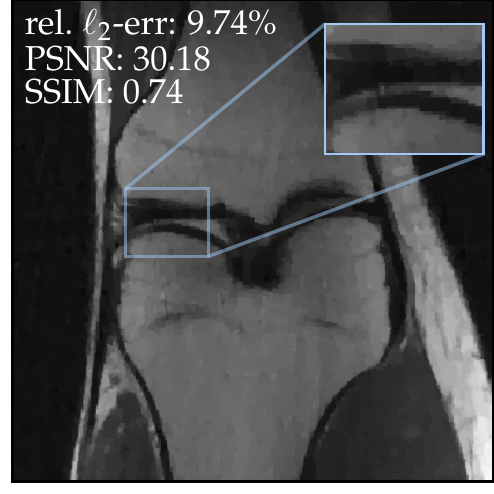} & \includegraphics[valign=c,width=0.15\textwidth]{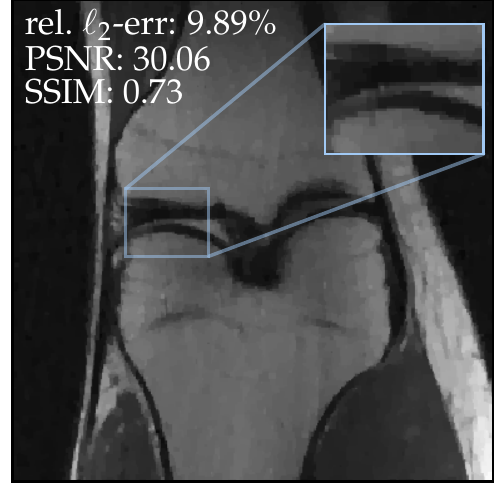} & \includegraphics[valign=c,width=0.15\textwidth]{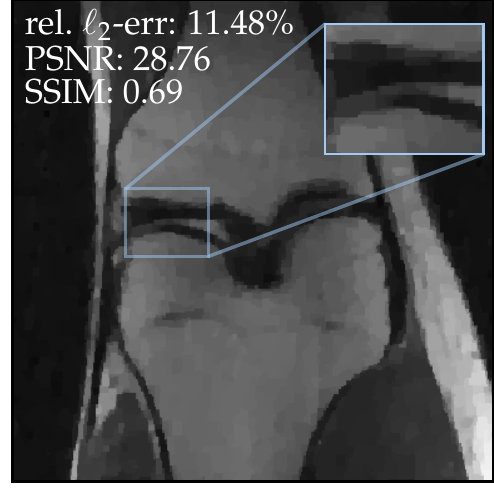} \\
&
\includegraphics[valign=c,width=0.15\textwidth]{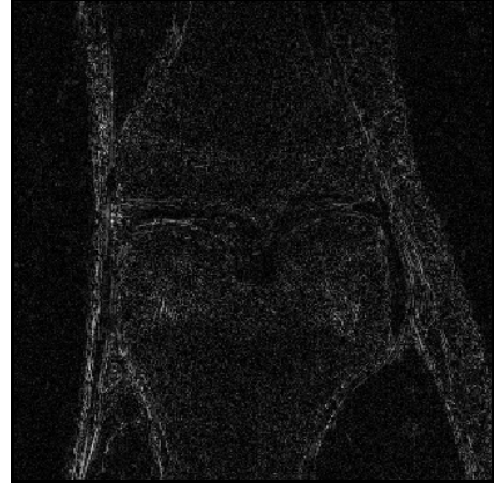} &
\includegraphics[valign=c,width=0.15\textwidth]{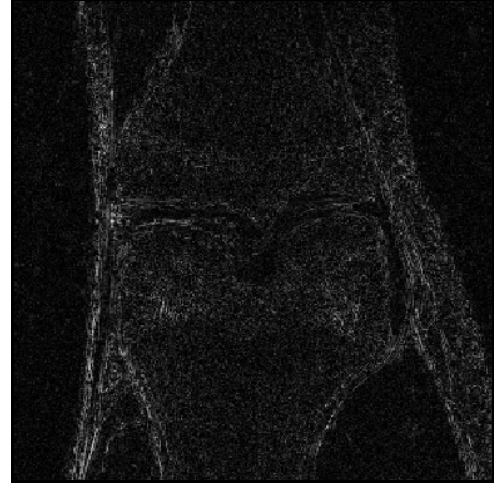} & \includegraphics[valign=c,width=0.15\textwidth]{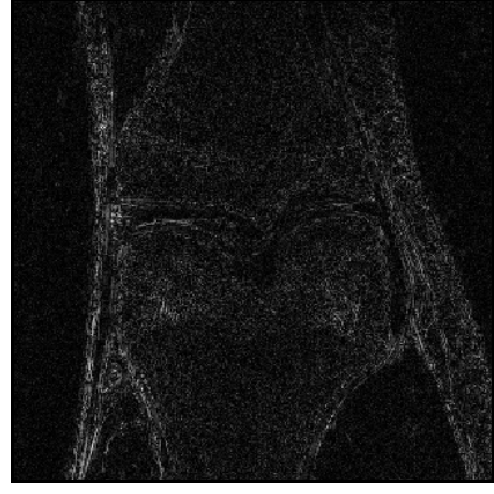} & \includegraphics[valign=c,width=0.15\textwidth]{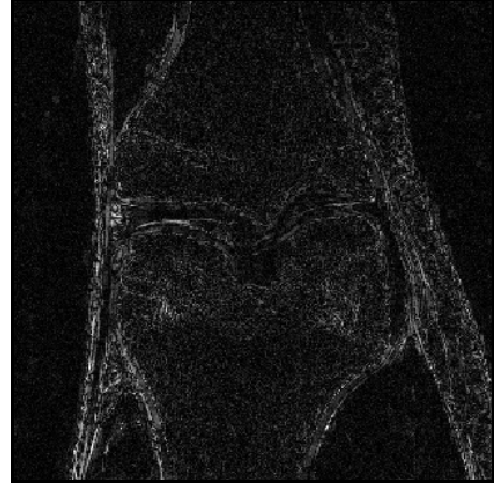} \\ \\

\rotatebox[origin=c]{90}{$\UNet$} &
\includegraphics[valign=c,width=0.15\textwidth]{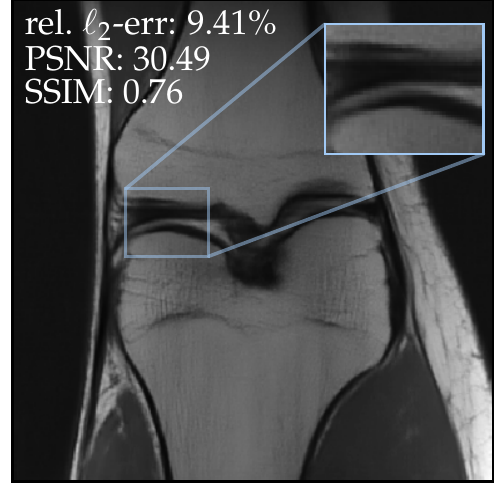} &
\includegraphics[valign=c,width=0.15\textwidth]{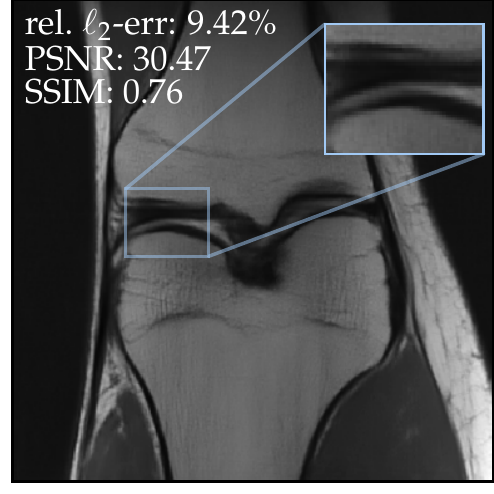} & \includegraphics[valign=c,width=0.15\textwidth]{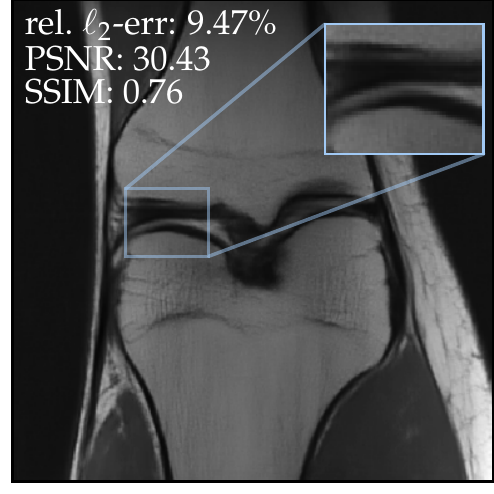} & \includegraphics[valign=c,width=0.15\textwidth]{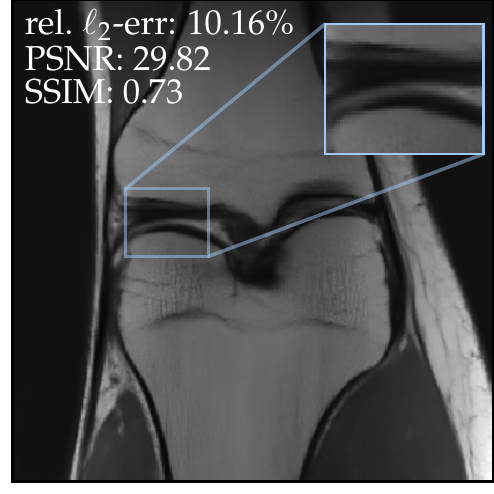} \\
&
\includegraphics[valign=c,width=0.15\textwidth]{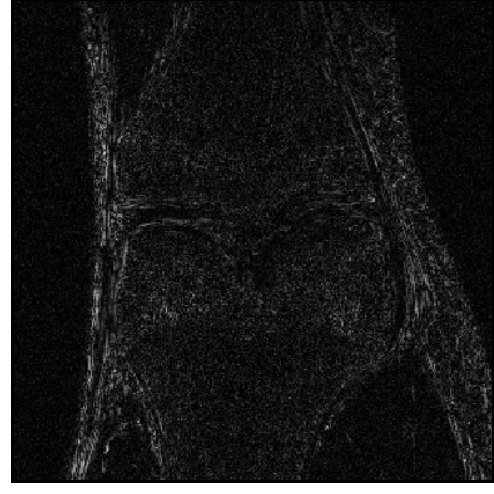} &
\includegraphics[valign=c,width=0.15\textwidth]{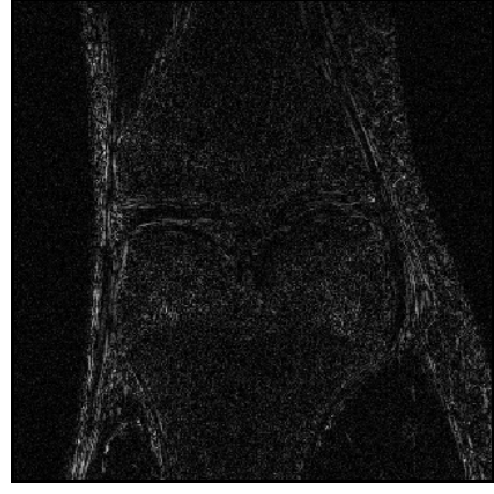} & \includegraphics[valign=c,width=0.15\textwidth]{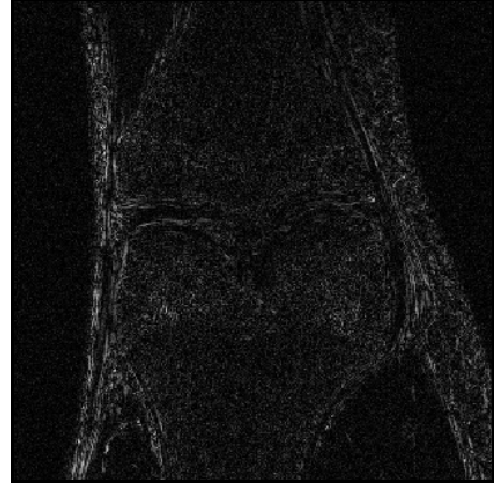} & \includegraphics[valign=c,width=0.15\textwidth]{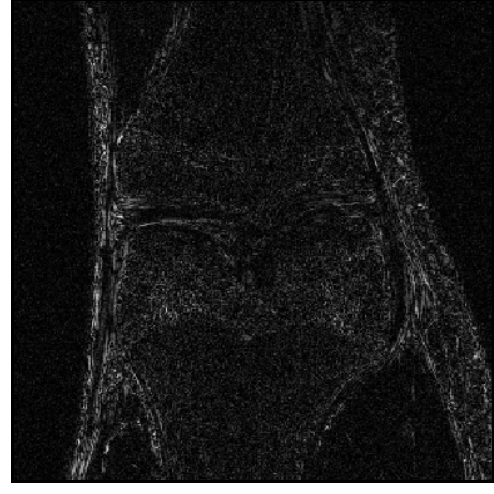} \\ \\

\rotatebox[origin=c]{90}{$\TiraFL$} &
\includegraphics[valign=c,width=0.15\textwidth]{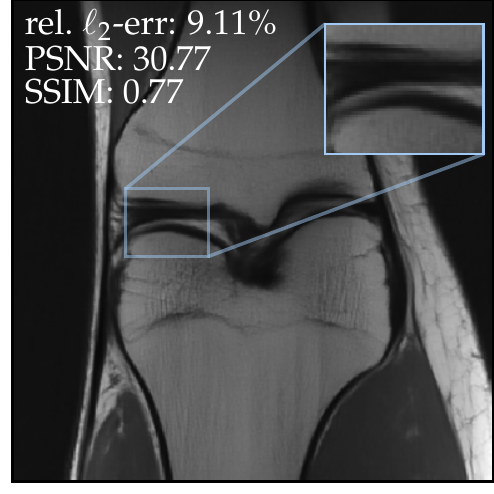} &
\includegraphics[valign=c,width=0.15\textwidth]{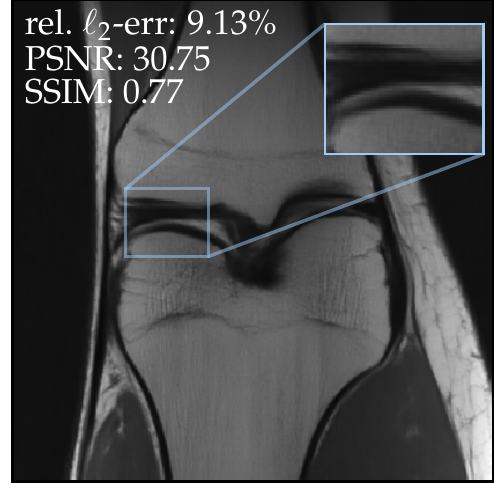} & \includegraphics[valign=c,width=0.15\textwidth]{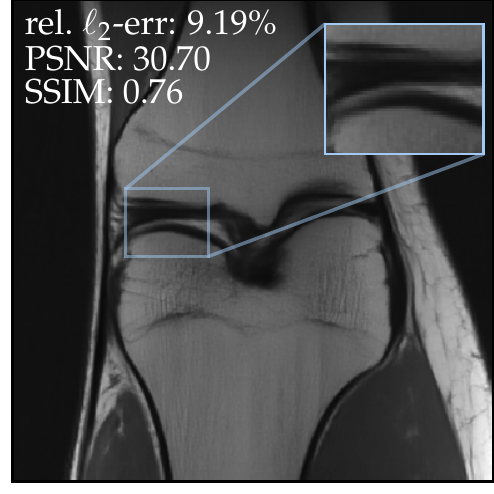} & \includegraphics[valign=c,width=0.15\textwidth]{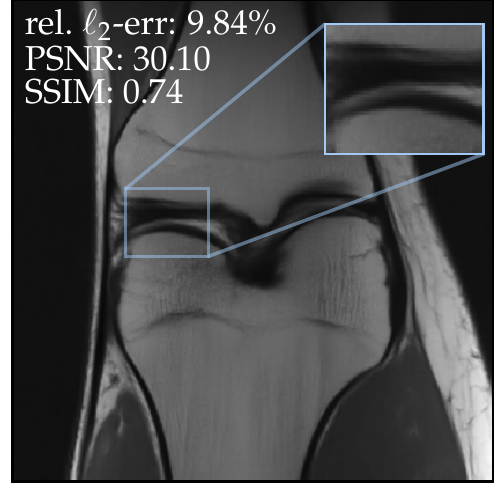} \\
&
\includegraphics[valign=c,width=0.15\textwidth]{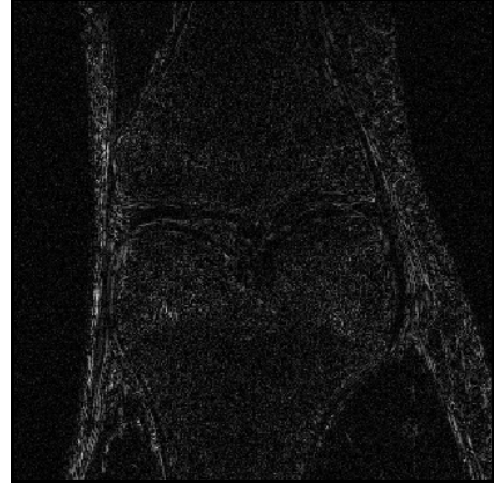} &
\includegraphics[valign=c,width=0.15\textwidth]{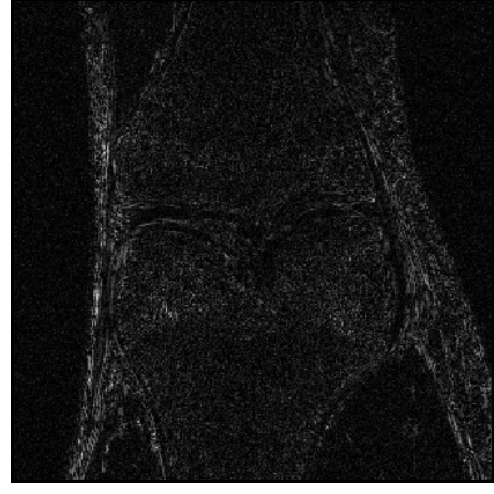} & \includegraphics[valign=c,width=0.15\textwidth]{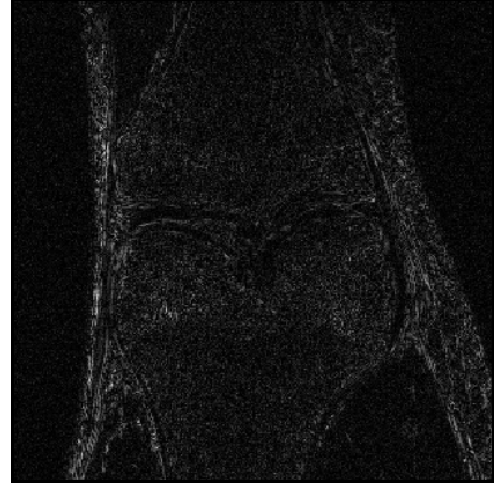} & \includegraphics[valign=c,width=0.15\textwidth]{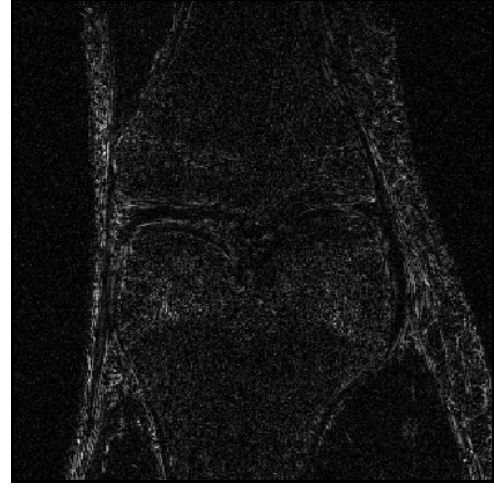} \\ \\

\rotatebox[origin=c]{90}{$\ItNet$} &
\includegraphics[valign=c,width=0.15\textwidth]{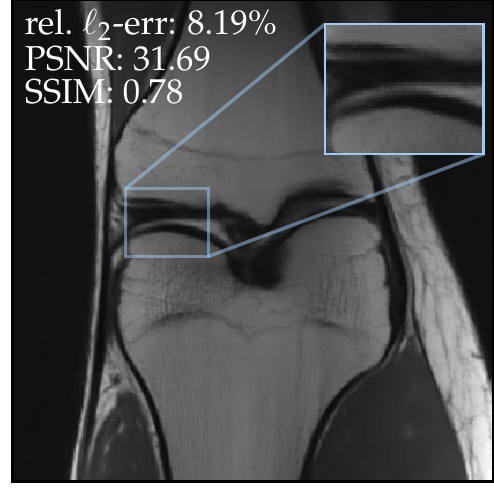} &
\includegraphics[valign=c,width=0.15\textwidth]{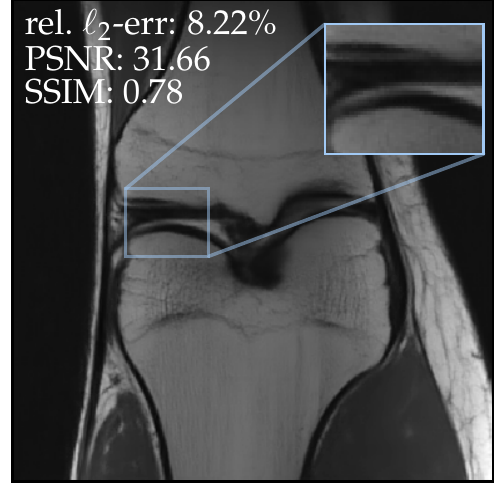} & \includegraphics[valign=c,width=0.15\textwidth]{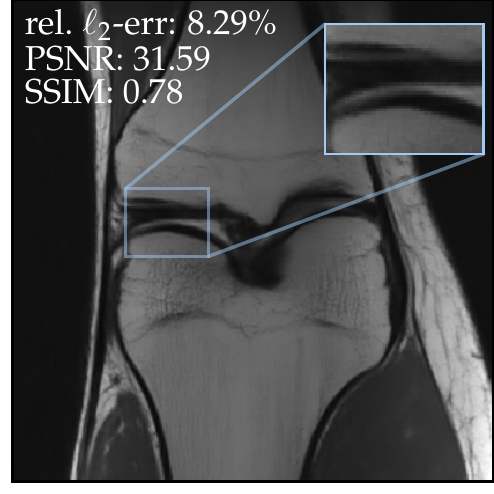} & \includegraphics[valign=c,width=0.15\textwidth]{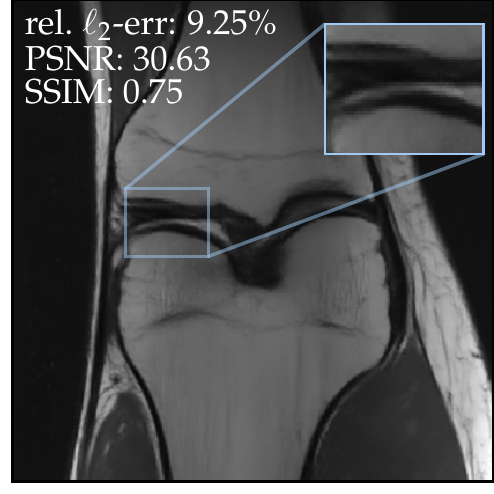} \\
&
\includegraphics[valign=c,width=0.15\textwidth]{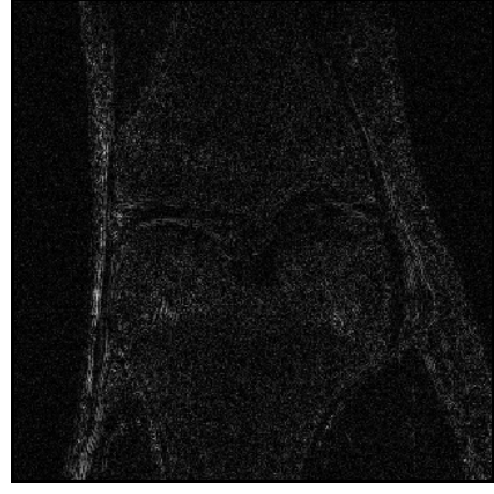} &
\includegraphics[valign=c,width=0.15\textwidth]{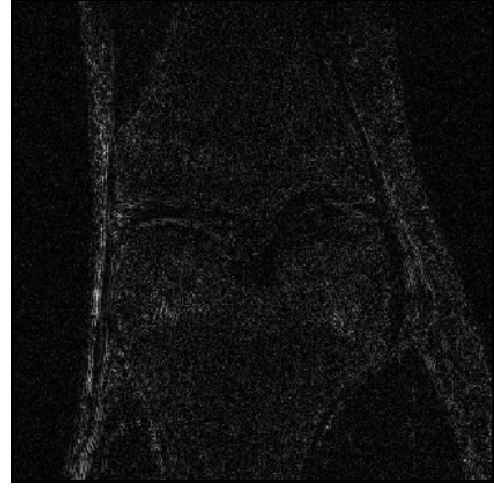} & \includegraphics[valign=c,width=0.15\textwidth]{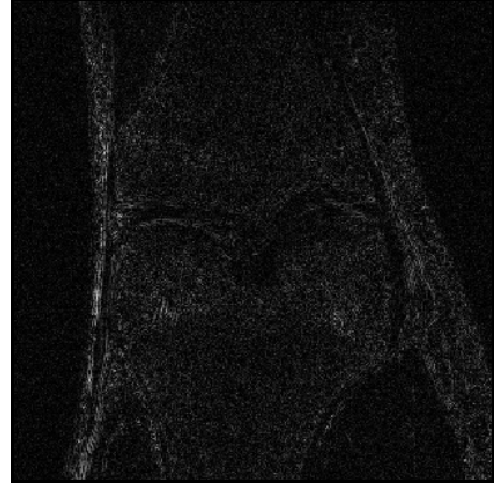} & \includegraphics[valign=c,width=0.15\textwidth]{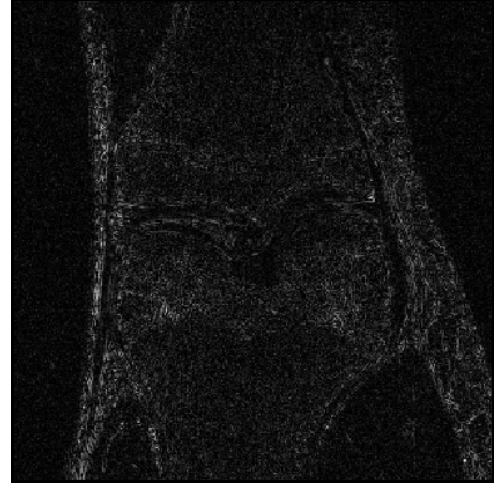}
\end{tabular}
\caption{\textbf{Case Study C -- fastMRI.} Individual reconstructions of the image from Fig.~\ref{fig:fastmri:example_adv} under Gaussian noise. The reconstructed images are displayed in the window $[0.05,4.50]$, which is also used for the computation of the PSNR and SSIM. The error plots shown below each reconstruction are displayed in the window $[0, 1.25]$. In favor of the more insightful noise level 10\%, we have omitted the noiseless case.}
\label{fig:fastmri:example_ref}
\end{figure}

\clearpage
\section{Supplementary Results for Section~\ref{sec:additional}}
\label{sec:supp:additional}

\begin{figure}[H]
	\centering
	\scriptsize
	\begin{tabular}{l|@{}c@{}c@{}c@{}c@{}c@{}c@{}c@{}c@{}c@{}c@{}c@{}c@{}c@{}c@{}c@{}c@{}l}
		& & it.~1 & & it.~2 & & it.~3 & & it.~4 & & it.~5 & & it.~6 & & it.~7 & & it.~8 & \\
		\rotatebox[origin=c]{90}{output $\UNetBB$} & \, $\inv\A\yadv \to $ &
		\includegraphics[valign=c,width=0.08\textwidth]{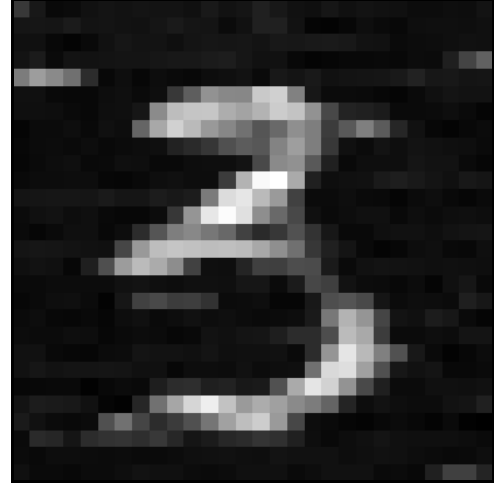} & &
		\includegraphics[valign=c,width=0.08\textwidth]{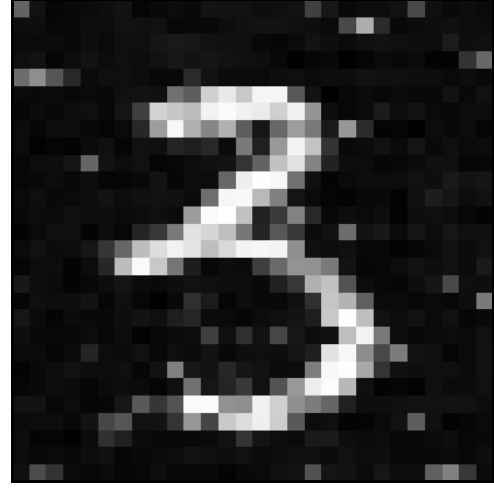} & &
		\includegraphics[valign=c,width=0.08\textwidth]{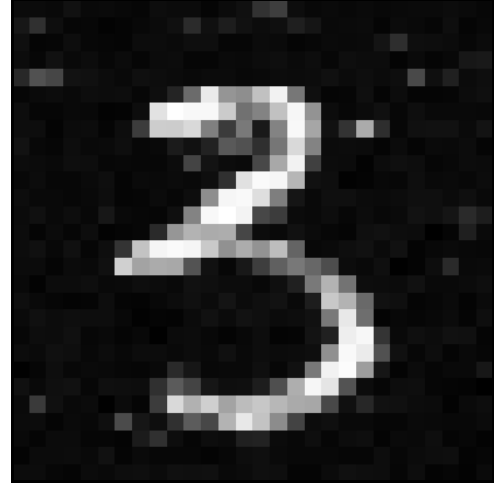} & &
		\includegraphics[valign=c,width=0.08\textwidth]{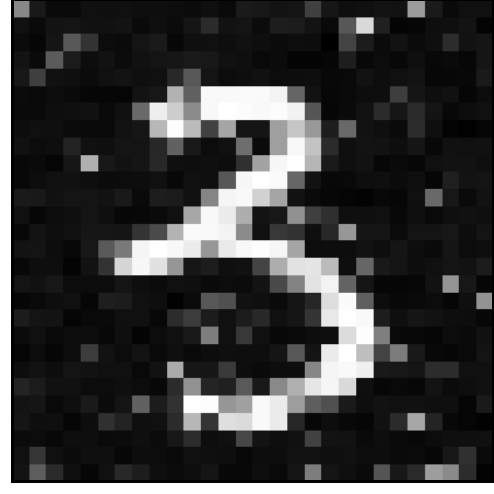} & &
		\includegraphics[valign=c,width=0.08\textwidth]{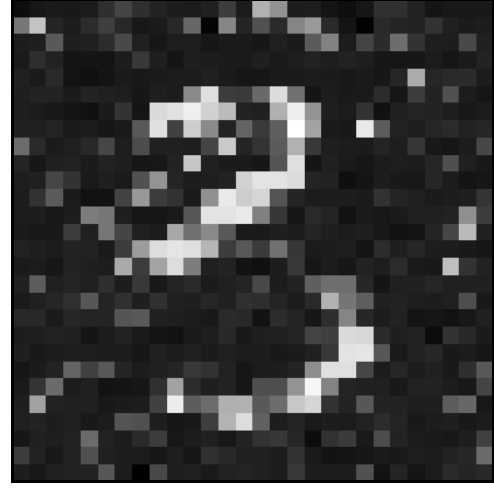} & &
		\includegraphics[valign=c,width=0.08\textwidth]{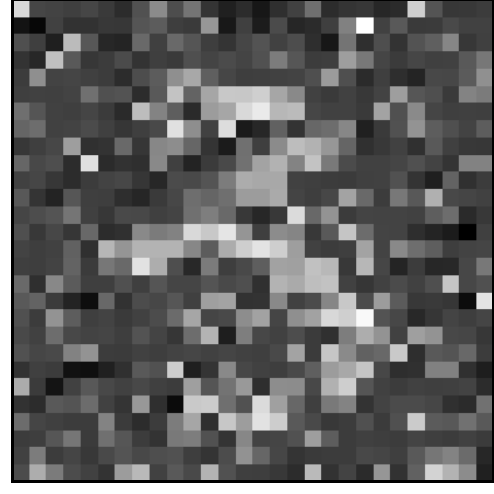} & &
		\includegraphics[valign=c,width=0.08\textwidth]{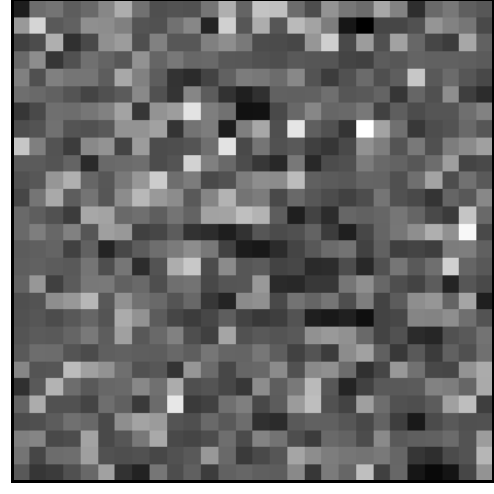} & &
		\includegraphics[valign=c,width=0.08\textwidth]{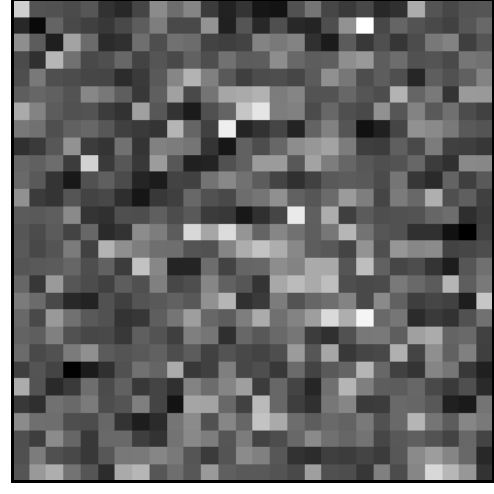} & \\
		& & $\downarrow$ & $\nearrow$ & $\downarrow$ & $\nearrow$ & $\downarrow$ & $\nearrow$ & $\downarrow$ & $\nearrow$ & $\downarrow$ & $\nearrow$ & $\downarrow$ & $\nearrow$ & $\downarrow$ & $\nearrow$ & $\downarrow$ & \\
		\rotatebox[origin=c]{90}{output $\DC$} & &
		\includegraphics[valign=c,width=0.08\textwidth]{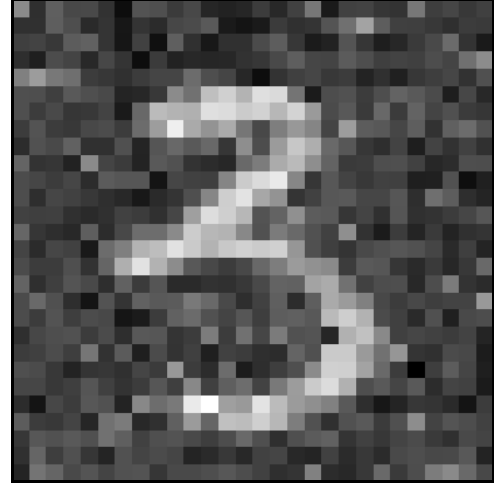} & &
		\includegraphics[valign=c,width=0.08\textwidth]{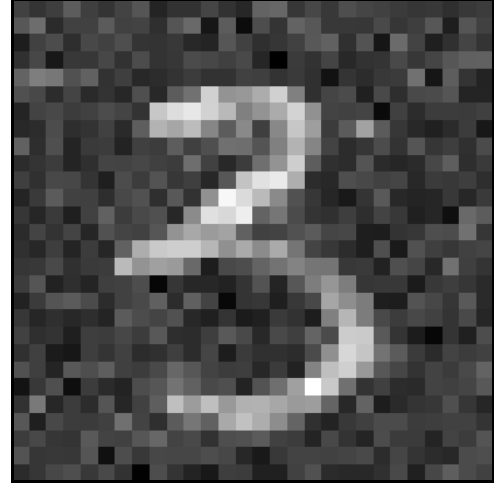} & &
		\includegraphics[valign=c,width=0.08\textwidth]{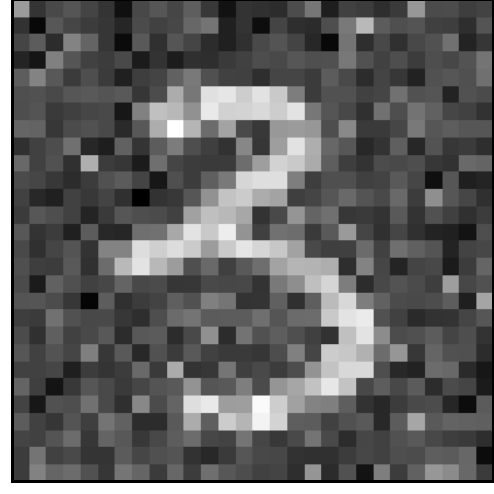} & &
		\includegraphics[valign=c,width=0.08\textwidth]{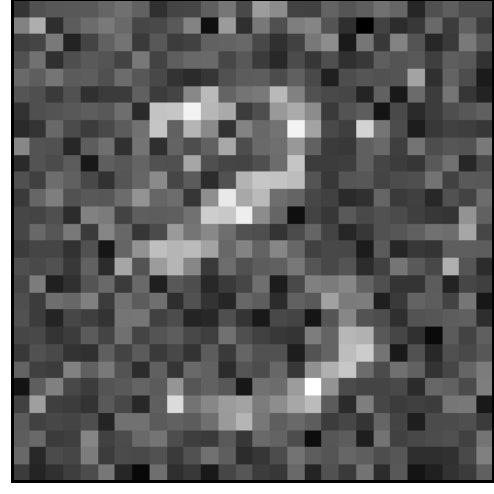} & &
		\includegraphics[valign=c,width=0.08\textwidth]{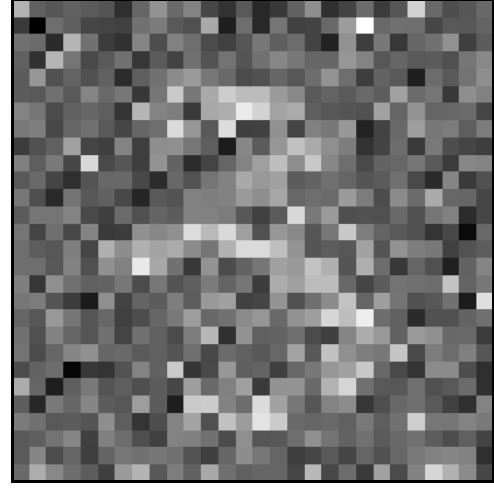} & &
		\includegraphics[valign=c,width=0.08\textwidth]{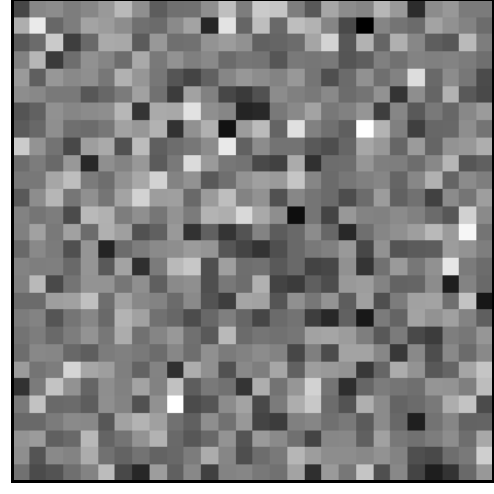} & &
		\includegraphics[valign=c,width=0.08\textwidth]{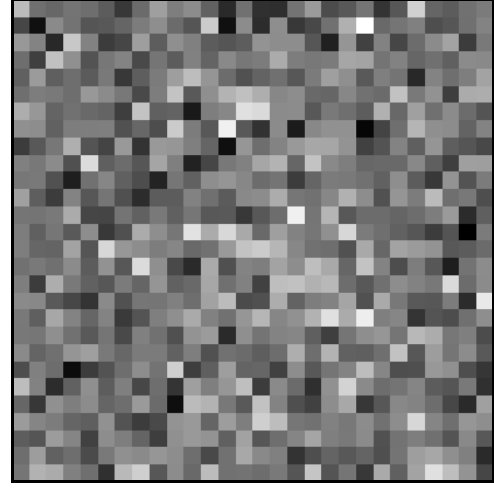} & &
		\includegraphics[valign=c,width=0.08\textwidth]{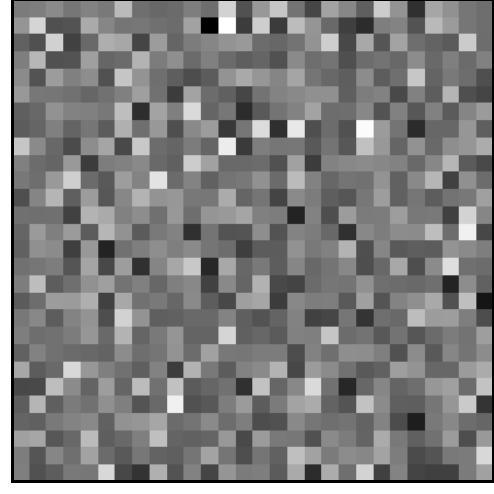} &
		\ \parbox{.1\textwidth}{$= \ItNet(\yadv)$ \\[.5em] w/o jittering}
	\end{tabular}
	\\[2em]

	\begin{tabular}{l|@{}c@{}c@{}c@{}c@{}c@{}c@{}c@{}c@{}c@{}c@{}c@{}c@{}c@{}c@{}c@{}c@{}l}
		& & it.~1 & & it.~2 & & it.~3 & & it.~4 & & it.~5 & & it.~6 & & it.~7 & & it.~8 & \\
		\rotatebox[origin=c]{90}{output $\UNetBB$} & \, $\inv\A\yadv \to $ &
		\includegraphics[valign=c,width=0.08\textwidth]{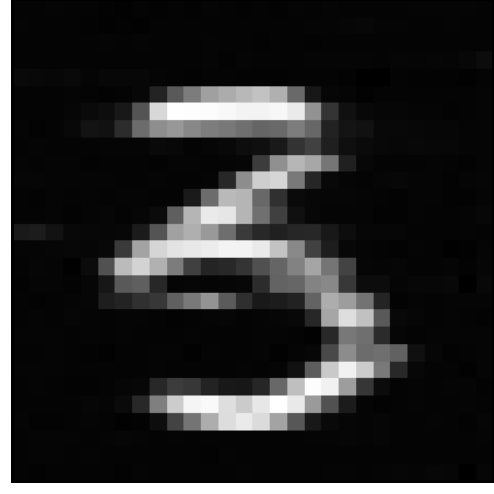} & &
		\includegraphics[valign=c,width=0.08\textwidth]{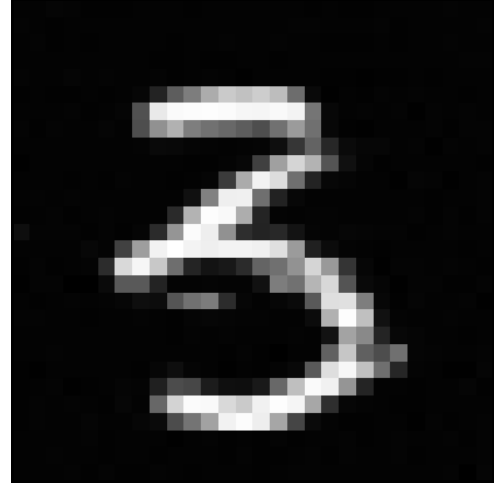} & &
		\includegraphics[valign=c,width=0.08\textwidth]{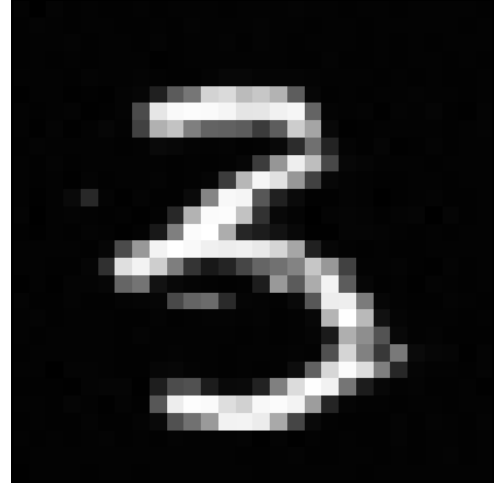} & &
		\includegraphics[valign=c,width=0.08\textwidth]{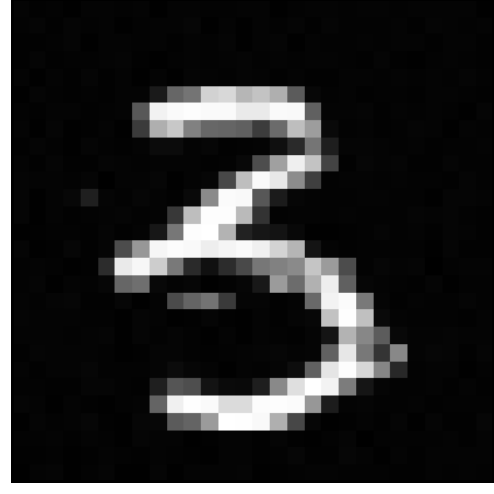} & &
		\includegraphics[valign=c,width=0.08\textwidth]{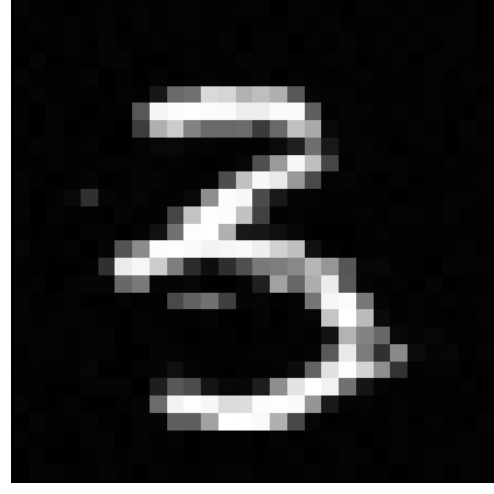} & &
		\includegraphics[valign=c,width=0.08\textwidth]{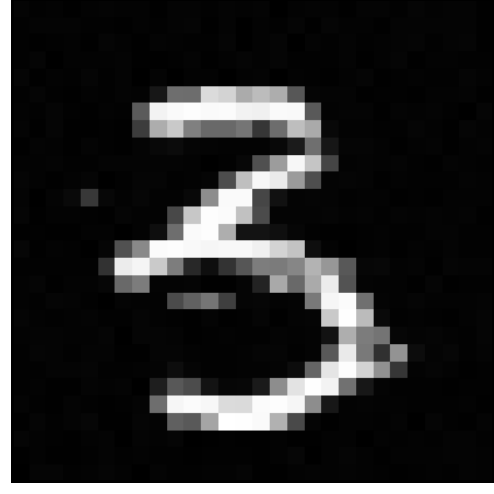} & &
		\includegraphics[valign=c,width=0.08\textwidth]{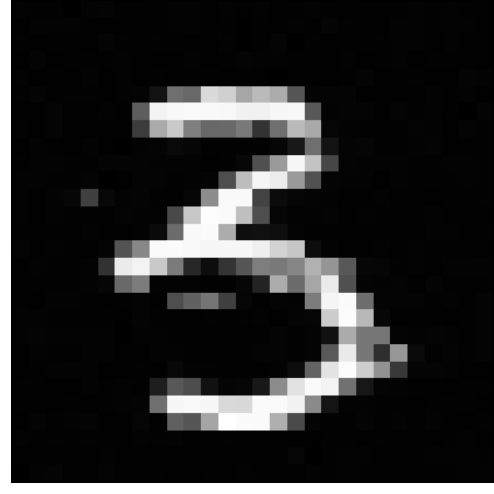} & &
		\includegraphics[valign=c,width=0.08\textwidth]{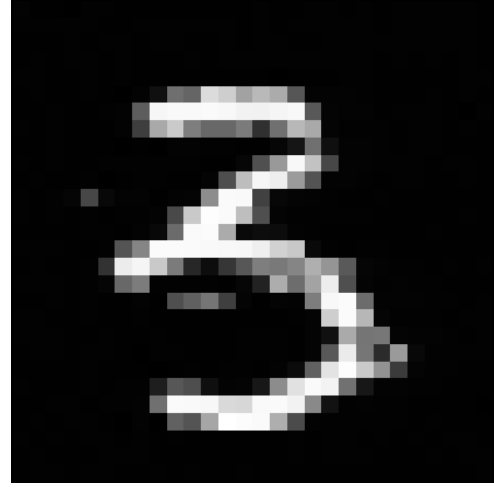} & \\
		& & $\downarrow$ & $\nearrow$ & $\downarrow$ & $\nearrow$ & $\downarrow$ & $\nearrow$ & $\downarrow$ & $\nearrow$ & $\downarrow$ & $\nearrow$ & $\downarrow$ & $\nearrow$ & $\downarrow$ & $\nearrow$ & $\downarrow$ & \\
		\rotatebox[origin=c]{90}{output $\DC$} & &
		\includegraphics[valign=c,width=0.08\textwidth]{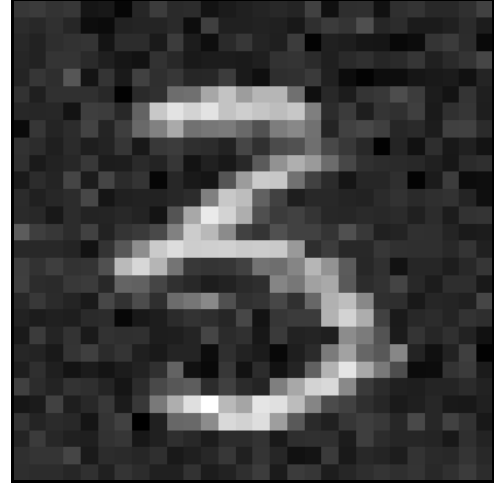} & &
		\includegraphics[valign=c,width=0.08\textwidth]{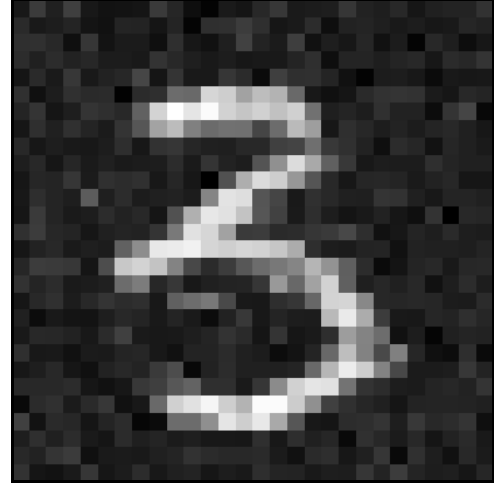} & &
		\includegraphics[valign=c,width=0.08\textwidth]{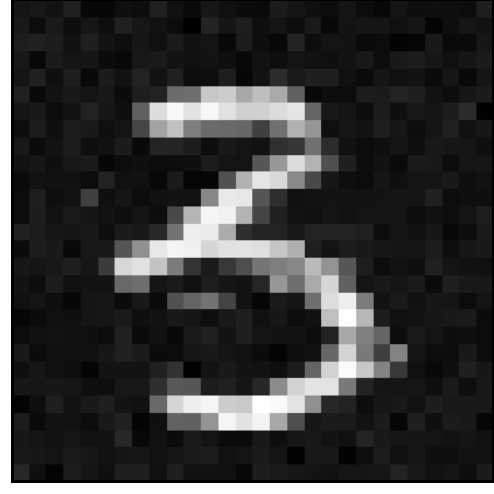} & &
		\includegraphics[valign=c,width=0.08\textwidth]{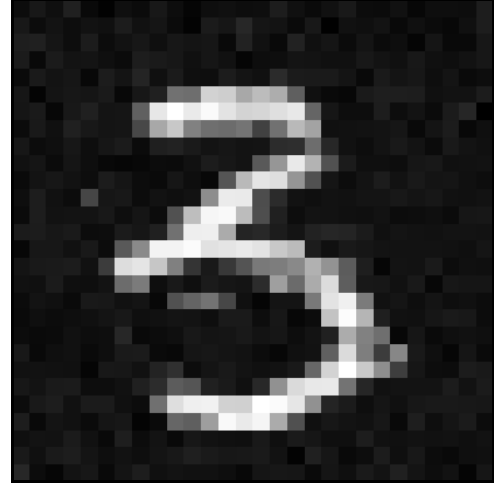} & &
		\includegraphics[valign=c,width=0.08\textwidth]{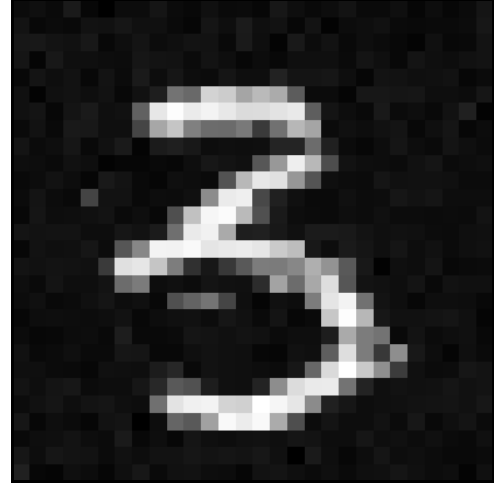} & &
		\includegraphics[valign=c,width=0.08\textwidth]{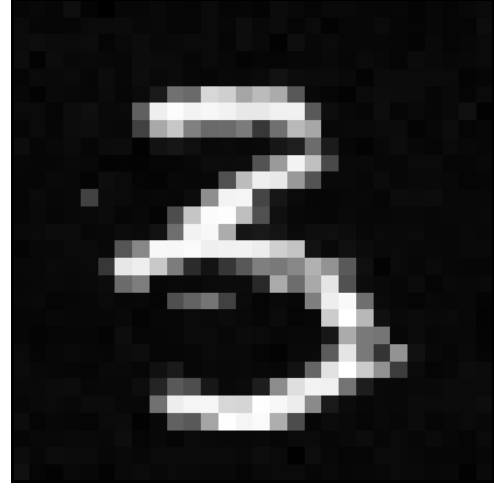} & &
		\includegraphics[valign=c,width=0.08\textwidth]{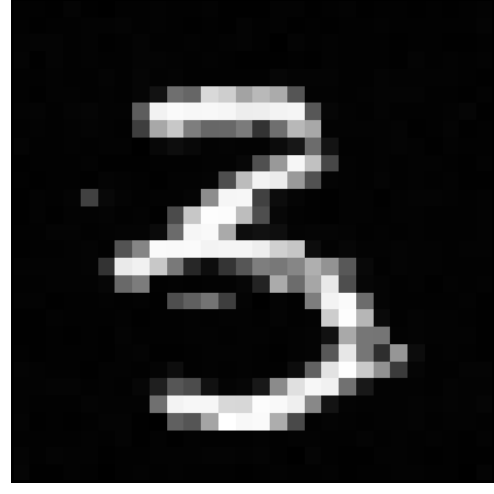} & &
		\includegraphics[valign=c,width=0.08\textwidth]{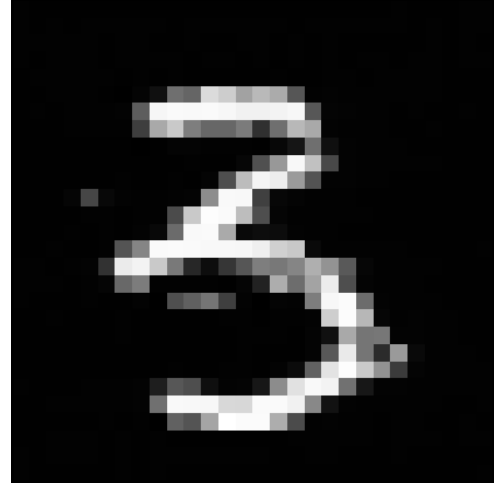} &
		\ \parbox{.1\textwidth}{$= \ItNet(\yadv)$ \\[.5em] w/ jittering}
	\end{tabular}
	\caption{\textbf{An inverse crime?} Intermediate steps performed by $\ItNet$ with and without jittering. The 20\%-adversarial perturbations correspond to the individual reconstructions shown in Fig.~\ref{fig:mnist:crime}.}
	\label{fig:crime:internal}
\end{figure}

\section{Choice of Hyper-Parameters}
\label{sec:supp:hyperparam}

Table~\ref{tab:networks:architectures} summarizes all hyper-parameters concerning the considered network architectures.
Table~\ref{tab:networks:training} shows the hyper-parameters selected for NN training.
Table~\ref{tab:networks:attack} contains relevant hyper-parameters for our adversarial attacks.

\begin{sidewaystable}
\centering
\footnotesize
\begin{tabular}{llllllll}
\toprule
 & & Piecewise Constant & MNIST & Ellipses (Fourier) & Ellipses (Radon) & fastMRI (radial) & fastMRI (challenge)  \\
\midrule
$\UNet$
    & inversion & Tikhonov ($0.02$) & Tikhonov ($0.02$) & $\adj\A$ & FBP (Hann-Filter) & $\adj\A$ & --\\
    & U-Net levels & 5 & 5 & 5 & 5 & 5\\
    & channels per level & (64, 128, 256, 512, 1024) & (64, 128, 256, 512, 1024) & (32, 64, 128, 256, 512) & (36, 72, 144, 288, 576) & (24, 48, 96, 192, 384) \\
\midrule
$\UNetFL$
    & inversion & learned, init Tikh. ($0.02$)  & learned, init Tikh. ($0.02$) & learned & -- & learned & -- \\
    & U-Net levels & 5 & 5 & 5 & & 5\\
    & channels per level  & (64, 128, 256, 512, 1024) & (64, 128, 256, 512, 1024) & (32, 64, 128, 256, 512) & & (24, 48, 96, 192, 384)\\
\midrule
$\Tira$
    & inversion & Tikhonov ($0.02$) & Tikhonov ($0.02$) & $\adj\A$ & -- & $\adj\A$ & $\adj\A$ \\
    & Tiramisu levels & 5 & 5 & 5 & & 5 & 5\\
    & dense blocks per level & (5, 7, 9, 12, 15) & (5, 7, 9, 12, 15) & (5, 7, 9, 12, 15) & & (5, 7, 9, 12, 15) & (6, 8, 10, 12, 14) \\
    & initial channels & 16 & 16 & 16 & & 12 & 16 \\
    & channel growth rate & 16 & 16 & 16 & & 12 & 12\\
    & bottleneck layers & 25 & 25 & 20 & & 18 & 20 \\
\midrule
$\TiraFL$
    & inversion & learned, init Tikh. ($0.02$) & learned, init Tikh. ($0.02$) & learned & -- & learned & -- \\
    & Tiramisu levels & 5 & 5 & 5 & & 5 \\
    & dense blocks per level & (5, 7, 9, 12, 15) & (5, 7, 9, 12, 15) & (5, 7, 9, 12, 15) & & (5, 7, 9, 12, 15) \\
    & initial channels & 16 & 16 & 16 & & 12\\
    & channel growth rate & 16 & 16 & 16 & & 12\\
    & bottleneck layers & 25 & 25 & 20 & & 18 \\
\midrule
$\ItNet$
    & inversion & Tikhonov ($0.02$) & Tikhonov ($0.02$) & $\adj\A$ & -- & $\adj\A$ & --\\
    & U-Net levels & 5 & 5 & 5 & & 5\\
    & channels per level & (64, 128, 256, 512, 1024) & (64, 128, 256, 512, 1024) & (32, 64, 128, 256, 512) & & (24, 96, 192, 384) \\
    & iterations & 8 & 8 & 8 & & 8\\
\midrule
$\ConvNet$
    & convolutional layers  & -- & 4 & -- & -- & -- & --\\
    & channels per layer & & (32, 32, 64, 64) \\
    & fully connected layers & &  3 \\
    & features per layer & &  (200, 200, 10) \\
    & remarks & & dropout ($p=0.5$) between fc layers \\
\bottomrule
\end{tabular}

\caption{Detailed description of hyper-parameters for all considered NN architectures. The convolution kernel sizes are $3$ or $3\times 3$, the max-pooling sizes are $2$ or $2\times 2$, and the activation functions are rectified linear units (ReLUs) for all networks.}
\label{tab:networks:architectures}
\end{sidewaystable}

\begin{sidewaystable}
\centering
\footnotesize
\begin{tabular}{llllllll}
\toprule
 & & Piecewise Constant & MNIST & Ellipses (Fourier) & Ellipses (Radon) & fastMRI (radial) & fastMRI (challenge)  \\
\midrule
$\UNet$
    & epochs  & (200, 75) & (200, 20) & (100, 10) & (155, 5) & (100) & -- \\
    & mini-batch size & (40, 40) & (40, 40) & (40, 40) & (40, 40) & (40) \\
    & learning rate & ($8\cdot 10^{-5}$, $5\cdot 10^{-5}$) & ($8\cdot 10^{-5}$, $5\cdot 10^{-5}$) & ($2\cdot 10^{-4}$, $5\cdot 10^{-5}$) & ($2\cdot 10^{-4}$, $5\cdot 10^{-5}$) & ($2\cdot 10^{-4}$) \\
    & weight decay & ($5\cdot 10^{-3}$, $5\cdot 10^{-3}$) & ($5\cdot 10^{-3}$, $5\cdot 10^{-3}$) & ($1\cdot 10^{-5}$, $1\cdot 10^{-5}$) & ($5\cdot 10^{-4}$, $1\cdot 10^{-4}$) & ($1\cdot 10^{-5}$)\\
    & Adam $\varepsilon$-parameter & ($1\cdot 10^{-5}$, $1\cdot 10^{-5}$) & ($1\cdot 10^{-5}$, $1\cdot 10^{-5}$) & ($1\cdot 10^{-5}$, $1\cdot 10^{-5}$) & ($1\cdot 10^{-4}$, $1\cdot 10^{-4}$) & ($1\cdot 10^{-4}$)\\
    & gradient accumulation & (1, 200) & (1, 200) & (1, 200) & (1, 200) & (1) \\
    & jittering level & 2 & 4 & 10 & 500 & 150 \\
\midrule
$\UNetFL$
    & epochs  & (250, 50) & (200, 75) & (100, 10) & -- & (45) & --\\
    & mini-batch size & (40, 40) & (40, 40) & (40, 40) & & (40)\\
    & learning rate & ($2\cdot 10^{-4}$, $5\cdot 10^{-5}$) & ($2\cdot 10^{-4}$, $5\cdot 10^{-5}$) & ($2\cdot 10^{-4}$, $5\cdot 10^{-5}$) & & ($8\cdot 10^{-5}$)\\
    & weight decay & ($5\cdot 10^{-4}$, $5\cdot 10^{-4}$) & ($5\cdot 10^{-3}$, $5\cdot 10^{-3}$) & ($1\cdot 10^{-3}$, $5\cdot 10^{-4}$) & & ($1\cdot 10^{-5}$) \\
    & Adam $\varepsilon$-parameter & ($1\cdot 10^{-5}$, $1\cdot 10^{-5}$) & ($1\cdot 10^{-5}$, $1\cdot 10^{-5}$) & ($1\cdot 10^{-5}$, $1\cdot 10^{-5}$) & & ($1\cdot 10^{-4}$) \\
    & gradient accumulation & (1, 200) & (1, 200) & (1, 200) & & (1) \\
    & jittering level & 2 & 4 & 10 & & 150 \\
    & remarks & & & & & init $\UNetBB$ from $\UNet$ \\
\midrule
$\Tira$
    & epochs  & (200, 50) & (200, 50) & (50, 10) & -- & (50, 16) & (40) \\
    & mini-batch size & (40, 40) & (40, 40) & (8, 5) & & (6, 6) & (4)  \\
    & learning rate & ($2\cdot 10^{-4}$, $5\cdot 10^{-5}$) & ($2\cdot 10^{-4}$, $5\cdot 10^{-5}$) & ($2\cdot 10^{-4}$, $5\cdot 10^{-5}$) & & ($8\cdot 10^{-5}$, $5\cdot 10^{-5}$) & ($1\cdot 10^{-4}$) \\
    & weight decay & ($5\cdot 10^{-4}$, $5\cdot 10^{-4}$) & ($1\cdot 10^{-5}$, $1\cdot 10^{-5}$) & ($1\cdot 10^{-5}$, $1\cdot 10^{-5}$) & & ($1\cdot 10^{-6}$, $1\cdot 10^{-6}$) & ($1\cdot 10^{-5}$) \\
    & Adam $\varepsilon$-parameter & ($1\cdot 10^{-5}$, $1\cdot 10^{-5}$) & ($1\cdot 10^{-5}$, $1\cdot 10^{-5}$) & ($1\cdot 10^{-5}$, $1\cdot 10^{-5}$) & & ($2\cdot 10^{-4}$, $2\cdot 10^{-4}$) & ($1\cdot 10^{-4}$) \\
    & gradient accumulation & (1, 200) & (1, 100) & (1, 200) & & (1, 1) & (1) \\
    & jittering level & 2 & 4 & 10 & & 150 & 10\\
\midrule
$\TiraFL$
    & epochs  & (200, 50) & (200, 50) & (30, 7) & -- & (50, 20) & --\\
    & mini-batch size & (40, 40) & (40, 40) & (10, 5) & & (6, 6)\\
    & learning rate & ($2\cdot 10^{-4}$, $5\cdot 10^{-5}$) & ($2\cdot 10^{-4}$, $5\cdot 10^{-5}$) & ($2\cdot 10^{-4}$, $5\cdot 10^{-5}$) & & ($8\cdot 10^{-5}$, $5\cdot 10^{-5}$)\\
    & weight decay & ($5\cdot 10^{-4}$, $5\cdot 10^{-4}$) & ($5\cdot 10^{-4}$, $5\cdot 10^{-4}$) & ($1\cdot 10^{-4}$, $1\cdot 10^{-5}$) & & ($1\cdot 10^{-6}$, $1\cdot 10^{-6}$)\\
    & Adam $\varepsilon$-parameter & ($1\cdot 10^{-5}$, $1\cdot 10^{-5}$) & ($1\cdot 10^{-5}$, $1\cdot 10^{-5}$) & ($2\cdot 10^{-4}$, $1\cdot 10^{-5}$) & & ($2\cdot 10^{-4}$, $2\cdot 10^{-4}$)\\
    & gradient accumulation & (1, 200) & (1, 200) & (1, 200) & & (1, 1)\\
    & jittering level & 2 & 4 & 10 & & 150 \\
    & remarks & & & init $\TiraBB$ from $\Tira$ \\
\midrule
$\ItNet$
    & epochs  & (100, 5) & (100, 10) & (35, 6) & -- & (15, 8) & --\\
    & mini-batch size & (40, 40) & (40, 40) & (15, 15) & & (10, 10)\\
    & learning rate & ($5\cdot 10^{-5}$, $2\cdot 10^{-5}$) & ($8\cdot 10^{-5}$, $5\cdot 10^{-5}$) & ($5\cdot 10^{-5}$, $5\cdot 10^{-5}$) & & ($5\cdot 10^{-5}$, $5\cdot 10^{-5}$)\\
    & weight decay & ($5\cdot 10^{-4}$, $5\cdot 10^{-4}$) & ($1\cdot 10^{-3}$, $1\cdot 10^{-3}$) & ($1\cdot 10^{-4}$, $1\cdot 10^{-4}$) & & ($1\cdot 10^{-6}$, $1\cdot 10^{-6}$)\\
    & Adam $\varepsilon$-parameter & ($1\cdot 10^{-5}$, $1\cdot 10^{-5}$) & ($1\cdot 10^{-5}$, $1\cdot 10^{-5}$) & ($2\cdot 10^{-4}$, $2\cdot 10^{-4}$) & & ($1\cdot 10^{-4}$, $1\cdot 10^{-4}$)\\
    & gradient accumulation & (1, 200) & (1, 200) & (1, 200) & & (1, 1)\\
    & jittering level & 2 & 4 & 10 & & 150 \\
    & remarks & & & init $\UNetBB$ from $\UNet$ & & init $\UNetBB$ from $\UNet$ \\
\midrule
$\ConvNet$
    & epochs  & -- & (20, 10) & -- & -- & -- & --\\
    & mini-batch size & & (40, 40)\\
    & learning rate & & ($2\cdot 10^{-4}$, $5\cdot 10^{-5}$) \\
    & weight decay & & ($1\cdot 10^{-5}$, $1\cdot 10^{-5}$) \\
    & Adam $\varepsilon$-parameter & & ($1\cdot 10^{-5}$, $1\cdot 10^{-5}$) \\
    & gradient accumulation & & (1, 1) \\
    & jittering level & & no \\
\bottomrule
\end{tabular}

\caption{Detailed description of hyper-parameters for all NN trainings. All networks are trained in 1 or 2 phases and respective parameters are shown per training phase. Default parameters for the Adam optimizer are used except for the $\varepsilon$ parameter, which is reported for all networks and training phases.}
\label{tab:networks:training}
\end{sidewaystable}

\begin{sidewaystable}
\centering
\footnotesize
\begin{tabular}{llllllll}
\toprule
 & & Piecewise Constant & MNIST & Ellipses (Fourier) & Ellipses (Radon) & fastMRI (radial) & fastMRI (challenge) \\
 \midrule
$\TV[\noisebnd]$
    & ADMM iterations (rec.)  & 50000 & 50000 & 5000 & 200 & 5000 & 5000 \\
    & ADMM iterations (init.) & 50000 & 50000 & 5000 & 200 & 5000 & 5000 \\
    & ADMM iterations (gradient) & 2000 & 2000 & 200 & 20 & 200 & 150\\
    & Adam iterations & 30 & 30 & 250 & 15  & 250 & 50\\
    & step size & $5\cdot 10^{0}$ & $5\cdot 10^{0}$ & $5\cdot 10^{0}$ & $5\cdot 10^{0}$ & $5\cdot 10^{0}$ & $5\cdot 10^{0}$ \\
    & random initializations & 200 & 100 & 6 & 6 & 6 & 6\\
\midrule
$\Net$
    & Adam iterations & 300  & 100 & 1000 & 500 & 250 & 200\\
    & step size & $5\cdot 10^{0}$ & $5\cdot 10^{0}$ & $5\cdot 10^{0}$ & $5\cdot 10^{0}$ & $5\cdot 10^{0}$ & $5\cdot 10^{0}$\\
    & random initializations & 200 & 100 & 6 & 6 & 6 & 6\\
\midrule
$\ConvNet \circ \TV[\noisebnd]$
    & ADMM iterations (rec.)  & -- & 50000 & -- & -- & -- & -- \\
    & ADMM iterations (init.) & & 50000 \\
    & ADMM iterations (gradient) & & 2000\\
    & Adam iterations & & 100 \\
    & step size & & $5\cdot 10^{-1}$ \\
    & random initializations & & 100 \\
\midrule
$\ConvNet \circ \Net$
        & Adam iterations & -- &  100 & -- & -- & -- & --\\
        & step size & & $5\cdot 10^{-1}$ \\
        & random initializations & & 100 \\
\bottomrule
\end{tabular}

\caption{Detailed description of hyper-parameters for finding adversarial perturbations. The parameters reported for $\Net$ apply equally to all network types.}
\label{tab:networks:attack}
\end{sidewaystable}


\end{document}